\documentclass[sn-mathphys,Numbered]{sn-jnl}

\usepackage{nicefrac}
\usepackage{graphicx}
\usepackage{color}
\usepackage{amsthm}
\usepackage{amsmath}
\usepackage{amsfonts}
\usepackage{enumerate}
\usepackage{algorithm, caption}
\usepackage[noend]{algpseudocode}
\usepackage{subcaption}
\usepackage{multirow}
\usepackage{booktabs}
\usepackage{dsfont}

\usepackage{comment}

\usepackage{tikz}
\usetikzlibrary{positioning}

\usepackage{tablefootnote}
\usepackage{subcaption}
\usepackage[section]{placeins}
\usepackage{hyperref}

\usepackage[open]{bookmark}
\usepackage{footmisc}

\usepackage{enumitem}

\usepackage{graphicx}%
\usepackage{graphicx,xcolor}%
\usepackage{multirow}%
\usepackage[load-configurations=version-1]{siunitx}
\usepackage{amsmath,amssymb,amsfonts}%
\usepackage{amsthm}%
\usepackage{mathrsfs}%
\usepackage[title]{appendix}%
\usepackage{xcolor}%
\usepackage{textcomp}%
\usepackage{manyfoot}%
\usepackage{booktabs}%
\usepackage{algorithm}%
\usepackage{algorithmicx}%
\usepackage{algpseudocode}%
\usepackage{listings}%

\theoremstyle{thmstyleone}%
\theoremstyle{thmstyletwo}%

\theoremstyle{thmstylethree}%

\usepackage{graphicx}
\usepackage{fancyhdr}
\usepackage{url}
\usepackage{color}
\usepackage{marvosym}

\usepackage{float}

\usepackage{caption}
\usepackage{subcaption}

\begin{document}

\title[]{Boosting t-SNE Efficiency for Sequencing Data: Insights from Kernel Selection}

\author[1]{\fnm{Avais} \sur{Jan}}\email{ajan3@student.gsu.edu}
\author[1]{\fnm{Prakash} \sur{Chourasia}}\email{pchourasia1@student.gsu.edu}
\author*[2]{\fnm{Sarwan} \sur{Ali}}\email{sa4559@cumc.columbia.edu}
\author*[1]{\fnm{Murray} \sur{Patterson}}\email{mpatterson30@gsu.edu}

\affil[1]{\orgdiv{Department of Computer Science}, \orgname{Georgia State University},\orgaddress{ \city{Atlanta}, \state{GA}, \country{USA}}}

\affil[2]{\orgdiv{Department of Neurology}, \orgname{Columbia University Irving Medical Center},\orgaddress{ \city{New York}, \state{NY}, \country{USA}}}



\abstract{ 

Dimensionality reduction techniques are essential for visualizing and analyzing high-dimensional biological sequencing data. t-distributed Stochastic Neighbor Embedding (t-SNE) is widely used for this purpose, traditionally employing the Gaussian kernel to compute pairwise similarities. However, the Gaussian kernel's lack of data-dependence and computational overhead limit its scalability and effectiveness for categorical biological sequences. Recent work proposed the isolation kernel as an alternative, yet it may not optimally capture sequence similarities. In this study, we comprehensively evaluate nine different kernel functions for t-SNE applied to molecular sequences, using three embedding methods: One-Hot Encoding, Spike2Vec, and minimizers. Through both subjective visualization and objective metrics (including neighborhood preservation scores), we demonstrate that the cosine similarity kernel in general outperforms other kernels, including Gaussian and isolation kernels, achieving superior runtime efficiency and better preservation of pairwise distances in low-dimensional space. We further validate our findings through extensive classification and clustering experiments across six diverse biological datasets (Spike7k, Host, ShortRead, Rabies, Genome, and Breast Cancer), employing multiple machine learning algorithms and evaluation metrics. Our results show that kernel selection significantly impacts not only visualization quality but also downstream analytical tasks, with the cosine similarity kernel providing the most robust performance across different data types and embedding strategies, making it particularly suitable for large-scale biological sequence analysis.

}



\maketitle

\section{Introduction}

The advent of high-throughput sequencing technologies has revolutionized biological research, generating unprecedented volumes of genomic data that require sophisticated computational methods for analysis and interpretation. The COVID-19 pandemic, caused by SARS-CoV-2, has particularly accelerated the generation of viral sequence data, with millions of genome sequences deposited in public databases~\cite{gisaid_website_url, elbe2017data}. This surge in sequencing data has pushed viral genomics into the ``Big Data'' realm, presenting significant challenges for traditional analytical approaches~\cite{stephens2015big}. Understanding the evolutionary dynamics and structural variations in viral proteins, especially the spike protein, which mediates viral entry into host cells, is crucial for tracking viral evolution, identifying variants of concern, and developing effective therapeutic strategies~\cite{harvey2021sars, walls2020structure}.

Machine learning (ML) techniques have emerged as powerful tools for analyzing biological sequence data, offering capabilities for classification, clustering, and pattern recognition~\cite{libbrecht2015machine, zou2019primer}. However, the high dimensionality of genomic data poses a fundamental challenge: biological sequences cannot be directly used as input to most ML models and must first be transformed into fixed-length numerical representations. This dimensionality challenge is further compounded when attempting to visualize and interpret the relationships between sequences. While various embedding methods such as k-mer based approaches~\cite{ali2021k}, word2vec-inspired techniques~\cite{asgari2015continuous, ali2021spike2vec}, and more recent deep learning methods~\cite{corso2021neural} have been proposed to convert sequences into numerical vectors, these representations often remain high-dimensional, necessitating further dimensionality reduction for effective visualization and analysis.

The t-distributed Stochastic Neighbor Embedding (t-SNE)~\cite{van2008visualizing} has become one of the most popular techniques for dimensionality reduction and visualization in computational biology. The method works by preserving pairwise similarities between data points when projecting from high-dimensional space to low-dimensional (typically 2D or 3D) representations. At its core, t-SNE relies on a kernel function to compute similarity matrices between pairs of sequences. The choice of kernel function is crucial as it determines how distances and relationships are measured in the original high-dimensional space. Traditionally, the Gaussian kernel (also known as the Radial Basis Function kernel) has been the default choice for t-SNE~\cite{van2008visualizing}. However, the Gaussian kernel is not data-dependent and uses Euclidean distance as its basis, which may not be optimal for categorical data such as biological sequences where edit distance or other sequence-specific metrics might be more appropriate~\cite{corso2021neural}.

Recent work by Zhu and Ting~\cite{zhu2021improving} proposed using the isolation kernel as an alternative to the Gaussian kernel for t-SNE, demonstrating improved computational efficiency and better preservation of local data structure. The isolation kernel is data-dependent and adapts to local density distributions, requiring only a single parameter for tuning. While this approach showed promise on standard benchmark datasets, its effectiveness for biological sequence data, which has unique characteristics such as a categorical nature and evolutionary relationships, remains understudied. Moreover, the isolation kernel may still not be optimal for capturing the specific types of similarities that are meaningful in biological contexts. The fundamental question remains: what is the most appropriate kernel function for t-SNE when working with biological sequence data?

The choice of kernel function is not merely a technical detail but can fundamentally impact the quality of dimensionality reduction and subsequent downstream analyses. Different kernels encode different notions of similarity, and the ``No Free Lunch'' theorem~\cite{wolpert1997no} reminds us that no single kernel systematically outperforms others across all domains and tasks. For biological sequences, the Euclidean distance underlying the Gaussian kernel may not capture the discrete, combinatorial nature of sequence dissimilarity~\cite{corso2021neural}. For instance, two sequences of different lengths might have large Euclidean distance despite small edit distance, potentially misrepresenting their biological similarity. This observation motivates a systematic investigation of alternative kernel functions that might better capture sequence relationships.

Beyond visualization, the choice of embedding method and kernel function can significantly impact downstream analytical tasks such as classification and clustering. Previous studies have shown that different embedding techniques yield varying performance in variant identification~\cite{ali2021k} and clustering of SARS-CoV-2 sequences~\cite{tayebi2021robust, melnyk2021alpha}. However, the interplay between embedding methods, kernel functions, and the performance of t-SNE for both visualization and subsequent machine learning tasks has not been comprehensively studied. Understanding these relationships is essential for developing robust analytical pipelines for biological sequence analysis.

In this paper, we present a comprehensive evaluation of nine different kernel functions for t-SNE applied to biological sequence data, with a focus on SARS-CoV-2 spike protein sequences and other genomic datasets. We evaluate these kernels using three different sequence embedding methods: One-Hot Encoding (OHE), Spike2Vec~\cite{ali2021spike2vec}, and minimizers. Our evaluation encompasses both subjective assessment through visual inspection of 2D scatter plots and objective quantification using neighborhood preservation metrics. Furthermore, we extend our analysis to assess how kernel choice impacts downstream classification and clustering tasks across six diverse biological datasets, providing insights into the practical implications of kernel selection for comprehensive sequence analysis pipelines. 

The contributions of this work are as follows:

\begin{itemize}
\item We conduct the first comprehensive evaluation of nine different kernel functions (cosine similarity, linear, polynomial, Gaussian, isolation, Laplacian, sigmoid, chi-squared, and additive-chi-squared) for t-SNE applied to biological sequence data, demonstrating that the cosine similarity kernel in general outperforms others, including the widely-used Gaussian kernel and the recently proposed isolation kernel.

\item We evaluate t-SNE performance using both subjective criteria (visual quality of 2D plots) and objective metrics (neighborhood preservation scores computed via k-ary neighborhood agreement), providing a rigorous assessment framework for dimensionality reduction quality.

\item We demonstrate that the cosine similarity kernel achieves superior computational efficiency compared to Gaussian and isolation kernels, with runtime analysis showing linear scaling behavior that makes it particularly suitable for large-scale biological sequence datasets.

\item We extend the evaluation beyond visualization to assess the impact of kernel choice on downstream analytical tasks, performing extensive classification experiments with seven ML algorithms (SVM, Naive Bayes, MLP, KNN, Random Forest, Logistic Regression, and Decision Tree) and clustering analysis with multiple evaluation metrics across six diverse biological datasets (Spike7k, Host, ShortRead, Rabies, Genome, and Breast Cancer).

\item We investigate the interaction between embedding methods and kernel functions, showing that while the cosine similarity kernel performs best across different embeddings, the choice of embedding method also significantly impacts overall performance, providing practical guidance for method selection in biological sequence analysis.
\end{itemize}

This work represents an extension of our preliminary conference paper~\cite{chourasia2023enhancing}, which initially explored kernel selection for t-SNE on SARS-CoV-2 spike sequences. The conference version is available online at \url{https://link.springer.com/chapter/10.1007/978-981-99-7074-2_35}.
The current manuscript extends that work by: 
(1) evaluating performance on five additional diverse biological datasets beyond the original Spike7k dataset, 
(2) conducting comprehensive classification and clustering experiments that were not included in the conference version, 
(3) providing detailed runtime analysis and scalability assessment, 
(4) including extensive experimental results with multiple machine learning algorithms and evaluation metrics, and 
(5) offering deeper insights into the interplay between embedding methods, kernel functions, and downstream analytical performance. 

\section{Related Work}
\label{sec_related_work}

Dimensionality reduction has been a fundamental challenge in machine learning and data analysis for decades. Principal Component Analysis (PCA)~\cite{wold1987principal1}, one of the earliest and most widely used techniques, performs linear dimensionality reduction by identifying orthogonal directions of maximum variance in the data. While computationally efficient and providing global structure preservation, PCA often fails to capture complex nonlinear relationships present in biological data~\cite{van2009dimensionality}. To address these limitations, various nonlinear dimensionality reduction techniques have been developed, including Isomap~\cite{tenenbaum2000global}, Locally Linear Embedding (LLE)~\cite{roweis2000nonlinear}, and Laplacian Eigenmaps~\cite{belkin2003laplacian}. These manifold learning methods attempt to preserve local neighborhood structure but often struggle with complex, high-dimensional biological datasets.

The t-distributed Stochastic Neighbor Embedding (t-SNE)~\cite{van2008visualizing} emerged as a significant advancement in dimensionality reduction, particularly for visualization purposes. Unlike its predecessor Stochastic Neighbor Embedding (SNE)~\cite{hinton2002stochastic}, t-SNE employs a Student t-distribution in the low-dimensional space to alleviate the crowding problem, resulting in more separated and interpretable clusters. The method has gained widespread adoption in computational biology, particularly for single-cell RNA sequencing analysis~\cite{kobak2019art}, flow cytometry data visualization~\cite{amir2013visne}, and genomic data exploration~\cite{platzer2013visualization}. However, t-SNE's computational complexity of $O(n^2)$ and sensitivity to hyperparameters have motivated various improvements and extensions.

Several variants of t-SNE have been proposed to address its limitations. Barnes-Hut t-SNE~\cite{van2014accelerating} reduces computational complexity to $O(n \log n)$ using spatial trees for approximating repulsive forces, enabling application to larger datasets. Parametric t-SNE~\cite{van2009learning} learns an explicit mapping function using neural networks, allowing for out-of-sample extension. Heavy-tailed Symmetric Stochastic Neighbor Embedding~\cite{yang2009heavy} explores different probability distributions in the embedding space. More recently, UMAP (Uniform Manifold Approximation and Projection)~\cite{mcinnes2018umap} has emerged as a competitive alternative to t-SNE, offering faster runtime and better preservation of global structure while maintaining local neighborhood relationships. Despite these advances, the fundamental question of optimal kernel selection for similarity computation in t-SNE remains relatively unexplored, particularly for domain-specific applications such as biological sequence analysis.


Kernel methods constitute a cornerstone of modern machine learning, providing a principled way to handle nonlinear relationships by implicitly mapping data into high-dimensional feature spaces~\cite{scholkopf2002learning, shawe2004kernel}. The kernel trick allows algorithms to operate in these high-dimensional spaces without explicitly computing the coordinates of the data in that space, requiring only the computation of inner products. Support Vector Machines (SVMs)~\cite{cortes1995support} popularized kernel methods, demonstrating their effectiveness for classification tasks. Various kernel functions have been proposed for different data types and applications, including polynomial kernels for capturing interactions between features~\cite{scholkopf2002learning}, Gaussian (RBF) kernels for smooth decision boundaries~\cite{scholkopf2002learning}, and string kernels specifically designed for sequence data~\cite{lodhi2002text}.

The choice of kernel function significantly impacts algorithm performance, yet this choice is often guided by empirical evaluation rather than theoretical guarantees. Kernel selection and optimization have been studied extensively in the context of SVMs~\cite{chapelle2002choosing} and Gaussian processes~\cite{rasmussen2006gaussian}. Multiple Kernel Learning (MKL)~\cite{gonen2011multiple} approaches attempt to automatically learn optimal kernel combinations from data. More recently, data-dependent kernels that adapt to local data characteristics have gained attention. The isolation kernel~\cite{ting2018isolation, zhu2021improving} represents a significant development in this direction, defining similarity based on the expected height at which two points are isolated in random partitioning trees, thereby adapting to local density distributions without requiring distance metric assumptions.

For biological sequence data, specialized kernels have been developed to capture domain-specific similarities. String kernels~\cite{lodhi2002text} measure similarity based on common subsequences, while spectrum kernels~\cite{leslie2002spectrum} have been specifically designed for protein and DNA sequences. Profile kernels~\cite{kuang2005profile} incorporate biological knowledge through position-specific scoring matrices. However, these specialized kernels have primarily been applied in supervised learning contexts (particularly for classification) and their effectiveness for unsupervised tasks like dimensionality reduction with t-SNE remains understudied.


The COVID-19 pandemic has generated an unprecedented volume of viral sequence data, with over 15 million SARS-CoV-2 genomes deposited in GISAID as of 2024~\cite{khare2021gisaid}. This data deluge has motivated extensive computational analysis efforts. Phylogenetic analysis has been the traditional approach for understanding viral evolution~\cite{hadfield2018a}, but the scale of available data has necessitated development of more scalable computational methods. Machine learning approaches have been increasingly applied to various aspects of SARS-CoV-2 analysis, including variant classification~\cite{ali2021k}, and host prediction~\cite{kuzmin2020machine}.

Clustering and classification of SARS-CoV-2 sequences have received particular attention. Multiple studies have explored different approaches to variant identification and lineage assignment~\cite{melnyk2021alpha, ali2021effective}. The Pangolin lineage assignment system~\cite{rambaut2020dynamic} provides a standardized nomenclature, while computational methods attempt to identify emerging variants and characterize their properties. Dimensionality reduction techniques, particularly t-SNE, have been employed to visualize the landscape of viral sequence diversity~\cite{ali2021k}, revealing patterns of viral evolution and geographic distribution of variants. However, these studies typically employ default t-SNE parameters and kernel functions without systematic evaluation of alternatives.

The spike protein of SARS-CoV-2 has been the focus of intense research due to its critical role in viral entry and its status as the primary target for vaccines and therapeutic antibodies~\cite{huang2020structural}. Mutations in the spike protein, particularly in the receptor-binding domain (RBD), can affect transmissibility, immune evasion, and disease severity~\cite{starr2020deep}. Understanding the relationships between spike protein sequences from different variants is crucial for pandemic surveillance and response. This motivates the need for effective visualization and analysis methods that can accurately represent sequence similarities and differences.


Converting biological sequences into numerical representations suitable for machine learning is a fundamental preprocessing step. Traditional approaches include one-hot encoding~\cite{kuzmin2020machine}, which creates binary vectors where each position corresponds to a specific residue, and physicochemical property encoding~\cite{meiler2001generation}, which represents amino acids by their biochemical properties. While simple and interpretable, these methods often produce high-dimensional sparse representations that may not capture sequence context.

K-mer based methods~\cite{leslie2002spectrum} have gained popularity for representing sequences through the frequencies of contiguous subsequences of length k. These methods have been successfully applied to various sequence analysis tasks~\cite{ali2021k} and form the basis for alignment-free sequence comparison~\cite{zielezinski2017alignment}. Minimizers~\cite{roberts-2004-minimizer, schleimer2003winnowing}, originally developed for efficient sequence indexing, select a representative subset of k-mers based on lexicographic ordering, providing a more compact representation while maintaining sequence similarity relationships~\cite{marccais2017locality}.

Deep learning-based representations have emerged as powerful alternatives. Word2vec-inspired approaches~\cite{asgari2015continuous} treat biological sequences as sentences and k-mers as words, learning distributed representations that capture semantic relationships. Protein language models such as ESM~\cite{rives2021biological}, ProtBERT~\cite{elnaggar2021prottrans}, and more recent foundation models~\cite{lin2023evolutionary} learn contextualized embeddings from large sequence databases, capturing evolutionary and functional information. For SARS-CoV-2 specifically, Spike2Vec~\cite{ali2021spike2vec} adapts the word2vec paradigm to learn embeddings from spike protein sequences. However, these sophisticated embedding methods still benefit from dimensionality reduction for visualization and exploratory analysis.


Assessing the quality of dimensionality reduction is challenging due to the inherent information loss when projecting from high to low dimensions. Evaluation approaches can be broadly categorized into subjective (visual) and objective (quantitative) methods. Visual assessment~\cite{van2008visualizing} examines the interpretability and separation of clusters in low-dimensional plots but is inherently subjective and can be influenced by perceptual biases. Quantitative metrics attempt to measure specific aspects of embedding quality.

Local structure preservation metrics focus on whether neighborhoods are maintained during dimensionality reduction. The k-ary neighborhood agreement~\cite{lee2015multi} measures the overlap between k-nearest neighbors in high and low dimensions. The trustworthiness and continuity metrics~\cite{venna2006local} quantify false neighbors (points that are neighbors in low dimension but not in high dimension) and missing neighbors (vice versa), respectively. The co-ranking matrix~\cite{lee2009quality} provides a comprehensive framework for evaluating both local and global structure preservation.

Global structure preservation can be assessed through metrics such as the preservation of pairwise distances or rank correlations between distance matrices~\cite{chen2009local}. The normalized stress~\cite{kruskal1964nonmetric} measures the difference between distances in the original and embedded spaces. 
The Area Under the RNX curve (AUC$_{RNX}$)~\cite{lee2013type} aggregates neighborhood preservation across multiple scales, providing a single comprehensive score. These objective metrics enable rigorous comparison of different dimensionality reduction methods and parameter settings.


While extensive research has been conducted on dimensionality reduction techniques, kernel methods, and biological sequence analysis, several important gaps remain. First, despite the widespread use of t-SNE in computational biology, the impact of kernel function selection on t-SNE performance for biological sequences has not been systematically investigated. Most studies employ the default Gaussian kernel without exploring alternatives that might be more appropriate for categorical sequence data. Second, while the isolation kernel has been proposed as an improvement over the Gaussian kernel for general datasets~\cite{zhu2021improving}, its effectiveness for biological sequences with their unique properties (discrete alphabet, evolutionary relationships, varying lengths) remains unexplored. Third, the interplay between sequence embedding methods and kernel functions has not been comprehensively studied, leaving practitioners without clear guidance on method selection. Finally, most prior work focuses solely on visualization quality without examining how kernel choice affects downstream analytical tasks such as classification and clustering, which are often the ultimate goals of sequence analysis pipelines. This paper addresses these gaps through systematic evaluation of nine kernel functions across multiple embedding methods and datasets, with assessment spanning visualization quality, computational efficiency, and downstream task performance.

\section{Kernel Matrix Computation}
\label{sec_proposed_approach}
In this section, we describe different methods to compute the kernel matrix.

\subsection{Cosine similarity}
Cosine similarity works on any two given row vectors by computing the L2-normalized dot product between them, 
\begin{equation}
    k(x, y)  = \frac{\|x\|\|y\| \times \mathrm{cos} (\theta)}{\|x\|\|y\|} =\frac{x.y}{\|x\|\|y\|}
\end{equation}
     
It is the cosine of the angles (a.k.a angle of separation) between the two points derived from the euclidean normalization of vectors. It uses the cosine value of the angle between two row vectors. The values range between 1 and 0. Where 1 denotes two row vectors are similar while 0 represents opposite. The kernel matrix $n \times n$ is computed using the cosine similarity-based method for n sequences.
    
\subsection{Linear Kernel}
The linear kernel is a specific case of the polynomial kernel with a degree $1$. Given two column vectors, it takes the inner product of these vectors and adds an optional constant $c$,
\begin{equation}
     k(x, y)= x^Ty + c
\end{equation}
It can be applied to linearly separable data.

\subsection{Polynomial Kernel}
Given two row vectors x and y, it gives the relationship between them across dimensions. Can be defined by,  
\begin{equation}
     k(x, y)=\left( x^T y + r\right)^{d}
\end{equation}    
where r is polynomial's coefficient and d represents the degree of the polynomial. And these parameters are determined using cross-validation.
The polynomial kernel is used to compute the relationship between observations in higher dimensions when data is non-linearly separable in lower dimensions.

\subsection{Gaussian Kernel} 
For any N points, the N-dimensional Gaussian kernel is computed by calculating the kernel coordinates of each point in the data space. For a point, the coordinates are determined by calculating its distance to the chosen points and taking the Gaussian function of these distances. In short, this kernel translates the dot product in N-dimensional space into the Gaussian function of the distance between points in the data space.
The Gaussian kernel in 2D is defined as,

\begin{equation}
\mathcal{K}\left(x, y\right)=\exp \left(\frac{-\left\|x-y\right\|^{2}}{2 \sigma^{2}}\right)
\end{equation} 

here $\sigma$ defines the width of the kernel. We use Squared Euclidean distance to compute the difference with perplexity value $250$ and we are tuning $\sigma$.

\subsection{Isolation Kernel}
It is a data-dependent kernel that is efficient to compute due to only one parameter being used. Unlike the Gaussian kernel, it adapts to local density distribution. The formal definition of isolated kernel given by authors of ~\cite{zhu2021improving} is as follows:

For points $x$, $y \in \mathrm{R}^d$, the Isolation kernel of $x$ and $y$ wrt $\mathrm{D}$ is defined to be the expectation taken over the probability distribution on all partitioning $\mathrm{H} \in \mathrm{H_\psi} (\mathrm{D})$ that both $x$ and $y$ fall into the same isolating partition $\theta[z]\in \mathrm{H}, z\in D$:
\begin{equation}
K_{\psi}(x, y \mid D)=\mathbb{E}_{\mathbb{H}_{\psi}(D)}[\mathbb{1}(x, y \in \theta[z] \mid \theta[z] \in H)]
\end{equation}

\subsection{Laplacian}
The laplacian kernel is a variant of RBF where the Manhattan distance between input feature vectors is used instead of Euclidean distance. Can be denoted by,  
\begin{equation}
k(x, y)=\exp \left(-\frac{1}{2 \sigma^2}\|x-y\|_{1}\right)
\end{equation} 
    
where $\mathrm{x}$ and $\mathrm{y}$ are the input vectors and $\|x-y\|_{1}$ is the Manhattan distance between the input vectors.
    
\subsection{Sigmoid Kernel}
Computes the sigmoid kernel between two input feature vectors $x$ and $y$. The sigmoid kernel is also known as a hyperbolic tangent or multilayer perceptron. It is defined as, 
\begin{equation}
k(x, y)=\tanh \left(\gamma x^{\top} y+c_{0}\right)
\end{equation}    
    
where $x$, $y$ are the input vectors, $\gamma$ is slope, and $c_0$ is intercept.
 
\subsection{Chi-squared Kernel}
The input feature vector is assumed to be non-negative and is often normalized to have an L1-norm of one. The normalization is rationalized with the connection to the chi-squared distance, which is the distance between discrete probability distributions. The chi-squared kernel is given by:
\begin{equation}
k(x, y)=\exp \left(-\gamma \sum_{i} \frac{(x_i-y_i)^{2}}{x_i+y_i}\right)
\end{equation}

\subsection{Additive-chi-squared (Additive-chi2) Kernel}
It is a Fourier approximation to chi-squared 
kernel. Given feature vectors $x$ and $y$, it computes the additive-chi2 kernel. Its idea is to consider $x$ and $y$ to be non-negative. The additive-chi2 kernel is given by:
\begin{equation}
k(x, y)=\sum_{i} \frac{2 x_{i} y_{i}}{x_{i}+y_{i}}
\end{equation}

\section{Applying t-SNE with Kernel Matrices}\label{sec_tsne_compute}
The t-SNE algorithm accepts an $n\times n$ kernel matrix as its input, with $n$ representing the count of sequences (embedding vectors). It generates a lower-dimensional representation with dimensionality $d$, where $d = \{1, 2, \cdots, n-1\}$. In essence, t-SNE aims to transform high-dimensional (HD) data into a low-dimensional (LD) space while maintaining the pairwise distances among embedding vectors. The objective is to ensure that inter-point distances in the LD space $Y$ mirror those in the HD space $X$ as faithfully as possible. The methodology proceeds through the following stages:

\subsection{Computing Pairwise Affinities}
Initially, t-SNE computes the Euclidean distance from each data point to every other point in the high-dimensional space. Various kernel functions can be employed for this purpose. These distances are subsequently transformed into conditional probabilities, referred to as pairwise affinities or similarity measures.

\subsection{Probability Distribution in High Dimensions}
The conditional probabilities are aggregated to form a joint distribution over point pairs, organized into a symmetric matrix. The joint probability is expressed as shown in Equation~\ref{eq_hd_jointprobability} (following~\cite{xue2020classification}): 

\begin{equation}
\label{eq_hd_jointprobability}
p_{j \mid i}=\frac{\exp \left(-\left\|x_i-x_j\right\|^2 / 2 \sigma_i^2\right)}{\sum_{k \neq i} \exp \left(-\left\|x_i-x_k\right\|^2 / 2 \sigma_i^2\right)}
\end{equation}

\subsection{Initialization of Low-Dimensional Solution}
The low-dimensional representation $Y$ is initialized with randomly generated values. These initial values undergo iterative optimization to achieve the optimal lower-dimensional mapping of the data.

\subsection{Probability Distribution in Low Dimensions}
Analogous to the high-dimensional case, conditional probabilities are computed for the low-dimensional space and aggregated into a joint probability distribution, defined as (per~\cite{saha2021privacy}):  

\begin{equation}
Q_{i j}= \begin{cases}0 & j=i \\ \frac{\left(1+\left\|\mathbf{y}_{i}-\mathbf{y}_{j}\right\|^{2}\right)^{-1}}{\sum_{k \neq l}\left(1+\| \mathbf{y}_{k}-\mathbf{y}_{l}||^{2}\right)^{-1}} & j \neq i\end{cases}
\label{eq_ld_jointprobability}
\end{equation}

\subsection{KL Divergence Calculation and Gradient Computation}
To quantify the discrepancy between the HD distribution $P$ and LD distribution $Q$, the Kullback-Leibler (KL) divergence is utilized. This metric captures the distributional differences in pairwise distances. The KL divergence is formulated as: 

\begin{equation}
\label{eq_compute_Kl_divergence}
J = min \hspace{0.3cm} KL(P \| Q)=\sum_{i} \sum_{j} p_{i j} \log \frac{p_{i j}}{q_{i j}}
\end{equation}

where $J$ denotes the objective function to be minimized, $P$ and $Q$ represent the HD and LD distributions respectively, and $y_i$, $y_j$ are coordinates in the LD space $Q$.

The gradient of the cost function is derived using Equation~\ref{eq_gradient_descent} (as presented in~\cite{xue2020classification}) to identify the minimum of $J$: 

\begin{equation} \label{eq_gradient_descent}
\nabla = \frac{\delta J}{\delta y_{i}}=4 \sum_{j}\left(p_{i j}-q_{i j}\right)\left(y_{i}-y_{j}\right)\left(1+\left\|y_{i}-y_{i}\right\|^{2}\right)^{-1}
\end{equation}

The LD coordinates $Y$ are iteratively refined to minimize $J$ according to Equation~\ref{eq_update_y}. The application of the t-distribution in this Stochastic Neighborhood Embedding (SNE) framework, specifically applied to the LD distribution $Q$, produces heavier tails that enhance visualization quality.

\begin{equation} \label{eq_update_y}
y^{(t)}=y^{(t-1)}+\eta \nabla +\alpha(t)\left(y^{(t-1)}-y^{(t-2)}\right)
\end{equation}
where $t \in T$ denotes the iteration number.

\begin{figure}[h!]
    \centering
    \includegraphics[scale=0.2]{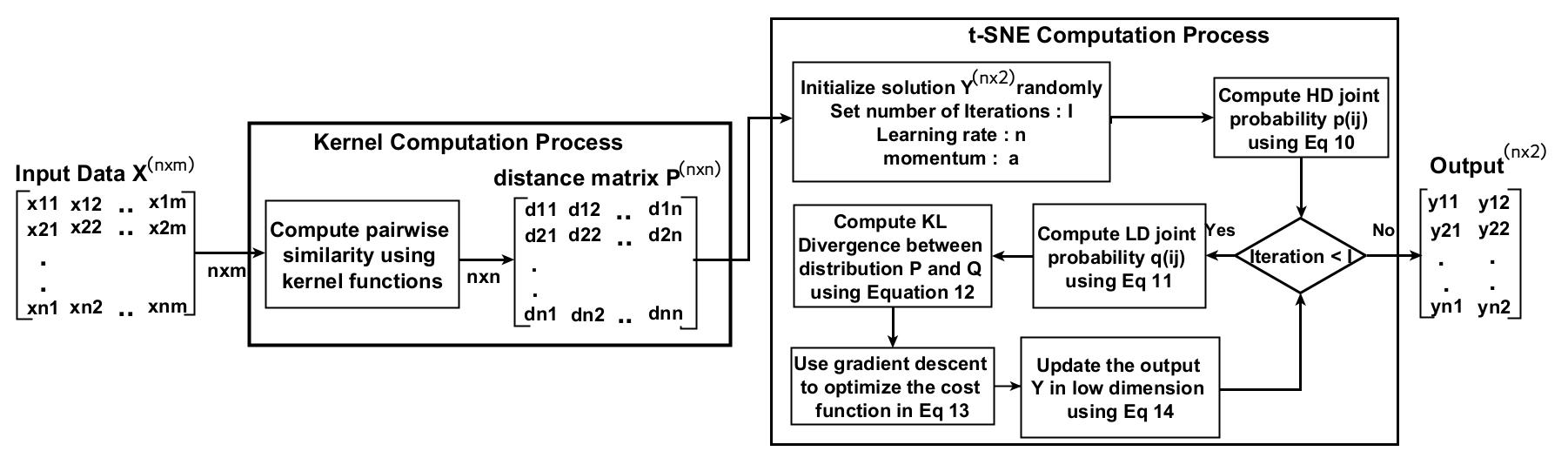}
    \caption{Overall architecture of the proposed approach. Starting from the input data, we compute a kernel matrix through various kernel functions, followed by the t-SNE algorithm to obtain the low-dimensional embedding.}
    \label{process_flow_chart}
\end{figure}

Figure~\ref{process_flow_chart} illustrates the t-SNE workflow. The procedure begins with computing pairwise affinities through kernel functions to construct a similarity matrix. Probability values are then derived from the distances for points $X$ (where $X \in R^n$, representing the high-dimensional space). Subsequently, the solution $Y$ is initialized randomly. The joint probabilities in both HD ($P_{ij}$) and LD ($Q_{ij}$) are computed. KL divergence measures the distributional discrepancy, and through gradient descent of this objective function, the optimal $Y$ emerges after multiple iterations. This optimized $Y$ constitutes the LD representation of the HD data points $X$.

The initial step of deriving probabilities from distances is fundamental. The calculated distances significantly influence how data distributes on the t-distribution for visualization and affects dimensionality reduction effectiveness. Kernel functions are crucial in determining inter-point distances in the original data space. Algorithm~\ref{algo_tsne} presents the detailed workflow for t-SNE computation using a kernel matrix as input.

\begin{algorithm}[h!]
 \scriptsize
    \caption{t-SNE Algorithm Workflow.
    }
    \label{algo_tsne}
	\begin{algorithmic}[1]

	\State \textbf{Input:} {$(KM, dim)$}  
	
	$KM =  x[i][j], x_{ij} \in R^n$ \Comment{KM represents the Kernel Matrix}
	
    dim: target dimensionality \Comment{desired output dimension}
    
	\State \textbf{Output:} $Y = [y_1,y_2,y_3, \ldots, y_N], y_i \in R^{dim}$ 
	
\State \textbf{Function:} tSNE($KM$, dim)  
        
        \State  $Y =  matrix(n, dim)$ \Comment{initialize output with random values}
        
        \State   //Initialize hyperparameters for optimization
        
        \State $I$ = 1000 \Comment{Maximum iterations}
        
        \State $\eta = 500$ \Comment{Step size parameter}
        
        \State $\alpha = 0.5$  \Comment{momentum coefficient}
        
        \State $P = matrix(n \times n)$  \Comment{HD probability distribution}
    
       \For{$i \leftarrow 1 \textup { to } n$} 
           \For{$j \leftarrow 1 \textup { to } n$}
            \State $P_{ij}$ = \Call{computeProbability}{$KM_{i,j}$} \Comment{via Eq~\ref{eq_hd_jointprobability}}
           \EndFor
        \EndFor
        
        \State $Q = matrix(n \times 2)$  \Comment{LD probability distribution}
        
        \For{$k \leftarrow 1 \textup { to } I$} 
        \Comment{Main optimization loop}
        
         \For{$i \leftarrow 1 \textup { to } n$}
           \For{$j \leftarrow 1 \textup { to } n$}
            
            \State $Q_{ij}$ = \Call{computeProbability}{$Y_{i,j}$} \Comment{via Eq.~\ref{eq_ld_jointprobability}}
            
           \EndFor
        \EndFor

        \State $J$ = \Call{computeKLDivergence}{$P,Q$}
        \\ \Comment{Calculate KL Divergence using Eq~\ref{eq_compute_Kl_divergence}}
        
        \State $\nabla$ = \Call{computeGradient}{$J$} \Comment{via Eq~\ref{eq_gradient_descent}}
        
        \State $Y$ = \Call{updateOutput}{$Y, \alpha, \eta, \nabla$} \Comment{via Eq~\ref{eq_update_y}}

        \If{($k$ == $250$)} 
           $\alpha$ =  $0.8$   \Comment{adjust momentum at iteration 250} 
        \EndIf
     \EndFor
\State \Return $Y$  
	\end{algorithmic}
\end{algorithm}

\section{Generation of Feature Embeddings}\label{sec_embeddings}
\begin{figure}[h!]
\centering
    \begin{subfigure}{0.5\textwidth}
        \centering
        \includegraphics[scale = 0.23] {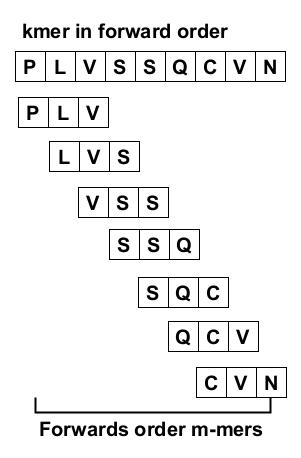}
        \caption{$k$mer in forward order}
        \label{fig_kmer_fwd}
    \end{subfigure}%
    \begin{subfigure}{0.5\textwidth}
        \centering
        \includegraphics[scale = 0.23] {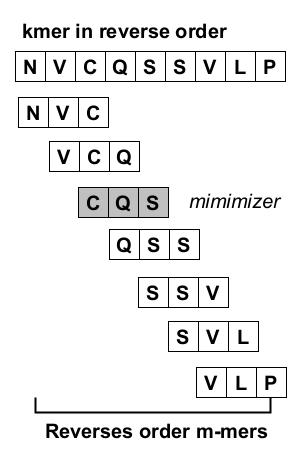}
        \caption{$k$mer in reverse order}
        \label{fig_kmer_rev}
    \end{subfigure}%
\caption{Computation of minimizer from $k$mer.}
\label{minimizer_generation}
\end{figure}

This section presents three distinct embedding techniques employed to transform biological sequences into fixed-dimensional numerical vectors.
\begin{enumerate}
\item One Hot Encoding (OHE) - We employ OHE~\cite{kuzmin2020machine,ali2021k} to encode amino acids into their numerical counterparts. 
\item Spike2Vec~\cite{ali2021spike2vec} - This approach produces fixed-dimensional numerical vectors by leveraging the $k$-mer (n-gram) methodology.
The technique utilizes a sliding window mechanism to extract substrings (termed mers) of length $k$ (corresponding to the window size) as demonstrated in Figure~\ref{kmer_generation}.
\begin{figure}[h!]
\centering
  \includegraphics[scale=0.27] {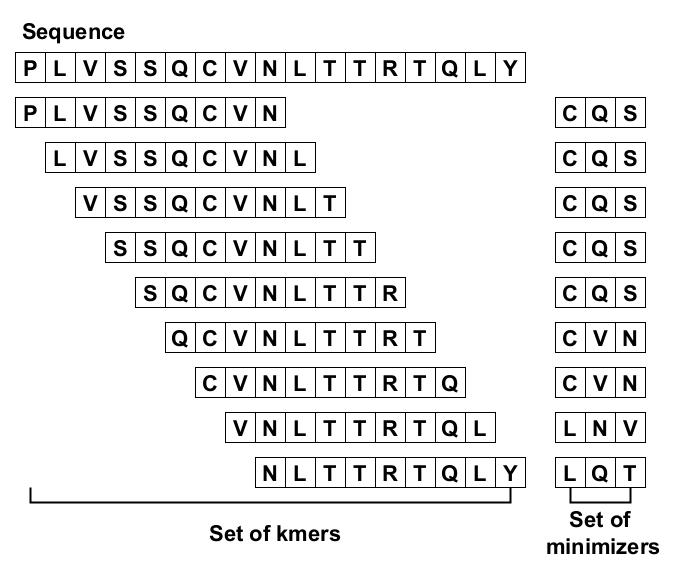}
  \caption{Illustration of $k$mers and minimizers for the amino acid sequence ``PLVSSQCVNLTTRTQLP".}
  \label{kmer_generation}
\end{figure}
\item Minimizers  - From each $k$mer, a minimizer is extracted as a substring (mer) with length $m$ (satisfying $m < k$) that represents the lexicographically smallest element when comparing both forward and reverse orientations of the $k$-mer. The minimizer computation process is depicted in Figure~\ref{minimizer_generation} and detailed in Algorithm~\ref{algo_minimizer}.
\item Spaced k-mers~\cite{singh2017gakco}  - This method generates a
non-contiguous variant of k-mers by initially constructing g-mers,
where g > k. Subsequently, the k-mer is derived from the g-mer
by selecting the first k amino acids/nucleotides when arranged
in lexicographic order considering both forward and reverse orientations. The final embedding is constructed based on the frequency
of these non-contiguous k-mers.
\end{enumerate}
\begin{algorithm}[h!]
	\caption{Algorithm for Minimizer Extraction}
        \label{algo_minimizer}
	\begin{algorithmic}[1]
	\State \textbf{Input:} Sequence $seq$, parameters $klen$ and $mlen$
	\State \textbf{Output:} Set of Minimizers
	
	\State minis = $\emptyset$
    \State q = [] \Comment{$ \text{queue for m-mers }$}
    \State ind = 0  \Comment{$ \text{index tracking current minimizer}$}
    \\
    \For{$i \leftarrow 1 \textup { to }|seq|-klen+1$}   
        \State kmer = $seq[i:i+klen]$

        \If{\textup{ind} $> 1$} 
            \State q.dequeue 
            \State mmer = $seq[i+klen-mlen:i+k]$  \Comment{$  \text{extract new m-mer}$}
            \State ind $\leftarrow$ ind $- 1$  \Comment{$  \text{adjust minimizer index}$}
            \State mmer = min(mmer, backward(mmer))  \Comment{$  \text{select lexicographic minimum}$}
            \State q.enqueue(mmer) 
            
            \If{\textup{mmer $<$ q[ind]}} 
                ind = $klen-mlen$ \Comment{$  \text{refresh minimizer position}$}
        \EndIf
        \Else
                
                \State q = [] \Comment{$  \text{reinitialize queue}$}
                \State ind = 0
                \For{$j \leftarrow 1 \textup{ to } klen-mlen+1$}
                    \State mmer = kmer$[j:j+mlen]$ \Comment{$ \text{generate m-mer}$}
                    \State mmer = min(mmer, reverse(mmer))
                    \State q.enqueue(mmer)
                    \If{\textup{mmer $<$ q[ind]}}
                        \State ind = $j$ \Comment{$ \text{track minimizer position}$}
                    \EndIf
                \EndFor
                
            \EndIf
        
        \State minis $\leftarrow$ minis $\cup$ q[ind] \Comment{$ \text{store identified minimizer}$}
        \EndFor
        \State \Return(minis)
	\end{algorithmic}
\end{algorithm}

\section{Experimental Setup}
\label{sec_experimental_setup}
In this section, we first discuss the dataset statistics followed by the goodness metrics used to evaluate the performance of t-SNE, classification, and clustering methods. All experiments are performed on Intel (R) Core i5 system with a 2.40 GHz processor and $32$ GB memory. Also, we use the $5$ fold cross-validation.
 
\subsection{Datasets}
\label{subsection_data_stats}

We use the following datasets in this study.

\subsubsection{Spike7k}
The dataset we use is Spike7k~\cite{ali2021k} consists of sequences of the SARS-CoV-2 virus. It has 22 unique lineages as the label with the following distribution:
B.1.1.7 ($3369$), B.1.617.2 ($875$), AY.4 ($593$), B.1.2 ($333$), B.1 ($292$), B.1.177 ($243$), P.1 ($194$), B.1.1 ($163$), B.1.429 ($107$), B.1.526 ($104$), AY.12 ($101$), B.1.160 ($92$), B.1.351 ($81$), B.1.427 ($65$), B.1.1.214 ($64$), B.1.1.519 ($56$), D.2 ($55$), B.1.221 ($52$), B.1.177.21 ($47$), B.1.258 ($46$), B.1.243  ($36$), R.1 ($32$).

\subsubsection{Host}
The NIAD Virus Pathogen Database and Analysis Resource(ViPR)~\cite{pickett2012vipr} and GISAID~\cite{gisaid_website_url} are used to retrieve Spike molecular sequences of CoVs for all hosts. 
The dataset comprises of $5558$ complete unaligned spike protein sequences with the $21$ host types with the following counts of sequences: 
Humans(1813), Rats(26), Cats(123) , Environment(1034), Pangolins(21), Bovines(88) ,Weasel(994), Hedgehog(15), Dogs(40), ,Swine(558), Dolphin(7), Python(2) ,Birds(374), Equine(5), Monkey(2), Camels(297), Fish(2), Cattle(1) ,Bats(153), Unknown(2), Turtle(1).


\subsubsection{ShortRead}

This dataset is generated by taking $10181$ SARS-CoV-2 nucleotide
sequences and simulating short reads from each sequence using
inSilicoSeq~\footnote{\url{https://github.com/HadrienG/InSilicoSeq}} with the
miseq model (all other settings were default settings). 
The lineages/class labels (count/distribution) in this dataset is as follows: B.1.1.7 (2587), B.1.617.2 (1198), AY.4 (1167), AY.43 (412), AY.25 (275), B.1.2 (253), AY.44 (249), B.1 (200), B.1.177 (184), AY.3 (166), P.1 (137), B.1.1 (128), B.1.526 (112), AY.9 (108), AY.5 (96), AY.29 (86), AY.39 (85), B.1.429 (81), B.1.160 (79), Others (2578). There are $496$ unique lineages for $10181$ sequences. 

\subsubsection{Rabies}
We used the labeled Nucleotide genome sequences for rabies virus hosts dataset from RABV-GLUE~\footnote{\url{http://rabv-glue.cvr.gla.ac.uk/\#/home}}, where the label is the name of the host for which we are classifying the sequences. The statistics of the rabies dataset are given in Table~\ref{tbl_data_summary_rabies}.


\begin{table}[ht!]
  \centering
    \resizebox{0.99
    \textwidth}{!}{
   \begin{tabular}{p{1.7cm}p{3cm}p{2.2cm}p{1.2cm}p{1cm}p{0.9cm}p{0.9cm}cp{0.9cm}} 
    \toprule
    \multirow{2}{*}{Name} & \multirow{2}{*}{Type} & \multirow{2}{*}{Source} & \multirow{2}{1.4cm}{Sequence Count} & \multirow{2}{*}{Classes} & \multicolumn{4}{c}{Sequence Length} \\
    \cmidrule{6-9}
    & & & & & Min & Max & Avg & Mode \\
    \midrule	\midrule	
    Rabies Virus Data & Nucleotide genome sequences for rabies virus hosts & RABV-GLUE~\citep{RABV_GLUE_website_url} & 20051 & 12 & 90 & 11930 & 1948.4 & 1353\\
    \bottomrule
  \end{tabular}
}
  \caption{Data Statistics.}
  \label{tbl_data_summary_rabies}
\end{table}

\subsubsection{Genome}
Using the well-known and widely used database of SARS-CoV-2, GISAID~\cite{gisaid_website_url}, we retrieve the full-length nucleotide sequences of the coronavirus. We include the COVID-19 variant information and $8220$ nucleotide sequences in our dataset. In our sample, there are 41 different versions altogether. The dataset statistics for the prepossessed genome data are provided in Table~\ref{tbl_dataset_statistics_Genome_data}. The length of the Spike2Vec-based feature vector is $5^3=125$ ($5$ is a unique character in the sequence ``ACGT-").

\begin{table}[ht!]
\centering
\resizebox{0.85\textwidth}{!}{
    \begin{tabular}{cp{1cm}|cp{1cm} | cp{1cm}|cp{1cm}}
    \toprule
        Lineages & Sequence count & Lineages & Sequence count & Lineages & Sequence count & Lineages & Sequence count \\
        \midrule \midrule
        AY.103 & 2271 & AY.121 & 40	& AY.4 & 100 & AY.114 & 15		\\
        AY.44 & 1416 & AY.75 & 37	&  AY.117 & 94 & AY.34 & 14        \\
        AY.100 & 717 & AY.3.1 & 30 & AY.113 & 94 & AY.125 & 14       \\
        AY.3 & 710 & AY.3.3 & 28 & AY.118 & 86 & AY.46.1 & 14      \\
        AY.25 & 585 & AY.107 & 27 & AY.43 & 85 & AY.92 & 13         \\
        AY.25.1 & 382 & AY.34.1 & 25 & AY.122 & 84 & AY.127 & 12   \\
        AY.39 & 248 & AY.46.6 & 21 & BA.1 & 79 & AY.46.4 & 12        \\
        AY.119 & 242 & AY.98.1 & 20	& AY.119.2 & 74 & AY.98 & 12      \\
        B.1.617.2 & 175 & AY.13 & 19 & AY.47 & 73 & AY.111 & 10    \\
        AY.20 & 130 & AY.116.1 & 18 & AY.39.1 & 70 & & \\
        AY.26 & 107 & AY.126 & 17 &  &  &  &            \\
        \midrule 
        & & & & & & \textbf{Total} &  \textbf{8220} \\
        \bottomrule
    \end{tabular}
    }
    \caption{Dataset Statistics for full length \textbf{Genome (nucleotide) data} (8000 sequences) comprised of 41 lineages (class labels).}
    \label{tbl_dataset_statistics_Genome_data}
\end{table}


\subsubsection{Breast Cancer}
The dataset on Membranolytic anticancer peptides (ACPs)~\cite{Grisoni2019} provides details regarding peptide sequences and their corresponding anticancer effectiveness against breast and lung cancer cell lines. The target labels are classified into four groups: ``very active," ``moderately active," ``experimental inactive," and ``virtual inactive." In total, the dataset comprises 949 peptide sequences for breast cancer. Table~\ref{tab_data_Stats_BC} shows the distribution.

\begin{table}[ht!]
      \centering
      \resizebox{0.65\textwidth}{!}{
         \begin{tabular}{lcccc}
    \toprule
    ACPs Category & Count & Min. & Max. & Average \\
    \midrule \midrule
        Inactive-Virtual & 750 & 8 & 30 & 16.64 \\
        Moderate Active & 98 & 10 & 38 & 18.44 \\
        Inactive-Experimental & 83 & 5 & 38 & 15.02 \\
        Very Active & 18 & 13 & 28 & 19.33 \\
        \midrule \midrule
        &  &  & Total  & 949 \\
        \bottomrule
    \end{tabular}
    }
 \caption{Dataset statistics for the Breast cancer data. Columns represent the min., max., and average lengths of the sequence.}
    \label{tab_data_Stats_BC}
\end{table}

\subsection{Clustering Algorithms Evaluation}
For clustering-based analysis of different embedding methods, we use standard $k$-means and $k$-modes algorithms. To get the optimal number of clusters, we use the elbow method~\cite{satopaa2011finding}. For this purpose, we measure the trade-off between clustering runtime and the sum of squared distances from each point to its center (also called distortion) for the different number of clusters (ranging from 2 to 14). The optimal number of clusters selected is $5$ from the elbow method.
To measure the goodness of clustering approaches, the first metric we use is silhouette coefficient~\cite{rousseeuw1987silhouettes}, which uses cohesion (within the cluster) and separation (between different clusters) combined to compute a goodness score for a given clustering. Its value is between  $-1$ to $1$ with a higher value being better.
The second metric that we are using is the Calinski--Harabasz index~\cite{calinski1974dendrite}, which measures the ratio of inter-cluster dispersion for all clusters and the sum of intra-clusters dispersion (higher value is better).
The third metric that we are using is Davies--Bouldin score~\cite{davies1979cluster}, which measures the similarity of each cluster with its most similar cluster, repeats this process for all clusters and reports the average (a lower value is better). We also use clustering computation runtime as the fourth evaluation metric.

\subsection{Classification Algorithms Evaluation}
For classification, we use Support Vector
Machine (SVM), Naive Bayes (NB), Multi-Layer Perceptron (MLP),
K-Nearest Neighbour (KNN) (with $K = 5$, which decided using standard validation set approach~\cite{validationSetApproach}), Random Forest (RF), Logistic
Regression (LR), and Decision Tree (DT) classifiers.
We use average accuracy, precision, recall, weighted, and ROC area under the curve (AUC) as evaluation metrics for measuring the goodness of classification algorithms.

\subsection{Evaluating t-SNE}
For objectively evaluation of the t-SNE model, we use a method called $k$-ary neighborhood agreement ($k$-ANA) method~\cite{zhu2021improving}.
The $k$-ANA method (for different $k$ nearest neighbors) checks the neighborhood agreement ($Q$) between HD and LD and takes the intersection on the numbers of neighbors. More formally:
\begin{equation}
\label{eqn_neignbor_agreement}
Q(k)=\sum_{i=1}^{n} \frac{1}{n k}\left \vert k N N\left(x_{i}\right) \cap k N N\left(x_{i}^{\prime}\right)\right \vert
\end{equation}

where $k$NN($x$) is set of $k$ nearest neighbours of $x$ in high-dimensional and $k$NN($x'$) is set of $k$ nearest neighbours of $x$ in  corresponding low-dimensional.

We use a quality assessment tool that quantifies the neighborhood preservation and is denoted by R($k$), which uses Equation~\ref{eqn_neignbor_agreement} to evaluate on scalar metric whether neighbors are preserved\cite{lee2015multi} in low dimensions. More formally: 

\begin{equation}
R(k)=\frac{(n-1) Q(k)-k}{n-1-k}
\end{equation}

R($k$) represents the measurement for $k$-ary neighborhood agreement. Its value lies between $0$ and $1$, the higher score represents better preservation of the neighborhood in LD space. In our experiment we computed $R(k)$ for $k \in {1,2,3,...,99}$ then considered the area under the curve (AUC) formed by   $k$ and R($k$).
Finally, to aggregate the performance for different k-ANN, we calculate the area under the $R(k)$ curve in the log plot ($AUC_{RNX}$)~\cite{lee2013type}. More formally:

\begin{equation}\label{eq_tsne_eval}
A U C_{RNX}=\frac{\Sigma_{k} \frac{R(k)}{k}}{\Sigma_{k} \frac{1}{k}}
\end{equation}

where $A U C_{R N X}$ denotes the average quality weight for $k$ nearest neighbors. 

\clearpage

\section{Subjective and Objective Evaluation of t-SNE}
\label{sec_evaluation}


The subjective evaluation for different embedding methods is shown below.

\subsection{OHE~\cite{kuzmin2020machine}}
\begin{figure}[h!]
  \centering
  \begin{subfigure}{0.25\textwidth}
  \centering
        \includegraphics[scale=0.085]{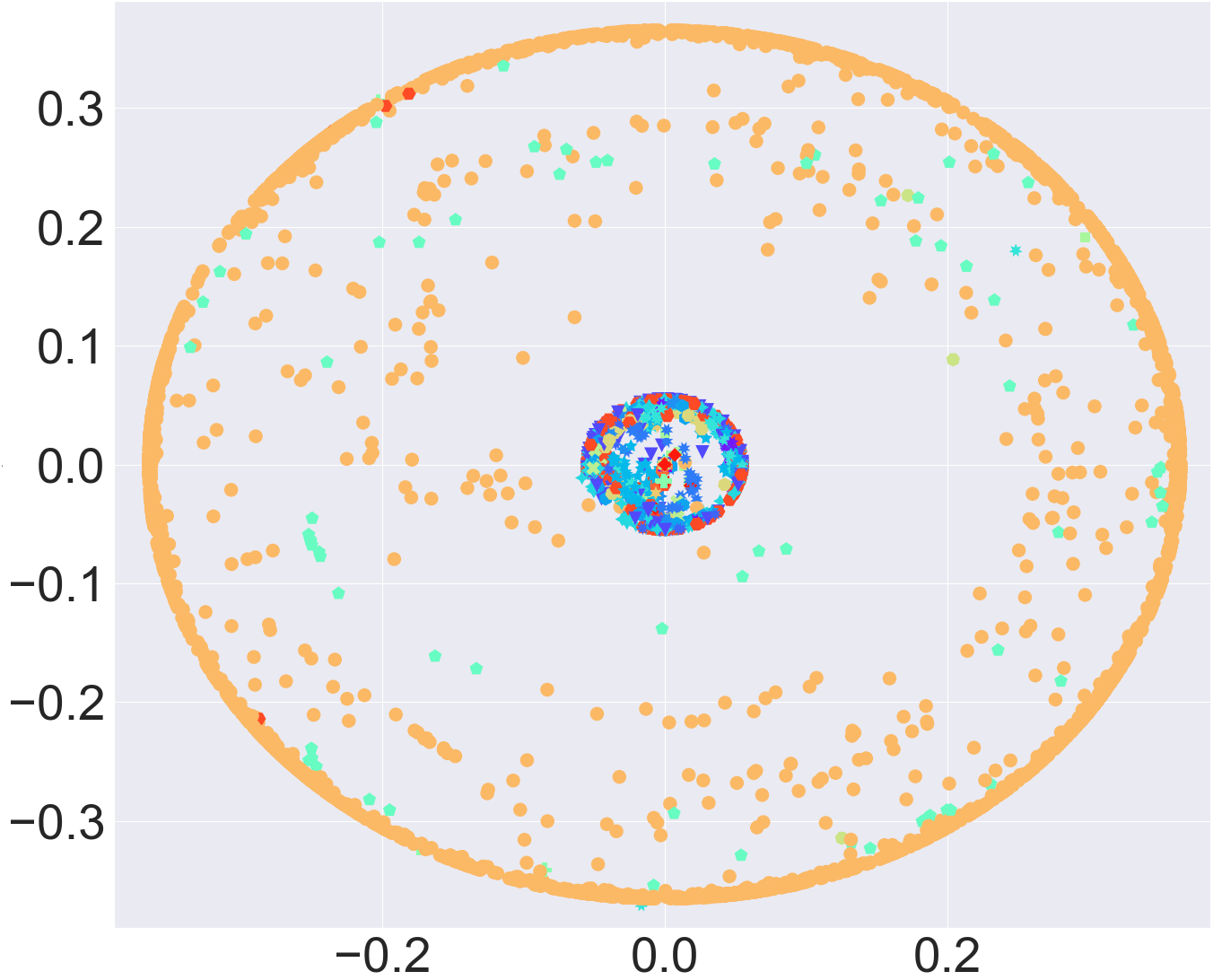}
        \caption{Additive-Chi2}
    \end{subfigure}%
    \begin{subfigure}{0.25\textwidth}
        \centering
        \includegraphics[scale=0.085]{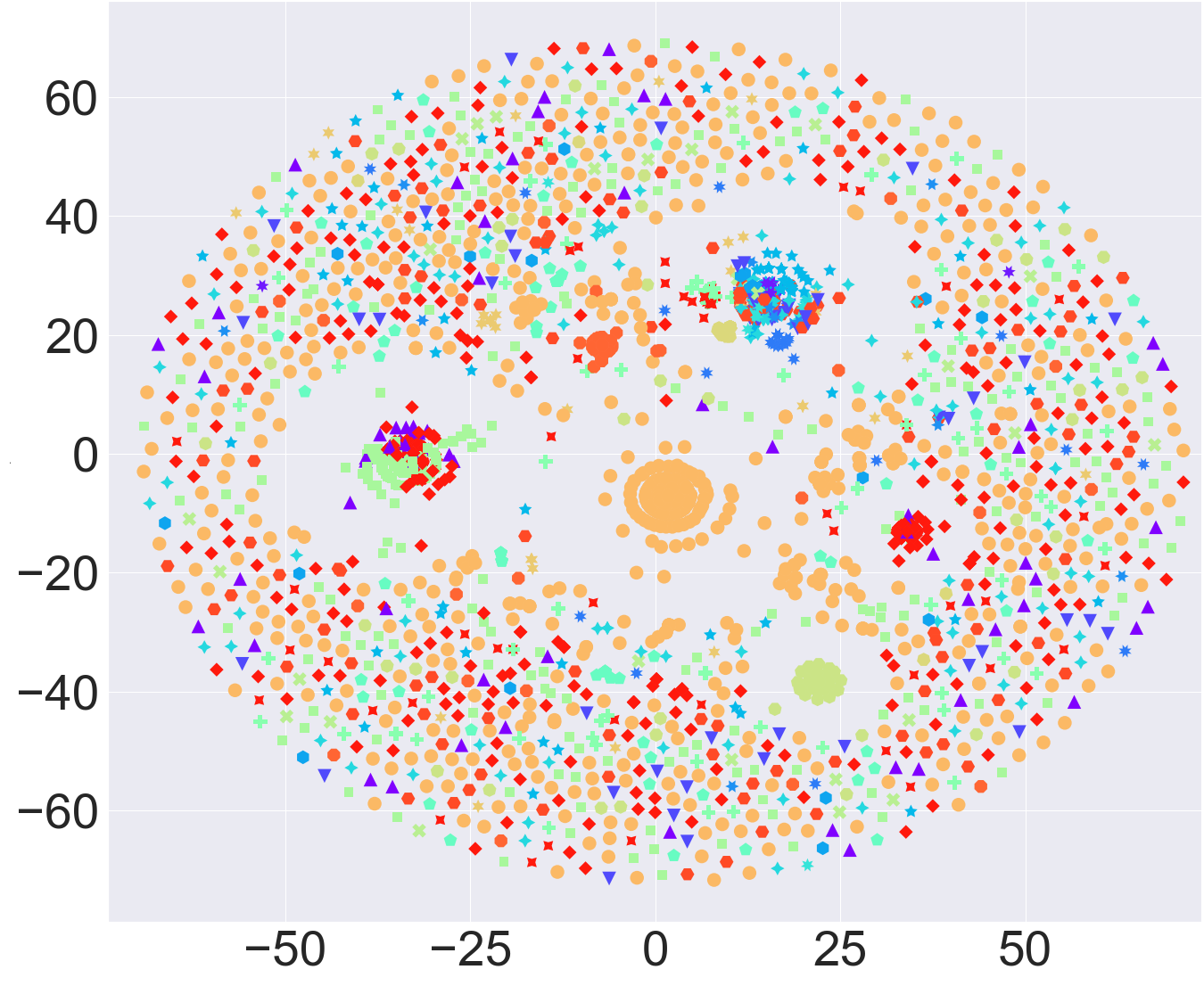}
        \caption{Chi-Squared}
    \end{subfigure}%
    \begin{subfigure}{0.25\textwidth}
        \centering
        \includegraphics[scale=0.085]{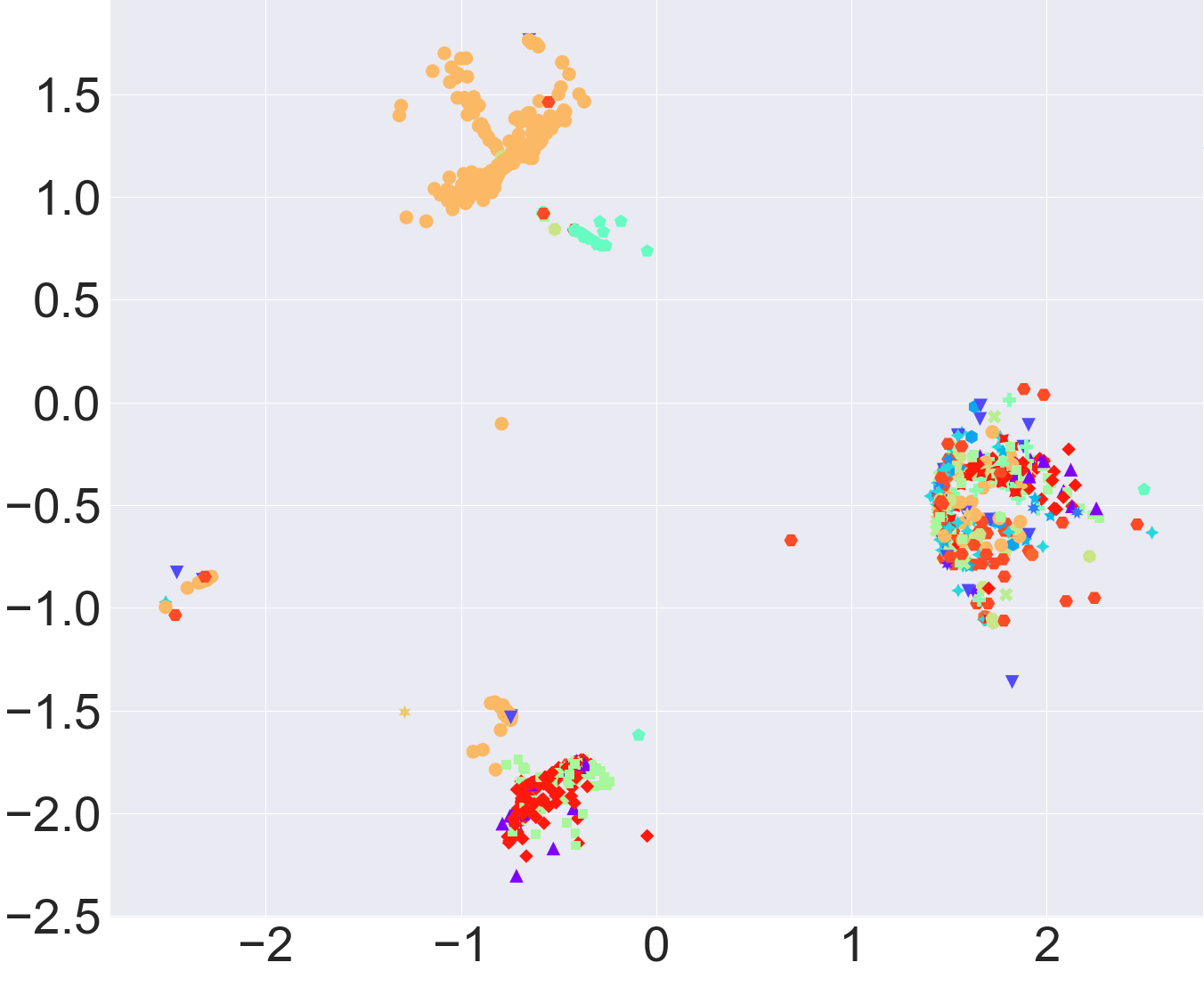}
        \caption{Cosine Similarity}
    \end{subfigure}%
    \begin{subfigure}{0.25\textwidth}
        \centering
        \includegraphics[scale=0.085]{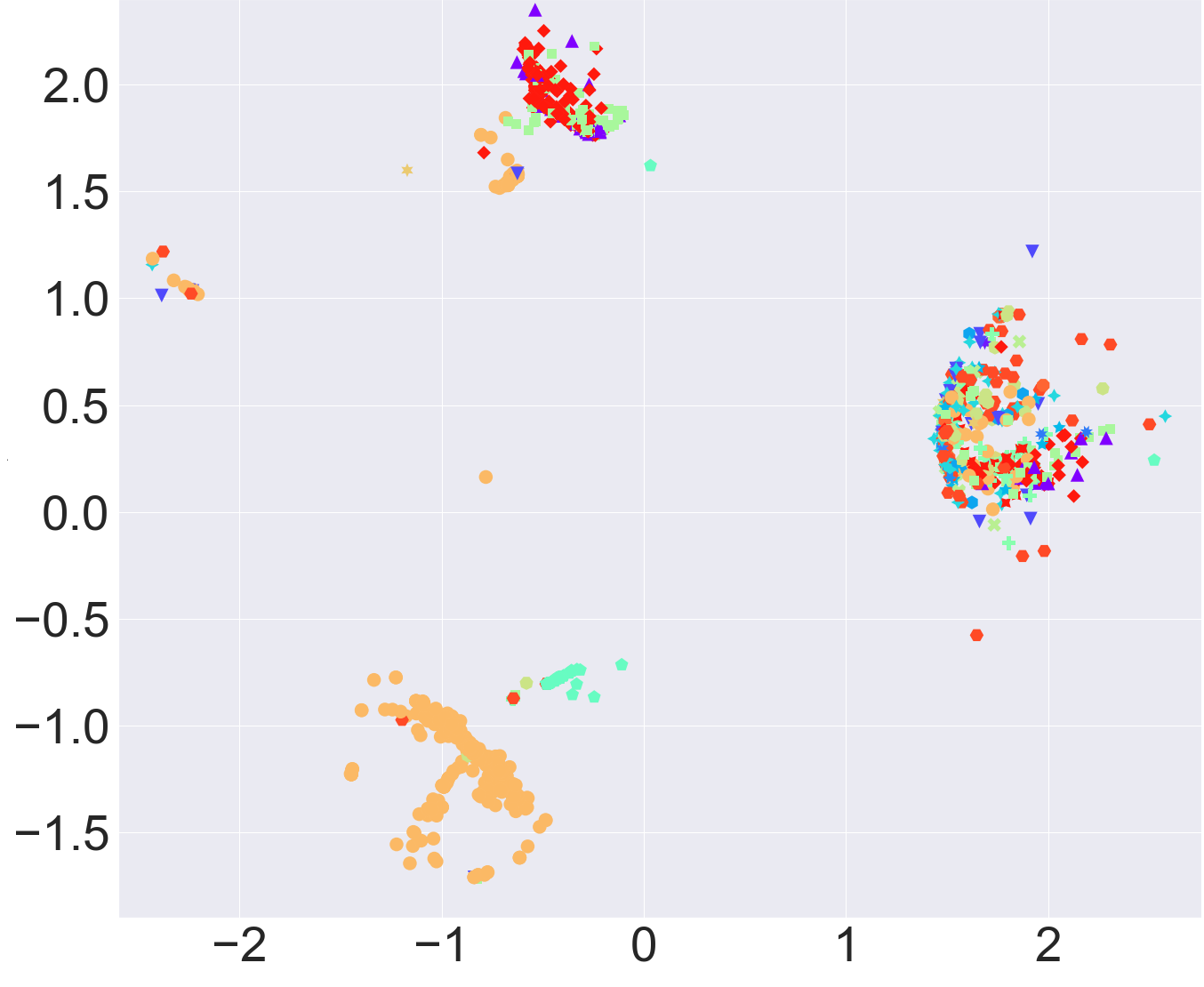}
        \caption{Linear}
    \end{subfigure}%
    \\
    \begin{subfigure}{0.2\textwidth}
        \centering
        \includegraphics[scale=0.07]{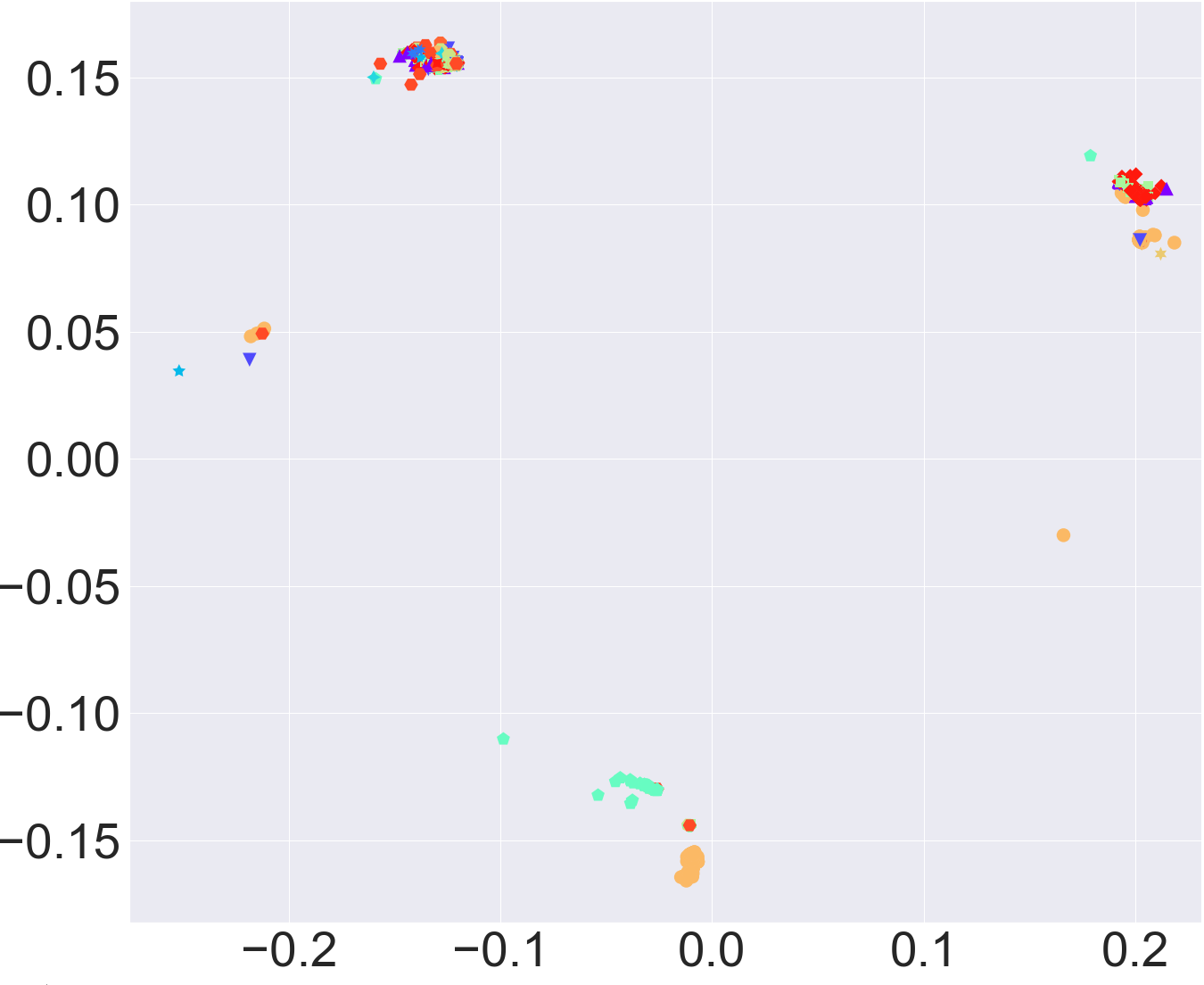}
        \caption{Polynomial}
    \end{subfigure}%
    \begin{subfigure}{0.2\textwidth}
        \centering
        \includegraphics[scale=0.07]{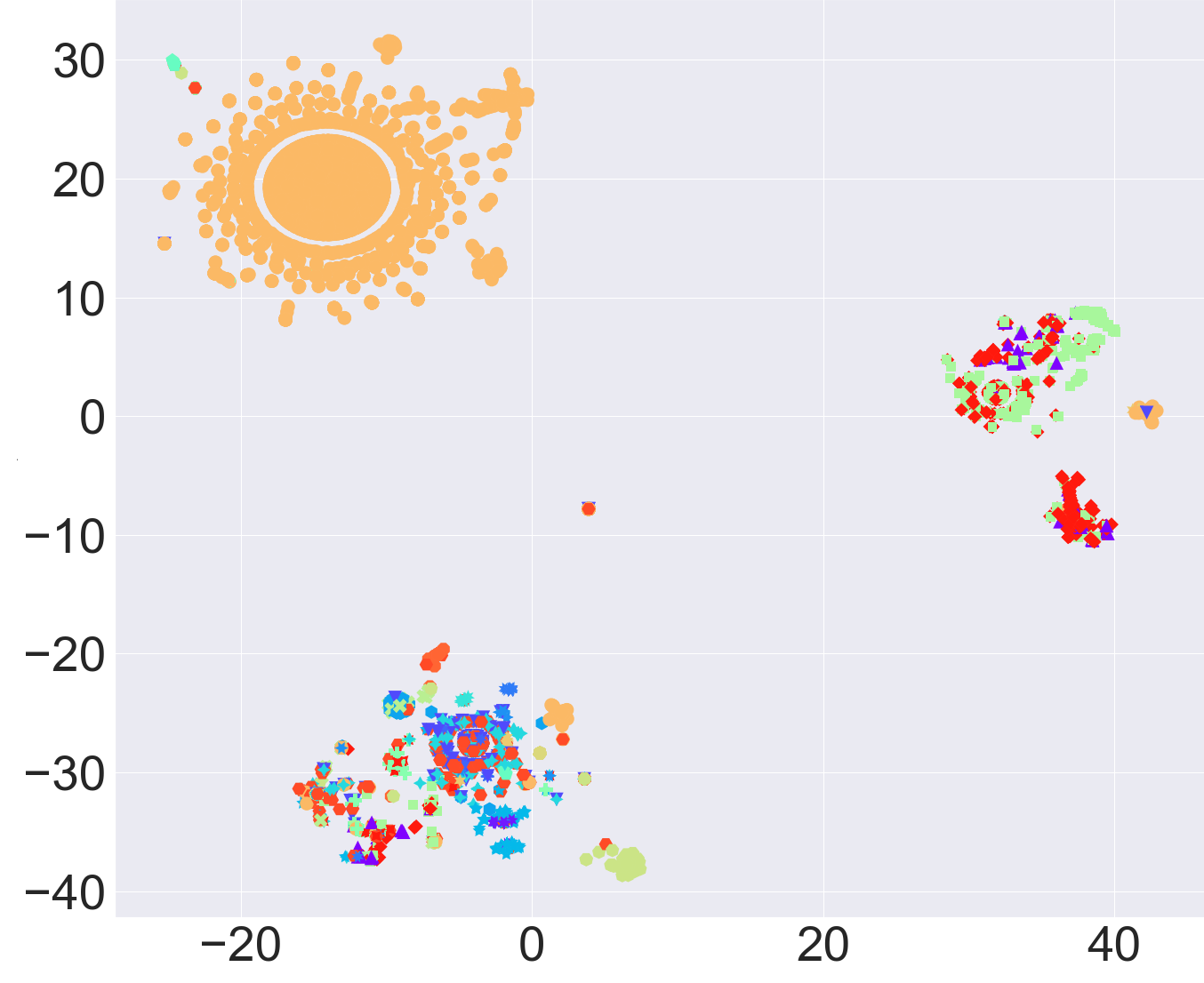}
        \caption{Gaussian}
    \end{subfigure}%
    \begin{subfigure}{0.2\textwidth}
        \centering
        \includegraphics[scale=0.07]{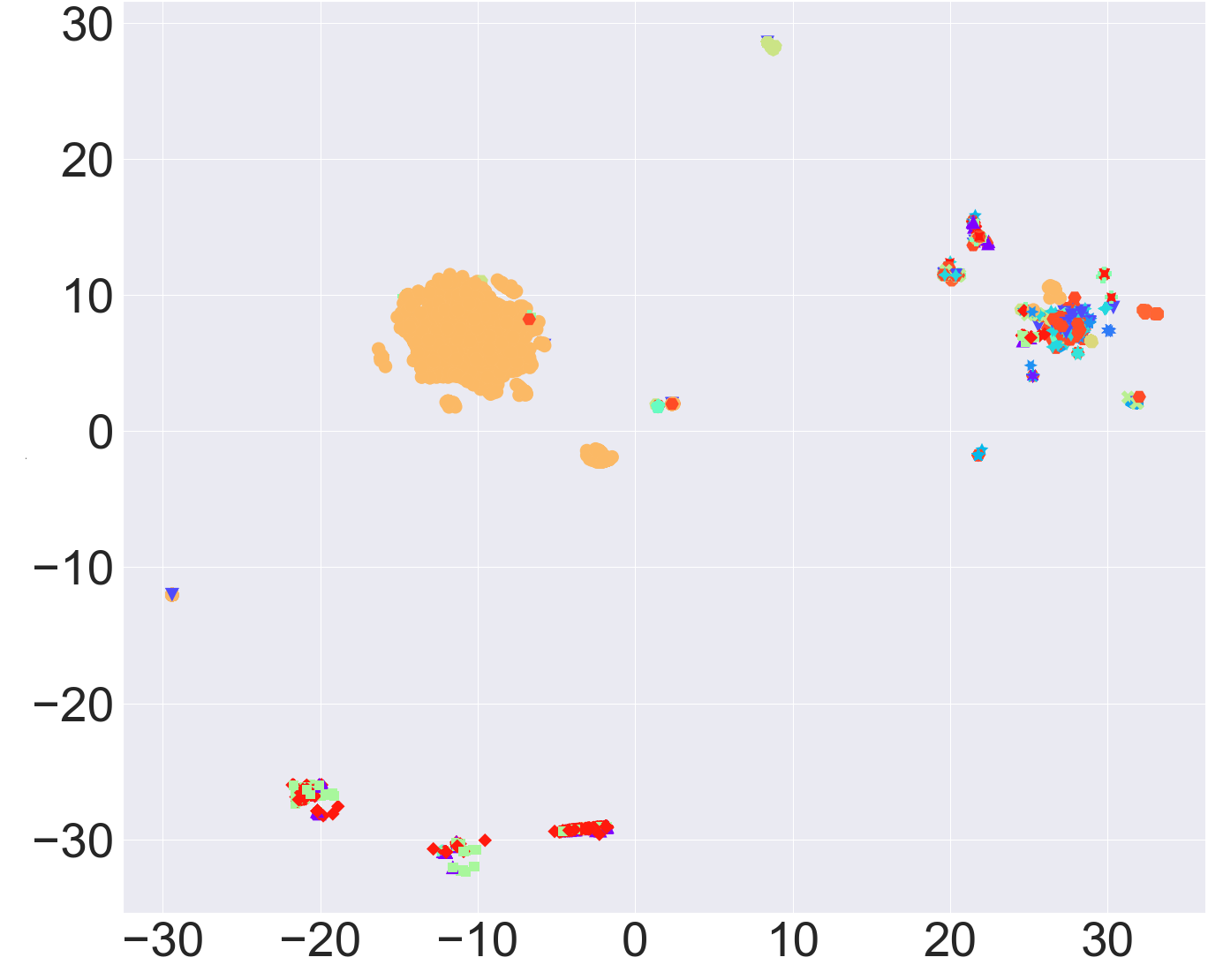}
        \caption{Isolation}
    \end{subfigure}%
    \begin{subfigure}{0.2\textwidth}
        \centering
        \includegraphics[scale=0.07]{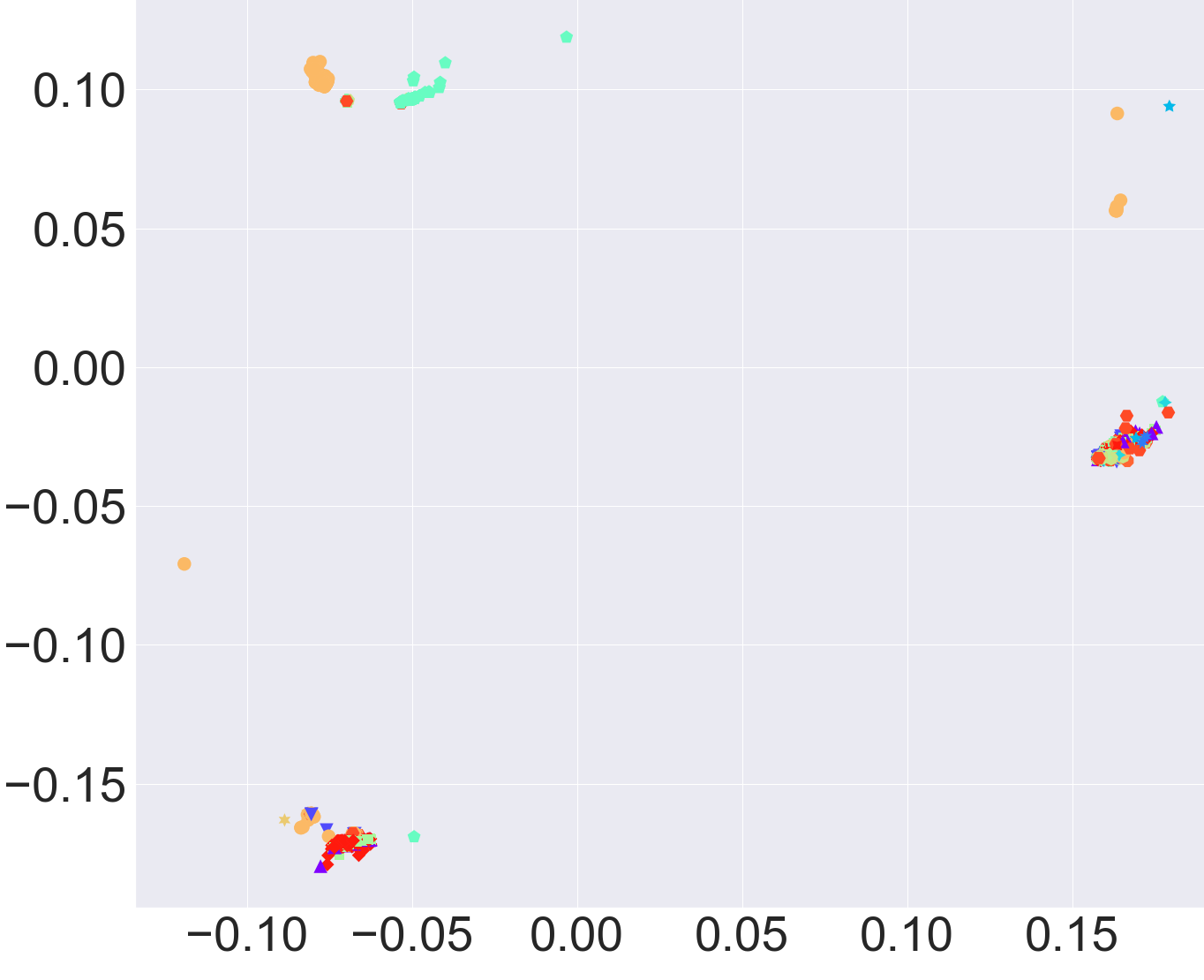}
        \caption{Laplacian}
    \end{subfigure}%
    \begin{subfigure}{0.2\textwidth}
        \centering
        \includegraphics[scale=0.07]{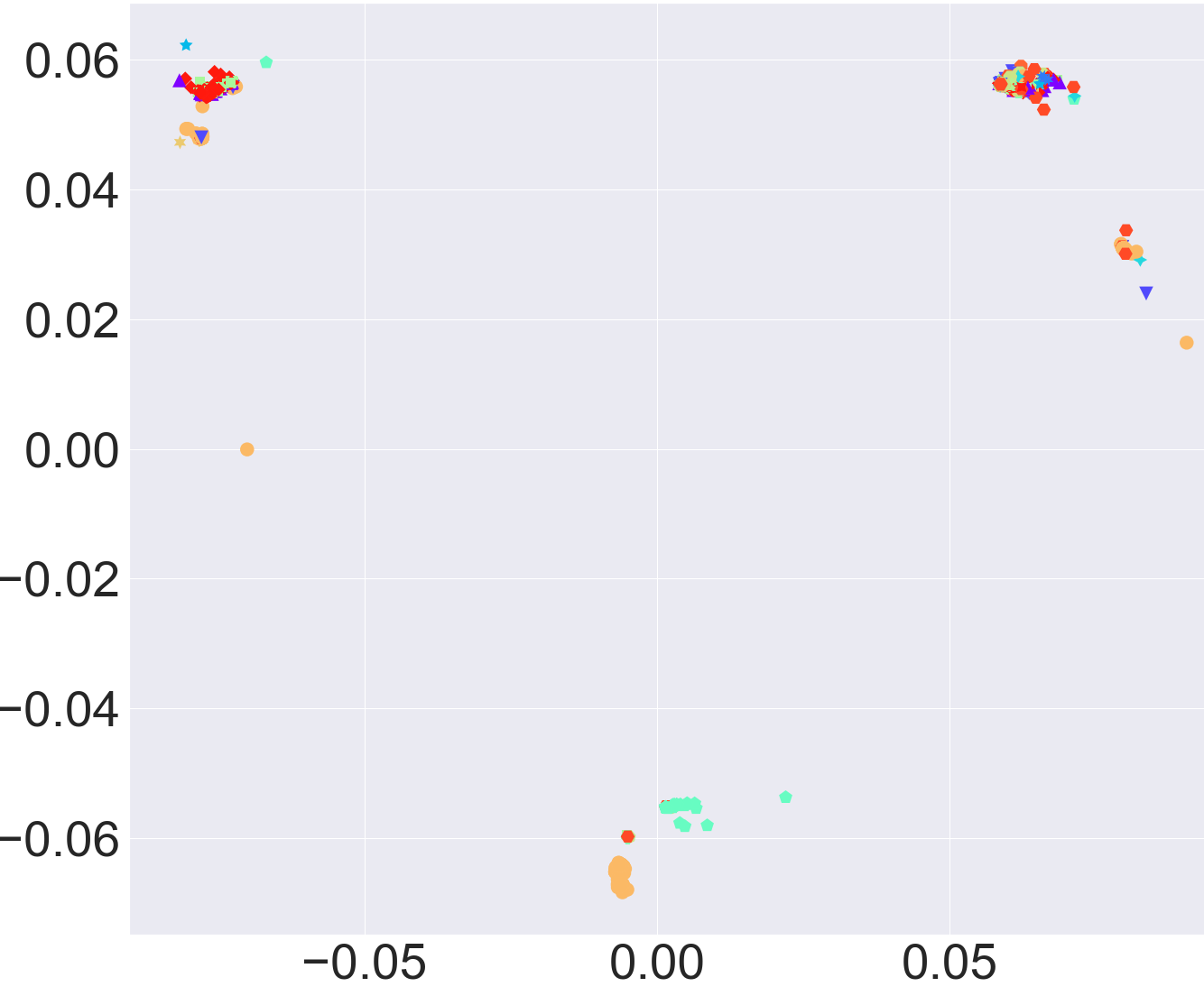}
        \caption{Sigmoid}
    \end{subfigure}%
    \\
    \begin{subfigure}{1\textwidth}
        \centering
        \includegraphics[scale=0.4]{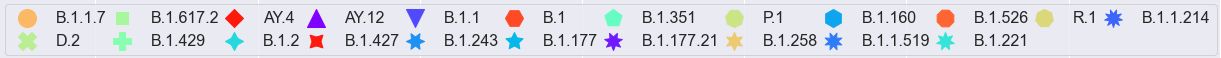}
    \end{subfigure}%
    \caption{t-SNE plots for different kernel methods using One-Hot Embedding. This figure is best seen in color.}
 \label{fig_tsne_ohe_plots}
\end{figure}

The t-SNE plots for one-hot embedding are given in Figure~\ref{fig_tsne_ohe_plots} for different kernel methods. For the Alpha variant (B.1.1.7), we can see that Gaussian 
, Isolation, Linear, and Cosine kernel are able to generate a clear grouping. However, for the other variants having small representation in the dataset (e.g. B.1.617.2 and AY.4), we can observe that the Cosine and linear kernels are better than the Gaussian 
and Isolation kernel. This could mean that the Gaussian 
and Isolation kernels tends to show biased behavior towards the more representative class (alpha variant).

\clearpage

\subsection{Spike2Vec~\cite{ali2021spike2vec}}
\begin{figure}[h!]
\begin{subfigure}{0.25\textwidth}
        \centering
        \includegraphics[scale=0.085]{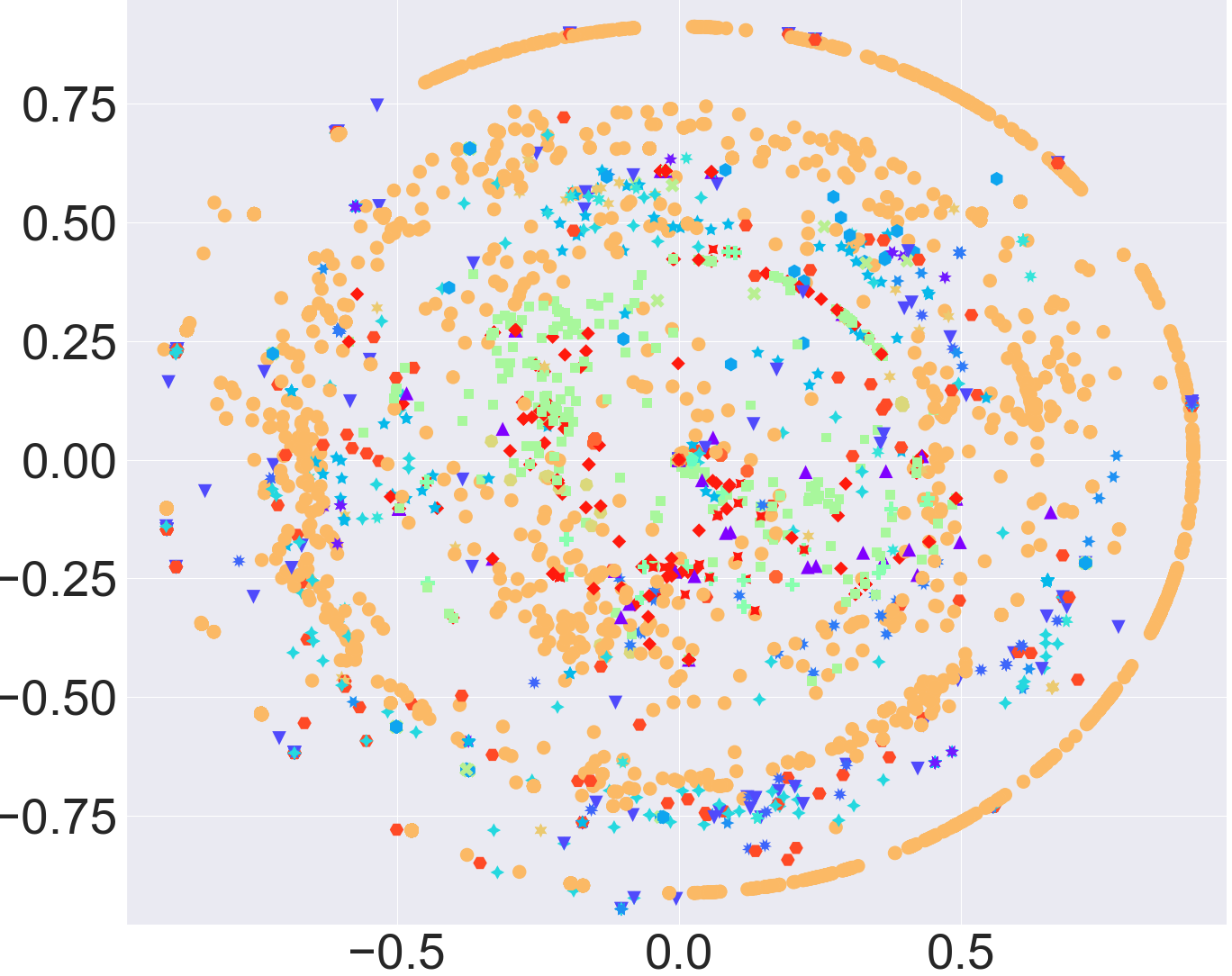}
    \caption{Additive-Chi2}
    \end{subfigure}%
    \begin{subfigure}{0.25\textwidth}
        \centering
        \includegraphics[scale=0.085]{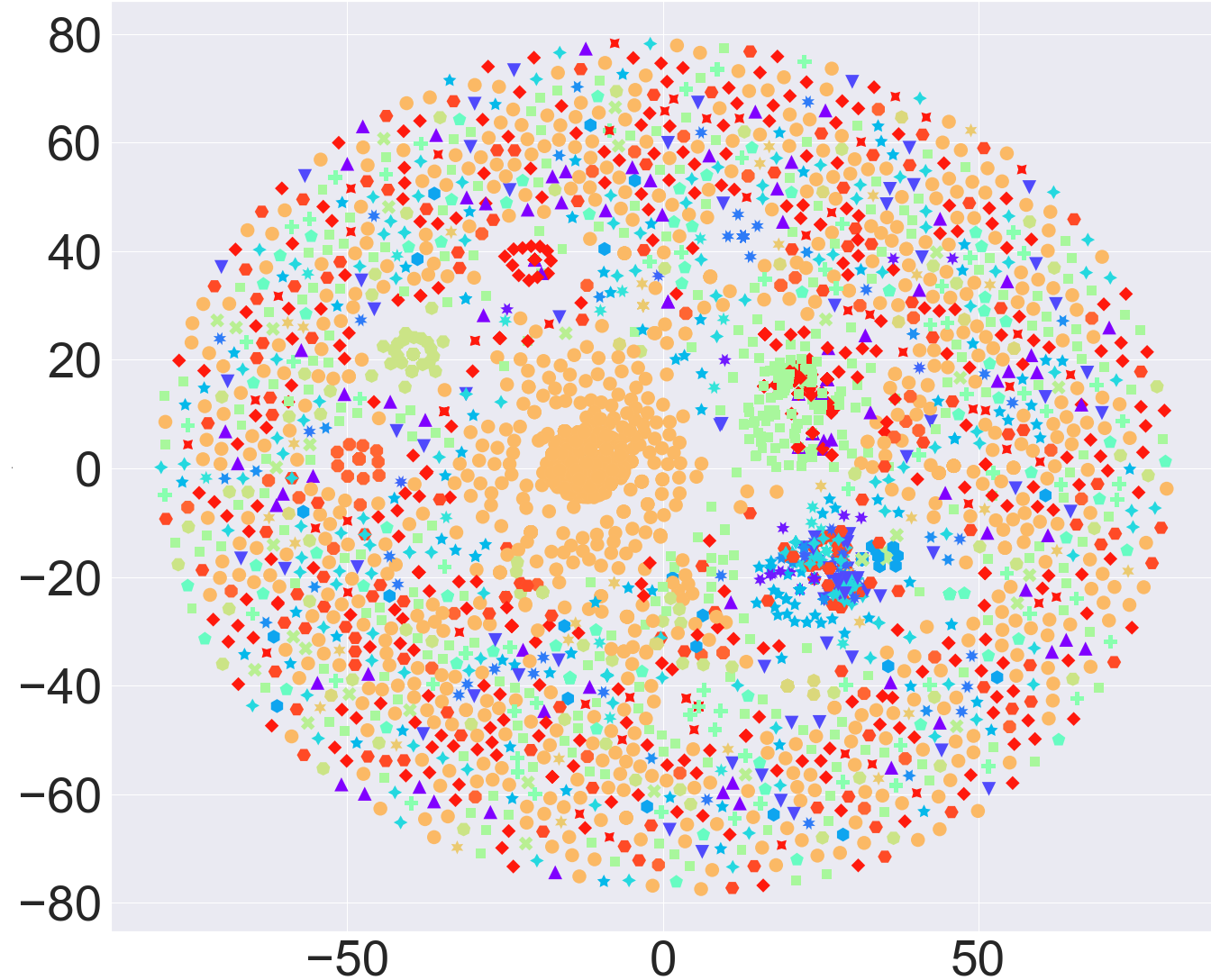}
    \caption{Chi-Squared}
    \end{subfigure}%
    \begin{subfigure}{0.25\textwidth}
        \centering
        \includegraphics[scale=0.085]{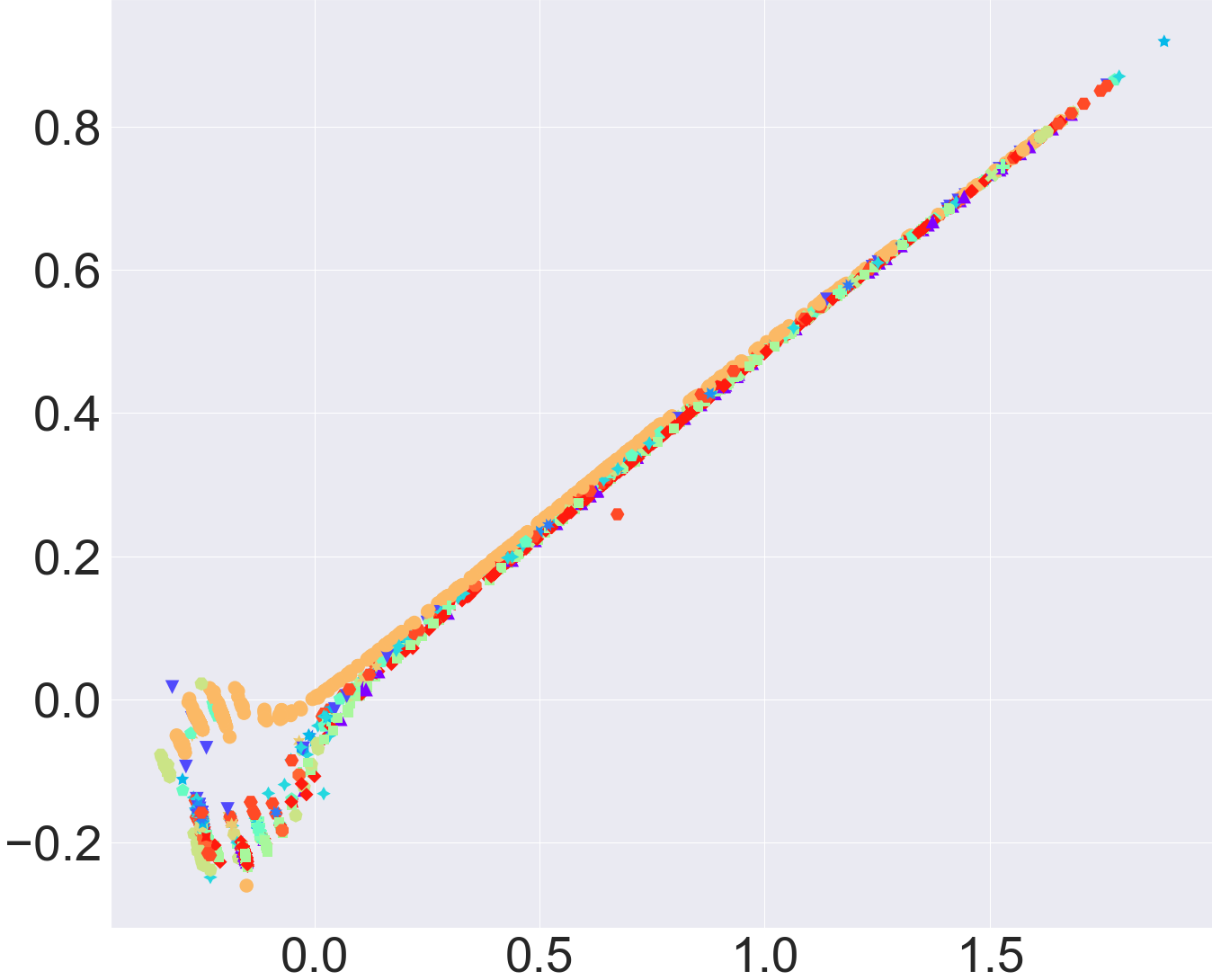}
    \caption{Cosine Similarity}
    \end{subfigure}%
    \begin{subfigure}{0.25\textwidth}
        \centering
        \includegraphics[scale=0.085]{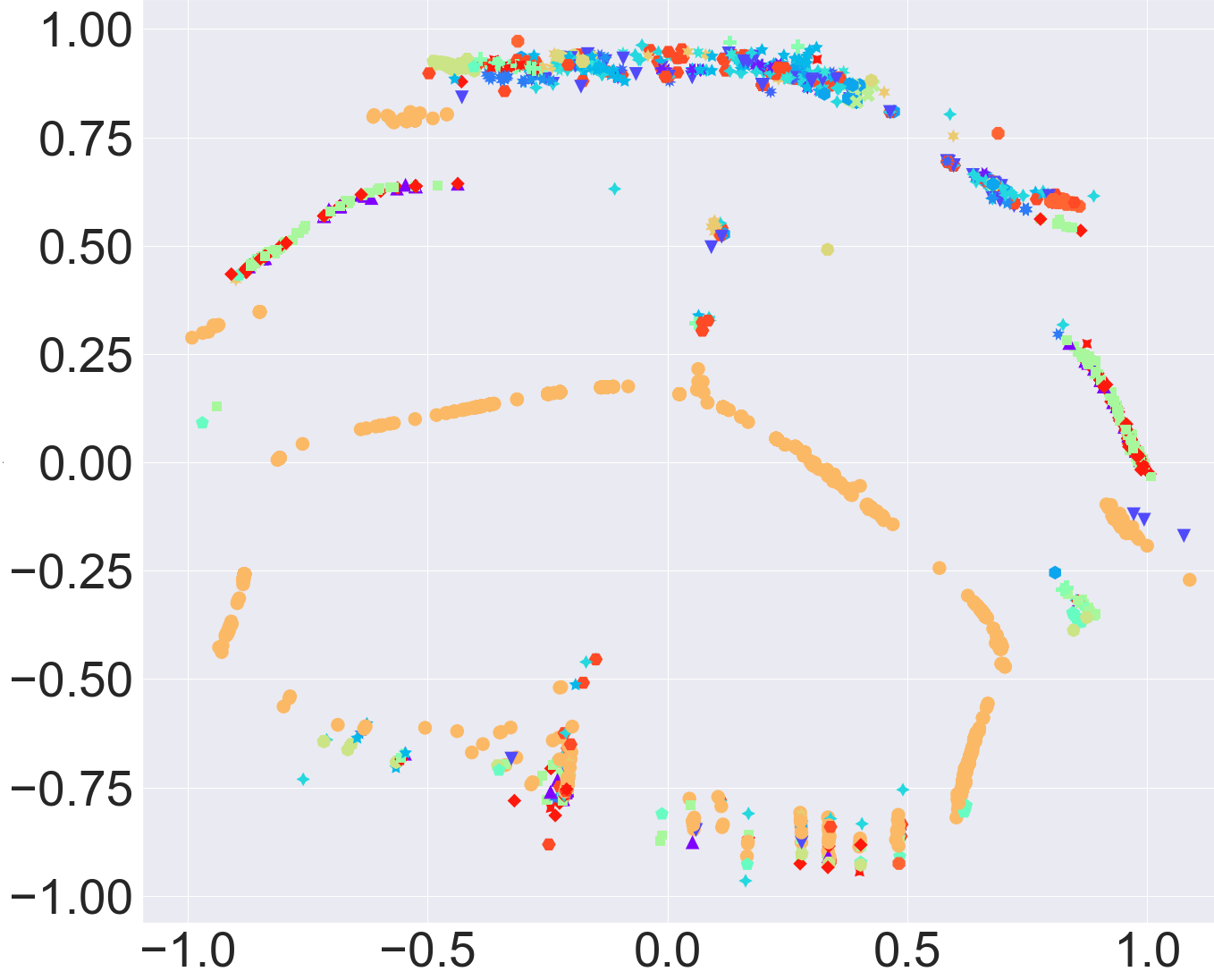}
    \caption{Linear}
    \end{subfigure}%
    \\
    \begin{subfigure}{0.2\textwidth}
        \centering
        \includegraphics[scale=0.07]{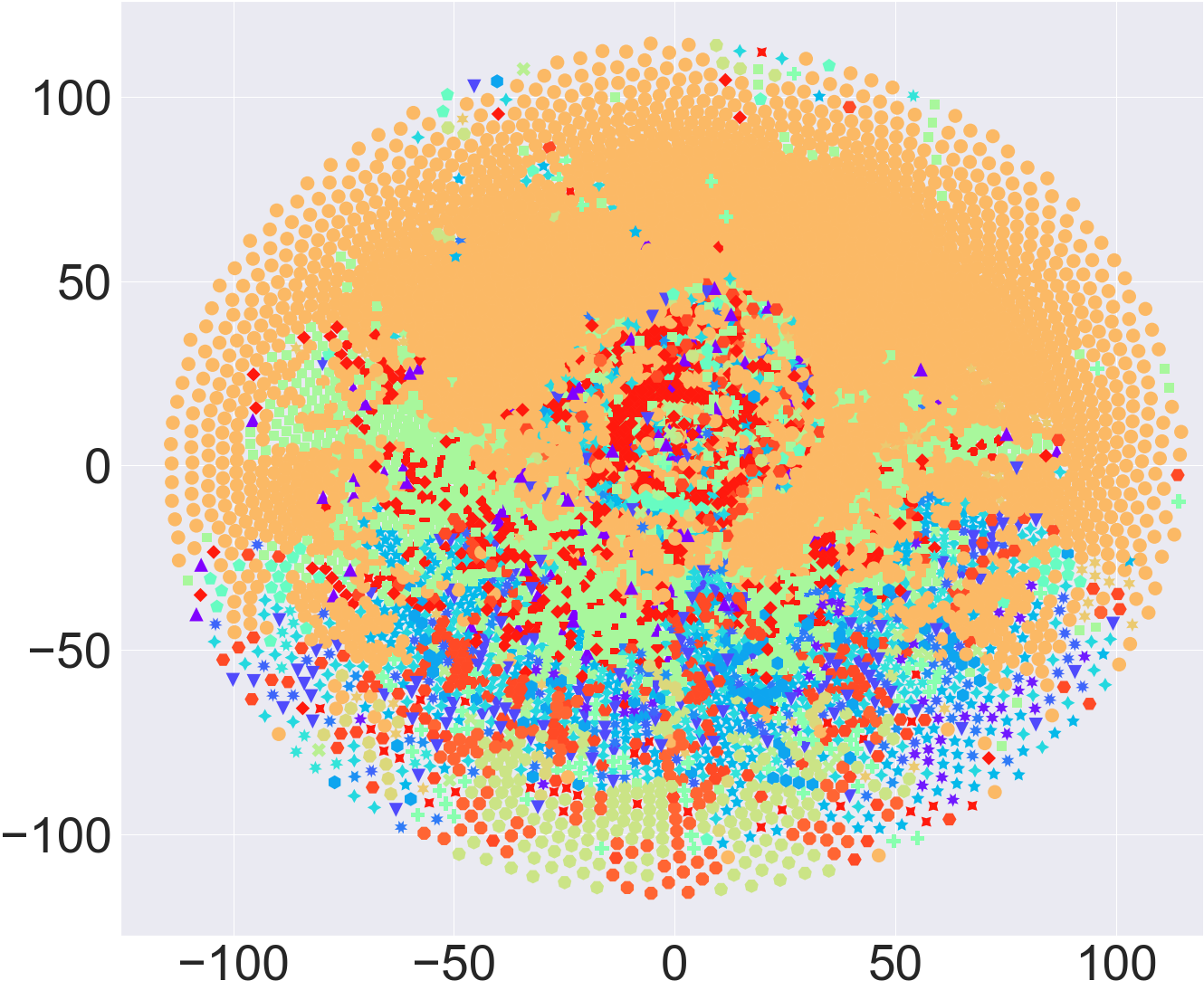}
    \caption{Polynomial}
    \end{subfigure}%
    \begin{subfigure}{0.2\textwidth}
        \centering
        \includegraphics[scale=0.07]{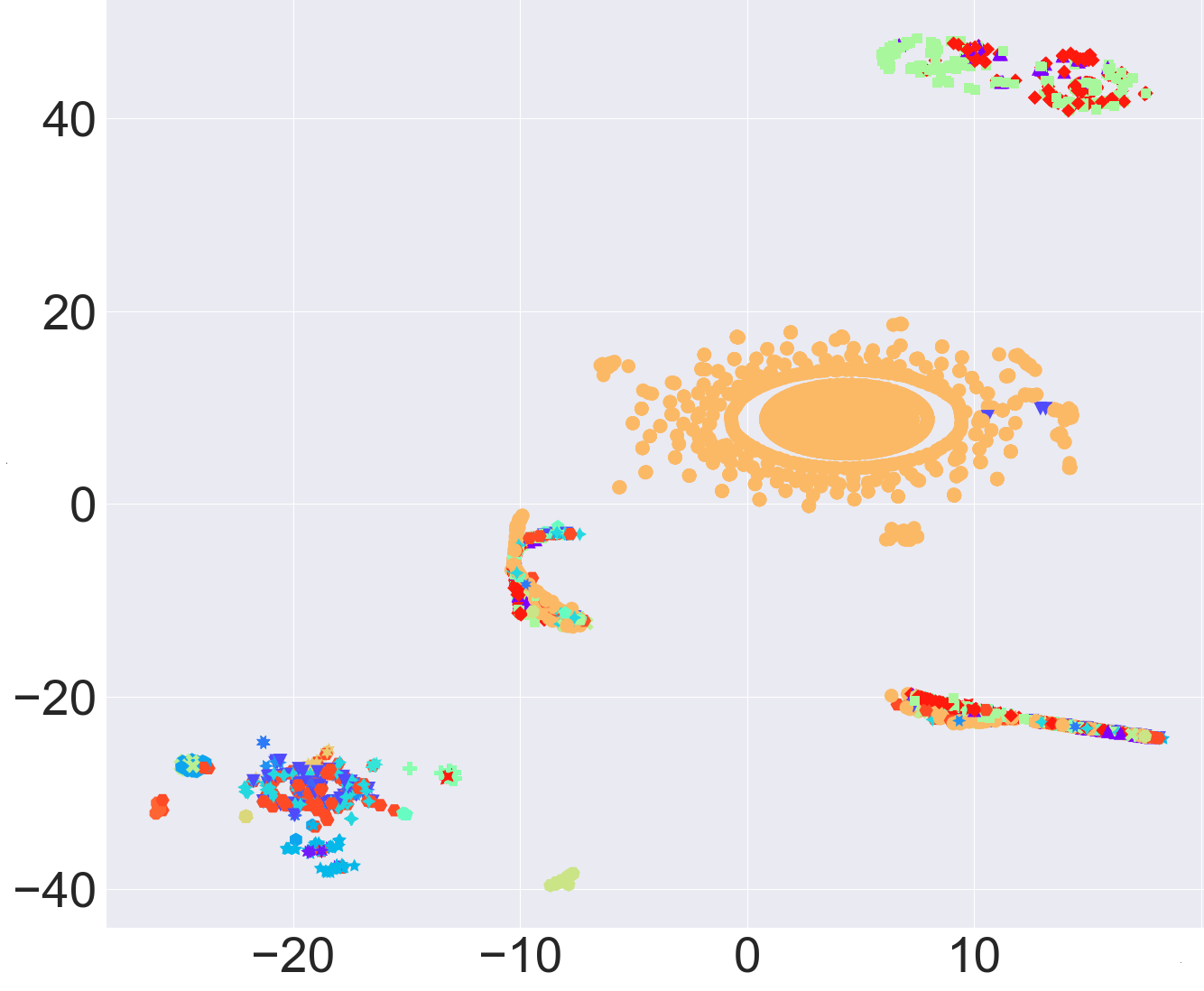}
    \caption{Gaussian}
    \end{subfigure}%
    \begin{subfigure}{0.2\textwidth}
        \centering
        \includegraphics[scale=0.07]{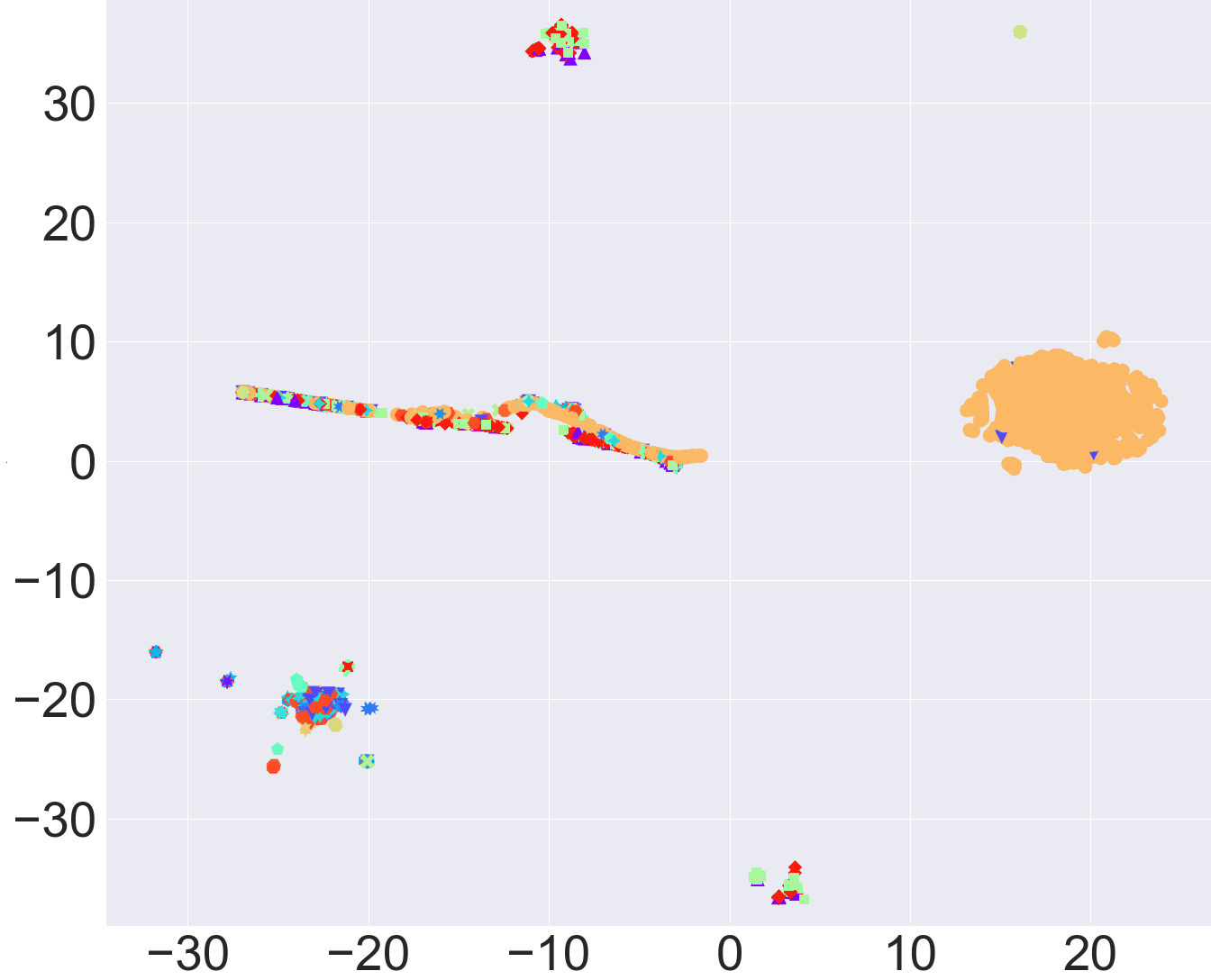}
    \caption{Isolation}
    \end{subfigure}%
    \begin{subfigure}{0.2\textwidth}
        \centering
        \includegraphics[scale=0.07]{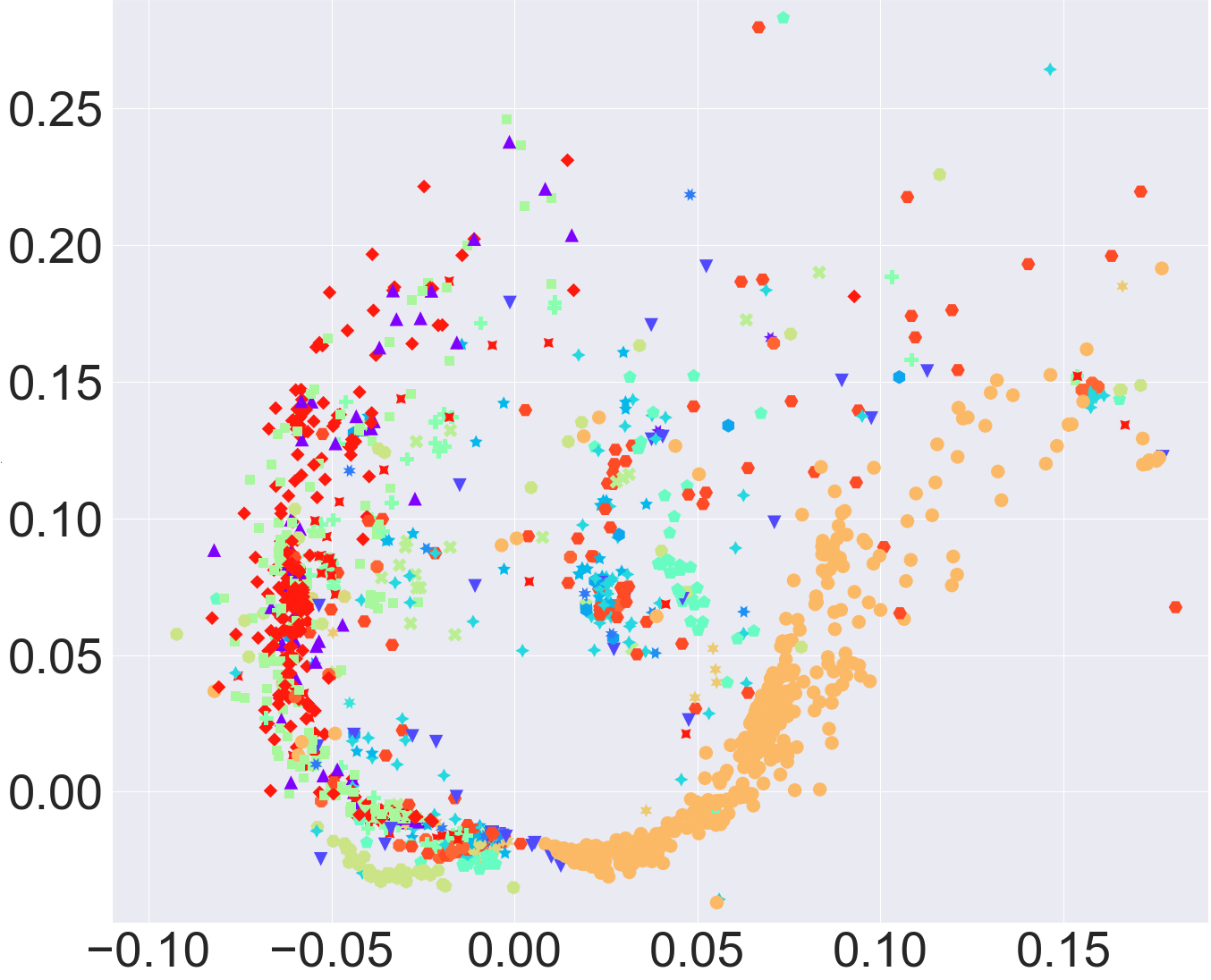}
    \caption{Laplacian}
    \end{subfigure}%
    \begin{subfigure}{0.2\textwidth}
        \centering
        \includegraphics[scale=0.07]{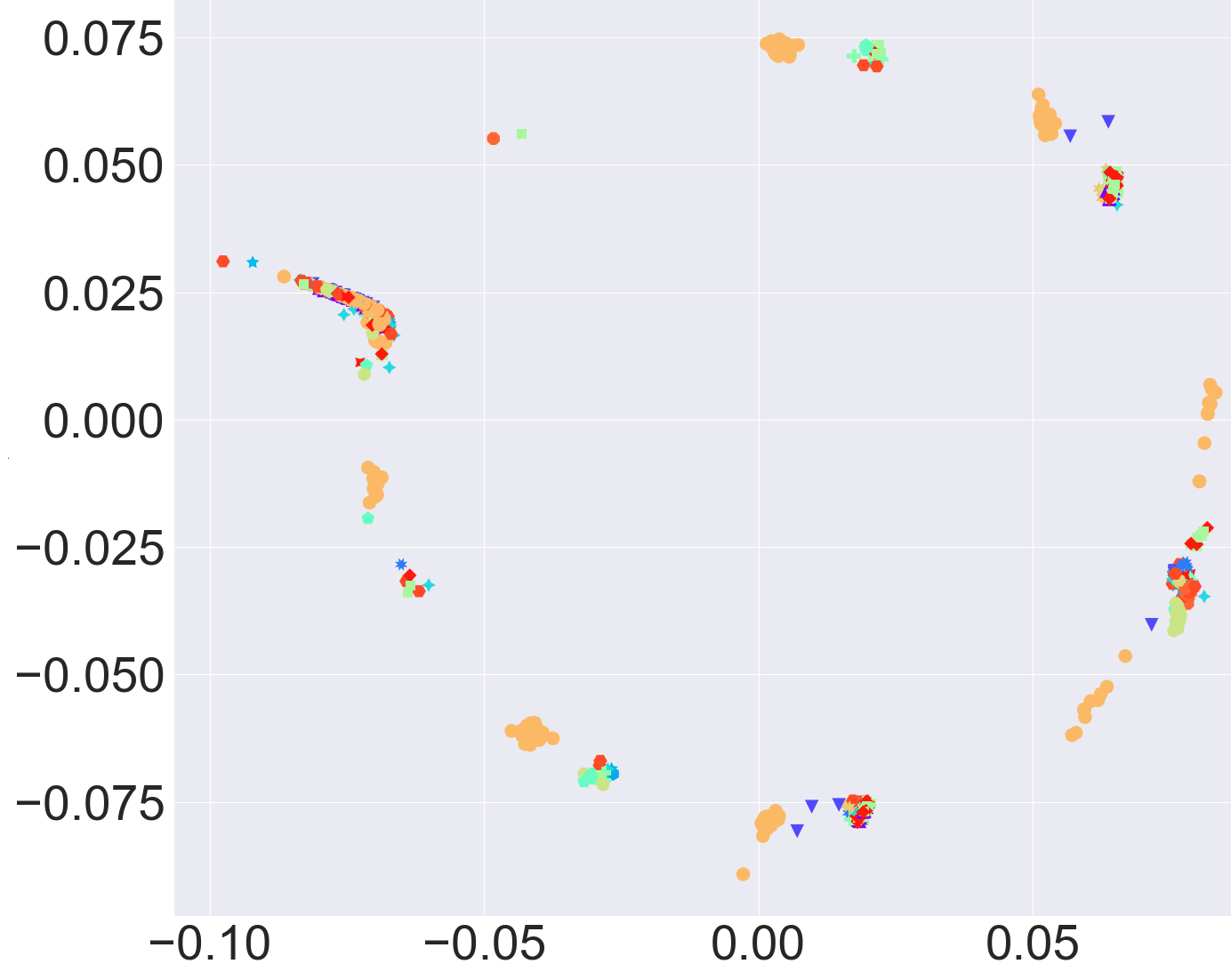}
    \caption{Sigmoid}
    \end{subfigure}%
    \\
    \begin{subfigure}{1\textwidth}
        \centering
       \includegraphics[scale=0.4]{Figures/legends_7000.png}
    \end{subfigure}%
    \caption{t-SNE plots for different kernel methods using Spike2Vec-based Embedding. This figure is best seen in color.}
    \label{fig_tsne_kmers_plots}
\end{figure}

The t-SNE plots for Spike2vec-based embeddings are given in Figure~\ref{fig_tsne_kmers_plots} using different kernel methods. It is similar behavior to OHE, where the Gaussian 
and isolation kernels group the alpha (B.1.1.7) variant almost perfectly, however, all other variants are scattered around in the plot.

\clearpage

\subsection{Minimizer}
\begin{figure}[h!]
\begin{subfigure}{0.25\textwidth}
        \centering
        \includegraphics[scale=0.085]{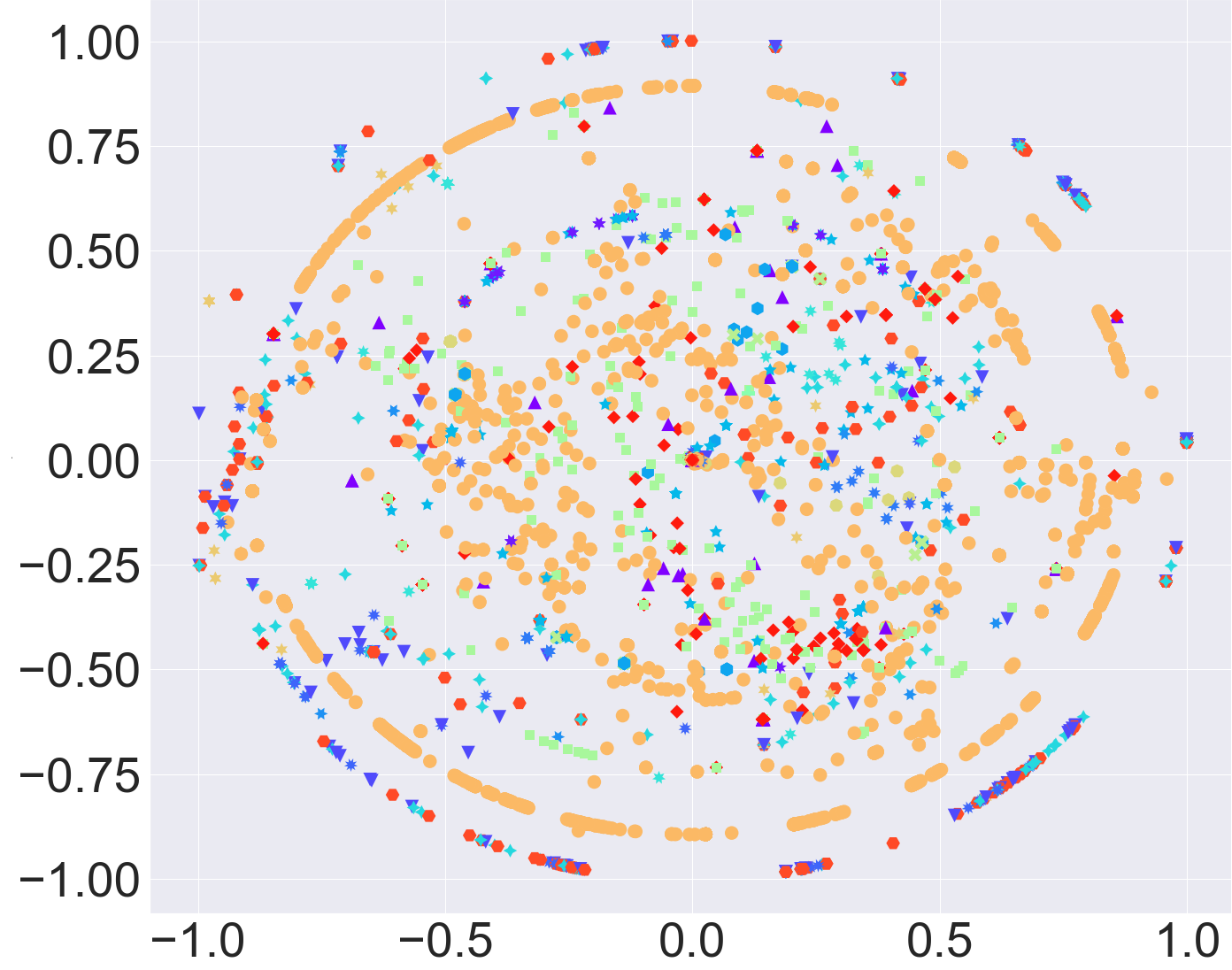}
    \caption{Additive-Chi2}
    \end{subfigure}%
    \begin{subfigure}{0.25\textwidth}
        \centering
        \includegraphics[scale=0.085]{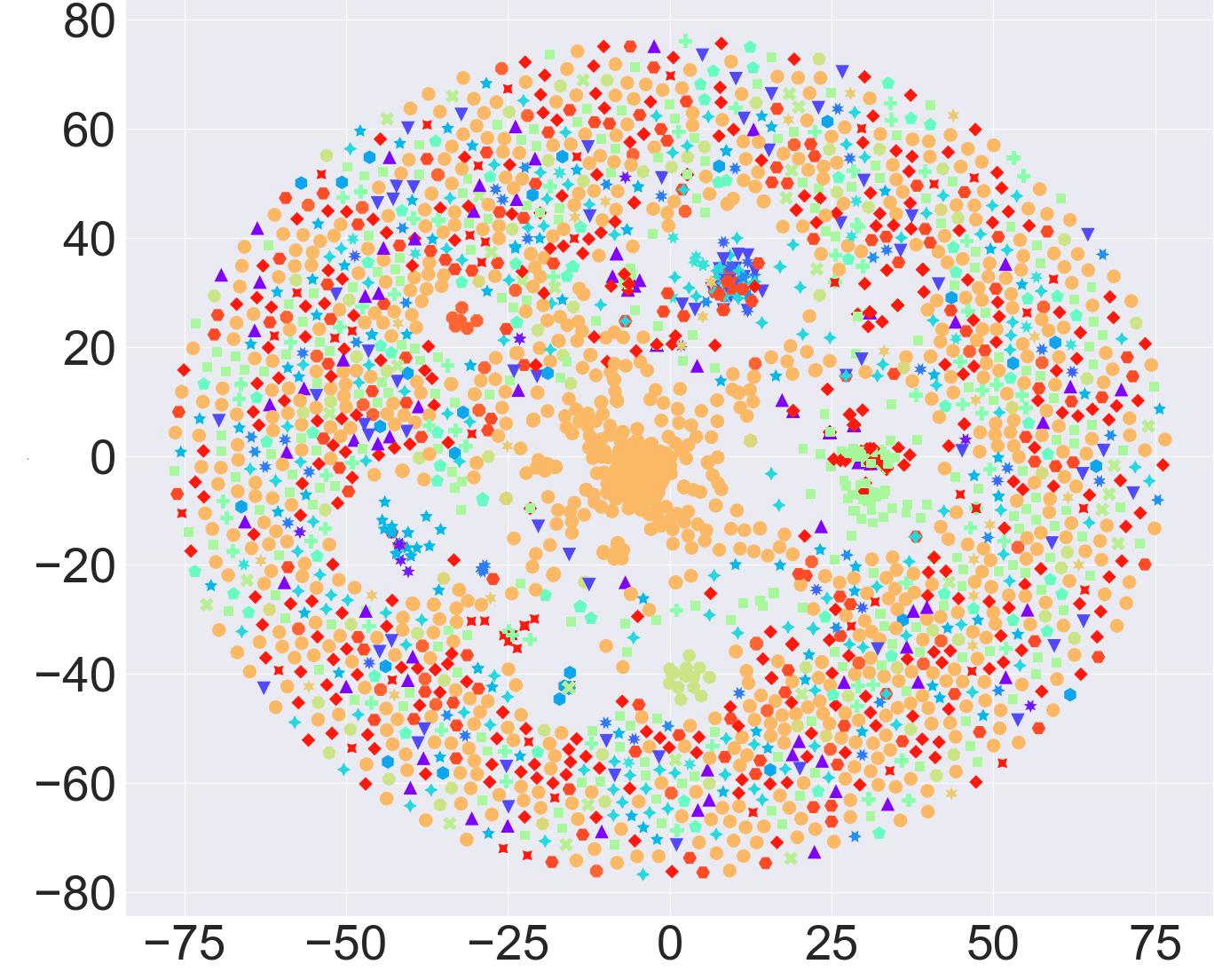}
    \caption{Chi-Squared}
    \end{subfigure}%
    \begin{subfigure}{0.25\textwidth}
        \centering
        \includegraphics[scale=0.085]{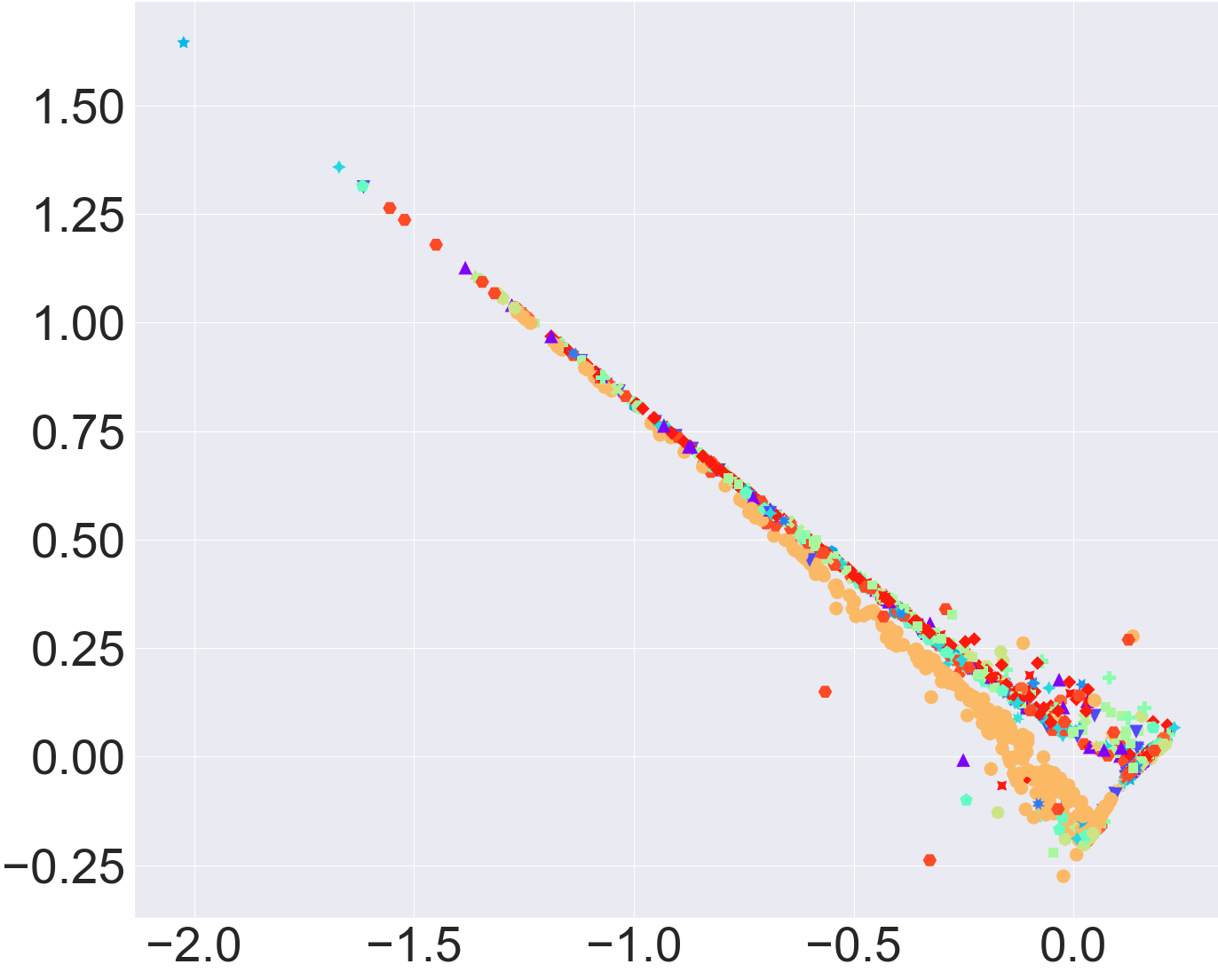}
    \caption{Cosine Similarity}
    \end{subfigure}%
    \begin{subfigure}{0.25\textwidth}
        \centering
        \includegraphics[scale=0.085]{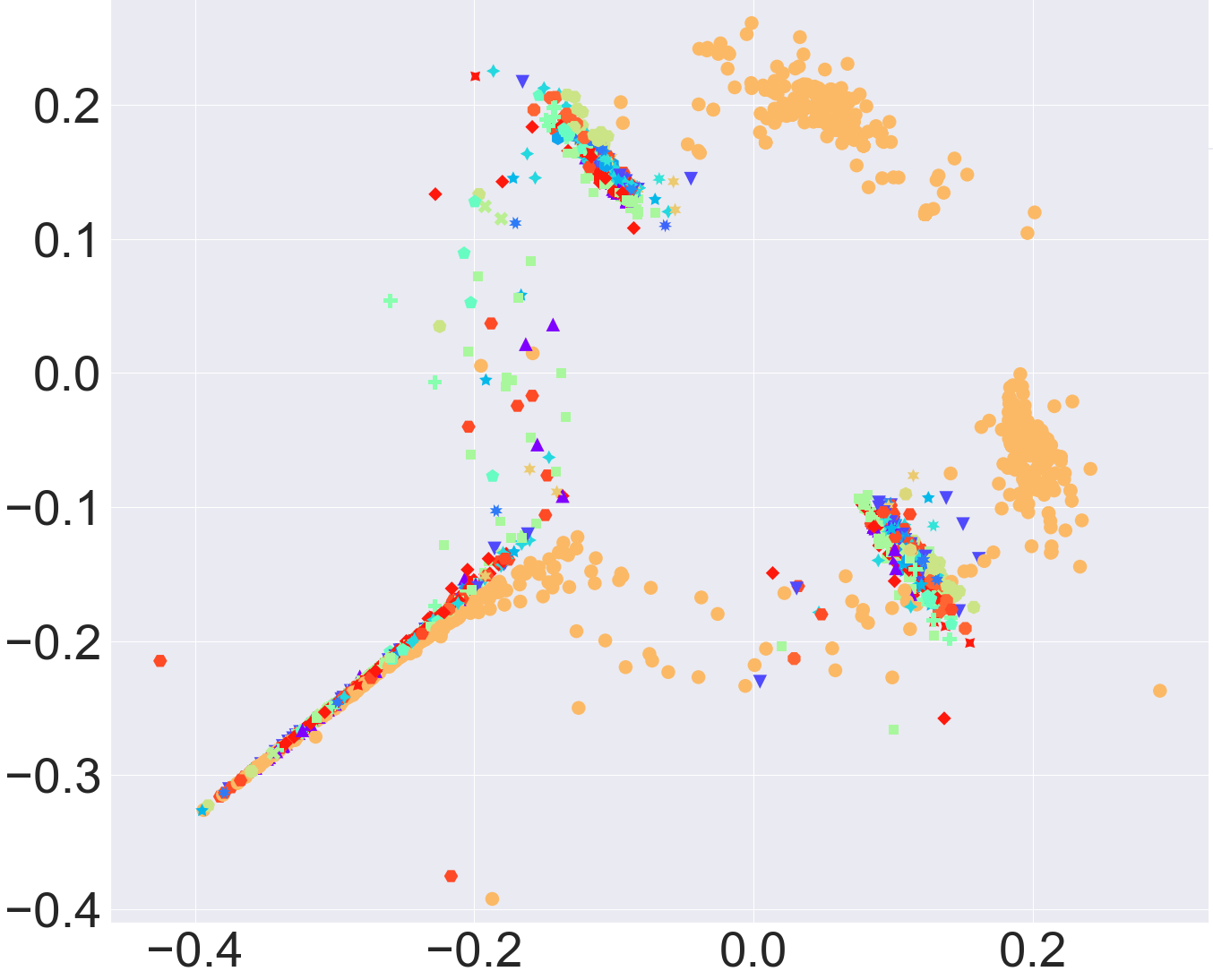}
    \caption{Linear}
    \end{subfigure}%
    \\
    \begin{subfigure}{0.2\textwidth}
        \centering
        \includegraphics[scale=0.07]{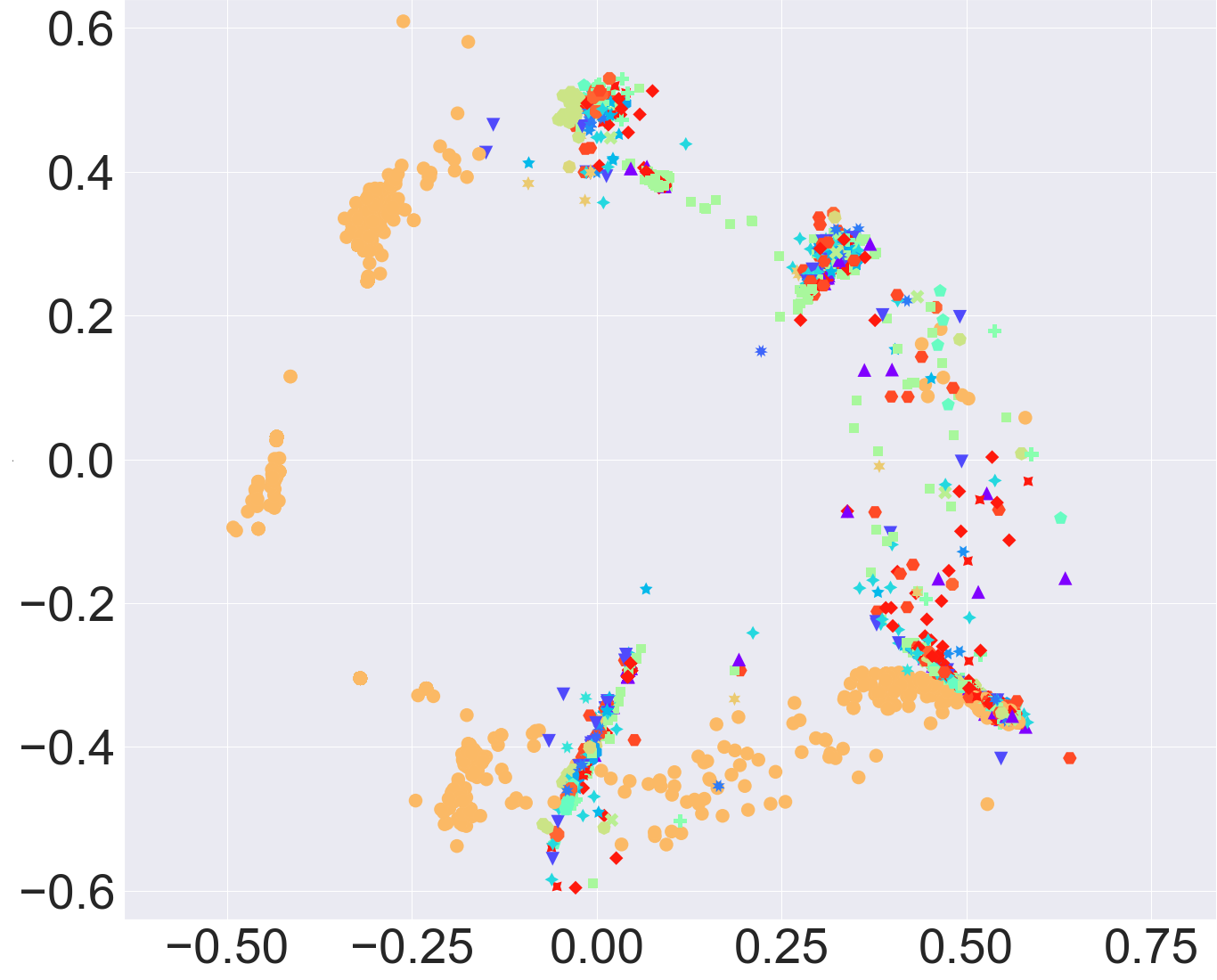}
    \caption{Polynomial}
    \end{subfigure}%
    \begin{subfigure}{0.2\textwidth}
        \centering
        \includegraphics[scale=0.07]{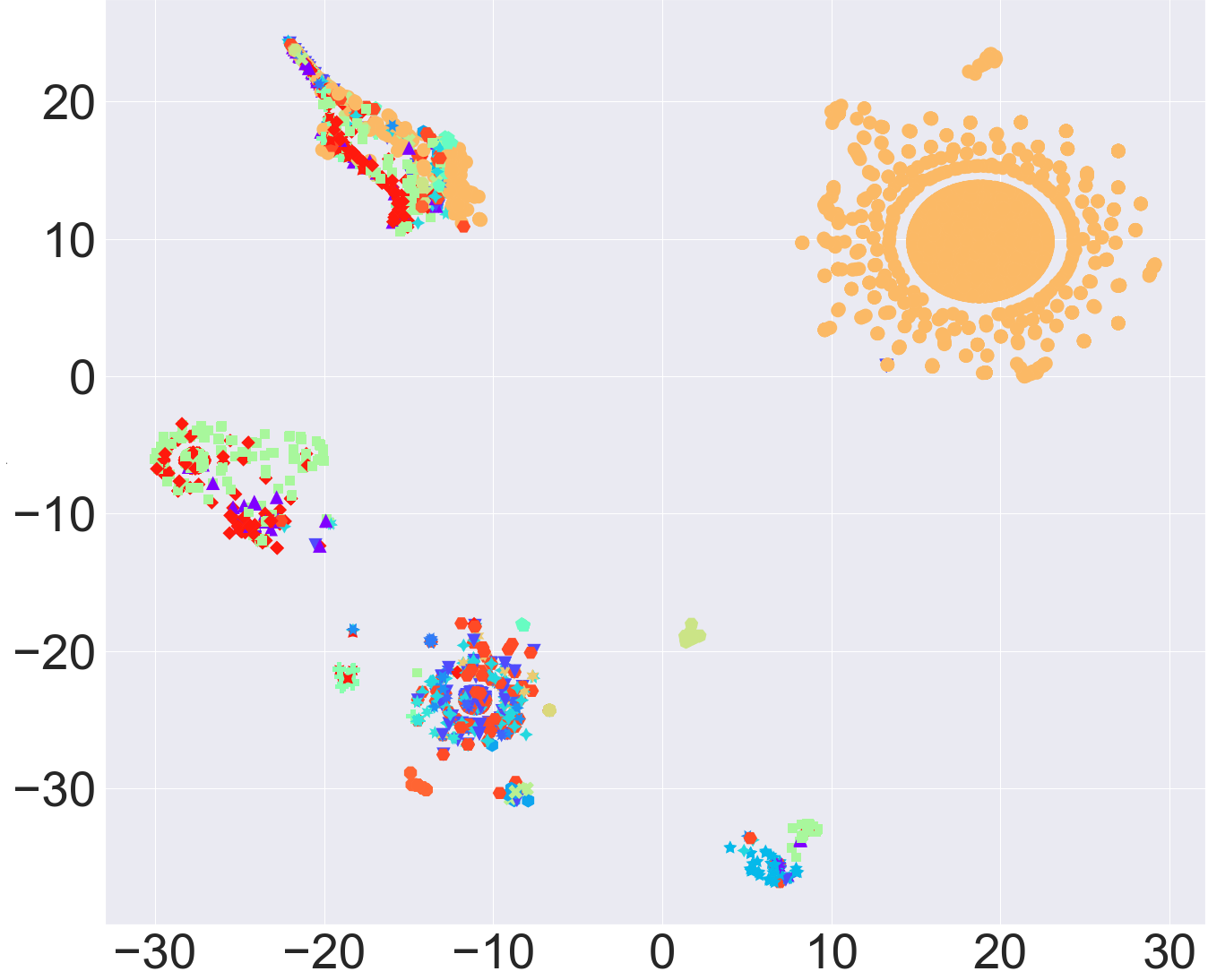}
    \caption{Gaussian}
    \end{subfigure}%
    \begin{subfigure}{0.2\textwidth}
        \centering
        \includegraphics[scale=0.07]{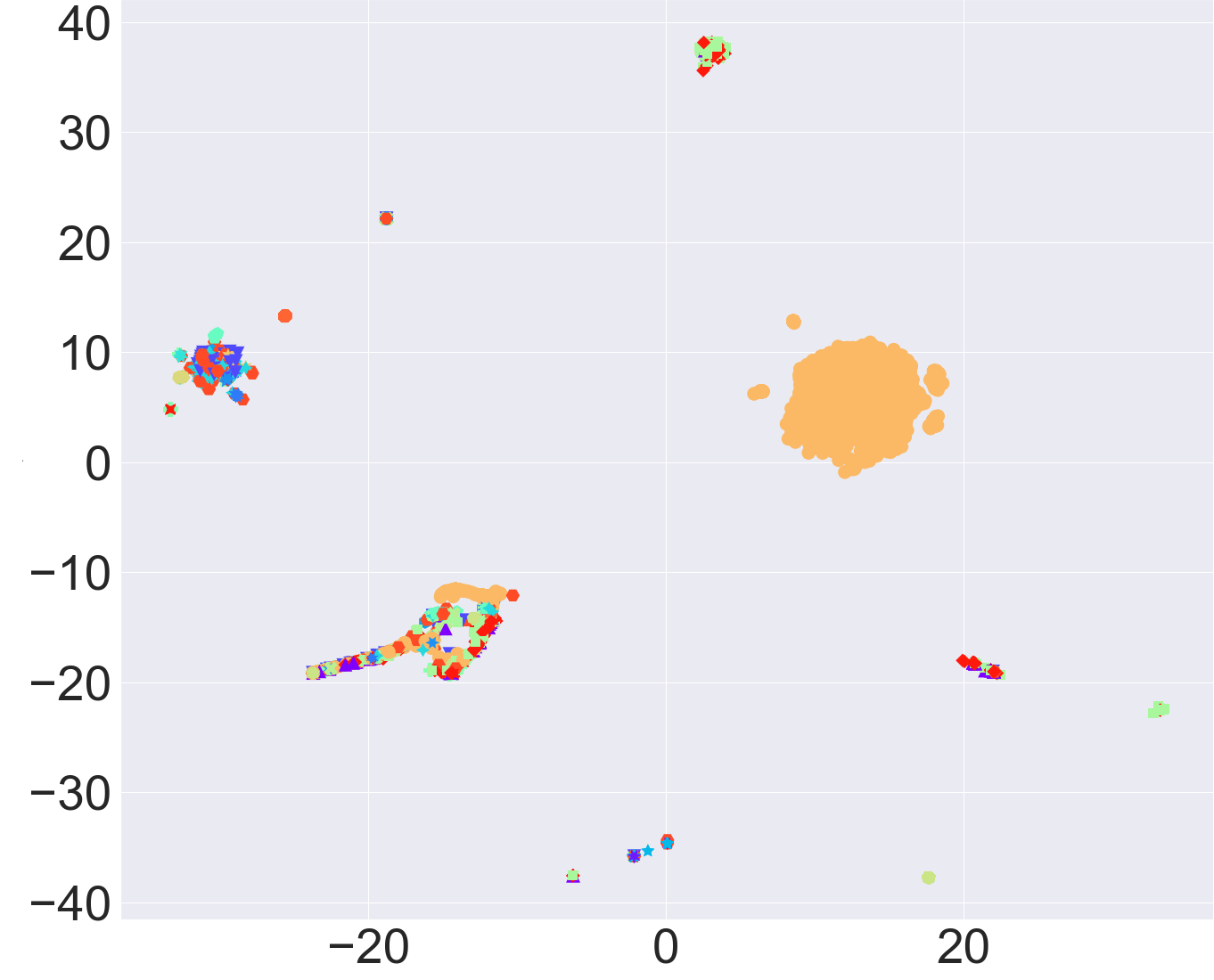}
    \caption{Isolation}
    \end{subfigure}%
    \begin{subfigure}{0.2\textwidth}
        \centering
        \includegraphics[scale=0.07]{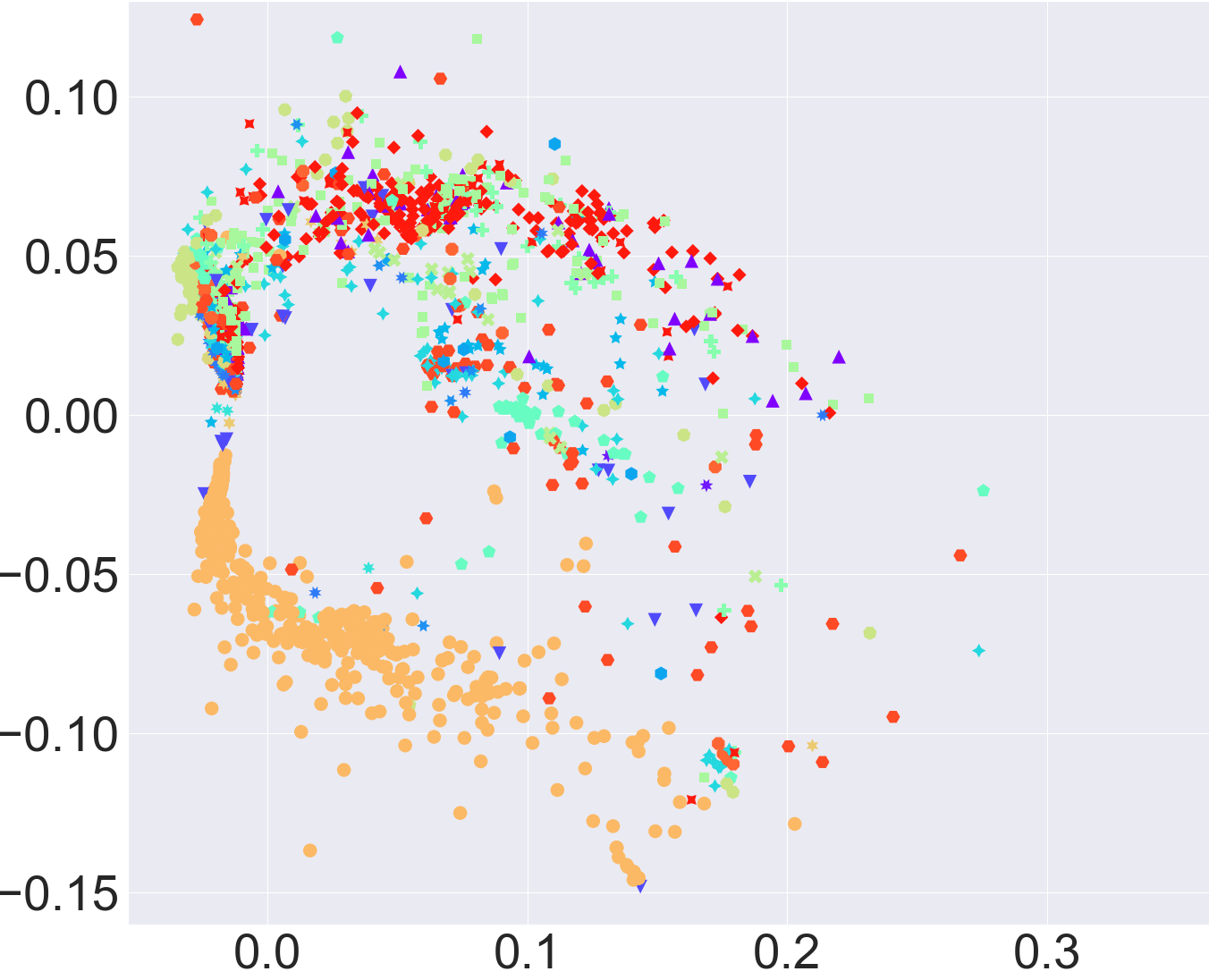}
    \caption{Laplacian}
    \end{subfigure}%
    \begin{subfigure}{0.2\textwidth}
        \centering
        \includegraphics[scale=0.07]{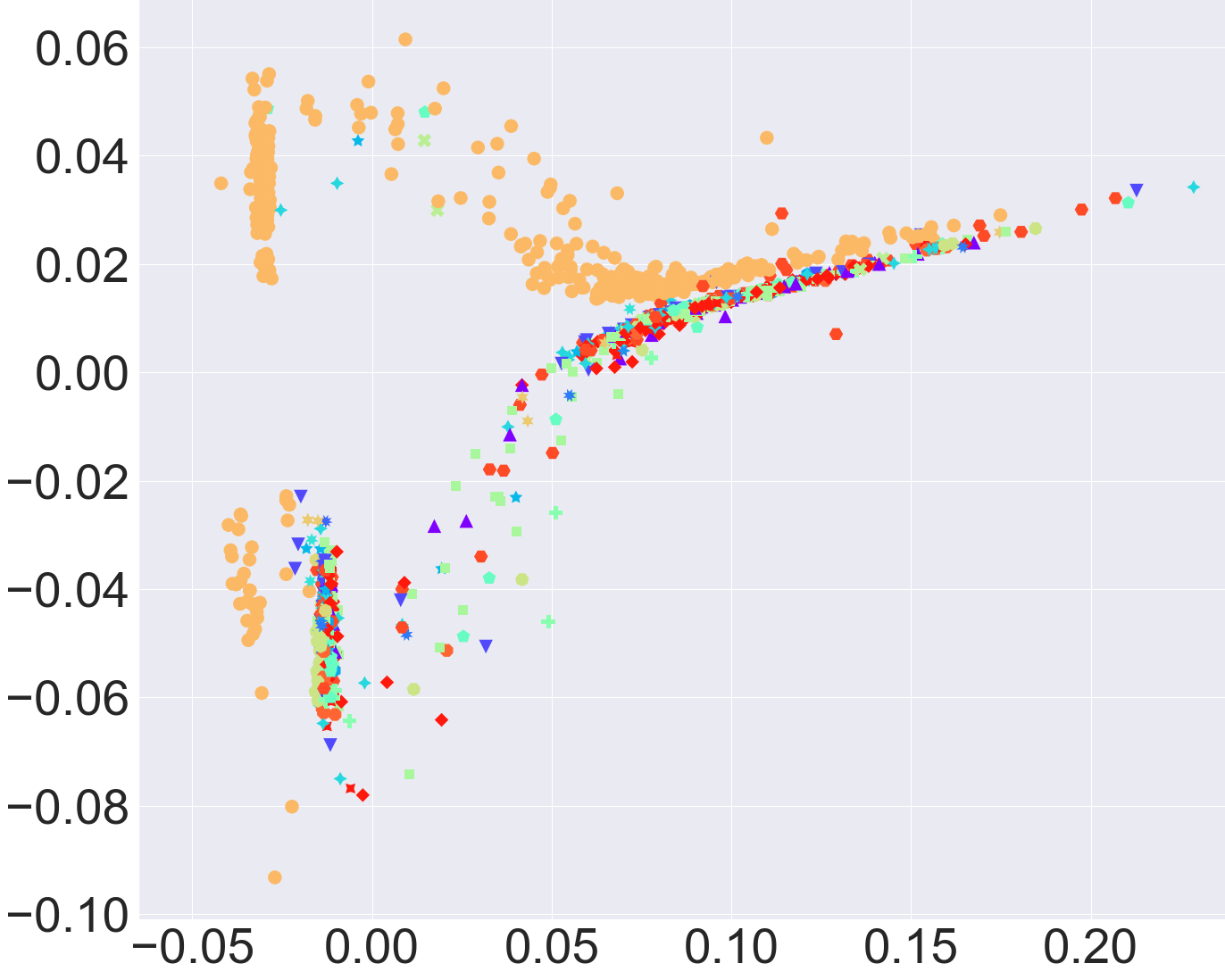}
    \caption{Sigmoid}
    \end{subfigure}%
    \\
    \begin{subfigure}{1\textwidth}
        \centering
       \includegraphics[scale=0.4]{Figures/legends_7000.png}
    \end{subfigure}%
   \caption{t-SNE plots for different kernel methods using minimizers-based Embedding.}
   \label{fig_tsne_minimizers_plots}
\end{figure}

\begin{figure}[h!]
\centering
\begin{subfigure}{.25\textwidth}
  \centering
  \includegraphics[scale=0.085]{Figures/o_cosine_7000.png}
  \caption{Cosine}
  \label{fig_tsne_ohe_cosine}
\end{subfigure}%
\begin{subfigure}{.25\textwidth}
  \centering
  \includegraphics[scale=0.085]{Figures/o_linear_7000.png}
  \caption{Linear}
  \label{fig_tsne_ohe_cosine}
\end{subfigure}%
\begin{subfigure}{.25\textwidth}
  \centering
  \includegraphics[scale=0.085]{Figures/o_gaussian_matlab_7000.png}
  \caption{Gaussian}
  \label{fig_tsne_ohe_cosine}
\end{subfigure}%
\begin{subfigure}{.25\textwidth}
  \centering
  \includegraphics[scale=0.085]{Figures/cosine_7000.png}
  \caption{Cosine}
  \label{fig_tsne_kmer_cosine}
\end{subfigure}%
\\
\begin{subfigure}{.2\textwidth}
  \centering
  \includegraphics[scale=0.07]{Figures/laplacian_7000.png}
  \caption{Laplacian}
  \label{fig_tsne_kmer_laplacian}
\end{subfigure}%
\begin{subfigure}{.2\textwidth}
  \centering
  \includegraphics[scale=0.07]{Figures/gaussian_matlab_7000.png}
  \caption{Gaussian}
  \label{fig_tsne_kmer_gaussian}
\end{subfigure}%
\begin{subfigure}{.2\textwidth}
  \centering
  \includegraphics[scale=0.07]{Figures/m_cosine_7000.png}
  \caption{Cosine}
  \label{fig_tsne__minimizer_cosine}
\end{subfigure}%
\begin{subfigure}{.2\textwidth}
  \centering
  \includegraphics[scale=0.07]{Figures/m_laplacian_7000.png}
  \caption{Laplacian}
  \label{fig_tsne_minimizer_laplacian}
\end{subfigure}%
\begin{subfigure}{0.2\textwidth}
  \centering
  \includegraphics[scale=0.07]{Figures/m_sigmoid_7000.png}
  \caption{Sigmoid}
  \label{fig_tsne_minimizer_sigmoid}
\end{subfigure}%
\\
\begin{subfigure}{1\textwidth}
  \centering
  \includegraphics[scale=0.38]{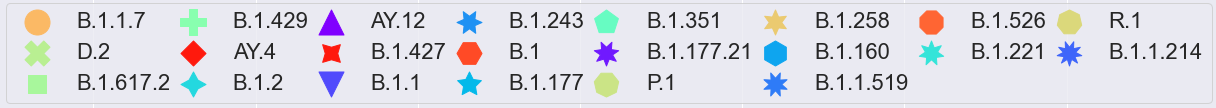}
\end{subfigure}%
\caption{t-SNE plots for top performing kernel methods for different embedding. This figure is best seen in color. (a) , (b) and (c) are from OHE. (d), (e), and (f) are from Spike2Vec. (g), (h) and (i) are from Minimizer encoding }
 \label{fig_top_performing_kernel_tsne}
\end{figure}

The t-SNE plots for one-hot embedding are given in Figure~\ref{fig_tsne_minimizers_plots} for different kernel methods. For the Gaussian 
and isolation kernel, we can observe similar behavior for the alpha (B.1.1.7) variant. 
Note that similar behavior is observed for the Spaced k-mers embedding. Moreover, Figure~\ref{fig_top_performing_kernel_tsne} shows the top 3 performing kernels in terms of $AUC_{RNX}$ score for the respective embedding.

\begin{figure}[h!]
\centering
\begin{subfigure}{.5\textwidth}
  \centering
        \includegraphics[scale=0.30]{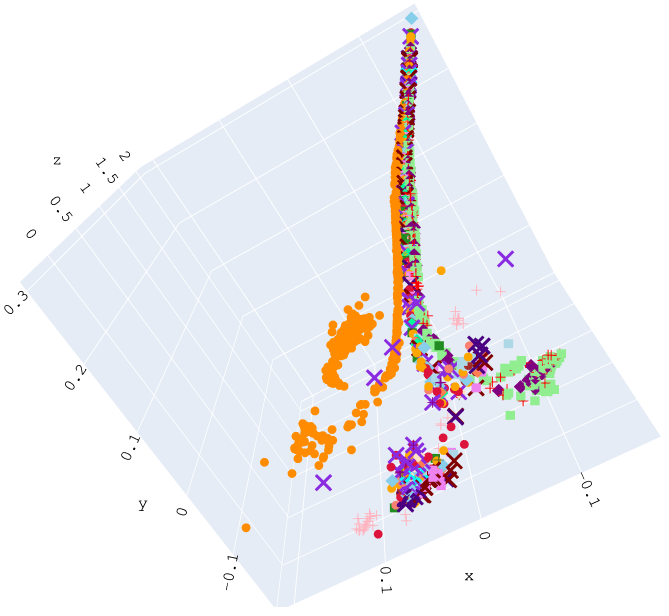}
        \caption{Spike2Vec}
        \label{}
\end{subfigure}%
\begin{subfigure}{.5\textwidth}
  \centering
        \includegraphics[scale=0.30]{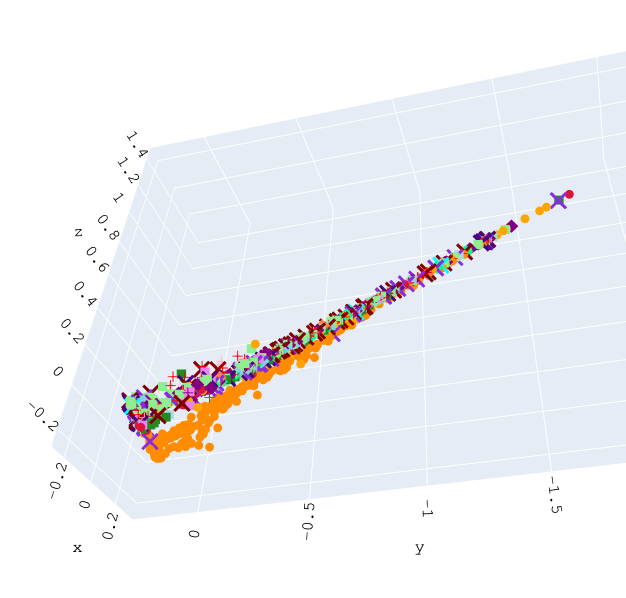}
        \caption{Minimizer}
        \label{}
\end{subfigure}%
\\
\begin{subfigure}{1\textwidth}
  \centering
     \includegraphics[scale=0.6]{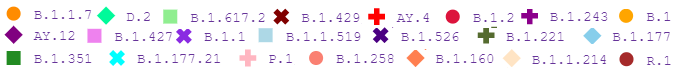}
\end{subfigure}%
    \caption{t-SNE 3d plots using Cosine similarity kernel. }
 \label{fig_tsne_3d_plots}
\end{figure}

We also show the 3D plots for t-SNE using the Cosine similarity kernel in Figure~\ref{fig_tsne_3d_plots} (for Spike2Vec and Minimizers-based embedding using Spike7k dataset). We can see for Spike2Vec, the Alpha variant shows clear grouping. Similarly, the delta and epsilon variant also contains a few small groups.

\clearpage

\section{Objective evaluation of t-SNE}

For objective evaluation of different kernel methods and embeddings on different datasets, we use the formulation mentioned in Equation~\ref{eq_tsne_eval}.
Table~\ref{tbl_kernel_auc_comparison_spike7k} demonstrates that cosine similarity achieves superior performance on the Spike7k dataset, particularly when combined with Spike2Vec encoding ($AUC_{RNX} = 0.331$). The computational efficiency of cosine similarity is notable, with kernel computation times substantially lower than chi-squared or Laplacian kernels. Spaced-Kmer encoding shows near-zero $AUC_{RNX}$ values across all kernels, suggesting poor neighborhood preservation for this representation on spike protein sequences.

\begin{table}[ht!]
    \centering
    \resizebox{0.99\textwidth}{!}{
    \begin{tabular}{@{\extracolsep{6pt}}cp{1.2cm}p{0.8cm}p{1.2cm}p{1cm}p{1.2cm}p{1cm} p{1.2cm}p{1cm}}
    \toprule
    \multirow{3}{*}{Kernel}& \multicolumn{2}{c}{OHE} & \multicolumn{2}{c}{Spike2Vec} & \multicolumn{2}{c}{Minimizer} & \multicolumn{2}{c}{Spaced-Kmer}\\
    \cmidrule{2-3} \cmidrule{4-5} \cmidrule{6-7} \cmidrule{8-9}
       & Kernel Comput. Time  & $AUC_{RNX}$ &  Kernel Comput. Time & $AUC_{RNX}$ &
      Kernel Comput. Time & $AUC_{RNX}$ & Kernel Comput. Time  & $AUC_{RNX}$ \\
      \midrule \midrule 
        Cosine similarity & \textbf{37.351} & \textbf{0.262}  &  \textbf{15.865} &  \textbf{0.331}  &  \textbf{17.912} &  \textbf{0.278}  & 92.516 & 0.00009   \\
		Linear  &  51.827 & 0.260  & 30.978 & 0.197  & 27.503 & 0.165 & 66.546 & 0.00006   \\
		Gaussian &   94.784 & 0.199  & 54.192 & 0.235 & 65.096 & 0.206  & 560469.5 & 0.00009   \\
		Sigmoid &   57.298 & 0.190  & 30.037 & 0.168   & 27.813 & 0.229 & 74.722 & 0.00002   \\
		Laplacian &   2250.30 & 0.184  & 721.834 & 0.260   & 983.26 & 0.242 & 59670.6 & 0.00021   \\
		Polynomial &   57.526 & 0.177  & 28.427 & 0.059  & 28.666 & 0.166& 79.763 & 0.00010   \\
		Isolation &   39.329 & 0.161  & 24.162 & 0.189  & 27.266 & 0.172  & 106.152 & 0.00005     \\
		Chi-squared &   1644.86 & 0.131 & 495.264 & 0.104 & 921.63 & 0.095  & 139165.3 & 0.00003   \\
		Additive-chi2 &   1882.45 & 0.110  & 495.221 & 0.121 & 576.92 & 0.125  & 128310.4 & 0.00007    \\

        \bottomrule
    \end{tabular}
    }
    \caption{$AUC_{RNX}$ values for t-SNE using different kernel and encoding methods on \textbf{Spike7k datasets}. The best values are shown in bold. 
        }
    \label{tbl_kernel_auc_comparison_spike7k}
\end{table}

For the Host dataset (Table~\ref{tbl_kernel_auc_comparison_host}), chi-squared and sigmoid kernels with Spike2Vec encoding yield the highest $AUC_{RNX}$ values (0.144 using Chi-squared and 0.143 using Sigmoid, respectively), though at considerable computational cost in case of Chi-squared kernel. 

\begin{table}[ht!]
    \centering
    \resizebox{0.99\textwidth}{!}{
    \begin{tabular}{@{\extracolsep{6pt}}cp{1.2cm}p{1cm}p{1.2cm}p{1cm}p{1.2cm}p{1cm} p{1.2cm}p{1cm}}
    \toprule
    \multirow{3}{*}{Kernel}& \multicolumn{2}{c}{OHE} & \multicolumn{2}{c}{Spike2Vec} & \multicolumn{2}{c}{Minimizer} & \multicolumn{2}{c}{Spaced-Kmer}\\
    \cmidrule{2-3} \cmidrule{4-5} \cmidrule{6-7} \cmidrule{8-9}
       & Kernel Comput. Time  & $AUC_{RNX}$ &  Kernel Comput. Time & $AUC_{RNX}$ &
      Kernel Comput. Time & $AUC_{RNX}$ & Kernel Comput. Time  & $AUC_{RNX}$ \\
      \midrule \midrule 
        Cosine similarity & 11.203 & 0.0205   & 5.072 & 0.126   &  2.781 & 0.094  & 174.508 & 0.129  \\
		Linear  &  7.936 & 0.0087   & 4.127 & 0.043   & 4.527 & 0.106  & 100.539 & 0.037   \\
		Gaussian &   7636.44 & 0.0180   & 197498.2 & 0.139   & 12847.7 & 0.122  & 44046.5 & 0.135   \\
		Sigmoid &   9.55 & 0.0206   & 6.16 & 0.143   & 4.477 & 0.0006  & 147.634 & 0.031  \\
		Laplacian &   5811.897 & 0.0214   & 1019.6 & 0.135   & 1252.822 & 0.131  & 24516.9 & 0.135   \\
		Polynomial &   9.366 & 0.0011   & 8.506 & 0.005   & 8.523 & 0.106  & 140.172 & 0.099   \\
		Isolation &   49.641 & 0.00004   & 66.745 & 0.0007   & 47.847 & 0.0001  & 66.431 & 0.011   \\
		Chi-squared &   19447.176 & 0.0201   & 3482.3 & 0.144   & 3581.766 & 0.105  & 57711.9 & 0.136   \\
		Additive-chi2 &    3241.723 & 0.0136  & 3547.94 & 0.048   & 3227.774 & 0.069  & 57408.6 & 0.041    \\

        \bottomrule
    \end{tabular}
    }
    \caption{$AUC_{RNX}$ values for t-SNE using different kernel and encoding methods on \textbf{Host datasets}. The best values are shown in bold. 
        }
    \label{tbl_kernel_auc_comparison_host}
\end{table}

Table~\ref{tbl_kernel_auc_comparison_short_read} reports $AUC_{RNX}$ values for different kernels and embeddings on short read dataset. The  near-zero values across most kernel-encoding combinations indicate poor neighbourhood preservation quality for short read sequences.

\begin{table}[ht!]
    \centering
    \resizebox{0.99\textwidth}{!}{
    \begin{tabular}{@{\extracolsep{6pt}}cp{1.2cm}p{1cm}p{1.2cm}p{1cm}p{1.2cm}p{1cm} p{1.2cm}p{1cm}}
    \toprule
    \multirow{3}{*}{Kernel}& \multicolumn{2}{c}{OHE} & \multicolumn{2}{c}{Spike2Vec} & \multicolumn{2}{c}{Minimizer} & \multicolumn{2}{c}{Spaced-Kmer}\\
    \cmidrule{2-3} \cmidrule{4-5} \cmidrule{6-7} \cmidrule{8-9}
       & Kernel Comput. Time  & $AUC_{RNX}$ &  Kernel Comput. Time & $AUC_{RNX}$ &
      Kernel Comput. Time & $AUC_{RNX}$ & Kernel Comput. Time  & $AUC_{RNX}$ \\
      \midrule \midrule 
Cosine similarity & 28.702 & 0.00009   & 1.271 & 0.0001   & 1.338 & 0.406  & 0.969 & 0.0625  \\
		Linear  &  20.649 & 0.00001   & 1.342 & 0.0007   & 1.296 & 0.00013  & 0.945 & 0.0001   \\
		Gaussian &   170685.3 & 0.00002   & 2494.6 & 0.0002   & 2545.1 & 2.681  & 3278.294 & 0.0007   \\
		Sigmoid &   22.484 & 0.00001   & 2.847 & 0.0004   & 2.574 & 0.0012  & 2.397 & 0.0002  \\
		Laplacian &   32735.7 & 0.00004   & 21.449 & 0.0001   & 21.176 & 0.0007  & 117.818 & 0.0005   \\
		Polynomial &   25.675 & 0.00008   & 16.387 & 0.0006   & 16.557 & 0.0001  & 16.645 & 0.0251   \\
		Isolation &   144.783 & 0.00006   & 139.55 & 0.0001   & 138.88 & 0.0005  & 138.307 & 0.0004    \\
		Chi-squared &   175684.8 & 0.00002   & 60.684 & 0.0009   & 50.787 & 0.0275  & 490.043 & 0.0009   \\
		Additive-chi2 &   176235.1 & 0.00011   & 57.273 & 0.0003   & 45.024 & 0.0942  & 371.799 & 0.0944    \\

        \bottomrule
    \end{tabular}
    }
    \caption{$AUC_{RNX}$ values for t-SNE using different kernel and encoding methods on \textbf{Short Read dataset}. The best values are shown in bold. 
        }
    \label{tbl_kernel_auc_comparison_short_read}
\end{table}

On the Rabies dataset (Table~\ref{tbl_kernel_auc_comparison_rabies}), Minimizer and Spaced-Kmer encodings demonstrate strong neighborhood preservation ($AUC_{RNX} \approx 0.13-0.15$) when paired with cosine similarity, Gaussian, or Laplacian kernels. These results are achieved with moderate computational overhead (other than gausian kernel), establishing an effective balance between accuracy and efficiency for viral genome analysis.

\begin{table}[ht!]
    \centering
    \resizebox{0.99\textwidth}{!}{
    \begin{tabular}{@{\extracolsep{6pt}}cp{1.2cm}p{1cm}p{1.2cm}p{1cm}p{1.2cm}p{1cm} p{1.2cm}p{1cm}}
    \toprule
    \multirow{3}{*}{Kernel}& \multicolumn{2}{c}{OHE} & \multicolumn{2}{c}{Spike2Vec} & \multicolumn{2}{c}{Minimizer} & \multicolumn{2}{c}{Spaced-Kmer}\\
    \cmidrule{2-3} \cmidrule{4-5} \cmidrule{6-7} \cmidrule{8-9}
       & Kernel Comput. Time  & $AUC_{RNX}$ &  Kernel Comput. Time & $AUC_{RNX}$ &
      Kernel Comput. Time & $AUC_{RNX}$ & Kernel Comput. Time  & $AUC_{RNX}$ \\
      \midrule \midrule 
Cosine similarity & 57.115 & 0.033630   & 32.92 & 0.029732   & 5.446 & 0.15034  & 7.833 & 0.134273  \\
		Linear  &  49.279 & 0.028748   & 25.18 & 0.005480   & 13.017 & 0.018272  & 11.082 & 0.014808   \\
		Gaussian &   284038.726 & 0.023793   & 102629.365 & 0.024123   & 4938.652 & 0.147844  & 6018.947 & 0.133738   \\
		Sigmoid &   59.17 & 0.004990   & 60.5 & 0.029248   & 21.495 & 0.000685  & 18.433 & 0.133057  \\
		Laplacian &   66975.297 & 0.031071   & 16292.126 & 0.031966   & 234.528 & 0.147133  & 549.362 & 0.133938   \\
		Polynomial &   74.54 & 0.023586   & 95.51 & 0.000480   & 131.311 & 0.001318  & 36.505 & 0.001871   \\
		Isolation &   1997.925 & 0.000023   & 1843.375 & 0.000071   & 1774.019 & 0.000054  & 2443.647 & 0.000071    \\
		Chi-squared &   80852.645 & 0.034822   & 20574.939 & 0.036115   & 375.659 & 0.147215  & 580.142 & 0.133544   \\
		Additive-chi2 &   125464.199 & 0.004369   & 18511.205 & 0.009503   & 332.14 & 0.028714  & 580.384 & 0.052454    \\

        \bottomrule
    \end{tabular}
    }
    \caption{$AUC_{RNX}$ values for t-SNE using different kernel and encoding methods on \textbf{Rabies datasets}. The best values are shown in bold. 
        }
    \label{tbl_kernel_auc_comparison_rabies}
\end{table}

Table~\ref{tbl_kernel_auc_comparison_genome} shows modest $AUC_{RNX}$ values across all configurations, with Laplacian and Gaussian kernels paired with Minimizer or Spaced-Kmer encodings achieving the best performance ($AUC_{RNX} \approx 0.034-0.038$). The substantially longer computation times for certain kernels suggest that genome-scale data requires careful consideration of the accuracy-efficiency tradeoff.

\begin{table}[ht!]
    \centering
    \resizebox{0.99\textwidth}{!}{
    \begin{tabular}{@{\extracolsep{6pt}}cp{1.2cm}p{1cm}p{1.2cm}p{1cm}p{1.2cm}p{1cm} p{1.2cm}p{1cm}}
    \toprule
    \multirow{3}{*}{Kernel}& \multicolumn{2}{c}{OHE} & \multicolumn{2}{c}{Spike2Vec} & \multicolumn{2}{c}{Minimizer} & \multicolumn{2}{c}{Spaced-Kmer}\\
    \cmidrule{2-3} \cmidrule{4-5} \cmidrule{6-7} \cmidrule{8-9}
       & Kernel Comput. Time  & $AUC_{RNX}$ &  Kernel Comput. Time & $AUC_{RNX}$ &
      Kernel Comput. Time & $AUC_{RNX}$ & Kernel Comput. Time  & $AUC_{RNX}$ \\
      \midrule \midrule 
Cosine similarity & 512.508 & 0.000363   & 2.584 & 0.002   & 2.131 & 0.003  & 3.015 & 0.012  \\
		Linear  &  376.152 & 0.000216   & 1.666 & 0.0301   & 1.795 & 0.0336  & 2.828 & 0.032   \\
		Gaussian &   959921.408 & 0.000196   & 4267.259 & 0.034   & 4004.788 & 0.0387  & 4305.981 & 0.034   \\
		Sigmoid &   466.266 & 0.000240   & 4.082 & 0.0001   & 2.803 & 0.0211  & 3.871 & 0.0003  \\
		Laplacian &   181529.777 & 0.000428   & 34.778 & 0.0336   & 28.244 & 0.0389  & 127.441 & 0.0351   \\
		Polynomial &   497.364 & 0.000424   & 26.636 & 0.0323   & 19.756 & 0.034  & 15.552 & 0.0341   \\
		Isolation &   191.014 & 0.000050   & 190.145 & 0.0225   & 231.071 & 0.0250  & 152.033 & 0.0001    \\
		Chi-squared &   455003.79 & 0.000560   & 91.718 & 0.0341   & 90.251 & 0.0379  & 422.979 & 0.034   \\
		Additive-chi2 &   449293.663 & 0.000364   & 74.021 & 0.004   & 62.237 & 0.0159  & 404.173 & 0.0047    \\

        \bottomrule
    \end{tabular}
    }
    \caption{$AUC_{RNX}$ values for t-SNE using different kernel and encoding methods on \textbf{Genome datasets}. The best values are shown in bold. 
        }
    \label{tbl_kernel_auc_comparison_genome}
\end{table}

For the Breast Cancer dataset (Table~\ref{tbl_kernel_auc_comparison_breast_cancer}), chi-squared kernel with Minimizer encoding achieves the highest $AUC_{RNX}$ (0.0458), though polynomial kernel with Spaced-Kmer also performs well (0.09228). The relatively low computation times across most configurations indicate this smaller dataset enables exploration of more computationally intensive kernel methods without prohibitive runtime penalties.

\begin{table}[ht!]
    \centering
    \resizebox{0.99\textwidth}{!}{
    \begin{tabular}{@{\extracolsep{6pt}}cp{1.2cm}p{1cm}p{1.2cm}p{1cm}p{1.2cm}p{1cm} p{1.2cm}p{1cm}}
    \toprule
    \multirow{3}{*}{Kernel}& \multicolumn{2}{c}{OHE} & \multicolumn{2}{c}{Spike2Vec} & \multicolumn{2}{c}{Minimizer} & \multicolumn{2}{c}{Spaced-Kmer}\\
    \cmidrule{2-3} \cmidrule{4-5} \cmidrule{6-7} \cmidrule{8-9}
       & Kernel Comput. Time  & $AUC_{RNX}$ &  Kernel Comput. Time & $AUC_{RNX}$ &
      Kernel Comput. Time & $AUC_{RNX}$ & Kernel Comput. Time  & $AUC_{RNX}$ \\
      \midrule \midrule 
Cosine similarity & 0.184 & 0.001059   & 1.841 & 0.0018   & 1.86 & 0.037  & 39.556 & 0.0226  \\
		Linear  &  0.168 & 0.000826   & 1.767 & 0.0015   & 1.599 & 0.0281  & 33.601 & 0.0136   \\
		Gaussian &   73.039 & 0.000581   & 139.648 & 0.0006   & 155.362 & 0.00955  & 3822.062 & 0.00521   \\
		Sigmoid &   0.37 & 0.001120   & 1.992 & 0.0009   & 2.985 & 0.0289  & 32.741 & 0.0002  \\
		Laplacian &   2.764 & 0.001558   & 31.888 & 0.0005   & 35.299 & 0.04002  & 1070.505 & 0.0136   \\
		Polynomial &   0.523 & 0.000187   & 1.686 & 0.0006   & 1.926 & 0.0398  & 31.912 & 0.09228  \\
		Isolation &   0.975 & 0.000812   & 0.807 & 0.0003   & 0.847 & 0.00057  & 5.048 & 0.00012    \\
		Chi-squared &   8.122 & 0.002648   & 57.782 & 0.0004   & 55.306 & 0.0458  & 1589.326 & 0.0193   \\
		Additive-chi2 &   8.623 & 0.000286   & 55.848 & 0.0001   & 57.097 & 0.0132  & 1779.769 & 0.00361    \\

        \bottomrule
    \end{tabular}
    }
    \caption{$AUC_{RNX}$ values for t-SNE using different kernel and encoding methods on \textbf{Breast Cancer datasets}. The best values are shown in bold. 
        }
    \label{tbl_kernel_auc_comparison_breast_cancer}
\end{table}

\clearpage

\section{Clustering Results}\label{sec_results}

We assess the quality of various embedding approaches by applying clustering algorithms with different kernel methods to the embeddings. Table~\ref{tbl_spike7k_clustering} presents the clustering performance (for both k-means and k-modes) measured through multiple internal validation metrics. The results reveal that for k-means clustering, the sigmoid kernel demonstrates superior performance according to Silhouette Coefficient and Calinski-Harabasz metrics, whereas the Gaussian kernel excels based on the Davies-Bouldin metric. For k-modes clustering, the chi-squared kernel achieves the best results across all three evaluation criteria.
Table~\ref{tbl_spike7k_clustering} also reports clustering performance for Spike2Vec embeddings. When applying k-means, the polynomial kernel yields optimal Silhouette Coefficient and Calinski-Harabasz values, while the Gaussian kernel produces the best Davies-Bouldin score. For k-modes clustering, the chi-squared kernel again dominates across all metrics, mirroring the pattern observed with OHE.
Clustering results for minimizer-based embeddings are presented in Table~\ref{tbl_spike7k_clustering}. The k-means algorithm exhibits the same kernel preference as Spike2Vec embeddings: polynomial kernel leads for Silhouette Coefficient and Calinski-Harabasz metrics, while Gaussian kernel performs best for Davies-Bouldin score. However, for k-modes clustering, the Gaussian kernel achieves superior Silhouette Coefficient and Calinski-Harabasz scores, whereas the additive-chi2 kernel provides the best Davies-Bouldin performance.

\begin{table}[h!]
  \centering
     \resizebox{0.99\textwidth}{!}{
   \begin{tabular}{@{\extracolsep{6pt}}p{0.5cm}p{1.8cm}
   p{1cm}p{1.2cm}p{0.6cm}p{1cm}
   p{1cm}p{1.2cm}p{0.8cm}p{1cm}
   p{1cm}p{1.2cm}p{0.8cm}p{1cm}
   p{1cm}p{1.2cm}p{0.8cm}p{1cm}
   }
    \toprule
     & & \multicolumn{4}{c}{OHE}  & \multicolumn{4}{c}{Spike2vec} & \multicolumn{4}{c}{Minimizer}& \multicolumn{4}{c}{Spaced k-mer} \\
    \cmidrule{3-6} \cmidrule{7-10} \cmidrule{11-14} \cmidrule{15-18}
    Algo. & Kernel & Silhouette Coefficient & Calinski Harabasz Score & Davies-Bouldin Score & Clustering Runtime (Sec.) & Silhouette Coefficient & Calinski Harabasz Score & Davies-Bouldin Score & Clustering Runtime (Sec.) & Silhouette Coefficient & Calinski Harabasz Score & Davies-Bouldin Score & Clustering Runtime (Sec.) & Silhouette Coefficient & Calinski Harabasz Score & Davies-Bouldin Score & Clustering Runtime (Sec.)\\
    \midrule	\midrule	
    \multirow{9}{*}{$k$-means} 
& Additive-chi2 & 0.841 & 31248.241 & 0.502 & 0.182 & 0.580 & 2880.192 & 1.273 & 0.164 & 0.527 & 2641.231 & 1.279 & 0.132 &0.549&1130.892&1.810&4.850\\
& Chi-squared & 0.485 & 2367.888 & 1.156 & 0.115 & 0.702 & 2926.907 & 0.342 & \textbf{0.091} & 0.677 & 2850.682 & 0.416 & 0.110 &0.527&1930.648&0.578&3.623  \\
& Cosine & 0.841 & 31248.241 & 0.502 & 0.116 & 0.565 & 30861.077 & 0.755 & 0.178 & 0.549 & 12474.205 & 0.949 & 0.187 &0.576&18573.876&0.984&3.072  \\
& Gaussian & 0.086 & 217.357 & 1.174 & 0.125 & 0.079 & 219.713 & 1.885 & 0.144  & 0.068 & 212.031 & 1.172 & 0.140 &0.503&1943.049&0.629&3.946 \\
& Isolation & 0.166 & 310.932 & 1.426 & 0.127 & 0.149 & 273.187 & 1.345 & 0.156 & 0.165 & 282.180 & 1.347 & 0.221 &0.847&281.357&0.628&6.754 \\
& Laplacian & 0.840 & 30573.981 & 0.501 & 0.166 & 0.528 & 2526.604 & 1.255 & 0.144 & 0.533 &0.530&1392.3 &1.553&20.909& 18692.820 & 0.724 & 6.863  \\
& Linear & 0.841 & 31248.241 & 0.502 & 0.167 & 0.843 & 27839.114 & 0.496 & 0.150  & 0.740 & 13683.208 & 0.655 & 0.163 &0.825&26567.483&0.385&11.527  \\
& Polynomial & 0.840 & 30592.941 & 0.501 & 0.107 & \textbf{0.919} &0.899&12497.3 &0.286&4.888 & \textbf{98941.860} & 0.409 & 0.171 & 0.928 & 14345.965 & 0.244 & 8.427  \\
& Sigmoid & \textbf{0.841} & \textbf{31719.471} & 0.502 & 0.189 & 0.557 & 26013.129 & 0.773 & 0.129 & 0.538 & 11183.672 & 0.910 & 0.187 &0.828&27887.420&0.392&2.298  \\
    
    \midrule
    \multirow{9}{*}{$k$-modes} 
& Additive-chi2 & -0.248 & 93.070 & 1.131 & 24.906  & -0.383 & 27.507 & 1.854 & 17.810 & -0.374 & 18.366 & \textbf{1.521} & 18.478 &-0.647&19.314&1.327&8405.023  \\
& Chi-squared & \textbf{0.019} & \textbf{250.247} & \textbf{0.751} & 23.813 & \textbf{0.058} & \textbf{191.749} & \textbf{0.714} & 17.869 & -0.351 & 10.807 & 2.136 & 22.274 &-0.023&233.120&0.825&2216.956   \\
& Cosine & -0.303 & 21.589 & 1.180 & 13.550 & -0.522 & 4.697 & 1.678 & 14.753 & -0.485 & 7.071 & 1.600 & 14.924 &-0.712&1.558&1.673&3404.789  \\
& Gaussian & -0.479 & 0.302 & 2.897 & 8.718 & -0.468 & 0.293 & 2.855 & 8.646 & -0.473 & 0.310 & 2.987 & 8.667 &-0.093&62.989&17.431&3755.273\\
& Isolation & -0.489 & 0.212 & 3.272 & 8.528 & -0.501 & 0.215 & 3.449 & 8.646 & -0.490 & 0.270 & 3.418 & 9.065 &-0.551&7.279&5.154&12523.985 \\
& Laplacian & -0.308 & 13.880 & 1.187 & 15.750 & -0.406 & 1.112 & 1.570 & 8.598 & -0.386 & 0.908 & 1.563 & 9.593 &-0.674&1.740&1.175&3167.464  \\
& Linear & -0.255 & 94.817 & 1.130 & 23.265 & -0.651 & 10.568 & 1.649 & 18.532 & -0.500 & 27.030 & 1.595 & 18.902 &-0.642&15.241&1.582&2580.823 \\
& Polynomial & -0.286 & 38.201 & 1.169 & 8.923 & -0.817 & 0.037 & 1.842 & 11.054 & -0.688 & 0.544 & 1.865 & 13.802 &-0.703&0.750&1.654&842.016 \\
& Sigmoid & -0.312 & 2.223 & 1.194 & 10.256 & -0.585 & 0.745 & 1.781 & \textbf{8.578} & -0.636 & 0.746 & 1.609 & 9.597 &-0.715&5.497&1.297&1320.494 \\
    \bottomrule
  \end{tabular}
  }
  \caption{Internal clustering quality metrics for different kernels and embeddings on \textbf{Spike7k dataset}.}
  \label{tbl_spike7k_clustering}
\end{table}

For the Host dataset (Table~\ref{tbl_host_clustering}), the polynomial kernel demonstrates exceptional performance with $k$-means clustering, achieving remarkably high Silhouette Coefficient (0.975) and Calinski-Harabasz scores (13378.873) for both OHE and Spike2Vec embeddings, along with favorable Davies-Bouldin scores (0.516). The isolation kernel also exhibits strong clustering quality across all embeddings, though with relatively longer computation times. In contrast, $k$-modes clustering shows predominantly negative Silhouette Coefficients across all kernel-embedding combinations, indicating poor cluster separation. The Minimizer embedding with chi-squared kernel achieves the best $k$-modes performance (Calinski-Harabasz: 146.595), though computational overhead remains substantial.

\begin{table}[h!]
  \centering
     \resizebox{0.99\textwidth}{!}{
   \begin{tabular}{@{\extracolsep{6pt}}ccp{1.2cm}p{1.5cm}p{0.8cm}p{1.2cm}p{1.2cm}p{1.6cm}p{0.8cm}p{1.2cm}p{1.2cm}p{1.5cm}p{0.8cm}p{1.2cm}p{1.2cm}p{1.5cm}p{0.8cm}p{1.2cm}}
    \toprule
     & & \multicolumn{4}{c}{OHE}  & \multicolumn{4}{c}{Spike2vec} & \multicolumn{4}{c}{Minimizer}& \multicolumn{4}{c}{Spaced k-mer} \\
    \cmidrule{3-6} \cmidrule{7-10} \cmidrule{11-14} \cmidrule{15-18}
    Algo. & Kernel & Silhouette Coefficient & Calinski Harabasz Score & Davies-Bouldin Score & Clustering Runtime (Sec.) & Silhouette Coefficient & Calinski Harabasz Score & Davies-Bouldin Score & Clustering Runtime (Sec.) & Silhouette Coefficient & Calinski Harabasz Score & Davies-Bouldin Score & Clustering Runtime (Sec.) & Silhouette Coefficient & Calinski Harabasz Score & Davies-Bouldin Score & Clustering Runtime (Sec.)\\
    \midrule	\midrule	
    \multirow{9}{*}{$k$-means} 
& Additive-chi2 &0.548&2100.474&1.279&2.012 &0.552&2221.759&1.368&2.113&0.603&1962.352&1.389&2.489 &0.647&2059.385&1.481&9.244
 \\
& Chi-squared &0.337&563.682&0.596&2.171&0.339&575.757&0.595&2.474 &0.410&698.090&0.789&5.490&0.346&570.304&0.587&4.504
 \\
& Cosine &-0.630&9.065&1.261&4699.372&0.667&2607.411&1.425&11.744 &0.595&2516.870&1.213&11.896 &0.413&1964.520&1.351&15.052 
 \\
& Gaussian &0.367&611.175&0.601&4.205 &0.348&518.095&0.609&3.591&0.419&819.097&0.654&4.225 &0.327&423.378&0.618&4.763
 \\
& Isolation &0.853&108.170&0.333&2.140 &0.778&120.465&0.930&3.653 &0.769&124.567&0.660&3.638&0.736&111.291&0.819&2.002 \\
& Laplacian &0.524&2031.987&1.320&9.128 &0.626&2095.452&1.432&6.957 &0.634&1875.474&1.468&5.106 &0.543&2073.552&1.539&8.310 
 \\
& Linear &0.806&14136.462&0.484&6.483 &0.806&14136.462&0.484&6.882 &0.633&3996.076&0.532&5.015&0.804&13973.043&0.488&4.474 
 \\
& Polynomial &0.975&13378.873&0.516&3.966 &0.975&13378.873&0.516&3.734 &0.715&12237.580&0.381&1.797&0.897&9566.441&0.547&3.320 
 \\
& Sigmoid &0.628&3565.245&1.226&4.095&0.604&3505.662&1.278&4.150&0.000&7.240&22.297&1.961 &0.712&13484.251&0.526&1.897
 \\
    
    \midrule
    \multirow{9}{*}{$k$-modes} 
& Additive-chi2 &-0.504&12.170&1.156&2522.139 &-0.491&16.360&1.837&3079.975
 &-0.400&12.037&1.801&4562.968 &-0.393&20.463&1.725&2536.756 
 \\
& Chi-squared &-0.409&5.339&5.736&2190.884 &-0.408&5.427&5.735&2754.262 &-0.347&146.595&2.551&3430.494 &-0.393&6.930&5.989&1844.897
 \\
& Cosine &-0.630&9.065&1.261&4699.372&-0.508&18.408&1.317&6415.113&-0.683&2.432&1.620&4327.268&-0.551&17.584&1.319&6613.378
 \\
& Gaussian &-0.154&16.150&21.112&2670.864
&-0.153&14.777&20.500&4486.247&-0.151&29.105&21.684&4323.309 &-0.152&14.544&19.867&3168.317
 \\
& Isolation &-0.368&10.444&6.477&9829.114 &-0.655&2.859&6.959&17960.518 &-0.410&3.236&6.799&17561.721 &-0.438&4.867&7.091&9805.705 \\
& Laplacian &-0.458&8.355&1.350&1600.797 &-0.476&3.668&1.446&1599.906 &-0.529&3.074&1.382&2317.782 &-0.438&2.529&1.167&1456.450 
 \\
& Linear &-0.517&14.229&1.752&847.966
 &-0.487&18.161&1.744&920.437 &-0.334&26.566&2.432&1815.314
 &-0.536&7.276&1.758&976.420
 \\
& Polynomial &-0.793&1.638&1.866&648.035 &-0.826&1.871&2.284&648.198
 &-0.563&31.853&2.093&1337.207 &-0.638&0.859&1.863&703.717
 \\
& Sigmoid &-0.578&4.035&1.392&5175.608&-0.626&4.447&1.407&6982.080 &-0.100&0.674&1.200&615.030 &-0.590&0.950&1.557&639.617 \\
    \bottomrule
  \end{tabular}
  }
  \caption{Internal clustering quality metrics for different kernels and embeddings on \textbf{Host dataset}. }
  \label{tbl_host_clustering}
\end{table}

The ShortRead dataset results (Table~\ref{tbl_shortRead_clustering}) reveal that the isolation kernel delivers superior clustering quality for $k$-means across all embeddings, with Silhouette Coefficients exceeding 0.90 and maintaining reasonable computation times. The Linear and Polynomial kernels paired with Spike2Vec, Minimizer, and Spaced k-mer embeddings also demonstrate strong performance, achieving notably high Calinski-Harabasz scores. For $k$-modes clustering, the Sigmoid kernel with Spike2Vec encoding yields the best results (Silhouette: 0.178, Davies-Bouldin: 0.669), though most configurations produce negative Silhouette values. The additive-chi2 kernel with Minimizer embedding presents an anomalous case where identical values appear in both $k$-means and $k$-modes results, suggesting potential data overlap or processing artifacts.

\begin{table}[h!]
  \centering
     \resizebox{0.99\textwidth}{!}{
   \begin{tabular}{@{\extracolsep{6pt}}ccp{1.2cm}p{1.5cm}p{0.8cm}p{1.2cm}p{1.2cm}p{1.6cm}p{0.8cm}p{1.2cm}p{1.2cm}p{1.5cm}p{0.8cm}p{1.2cm}p{1.2cm}p{1.5cm}p{0.8cm}p{1.2cm}}
    \toprule
     & & \multicolumn{4}{c}{OHE}  & \multicolumn{4}{c}{Spike2vec} & \multicolumn{4}{c}{Minimizer}& \multicolumn{4}{c}{Spaced k-mer} \\
    \cmidrule{3-6} \cmidrule{7-10} \cmidrule{11-14} \cmidrule{15-18}
    Algo. & Kernel & Silhouette Coefficient & Calinski Harabasz Score & Davies-Bouldin Score & Clustering Runtime (Sec.) & Silhouette Coefficient & Calinski Harabasz Score & Davies-Bouldin Score & Clustering Runtime (Sec.) & Silhouette Coefficient & Calinski Harabasz Score & Davies-Bouldin Score & Clustering Runtime (Sec.) & Silhouette Coefficient & Calinski Harabasz Score & Davies-Bouldin Score & Clustering Runtime (Sec.)\\
    \midrule	\midrule	
    \multirow{9}{*}{$k$-means} 
& Additive-chi2 &0.465&5445.396&0.619&3.830 &0.684&22545.943&0.727&2.206 &0.653&19207.977&0.757&26.296 &0.644&16504.801&0.875&2.817 
 \\
& Chi-squared &0.673&57.336&0.769&10.221 &0.447&14.721&0.937&24.267 &0.137&13.188&3.602&37.869 &0.568&14.211&0.504&3.983 
 \\
& Cosine &0.465&5445.396&0.619&5.398 &0.745&93896.412&0.529&23.620 &0.729&49085.719&0.524&33.267&0.663&114154.429&0.614&11.227
 \\
& Gaussian &-0.025&76.595&15.802&10.812 &0.500&14.124&2.110&34.046 &0.467&13.952&2.640&51.376&0.469&13.817&0.739&19.827
 \\
& Isolation &0.917&229.720&0.982&7.165 &0.905&100.900&0.856&7.853 &0.908&97.650&0.786&5.642 &0.914&138.229&0.812&12.221 \\
& Laplacian &0.465&5365.724&0.620&10.796 &0.388&13.676&3.469&26.740 &0.001&6.925&15.816&4.828 &0.480&13.847&0.998&16.371 
 \\
& Linear &0.465&5445.396&0.619&7.775 &0.780&61422.728&0.472&9.831 &0.730&48515.324&0.527&12.310&0.775&60533.842&0.472&6.499
 \\
& Polynomial &0.465&5380.650&0.620&8.783 &0.779&59049.256&0.476&2.683 &0.730&48101.028&0.534&2.831 &0.775&52560.712&0.464&3.371
 \\
& Sigmoid &0.465&5515.529&0.618&6.184 &0.684&22545.943&0.727&24.429&-0.001&10.290&22.450&2.487 &-0.002&10.827&20.230&2.433 
 \\
    
    \midrule
    \multirow{9}{*}{$k$-modes} 
& Additive-chi2 &-0.447&0.990&1.181&2527.511
&-0.541&0.408&1.367&1283.833 &0.653&19207.977&0.757&26.296 &-0.063&0.752&1.092&1652.220 
 \\
& Chi-squared &-0.679&0.045&9.109&5811.840 &-0.415&0.622&2.556&7327.263 &-0.430&1.301&2.110&5887.333
 &-0.459&0.753&2.882&5782.701
 \\
& Cosine &-0.185&0.400&1.212&3207.832
 &-0.863&1.511&3.131&4616.545 &-0.054&0.330&1.853&1783.393&-0.834&0.409&2.012&3943.872
 \\
& Gaussian &-0.076&1.401&56.106&5515.107 &-0.397&0.855&1.595&7077.646 &-0.434&0.343&2.055&5302.751
 &-0.422&0.450&2.254&5336.310
 \\
& Isolation &-0.563&11.948&6.255&8625.942
&-0.363&3.337&9.337&8424.060 &-0.395&3.563&9.617&7149.288 &-0.253&2.693&8.415&30170.856
 \\
& Laplacian &-0.395&0.578&1.028&2823.894 &-0.440&1.871&2.335&6947.622
 &-0.036&1.017&0.997&3075.104
 &-0.451&0.584&2.168&4978.428 
 \\
& Linear &-0.351&0.741&1.021&2620.750 &-0.703&5.002&1.383&3188.894
 &-0.578&0.312&1.320&1527.236
&-0.702&1.577&1.097&1365.664 \\
& Polynomial &-0.469&0.507&1.150&2557.172 &-0.808&0.360&1.922&3272.108 &-0.739&0.228&1.899&1538.851 &-0.786&0.218&1.724&1347.649
 \\
& Sigmoid &-0.420&0.514&1.106&2509.769 &0.178&2.155&0.669&2133.886 &-0.146&0.552&1.344&880.223 &-0.145&0.566&1.333&850.938 \\
    \bottomrule
  \end{tabular}
  }
  \caption{Internal clustering quality metrics for different kernels and embeddings on \textbf{ShortRead dataset}. }
  \label{tbl_shortRead_clustering}
\end{table}

Table~\ref{tbl_rabies_clustering} demonstrates that the Polynomial kernel excels with $k$-means clustering across Spike2Vec, Minimizer, and Spaced k-mer embeddings, achieving exceptionally high Silhouette Coefficients (0.985-0.996) and Calinski-Harabasz scores reaching beyond 600,000 for Minimizer encoding. The isolation kernel shows good performance across different embeddings with efficient computation times. For $k$-modes clustering, performance varies significantly across configurations, with chi-squared kernel and Minimizer encoding achieving the highest Calinski-Harabasz score (91.468) and a positive Silhouette Coefficient (0.301). The Sigmoid kernel with Spaced k-mer encoding produces an exceptionally high Calinski-Harabasz score (2926.083) for $k$-modes, suggesting potentially well-separated clusters, though this requires careful interpretation given the unusual magnitude.

\begin{table}[h!]
  \centering
     \resizebox{0.99\textwidth}{!}{
\begin{tabular}{@{\extracolsep{6pt}}ccp{1.2cm}p{1.5cm}p{0.8cm}p{1.2cm}p{1.2cm}p{1.6cm}p{0.8cm}p{1.2cm}p{1.2cm}p{1.5cm}p{0.8cm}p{1.2cm}p{1.2cm}p{1.5cm}p{0.8cm}p{1.2cm}}
    \toprule
     & & \multicolumn{4}{c}{OHE}  & \multicolumn{4}{c}{Spike2vec} & \multicolumn{4}{c}{Minimizer}& \multicolumn{4}{c}{Spaced k-mer} \\
    \cmidrule{3-6} \cmidrule{7-10} \cmidrule{11-14} \cmidrule{15-18}
    Algo. & Kernel & Silhouette Coefficient & Calinski Harabasz Score & Davies-Bouldin Score & Clustering Runtime (Sec.) & Silhouette Coefficient & Calinski Harabasz Score & Davies-Bouldin Score & Clustering Runtime (Sec.) & Silhouette Coefficient & Calinski Harabasz Score & Davies-Bouldin Score & Clustering Runtime (Sec.) & Silhouette Coefficient & Calinski Harabasz Score & Davies-Bouldin Score & Clustering Runtime (Sec.)\\
    \midrule	\midrule	
    \multirow{9}{*}{$k$-means} 
& Additive-chi2 &0.187&4002.000&1.200&31.988&0.577&112867.481&0.439&49.925 &0.533&89547.624&0.512&15.939&0.346&46651.675&1.070&90.265
 \\
& Chi-squared&0.531&216.896&1.249&4.319 &0.044&210.251&1.662&62.677 &0.209&222.433&1.281&121.588 &0.492&410.654&0.382&11.122 
 \\
& Cosine &0.170&2565.670&2.204&21.132 &0.235&2711.211&1.586&29.112&0.227&3507.135&1.474&252.261&0.205&2112.364&1.691&180.789 
 \\
& Gaussian &0.646&459.827&0.241&8.027 &0.646&464.056&0.241&38.197&0.603&492.633&0.274&67.811 &0.644&494.341&0.243&75.072
 \\
& Isolation &0.939&275.123&0.568&10.977 &0.957&249.999&0.507&7.279 &0.955&259.133&0.520&6.612 &0.944&288.276&0.515&12.152 \\
& Laplacian &0.238&3430.153&1.357&27.947&0.457&31070.489&0.693&109.818 &0.087&1250.415&3.408&179.241&0.205&3378.445&1.897&165.092
 \\
& Linear &0.345&3528.133&1.145&24.403 &0.633&316957.691&0.441&37.924&0.639&432955.835&0.410&56.297 &0.911&124880.561&0.344&85.801
 \\
& Polynomial &0.656&5447.339&1.207&4.110 &0.986&325677.263&0.382&40.584 &0.985&624187.091&0.122&50.107&0.996&193683.606&0.197&33.559
 \\
& Sigmoid &0.318&3768.843&1.087&13.041
 &0.333&13727.034&1.459&50.409 &0.999&27151.997&0.278&50.842&-0.474&12.827&1.508&8012.191 
 \\
    
    \midrule
    \multirow{9}{*}{$k$-modes} 
& Additive-chi2 &-0.401&10.868&1.228&13098.226 &-0.460&18.621&1.426&6222.205
&-0.538&1.883&1.900&6769.886&-0.361&8.152&1.351&6410.721
 \\
& Chi-squared &-0.635&0.685&15.698&9785.699 &-0.076&28.379&0.910&5776.951 &0.301&91.468&0.534&11234.866&-0.608&0.203&16.393&3556.649
 \\
& Cosine &0.024&92.713&0.931&6019.710 &-0.348&16.788&1.783&12091.892&-0.302&6.576&1.151&8389.929 &-0.182&64.033&0.889&13641.548 
 \\
& Gaussian &-0.559&3.748&16.298&11464.403
&-0.516&3.515&16.536&12086.981
&-0.536&3.192&16.392&16892.643&-0.373&6.566&15.808&20405.049
\\
& Isolation &-0.443&4.339&9.533&35971.307 &-0.530&2.534&9.768&13983.651 &-0.454&3.383&9.747&7021.328 &-0.719&1.412&9.601&48983.839 \\
& Laplacian &-0.397&5.838&1.298&12115.963 &-0.404&2.864&1.284&6018.343
&-0.033&62.789&0.982&6854.140&-0.062&18.401&1.032&10257.529 
 \\
& Linear &-0.379&4.600&1.434&10764.303&-0.726&2.249&1.962&6138.794&-0.750&7.149&1.213&5767.488&-0.565&18.341&1.430&4496.649\\
& Polynomial &-0.232&5.180&1.361&2447.411
 &-0.886&0.589&1.851&6659.856 &-0.971&7.111&1.258&6167.346
 &-0.891&0.040&2.079&3939.441 
 \\
& Sigmoid &-0.380&4.953&1.219&6420.527 &-0.141&34.853&0.895&4379.701 &-0.758&0.036&6.288&6125.163 &0.383&2926.083&1.892&108.023  \\
    \bottomrule
  \end{tabular}
  }
  \caption{Internal clustering quality metrics for different kernels and embeddings on \textbf{Rabies dataset}. }
  \label{tbl_rabies_clustering}
\end{table}

For the Genome dataset (Table~\ref{tbl_genome_clustering}), the isolation kernel outperforms other methods in $k$-means clustering, achieving Silhouette Coefficients above 0.89 across all embeddings with minimal computational cost. The Cosine and Sigmoid kernels with Spaced k-mer encoding also demonstrate strong clustering quality (Silhouette: 0.886 and Davies-Bouldin: 0.333). The $k$-modes results show mixed performance, with most configurations yielding negative Silhouette Coefficients. Notably, the Laplacian kernel with Spike2Vec encoding achieves a positive Silhouette (0.137) and competitive Davies-Bouldin score (0.722), while chi-squared kernel with Spaced k-mer produces the highest Silhouette (0.056) among $k$-modes configurations. The additive-chi2 kernel again shows identical values between $k$-means and $k$-modes for Minimizer embedding.

\begin{table}[h!]
  \centering
     \resizebox{0.99\textwidth}{!}{
 \begin{tabular}{@{\extracolsep{6pt}}ccp{1.2cm}p{1.5cm}p{0.8cm}p{1.2cm}p{1.2cm}p{1.6cm}p{0.8cm}p{1.2cm}p{1.2cm}p{1.5cm}p{0.8cm}p{1.2cm}p{1.2cm}p{1.5cm}p{0.8cm}p{1.2cm}}
    \toprule
     & & \multicolumn{4}{c}{OHE}  & \multicolumn{4}{c}{Spike2vec} & \multicolumn{4}{c}{Minimizer}& \multicolumn{4}{c}{Spaced k-mer} \\
    \cmidrule{3-6} \cmidrule{7-10} \cmidrule{11-14} \cmidrule{15-18}
    Algo. & Kernel & Silhouette Coefficient & Calinski Harabasz Score & Davies-Bouldin Score & Clustering Runtime (Sec.) & Silhouette Coefficient & Calinski Harabasz Score & Davies-Bouldin Score & Clustering Runtime (Sec.) & Silhouette Coefficient & Calinski Harabasz Score & Davies-Bouldin Score & Clustering Runtime (Sec.) & Silhouette Coefficient & Calinski Harabasz Score & Davies-Bouldin Score & Clustering Runtime (Sec.)\\
    \midrule	\midrule	
    \multirow{9}{*}{$k$-means} 
& Additive-chi2 &0.701&1981.100&1.306&1.824 &0.667&20328.402&0.751&51.845 &0.594&13106.837&0.863&107.852&0.824&65184.258&0.446&3.708 
 \\
& Chi-squared &0.430&952.541&1.280&1.969 &0.423&4474.053&1.104&41.280 &0.360&3518.464&1.238&72.189 &0.780&53495.827&0.304&3.029 
 \\
& Cosine &0.754&1741.799&1.620&3.593&0.617&48306.107&0.783&51.435 &0.540&20457.328&1.034&94.370&0.886&196623.437&0.333&3.559 
 \\
& Gaussian &0.719&35.720&3.050&5.708 &0.699&33.868&0.202&99.454 &0.689&34.808&0.208&58.513&0.788&96.620&0.128&4.522
 \\
& Isolation &0.902&98.655&0.565&2.112 &0.915&95.627&0.635&1.919 &0.903&94.709&0.531&4.025&0.894&118.551&0.672&2.219 \\
& Laplacian &0.757&1775.133&1.475&5.406 &0.136&1157.928&2.443&74.059 &0.117&955.283&2.690&121.627&0.581&31109.936&0.676&6.853 
 \\
& Linear &0.694&1913.539&1.810&4.883 &0.751&48302.567&0.566&70.012 &0.572&21631.400&0.801&83.521 &0.645&23308.594&0.532&9.783 
 \\
& Polynomial &0.757&1772.842&1.321&3.014 &0.733&40629.911&0.596&67.203&0.406&15132.253&1.246&113.842 &0.645&23442.238&0.530&5.150
 \\
& Sigmoid &0.758&1787.739&1.274&2.085 &0.001&10.775&22.667&5.668 &0.001&10.758&22.784&5.813 &0.001&10.724&22.771&7.329 
 \\
    
    \midrule
    \multirow{9}{*}{$k$-modes} 
& Additive-chi2 &-0.585&1.224&1.085&3684.682 &-0.469&0.603&1.541&2881.500 &0.594&13106.837&0.863&107.852 &-0.658&0.829&1.511&1504.120
 \\
& Chi-squared &-0.455&1.767&3.230&3366.333 &-0.045&1.084&0.953&2918.957&-0.235&3.929&0.969&4984.650 &0.056&6.830&0.739&1313.927 
 \\
& Cosine &-0.532&1.398&0.996&5602.487&-0.683&1.754&1.275&6654.192&-0.375&0.399&1.287&5611.547&-0.733&0.773&1.575&2734.444
 \\
& Gaussian &-0.278&0.818&26.718&15240.669 &-0.281&1.065&25.428&36842.879&-0.184&1.107&36.419&9057.544&-0.417&44.183&1.712&5243.214
 \\
& Isolation &-0.271&4.017&9.406&3988.195 &-0.457&3.005&9.419&9011.211
 &-0.349&2.357&9.342&24640.874 &-0.287&5.380&8.711&9811.822 \\
& Laplacian &-0.608&1.019&1.070&5092.798&0.137&6.494&0.722&10395.885 &-0.062&1.305&1.064&3796.323&-0.477&2.187&2.074&1199.087
 \\
& Linear &-0.439&1.203&1.086&3655.515
&-0.562&0.272&1.489&10262.275
&-0.436&0.796&1.417&3495.170 &-0.547&1.368&1.467&1878.157
 \\
& Polynomial &-0.628&1.155&1.003&5334.133&-0.661&2.377&1.087&8678.394&-0.569&1.783&1.944&3501.511&-0.591&4.795&1.122&1945.527
 \\
& Sigmoid&-0.597&1.096&1.028&5956.777 &-0.039&1.037&1.022&1130.992 &-0.026&0.941&1.028&987.253 &-0.029&0.942&1.035&1010.187 
\\
    \bottomrule
  \end{tabular}
  }
  \caption{Internal clustering quality metrics for different kernels and embeddings on \textbf{Genome dataset}. }
  \label{tbl_genome_clustering}
\end{table}

The Breast Cancer dataset (Table~\ref{tbl_BreastCancer_clustering}) exhibits moderate clustering performance overall, with the isolation kernel providing the best $k$-means results across all embeddings (Silhouette Coefficients: 0.325-0.659), despite relatively modest Calinski-Harabasz scores. The Sigmoid kernel with Minimizer encoding achieves the highest Silhouette Coefficient (0.267) among other kernels, combined with favorable Davies-Bouldin scores. For $k$-modes clustering, the Cosine kernel with Spaced k-mer encoding yields the best Silhouette Coefficient (0.027) and highest Davies-Bouldin score (11.379), though values remain low overall. The Polynomial kernel with Spike2Vec shows slight improvement (Silhouette: 0.011), while most other configurations produce negative Silhouette values, indicating challenging cluster structure in this dataset regardless of the chosen embedding method.

\begin{table}[h!]
  \centering
     \resizebox{0.99\textwidth}{!}{
\begin{tabular}{@{\extracolsep{6pt}}ccp{1.2cm}p{1.5cm}p{0.8cm}p{1.2cm}p{1.2cm}p{1.6cm}p{0.8cm}p{1.2cm}p{1.2cm}p{1.5cm}p{0.8cm}p{1.2cm}p{1.2cm}p{1.5cm}p{0.8cm}p{1.2cm}}
    \toprule
     & & \multicolumn{4}{c}{OHE}  & \multicolumn{4}{c}{Spike2vec} & \multicolumn{4}{c}{Minimizer}& \multicolumn{4}{c}{Spaced k-mer} \\
    \cmidrule{3-6} \cmidrule{7-10} \cmidrule{11-14} \cmidrule{15-18}
    Algo. & Kernel & Silhouette Coefficient & Calinski Harabasz Score & Davies-Bouldin Score & Clustering Runtime (Sec.) & Silhouette Coefficient & Calinski Harabasz Score & Davies-Bouldin Score & Clustering Runtime (Sec.) & Silhouette Coefficient & Calinski Harabasz Score & Davies-Bouldin Score & Clustering Runtime (Sec.) & Silhouette Coefficient & Calinski Harabasz Score & Davies-Bouldin Score & Clustering Runtime (Sec.)\\
    \midrule	\midrule	
    \multirow{9}{*}{$k$-means} 
& Additive-chi2 &0.019&17.447&5.123&0.487&0.013&23.777&2.207&0.144 &0.050&29.455&1.915&0.165&0.066&15.256&1.547&0.112 
 \\
& Chi-squared&-0.055&2.450&3.136&0.458 &-0.056&2.347&5.818&0.945&0.023&12.428&2.763&0.274&0.128&7.764&1.224&0.081
 \\
& Cosine &0.015&18.778&5.641&0.824 &0.047&30.859&2.269&0.229 &0.052&34.805&2.442&0.142&0.079&20.133&2.088&0.181 
 \\
& Gaussian &0.065&1.462&0.816&0.323 &-0.005&1.571&0.962&0.559 &0.028&3.028&0.870&0.137&0.113&4.578&0.744&0.545
 \\
& Isolation &0.581&28.520&0.927&0.350 &0.659&36.862&1.170&0.116&0.438&23.250&0.818&0.125&0.325&75.683&0.998&0.414 \\
& Laplacian &0.012&16.594&6.836&1.046 &0.025&24.121&2.432&0.139&0.059&26.529&2.213&0.119&0.146&14.099&1.819&0.124
 \\
& Linear &0.014&17.142&6.123&0.558 &0.110&38.117&2.255&0.534&0.119&29.624&2.340&0.520&0.074&26.299&1.653&0.121
 \\
& Polynomial &0.019&19.665&7.158&0.748 &0.179&40.006&1.823&0.165&0.099&40.198&1.527&0.121 &0.242&19.632&1.052&0.121 
 \\
& Sigmoid &-0.056&13.129&7.465&0.561 &0.149&41.463&1.755&0.609 &0.267&49.654&1.182&0.391 &0.085&26.031&1.440&0.403 
 \\
    
    \midrule
    \multirow{9}{*}{$k$-modes} 
& Additive-chi2 &-0.077&1.226&0.987&150.826 &-0.156&0.910&1.256&106.128 &-0.318&1.201&6.193&106.282 &-0.271&0.176&14.237&125.576 
 \\
& Chi-squared &-0.090&1.109&0.982&608.331 &-0.127&0.984&1.047&655.013 &-0.286&0.710&2.573&159.327 &-0.352&0.851&4.461&210.192
 \\
& Cosine &-0.044&0.985&1.017&163.095 &0.001&1.082&0.966&130.729 &-0.019&1.104&5.278&117.911 &0.027&1.464&11.379&230.816 
 \\
& Gaussian &-0.194&0.976&1.111&493.225 &-0.230&1.052&1.125&464.578&-0.003&1.262&21.973&257.913 &-0.139&1.077&10.865&848.657
 \\
& Isolation &-0.344&2.699&5.046&431.237&-0.389&2.912&3.908&264.000 &-0.381&1.396&5.388&982.025 &-0.464&2.410&2.892&1188.255 \\
& Laplacian &-0.106&0.781&1.159&157.846 &-0.147&1.649&1.141&132.096 &-0.214&0.511&1.557&127.406 &-0.439&1.611&9.283&112.490
 \\
& Linear &-0.014&1.109&0.960&151.999 &-0.171&0.966&1.276&127.811&-0.351&0.381&4.813&216.572&-0.347&0.561&12.109&144.863 \\
& Polynomial &-0.033&1.004&1.017&104.546 &0.011&1.102&0.931&117.900&-0.273&0.208&2.111&112.545&-0.437&0.547&9.841&113.135 
 \\
& Sigmoid &-0.163&0.657&1.353&104.484 &-0.113&0.731&1.184&114.074 &-0.153&0.727&1.208&103.679&-0.397&0.765&3.647&100.370 
 \\
    \bottomrule
  \end{tabular}
  }
  \caption{Internal clustering quality metrics for different kernels and embeddings on \textbf{Breast Cancer dataset}. }
  \label{tbl_BreastCancer_clustering}
\end{table}

\clearpage

\section{Classification Results}
In this section, we report the classification results for different kernels, embeddings, classifiers, and datasets.


\subsection{Results for Spike7k Data}


Table~\ref{tbl_variant_classification_ohe} presents the classification performance on the Spike7k dataset using OHE encoding across various classifiers and evaluation metrics. The classification objective involves categorizing SARS-CoV-2 viral sequences into $22$ distinct lineages (class distribution detailed in Section~\ref{subsection_data_stats}). Results indicate that the linear kernel combined with random forest achieves superior performance across nearly all metrics, with the exception of ROC-AUC, where the polynomial kernel paired with KNN demonstrates the best results.
The classification outcomes for $k$-mers-based representations are detailed in Table~\ref{tbl_variant_classification_kmers}, while minimizer-based embedding results appear in Table~\ref{tbl_variant_classification_minimizer}. These results reveal that linear and cosine kernels, when combined with either logistic regression or random forest classifiers, yield the strongest predictive capabilities compared to alternative kernel-classifier combinations. For spaced k-mers embeddings, Table~\ref{tbl_variant_classification_spike7k_spaced_kmer} shows that the Additive-chi2 kernel with logistic regression attains the peak accuracy of 85.7\%.

\begin{table}[h!]
  \resizebox{0.99\textwidth}{!}{
  \begin{tabular}{p{1.7cm}cccccp{1.2cm} p{1.5cm} | p{1.7cm}cccccp{1.2cm} p{1.5cm}}
    \toprule
   
    \multirow{2}{*}{Kernels} & 
    \multirow{2}{0.7cm}{ML Algo.} & \multirow{2}{*}{Acc.} & \multirow{2}{*}{Prec.} & \multirow{2}{*}{Recall} & \multirow{2}{0.9cm}{F1 weigh.}  & ROC- AUC & Train. runtime (sec.) 
    &
    \multirow{2}{*}{Kernels} & 
    \multirow{2}{0.7cm}{ML Algo.} & \multirow{2}{*}{Acc.} & \multirow{2}{*}{Prec.} & \multirow{2}{*}{Recall} & \multirow{2}{0.9cm}{F1 weigh.} &  ROC- AUC & Train. runtime (sec.)
    \\	
    \midrule	\midrule
    
    \multirow{7}{2cm}{Additive-chi2}
 & SVM & 0.786 & 0.773 & 0.786 & 0.763 &  0.735 & 0.280 & \multirow{7}{2cm}{Laplacian} 
 & SVM & 0.614 & 0.536 & 0.614 & 0.554 &  0.555 & 0.312 \\
 & NB & 0.700 & 0.763 & 0.700 & 0.717 &   0.755 & 0.408 & &
 NB & 0.637 & 0.698 & 0.637 & 0.640 &   0.743 & 0.322 \\
 & MLP & 0.764 & 0.761 & 0.764 & 0.757 &   0.753 & 2.568 & &
 MLP & 0.767 & 0.746 & 0.767 & 0.749 &   0.762 & 2.077 \\
 & KNN & 0.790 & 0.792 & 0.790 & 0.781 &   0.771 & 0.410 & &
 KNN & 0.792 & 0.780 & 0.792 & 0.778 &   0.761 & 0.403 \\
 & RF & 0.799 & 0.799 & 0.799 & 0.792 &   0.790 & 0.509  & &
 RF & 0.803 & 0.776 & 0.803 & 0.778 &   0.786 & 0.504 \\
 & LR & 0.772 & 0.757 & 0.772 & 0.750 &   0.701 & 0.356 & &
 LR & 0.579 & 0.350 & 0.579 & 0.436 &   0.526 & 0.261 \\
 & DT & 0.767 & 0.773 & 0.767 & 0.763 &   0.766 & 0.219 & &
 DT & 0.776 & 0.753 & 0.776 & 0.758 &   0.769 & 0.254 \\
\midrule
\multirow{7}{2cm}{Chi-squared}
 & SVM & 0.676 & 0.631 & 0.676 & 0.610 &   0.656 & 0.294 &
 \multirow{7}{2cm}{Linear}
 & SVM & 0.800 & 0.767 & 0.800 & 0.772 &  0.748 & 0.290 \\
 & NB & 0.532 & 0.758 & 0.532 & 0.604 &   0.731 & 0.325 & &
 NB & 0.692 & 0.780 & 0.692 & 0.717 &  0.776 & 0.333 \\
 & MLP & 0.720 & 0.685 & 0.720 & 0.677 &   0.718 & 1.826 & &
 MLP & 0.771 & 0.772 & 0.771 & 0.767 &   0.756 & 2.251 \\
 & KNN & 0.705 & 0.708 & 0.705 & 0.699 &   0.714 & 0.387 &  & 
 KNN & 0.802 & 0.807 & 0.802 & 0.800 &   0.780 & 0.386 \\
 & RF & 0.723 & 0.698 & 0.723 & 0.692 &   0.724 & 0.477 & & 
 RF & \textbf{0.817} & \textbf{0.809} & \textbf{0.817} & \textbf{0.808}   & 0.795 & 0.461 \\
 & LR & 0.658 & 0.577 & 0.658 & 0.577 &   0.605 & 0.233 & & 
 LR & 0.791 & 0.765 & 0.791 & 0.764 &   0.725 & 0.306 \\
 & DT & 0.716 & 0.689 & 0.716 & 0.689 &  0.726 & 0.230 & & 
 DT & 0.789 & 0.787 & 0.789 & 0.781 &   0.783 & 0.215 \\
\midrule
\multirow{7}{2cm}{Cosine}
 & SVM & 0.653 & 0.630 & 0.653 & 0.625 &   0.586 & 0.301 &
 \multirow{7}{2cm}{Polynomial} 
 & SVM & 0.612 & 0.537 & 0.612 & 0.554 &  0.555 & 0.308 \\
 & NB & 0.697 & 0.728 & 0.697 & 0.700 &   0.751 & 0.378 &  & 
 NB & 0.707 & 0.769 & 0.707 & 0.726 &  0.765 & 0.396 \\
 & MLP & 0.775 & 0.760 & 0.775 & 0.760 &  0.762 & 2.149 & & 
 MLP & 0.776 & 0.760 & 0.776 & 0.759 &  0.764 & 2.272 \\
 & KNN & 0.787 & 0.778 & 0.787 & 0.774 &   0.753 & 0.386 & & 
 KNN & 0.801 & 0.799 & 0.801 & 0.788 &  \textbf{0.796} & 0.407 \\
 & RF & 0.803 & 0.802 & 0.803 & 0.795 &   0.787 & 0.468 & & 
 RF & 0.806 & 0.794 & 0.806 & 0.785 &  0.788 & 0.510 \\
 & LR & 0.638 & 0.610 & 0.638 & 0.610 &   0.563 & 0.238 &  & 
 LR & 0.611 & 0.520 & 0.611 & 0.546 &   0.554 & 0.231 \\
 & DT & 0.770 & 0.778 & 0.770 & 0.767 &  0.774 & 0.214 & & 
 DT & 0.770 & 0.762 & 0.770 & 0.761 &   0.770 & 0.228 \\
\midrule
\multirow{7}{2cm}{Isolation}
 & SVM & 0.483 & 0.233 & 0.483 & 0.314 &  0.500 & 0.328 & 
 \multirow{7}{2cm}{Sigmoid} 
 & SVM & 0.582 & 0.354 & 0.582 & 0.440 &   0.526 & 0.314 \\
 & NB & 0.747 & 0.770 & 0.747 & 0.749 &   0.768 & 0.303 & & 
 NB & 0.701 & 0.757 & 0.701 & 0.712 &  0.731 & 0.278 \\
 & MLP & 0.786 & 0.764 & 0.786 & 0.764 &  0.750 & 2.232 & & 
 MLP & 0.788 & 0.764 & 0.788 & 0.769 &   0.773 & 1.685 \\
 & KNN & 0.795 & 0.773 & 0.795 & 0.776 &   0.760 & 0.388 & & 
 KNN & 0.791 & 0.766 & 0.791 & 0.774 &   0.736 & 0.384 \\
 & RF & 0.797 & 0.791 & 0.797 & 0.784 &  0.766 & 0.530 &  & 
 RF & 0.805 & 0.783 & 0.805 & 0.781 &   0.780 & 0.497 \\
 & LR & 0.483 & 0.233 & 0.483 & 0.314 &   0.500 & \textbf{0.212} &  & 
 LR & 0.481 & 0.232 & 0.481 & 0.313 &   0.500 & 0.227 \\
 & DT & 0.770 & 0.772 & 0.770 & 0.767 &   0.759 & 0.217 &  & 
 DT & 0.784 & 0.772 & 0.784 & 0.771 &   0.767 & 0.248 \\
\midrule

\multirow{7}{2cm}{Gaussian}
 & SVM & 0.479 & 0.229 & 0.479 & 0.310 &   0.500 & 0.381 &
 
 &  &  &  &  &  &  &  \\
 & NB & 0.528 & 0.718 & 0.528 & 0.577 &   0.764 & 0.495 &  & 
   &  &  &  &  &  &  \\
 & MLP & 0.780 & 0.788 & 0.780 & 0.778 &   0.762 & 3.272 &  & 
   &  &  &  &  &  &  \\
 & KNN & 0.773 & 0.781 & 0.773 & 0.773 &   0.736 & 0.580 &  & 
   &  &  &  &  &  &  \\
 & RF & 0.777 & 0.778 & 0.777 & 0.774 &   0.755 & 0.632 &  & 
   &  &  &  &  &  &  \\
 & LR & 0.479 & 0.229 & 0.479 & 0.310 &   0.500 & 0.239 &   &
   &  &  &  &  &  &  \\
 & DT & 0.743 & 0.749 & 0.743 & 0.743 &   0.720 & 0.249 &  &  
   &  &  &  &  &  &  \\
\bottomrule

  \end{tabular}
  }
  \caption{Classification comparing for different kernel methods and classifiers on \textbf{OHE-based embeddings for Spike7k dataset}. The best values are shown in bold.
}
  \label{tbl_variant_classification_ohe}
\end{table}

\begin{table}[h!]
  \centering
  \resizebox{0.99\textwidth}{!}{
  \begin{tabular}{p{1.7cm}cccccp{1.2cm} p{1.5cm} | p{1.7cm}cccccp{1.2cm} p{1.5cm}}
    \toprule
   
    \multirow{2}{*}{Kernels} & 
    \multirow{2}{0.7cm}{ML Algo.} & \multirow{2}{*}{Acc.} & \multirow{2}{*}{Prec.} & \multirow{2}{*}{Recall} & \multirow{2}{0.9cm}{F1 weigh.} &  ROC- AUC & Train. runtime (sec.) 
    &
    \multirow{2}{*}{Kernels} & 
    \multirow{2}{0.7cm}{ML Algo.} & \multirow{2}{*}{Acc.} & \multirow{2}{*}{Prec.} & \multirow{2}{*}{Recall} & \multirow{2}{0.9cm}{F1 weigh.} &   ROC- AUC & Train. runtime (sec.)
    \\	
    \midrule	\midrule
\multirow{7}{2cm}{Additive-chi2}
 & SVM & 0.835 & 0.811 & 0.835 & 0.817 &  0.802 & 0.289 & \multirow{7}{2cm}{Laplacian}
 & SVM & 0.606 & 0.425 & 0.606 & 0.467 &   0.530 & 0.355 \\
 & NB & 0.719 & 0.788 & 0.719 & 0.744 &   0.776 & 0.336 &  & 
 NB & 0.680 & 0.735 & 0.680 & 0.696 &   0.726 & 0.338 \\
 & MLP & 0.786 & 0.797 & 0.786 & 0.780 &  0.781 & 1.859 &  & 
 MLP & 0.780 & 0.770 & 0.780 & 0.771 &   0.765 & 1.958 \\
 & KNN & 0.800 & 0.798 & 0.800 & 0.793 &   0.757 & 0.407 &  & 
 KNN & 0.793 & 0.780 & 0.793 & 0.782 &   0.724 & 0.395 \\
 & RF & 0.820 & 0.813 & 0.820 & 0.806 &   0.794 & 0.483 &   & 
 RF & 0.811 & 0.792 & 0.811 & 0.794 &   0.781 & 0.475 \\
 & LR & \textbf{0.838} & 0.810 & \textbf{0.838} & 0.816   & 0.799 & 0.308  &  & 
 LR & 0.485 & 0.235 & 0.485 & 0.317 &   0.500 & 0.220 \\
 & DT & 0.790 & 0.794 & 0.790 & 0.785 &   0.776 & 0.216  &   & 
 DT & 0.784 & 0.778 & 0.784 & 0.777 &   0.762 & 0.217 \\
\midrule
\multirow{7}{2cm}{Chi-squared}
 & SVM & 0.653 & 0.610 & 0.653 & 0.583 &   0.639 & 0.325 & \multirow{7}{2cm}{Linear}
 & SVM & 0.835 & \textbf{0.818} & 0.835 & \textbf{0.819} &   \textbf{0.812} & 0.340 \\
 & NB & 0.480 & 0.735 & 0.480 & 0.558 &   0.684 & 0.340 &  & 
 NB & 0.730 & 0.810 & 0.730 & 0.761 &   0.793 & 0.420 \\
 & MLP & 0.718 & 0.682 & 0.718 & 0.672 &   0.695 & 1.785 &   & 
 MLP & 0.769 & 0.761 & 0.769 & 0.759 &   0.780 & 1.986 \\
 & KNN & 0.715 & 0.715 & 0.715 & 0.707 &   0.710 & 0.419  &   & 
 KNN & 0.768 & 0.767 & 0.768 & 0.765 &  0.766 & 0.518 \\
 & RF & 0.735 & 0.707 & 0.735 & 0.706 &   0.718 & 0.591 &   & 
 RF & 0.811 & 0.802 & 0.811 & 0.800 &   0.793 & 0.555 \\
 & LR & 0.636 & 0.568 & 0.636 & 0.556 &  0.599 & 0.293 &   & 
 LR & 0.838 & 0.815 & 0.838 & 0.817 &   0.804 & 0.322 \\
 & DT & 0.722 & 0.693 & 0.722 & 0.698 &   0.718 & 0.243 &   & 
 DT & 0.776 & 0.784 & 0.776 & 0.773 &  0.776 & 0.221 \\
\midrule
\multirow{7}{2cm}{Cosine}
 & SVM & 0.619 & 0.509 & 0.619 & 0.545 &   0.544 & 0.329  & \multirow{7}{2cm}{Polynomial}
 & SVM & 0.636 & 0.492 & 0.636 & 0.522 &   0.542 & 0.343 \\
 & NB & 0.694 & 0.778 & 0.694 & 0.725 &  0.749 & 0.341 &   & 
 NB & 0.687 & 0.725 & 0.687 & 0.685 &   0.705 & 0.355 \\
 & MLP & 0.740 & 0.746 & 0.740 & 0.736 &   0.749 & 1.879 &   & 
 MLP & 0.805 & 0.776 & 0.805 & 0.781 &   0.779 & 2.389 \\
 & KNN & 0.787 & 0.804 & 0.787 & 0.780 &  0.747 & 0.399 &   & 
 KNN & 0.762 & 0.757 & 0.762 & 0.752 &   0.755 & 0.411 \\
 & RF & 0.819 & 0.815 & 0.819 & 0.802 &   0.787 & 0.536 &   & 
 RF & 0.818 & 0.795 & 0.818 & 0.793 &   0.780 & 0.485 \\
 & LR & 0.631 & 0.436 & 0.631 & 0.514 &  0.540 & 0.235 &   & 
 LR & 0.523 & 0.314 & 0.523 & 0.378 &  0.515 & 0.249 \\
 & DT & 0.783 & 0.775 & 0.783 & 0.774 &   0.769 & 0.235 &   & 
 DT & 0.781 & 0.768 & 0.781 & 0.770 &   0.756 & 0.215 \\
\midrule

\multirow{7}{2cm}{Isolation}
 & SVM & 0.481 & 0.232 & 0.481 & 0.313 &   0.500 & 0.330 & \multirow{7}{2cm}{Sigmoid}
 & SVM & 0.481 & 0.231 & 0.481 & 0.312 &  0.500 & 0.320 \\
 & NB & 0.614 & 0.718 & 0.614 & 0.639 &   0.702 & 0.432 &   & 
 NB & 0.703 & 0.779 & 0.703 & 0.727 &   0.766 & 0.331 \\
 & MLP & 0.764 & 0.739 & 0.764 & 0.747 &  0.730 & 2.628 &  & 
 MLP & 0.772 & 0.767 & 0.772 & 0.762 &  0.762 & 1.987 \\
 & KNN & 0.753 & 0.721 & 0.753 & 0.735 &   0.709 & 0.421 &   & 
 KNN & 0.773 & 0.772 & 0.773 & 0.769 &   0.735 & 0.391 \\
 & RF & 0.758 & 0.737 & 0.758 & 0.745 &   0.722 & 0.525 &  & 
 RF & 0.813 & 0.788 & 0.813 & 0.792 &   0.775 & 0.486 \\
 & LR & 0.481 & 0.232 & 0.481 & 0.313 &   0.500 & 0.214 &  & 
 LR & 0.481 & 0.231 & 0.481 & 0.312 &   0.500 & 0.218 \\
 & DT & 0.732 & 0.723 & 0.732 & 0.725 &  0.700 & 0.224 &  & 
 DT & 0.770 & 0.773 & 0.770 & 0.766 &   0.751 & \textbf{0.210} \\
\midrule

\multirow{7}{2cm}{Gaussian}
 & SVM & 0.481 & 0.231 & 0.481 & 0.312 &   0.500 & 0.376 & 
 \\
 & NB & 0.500 & 0.697 & 0.500 & 0.553 &   0.732 & 0.403 &  & 
 \\
 & MLP & 0.732 & 0.718 & 0.732 & 0.719 &   0.742 & 2.594 &   & 
 \\
 & KNN & 0.747 & 0.734 & 0.747 & 0.733 &   0.749 & 0.454 &   & 
 \\
 & RF & 0.747 & 0.733 & 0.747 & 0.737 &   0.747 & 0.636 &   &
 \\
 & LR & 0.481 & 0.231 & 0.481 & 0.312 &   0.500 & 0.253 &   & 
 \\
 & DT & 0.715 & 0.725 & 0.715 & 0.717 &   0.734 & 0.263 &    & 
 \\
\bottomrule

  \end{tabular}
  }
  \caption{Classification comparison for different kernel methods and classifiers on \textbf{$k$-mer-based embeddings for Spike7k dataset}. The best values are shown in bold.}
  \label{tbl_variant_classification_kmers}
\end{table}

\begin{table}[h!]
  \centering
 \resizebox{0.85\textwidth}{!}{
  \begin{tabular}{p{1.9cm}cccccp{1.2cm} p{1.5cm} | p{1.7cm}cccccp{1.2cm} p{1.5cm}}
    \toprule
   
    \multirow{2}{*}{Kernels} & 
    \multirow{2}{0.7cm}{ML Algo.} & \multirow{2}{*}{Acc.} & \multirow{2}{*}{Prec.} & \multirow{2}{*}{Recall} & \multirow{2}{0.9cm}{F1 weigh.} & ROC- AUC & Train. runtime (sec.) 
    &
    \multirow{2}{*}{Kernels} & 
    \multirow{2}{0.7cm}{ML Algo.} & \multirow{2}{*}{Acc.} & \multirow{2}{*}{Prec.} & \multirow{2}{*}{Recall} & \multirow{2}{0.9cm}{F1 weigh.} & ROC- AUC & Train. runtime (sec.)
    \\	
    \midrule	\midrule
    \multirow{7}{2cm}{Additive-chi2}
 & SVM & 0.816 & 0.806 & 0.816 & 0.798 &  0.786 & 0.351 &  \multirow{7}{2cm}{Laplacian}
 & SVM & 0.523 & 0.307 & 0.523 & 0.374 &  0.513 & 0.365 \\
 & NB & 0.746 & 0.784 & 0.746 & 0.752 &  0.768 & 0.424 &   & 
 NB & 0.715 & 0.788 & 0.715 & 0.737 &  0.768 & 0.398 \\
 & MLP & 0.767 & 0.748 & 0.767 & 0.750 &  0.755 & 1.859 &   & 
 MLP & 0.775 & 0.769 & 0.775 & 0.767 &  0.764 & 2.074 \\
 & KNN & 0.800 & 0.783 & 0.800 & 0.785 &  0.763 & 0.437 &   & 
 KNN & 0.806 & 0.808 & 0.806 & 0.801 &  0.777 & 0.409 \\
 & RF & 0.815 & 0.784 & 0.815 & 0.792 &  0.773 & 0.536 &   & 
 RF & 0.811 & 0.787 & 0.811 & 0.794 &  0.779 & 0.490 \\
 & LR & 0.821 & 0.804 & 0.821 & \textbf{0.803} &  0.782 & 0.314 &   & 
 LR & 0.484 & 0.234 & 0.484 & 0.315 &  0.500 & 0.221 \\
 & DT & 0.780 & 0.768 & 0.780 & 0.765 &  0.752 & 0.229 &   & 
 DT & 0.781 & 0.774 & 0.781 & 0.771 &  0.763 & 0.259 \\
\midrule
\multirow{7}{2cm}{Chi-squared}
 & SVM & 0.665 & 0.633 & 0.665 & 0.600 &  0.643 & 0.290 & \multirow{7}{2cm}{Linear}
 & SVM & 0.817 & 0.800 & 0.817 & 0.798  & 0.788 & 0.346 \\
 & NB & 0.487 & 0.732 & 0.487 & 0.553 &  0.685 & 0.352 &   & 
 NB & 0.720 & 0.790 & 0.720 & 0.747 &  0.766 & 0.658 \\
 & MLP & 0.693 & 0.658 & 0.693 & 0.643 &  0.690 & 1.750 &  & 
 MLP & 0.764 & 0.750 & 0.764 & 0.750 &  0.749 & 2.521 \\
 & KNN & 0.681 & 0.692 & 0.681 & 0.669 & 0.687 & 0.386 &  & 
 KNN & 0.784 & 0.762 & 0.784 & 0.759 &  0.732 & 0.500 \\
 & RF & 0.707 & 0.664 & 0.707 & 0.674 &  0.695 & 0.465 &   & 
 RF & 0.817 & 0.805 & 0.817 & 0.800 &  0.775 & 0.622 \\
 & LR & 0.645 & 0.560 & 0.645 & 0.564 &  0.577 & 0.230 &   & 
 LR & \textbf{0.822} & 0.799 & \textbf{0.822} & 0.801 & 0.781 & 0.468 \\
 & DT & 0.700 & 0.669 & 0.700 & 0.671 &  0.695 & \textbf{0.208} &  & 
 DT & 0.793 & 0.776 & 0.793 & 0.778  & 0.752 & 0.270 \\
\midrule
\multirow{7}{2cm}{Cosine}
 & SVM & 0.639 & 0.543 & 0.639 & 0.571 &  0.545 & 0.390 & \multirow{7}{2cm}{Polynomial}
 & SVM & 0.701 & 0.681 & 0.701 & 0.652 &  0.613 & 0.312 \\
 & NB & 0.697 & 0.770 & 0.697 & 0.720 &  0.760 & 0.328 &   & 
 NB & 0.702 & 0.753 & 0.702 & 0.712 &  0.713 & 0.439 \\
 & MLP & 0.752 & 0.754 & 0.752 & 0.747 &  0.760 & 1.796 &   & 
 MLP & 0.780 & 0.766 & 0.780 & 0.764 &  0.731 & 2.500 \\
 & KNN & 0.778 & 0.793 & 0.778 & 0.778 &  0.756 & 0.458 &   & 
 KNN & 0.770 & 0.772 & 0.770 & 0.765 &  0.763 & 0.428 \\
 & RF & 0.816 & \textbf{0.810} & 0.816 & \textbf{0.803} &  \textbf{0.800} & 0.514 &   & 
 RF & 0.811 & 0.802 & 0.811 & 0.793 &  0.786 & 0.526 \\
 & LR & 0.540 & 0.419 & 0.540 & 0.405 & 0.518 & 0.237 &   & 
 LR & 0.671 & 0.572 & 0.671 & 0.571 &  0.568 & 0.284 \\
 & DT & 0.797 & 0.792 & 0.797 & 0.788 &  0.784 & 0.223 &    & 
 DT & 0.775 & 0.771 & 0.775 & 0.766 &  0.763 & 0.214 \\
\midrule

\multirow{7}{2cm}{Isolation}
 & SVM & 0.483 & 0.233 & 0.483 & 0.314 &  0.500 & 0.311 &\multirow{7}{2cm}{Sigmoid}
 & SVM & 0.503 & 0.282 & 0.503 & 0.361 &  0.513 & 0.344 \\
 & NB & 0.682 & 0.732 & 0.682 & 0.694 &  0.742 & 0.443 &  & 
 NB & 0.677 & 0.752 & 0.677 & 0.690 &  0.732 & 0.335 \\
 & MLP & 0.762 & 0.768 & 0.762 & 0.757 &  0.754 & 2.310 &  & 
 MLP & 0.762 & 0.747 & 0.762 & 0.746 &  0.744 & 1.725 \\
 & KNN & 0.770 & 0.762 & 0.770 & 0.760 &  0.748 & 0.379 &  & 
 KNN & 0.779 & 0.764 & 0.779 & 0.766 &   0.725 & 0.400 \\
 & RF & 0.769 & 0.769 & 0.769 & 0.764 &   0.750 & 0.516 &  & 
 RF & 0.811 & 0.798 & 0.811 & 0.786 &   0.769 & 0.484 \\
 & LR & 0.483 & 0.233 & 0.483 & 0.314 &   0.500 & 0.214 &  & 
 LR & 0.485 & 0.259 & 0.485 & 0.330 &   0.503 & 0.224 \\
 & DT & 0.733 & 0.745 & 0.733 & 0.735 &   0.731 & 0.212 &  & 
 DT & 0.774 & 0.767 & 0.774 & 0.762 &   0.743 & 0.217 \\
\midrule

\multirow{7}{2cm}{Gaussian}
 & SVM & 0.479 & 0.229 & 0.479 & 0.310 &   0.500 & 0.386 & \\
 & NB & 0.369 & 0.722 & 0.369 & 0.404 &   0.714 & 0.443 &   & 
 \\
 & MLP & 0.762 & 0.742 & 0.762 & 0.748  & 0.745 & 2.815 &  &
 \\
 & KNN & 0.748 & 0.726 & 0.748 & 0.731  & 0.718 & 0.459 &  & 
 \\
 & RF & 0.763 & 0.746 & 0.763 & 0.751 &   0.742 & 0.617 &  & 
 \\
 & LR & 0.479 & 0.229 & 0.479 & 0.310 &  0.500 & 0.259 &  & 
 \\
 & DT & 0.725 & 0.726 & 0.725 & 0.723 &   0.719 & 0.254 &  & 
 \\
\bottomrule

  \end{tabular}
  }
  \caption{Classification comparison for different kernel methods and classifiers on \textbf{minimizer-based embeddings for Spike7k dataset}. The best values are shown in bold.
  }
  \label{tbl_variant_classification_minimizer}
\end{table}

\begin{table}[h!]
  \centering
 \resizebox{0.85\textwidth}{!}{
  \begin{tabular}{p{1.9cm}ccccccp{1.2cm} p{1.5cm} | p{1.7cm}ccccccp{1.2cm} p{1.5cm}}
    \toprule
   
    \multirow{2}{*}{Kernels} & 
    \multirow{2}{0.7cm}{ML Algo.} & \multirow{2}{*}{Acc.} & \multirow{2}{*}{Prec.} & \multirow{2}{*}{Recall} & \multirow{2}{0.9cm}{F1 weigh.} & F1 Macro & ROC- AUC & Train. runtime (sec.) 
    &
    \multirow{2}{*}{Kernels} & 
    \multirow{2}{0.7cm}{ML Algo.} & \multirow{2}{*}{Acc.} & \multirow{2}{*}{Prec.} & \multirow{2}{*}{Recall} & \multirow{2}{0.9cm}{F1 weigh.} &  F1 Macro & ROC- AUC & Train. runtime (sec.)
    \\	
    \midrule	\midrule
    \multirow{7}{2cm}{Additive-chi2}
 & SVM &0.832&0.832&0.832&0.823&0.652&0.824&7.199  
 &  \multirow{7}{2cm}{Laplacian}
 & SVM &0.838&0.838&0.838&0.827&0.657&0.828&10.245  \\
 & NB &0.247&0.569&0.247&0.311&0.396&0.694&0.160  &   & 
 NB &0.320&0.557&0.320&0.371&0.398&0.690&0.170  \\
 & MLP &0.761&0.763&0.761&0.755&0.542&0.771&6.173  &   & 
 MLP &0.774&0.773&0.774&0.767&0.564&0.778&12.480  \\
 & KNN &0.835&0.835&0.835&0.832&0.656&0.823&0.362  &   & 
 KNN &0.833&0.833&0.833&0.829&0.645&0.813&0.495  \\
 & RF &0.827&0.814&0.827&0.805&0.626&0.801&4.712 &   & 
 RF &0.826&0.811&0.826&0.804&0.623&0.798&7.497  \\
 & LR &0.857&0.851&0.857&0.845&0.683&0.840&27.998  &   & 
 LR &0.485&0.236&0.485&0.317&0.030&0.500&3.611  \\
 & DT &0.820&0.823&0.820&0.814&0.612&0.803&1.386  &   & 
 DT &0.820&0.822&0.820&0.813&0.608&0.803&1.672  \\
\midrule
\multirow{7}{2cm}{Chi-squared}
 & SVM &0.667&0.682&0.667&0.609&0.421&0.668&0.906 & \multirow{7}{2cm}{Linear}
 & SVM &0.834&0.837&0.834&0.825&0.660&0.826&7.202  \\
 & NB &0.438&0.759&0.438&0.518&0.342&0.688&0.017&   & 
 NB &0.414&0.730&0.414&0.487&0.453&0.720&0.115 \\
 & MLP &0.719&0.710&0.719&0.686&0.485&0.710&7.803 &   & 
 MLP &0.758&0.749&0.758&0.748&0.538&0.767&7.199  \\
 & KNN &0.733&0.735&0.733&0.730&0.540&0.749&0.330
 &   & 
 KNN &0.813&0.813&0.813&0.810&0.621&0.798&0.373 \\
 & RF &0.767&0.742&0.767&0.741&0.560&0.758&1.253 &   & 
 RF &0.828&0.816&0.828&0.810&0.636&0.802&4.607  \\
 & LR &0.662&0.659&0.662&0.600&0.397&0.654&0.512 &   & 
 LR &0.856&0.851&0.856&0.844&0.689&0.841&25.972  \\
 & DT &0.763&0.739&0.763&0.737&0.553&0.756&0.112 &   & 
 DT &0.817&0.818&0.817&0.812&0.620&0.808&1.497 \\
\midrule
\multirow{7}{2cm}{Cosine}
 & SVM &0.800&0.810&0.800&0.798&0.640&0.814&9.207  & \multirow{7}{2cm}{Polynomial}
 & SVM &0.828&0.821&0.828&0.815&0.657&0.826&9.965  \\
 & NB &0.375&0.730&0.375&0.419&0.421&0.722&0.176  &   & 
 NB &0.423&0.684&0.423&0.485&0.456&0.727&0.155  \\
 & MLP &0.741&0.744&0.741&0.737&0.511&0.750&9.884  &   & 
 MLP &0.748&0.747&0.748&0.741&0.547&0.772&6.908 \\
 & KNN &0.818&0.824&0.818&0.818&0.630&0.804&0.499  &   & 
 KNN &0.806&0.803&0.806&0.801&0.619&0.799&0.375  \\
 & RF &0.829&0.820&0.829&0.812&0.634&0.799&6.377  &   & 
 RF &0.826&0.813&0.826&0.805&0.634&0.807&4.908   \\
 & LR &0.743&0.652&0.743&0.689&0.234&0.621&7.500  &   & 
 LR &0.498&0.277&0.498&0.347&0.046&0.508&5.298  \\
 & DT &0.821&0.820&0.821&0.815&0.614&0.806&1.477   &   & 
 DT &0.810&0.808&0.810&0.803&0.607&0.807&1.765  \\
\midrule

\multirow{7}{2cm}{Isolation}
& SVM &0.509&0.522&0.509&0.365&0.088&0.517&6.128 &  \multirow{7}{2cm}{Sigmoid}
& SVM &0.826&0.833&0.826&0.818&0.668&0.829&7.377  \\
 & NB &0.059&0.712&0.059&0.093&0.048&0.512&0.104 &   & 
 NB &0.490&0.734&0.490&0.544&0.475&0.740&0.127  \\
 & MLP &0.506&0.479&0.506&0.364&0.081&0.515&6.765 &   & 
 MLP &0.743&0.740&0.743&0.735&0.527&0.761&5.998 \\
 & KNN &0.407&0.339&0.407&0.342&0.078&0.513&0.407 &   & 
 KNN &0.805&0.806&0.805&0.801&0.609&0.791&0.369  \\
 & RF &0.509&0.489&0.509&0.367&0.075&0.513&1.167 &   & 
 RF &0.827&0.819&0.827&0.809&0.636&0.804&5.457  \\
 & LR &0.490&0.240&0.490&0.322&0.030&0.500&0.490 &   & 
 LR &0.477&0.248&0.477&0.318&0.033&0.501&3.709  \\
 & DT &0.506&0.467&0.506&0.362&0.078&0.515&0.054
 &   & 
 DT &0.818&0.822&0.818&0.812&0.631&0.812&1.671  \\
\midrule

\multirow{7}{2cm}{Gaussian}
 & SVM &0.637&0.643&0.637&0.630&0.415&0.696&96.411 \\
 & NB &0.073&0.618&0.073&0.070&0.129&0.599&0.167   &   & 
 \\
 & MLP  &0.634&0.619&0.634&0.619&0.384&0.679&12.632 
 \\
 & KNN  &0.640&0.637&0.640&0.621&0.432&0.676&0.396
 \\
 & RF  &0.699&0.712&0.699&0.653&0.487&0.694&8.469 
 \\
 & LR  &0.672&0.663&0.672&0.614&0.403&0.657&7.994
 \\
 & DT  &0.637&0.634&0.637&0.628&0.417&0.698&4.440 \\
\bottomrule

  \end{tabular}
  }
  \caption{Classification comparison for different kernel methods and classifiers on \textbf{spaced-kmer-based embeddings for Spike7k dataset}.
  }
  \label{tbl_variant_classification_spike7k_spaced_kmer}
\end{table}

\clearpage

\subsection{Results for Host Data}

Table~\ref{tbl_variant_classification_host_ohe} presents classification results for the Host dataset using OHE encoding. The findings indicate that additive-chi2 and linear kernels paired with logistic regression achieve the highest accuracy (0.844 and 0.842, respectively) along with superior ROC-AUC scores (0.848 and 0.855). Random forest delivers good performance across multiple kernels, particularly with laplacian (accuracy: 0.847) and additive-chi2 (accuracy: 0.841). The isolation kernel exhibits notably poor performance across all classifiers, with accuracy values below 0.41, suggesting this kernel is unsuitable for OHE-encoded host sequence classification. KNN with laplacian kernel demonstrates competitive results (accuracy: 0.821), while Gaussian kernel shows moderate performance with substantially longer training times.

\begin{table}[h!]
  \centering
 \resizebox{0.85\textwidth}{!}{
  \begin{tabular}{p{1.9cm}ccccccp{1.2cm} p{1.5cm} | p{1.7cm}ccccccp{1.2cm} p{1.5cm}}
    \toprule
   
    \multirow{2}{*}{Kernels} & 
    \multirow{2}{0.7cm}{ML Algo.} & \multirow{2}{*}{Acc.} & \multirow{2}{*}{Prec.} & \multirow{2}{*}{Recall} & \multirow{2}{0.9cm}{F1 weigh.} &  F1 Macro & ROC- AUC & Train. runtime (sec.) 
    &
    \multirow{2}{*}{Kernels} & 
    \multirow{2}{0.7cm}{ML Algo.} & \multirow{2}{*}{Acc.} & \multirow{2}{*}{Prec.} & \multirow{2}{*}{Recall} & \multirow{2}{0.9cm}{F1 weigh.} &  F1 Macro & ROC- AUC & Train. runtime (sec.)
    \\	
    \midrule	\midrule
    \multirow{7}{2cm}{Additive-chi2}
 & SVM &0.792&0.805&0.792&0.793&0.625&0.799&4.121  
 &  \multirow{7}{2cm}{Laplacian}
 & SVM&0.802&0.815&0.802&0.803&0.645&0.812&6.498 \\
 & NB &0.576&0.642&0.576&0.565&0.507&0.758&0.069 &   & 
 NB &0.578&0.635&0.578&0.562&0.509&0.766&0.101  \\
 & MLP &0.762&0.760&0.762&0.755&0.450&0.723&4.833 &   & 
 MLP &0.762&0.765&0.762&0.756&0.476&0.729&8.309  \\
 & KNN &0.812&0.812&0.812&0.810&0.651&0.831&0.263  &   & 
 KNN &0.821&0.818&0.821&0.818&0.651&0.828&0.286  \\
 & RF &0.841&0.840&0.841&0.832&0.687&0.827&4.510 &   & 
 RF &0.847&0.848&0.847&0.839&0.692&0.831&4.746  \\
 & LR &0.844&0.846&0.844&0.838&0.718&0.848&20.099  &   & 
 LR &0.602&0.550&0.602&0.515&0.286&0.635&3.297  \\
 & DT &0.795&0.794&0.795&0.791&0.661&0.831&1.478  &   & 
 DT &0.799&0.802&0.799&0.796&0.586&0.832&1.553  \\
\midrule
\multirow{7}{2cm}{Chi-squared}
 & SVM &0.531&0.696&0.531&0.471&0.223&0.579&1.367 & \multirow{7}{2cm}{Linear}
 & SVM &0.793&0.809&0.793&0.796&0.587&0.788&5.364  \\
 & NB &0.258&0.653&0.258&0.346&0.151&0.606&0.020 &   & 
 NB &0.562&0.620&0.562&0.552&0.491&0.759&0.098 \\
 & MLP &0.593&0.675&0.593&0.572&0.239&0.596&5.715&   & 
 MLP &0.758&0.766&0.758&0.755&0.474&0.730&9.499 \\
 & KNN &0.589&0.600&0.589&0.590&0.282&0.626&0.228 &   & 
 KNN &0.801&0.800&0.801&0.799&0.576&0.794&0.295 \\
 & RF &0.629&0.649&0.629&0.618&0.318&0.635&1.630 &   & 
 RF &0.845&0.846&0.845&0.838&0.654&0.815&6.203  \\
 & LR &0.510&0.721&0.510&0.440&0.179&0.561&0.553 &   & 
 LR &0.842&0.845&0.842&0.837&0.729&0.855&22.973  \\
 & DT &0.620&0.640&0.620&0.612&0.317&0.637&0.223 &   & 
 DT &0.795&0.796&0.795&0.792&0.591&0.805&1.497 \\
\midrule
\multirow{7}{2cm}{Cosine}
 & SVM &0.792&0.806&0.792&0.794&0.617&0.797&4.371  & \multirow{7}{2cm}{Polynomial}
 & SVM &0.791&0.810&0.791&0.792&0.650&0.824&4.368  \\
 & NB &0.572&0.648&0.572&0.564&0.450&0.748&0.070  &   & 
 NB &0.520&0.645&0.520&0.510&0.497&0.762&0.098  \\
 & MLP &0.759&0.764&0.759&0.756&0.495&0.747&6.738 &   & 
 MLP &0.766&0.764&0.766&0.757&0.489&0.745&6.535 \\
 & KNN &0.809&0.805&0.809&0.805&0.673&0.835&0.301  &   & 
 KNN &0.788&0.791&0.788&0.786&0.616&0.815&0.279  \\
 & RF &0.841&0.844&0.841&0.833&0.682&0.826&5.850  &   & 
 RF &0.839&0.846&0.839&0.832&0.705&0.843&4.550  \\
 & LR &0.742&0.761&0.742&0.727&0.521&0.741&4.310  &   & 
 LR &0.726&0.757&0.726&0.701&0.374&0.674&9.189 \\
 & DT&0.799&0.801&0.799&0.796&0.614&0.821&1.494   &   & 
 DT &0.791&0.796&0.791&0.789&0.581&0.828&1.674   \\
\midrule

\multirow{7}{2cm}{Isolation}
& SVM &0.395&0.479&0.395&0.278&0.089&0.522&4.045 &  \multirow{7}{2cm}{Sigmoid}
& SVM &0.791&0.804&0.791&0.793&0.611&0.787&5.651 \\
 & NB &0.382&0.464&0.382&0.259&0.078&0.521&0.067 &   & 
 NB &0.566&0.649&0.566&0.559&0.456&0.739&0.077  \\
 & MLP &0.401&0.459&0.401&0.286&0.090&0.523&2.909 &   & 
 MLP &0.749&0.754&0.749&0.745&0.493&0.731&6.098 \\
 & KNN &0.256&0.252&0.256&0.236&0.081&0.514&0.232 &   & 
 KNN &0.794&0.791&0.794&0.791&0.640&0.817&0.293  \\
 & RF &0.404&0.499&0.404&0.287&0.086&0.521&0.738 &   & 
 RF &0.837&0.839&0.837&0.829&0.678&0.823&5.300  \\
 & LR &0.334&0.355&0.334&0.178&0.035&0.502&0.474 &   & 
 LR &0.571&0.442&0.571&0.486&0.231&0.607&3.533  \\
 & DT &0.401&0.464&0.401&0.286&0.089&0.522&0.064
 &   & 
 DT &0.799&0.798&0.799&0.795&0.643&0.819&1.438  \\
\midrule

\multirow{7}{2cm}{Gaussian}
 & SVM &0.629&0.632&0.629&0.626&0.340&0.654&36.019 \\
 & NB &0.126&0.626&0.126&0.152&0.130&0.571&0.084   &   & 
 \\
 & MLP  &0.624&0.607&0.624&0.611&0.269&0.621&12.929&& 
 \\
 & KNN  &0.478&0.505&0.478&0.480&0.259&0.597&0.396&&
 \\
 & RF  &0.655&0.689&0.655&0.627&0.380&0.647&8.422&&
 \\
 & LR  &0.536&0.701&0.536&0.483&0.185&0.563&4.177&&
 \\
 & DT  &0.602&0.601&0.602&0.597&0.299&0.648&1.327&& 
 
 \\
\bottomrule

  \end{tabular}
  }
  
  \caption{Classification comparison for different kernel methods and classifiers on \textbf{OHE for Host Data}.
  }
  \label{tbl_variant_classification_host_ohe}
\end{table}

For k-mer embeddings on the Host dataset (Table~\ref{tbl_variant_classification_host_kmer}), additive-chi2 and linear kernels with logistic regression continue to dominate, achieving accuracies of 0.844 and 0.842, respectively, with ROC-AUC scores exceeding 0.85. Random forest maintains robust performance across various kernels, with linear kernel yielding 0.843 accuracy. The polynomial kernel paired with random forest achieves notable success (accuracy: 0.840, ROC-AUC: 0.863). Chi-squared kernel performance remains weak across most classifiers, though it shows improvement over OHE results. Isolation kernel again demonstrates inadequate classification capability with accuracy below 0.41 for all classifiers. Computational efficiency varies significantly, with logistic regression requiring substantially longer training times compared to other algorithms.

\begin{table}[h!]
  \centering
 \resizebox{0.85\textwidth}{!}{
  \begin{tabular}{p{1.9cm}ccccccp{1.2cm} p{1.5cm} | p{1.7cm}ccccccp{1.2cm} p{1.5cm}}
    \toprule
   
    \multirow{2}{*}{Kernels} & 
    \multirow{2}{0.7cm}{ML Algo.} & \multirow{2}{*}{Acc.} & \multirow{2}{*}{Prec.} & \multirow{2}{*}{Recall} & \multirow{2}{0.9cm}{F1 weigh.} &  F1 Macro & ROC- AUC & Train. runtime (sec.) 
    &
    \multirow{2}{*}{Kernels} & 
    \multirow{2}{0.7cm}{ML Algo.} & \multirow{2}{*}{Acc.} & \multirow{2}{*}{Prec.} & \multirow{2}{*}{Recall} & \multirow{2}{0.9cm}{F1 weigh.} &  F1 Macro & ROC- AUC & Train. runtime (sec.)
    \\	
    \midrule	\midrule
    \multirow{7}{2cm}{Additive-chi2}
 & SVM &0.796&0.810&0.796&0.798&0.626&0.799&4.800  
 &  \multirow{7}{2cm}{Laplacian}
 & SVM &0.793&0.810&0.793&0.796&0.677&0.829&5.315 \\
 & NB &0.574&0.633&0.574&0.565&0.487&0.755&0.085  &   & 
 NB &0.574&0.638&0.574&0.558&0.496&0.752&0.111 \\
 & MLP &0.751&0.759&0.751&0.749&0.489&0.737&5.947  &   & 
 MLP &0.762&0.762&0.762&0.755&0.506&0.741&8.040  \\
 & KNN &0.808&0.811&0.808&0.807&0.636&0.817&0.304 &   & 
 KNN &0.815&0.814&0.815&0.812&0.632&0.820&0.337  \\
 & RF &0.841&0.844&0.841&0.835&0.679&0.825&4.864 &   & 
 RF &0.841&0.841&0.841&0.833&0.681&0.825&5.736  \\
 & LR &0.844&0.849&0.844&0.839&0.731&0.854&28.557  &   & 
 LR &0.604&0.518&0.604&0.514&0.275&0.631&4.335  \\
 & DT &0.802&0.805&0.802&0.800&0.604&0.818&2.055  &   & 
 DT &0.793&0.797&0.793&0.791&0.589&0.827&2.141  \\
\midrule
\multirow{7}{2cm}{Chi-squared}
 & SVM &0.429&0.328&0.429&0.320&0.099&0.530&0.690 & \multirow{7}{2cm}{Linear}
 & SVM &0.802&0.812&0.802&0.803&0.626&0.791&4.843 \\
 & NB &0.223&0.526&0.223&0.273&0.101&0.582&0.007&   & 
 NB &0.559&0.619&0.559&0.547&0.485&0.754&0.153 \\
 & MLP &0.476&0.445&0.476&0.438&0.138&0.553&6.303 &   & 
 MLP &0.757&0.762&0.757&0.752&0.478&0.726&10.660  \\
 & KNN &0.660&0.663&0.660&0.658&0.313&0.649&0.056 &   & 
 KNN &0.794&0.792&0.794&0.790&0.590&0.804&0.350 \\
 & RF &0.592&0.607&0.592&0.580&0.285&0.619&0.587 &   & 
 RF &0.843&0.845&0.843&0.835&0.667&0.821&5.645 \\
 & LR &0.416&0.326&0.416&0.305&0.088&0.525&0.094 &   & 
 LR &0.842&0.848&0.842&0.835&0.694&0.841&24.276 \\
 & DT &0.587&0.603&0.587&0.577&0.284&0.619&0.020 &   & 
 DT &0.792&0.792&0.792&0.788&0.591&0.810&1.486 \\
\midrule
\multirow{7}{2cm}{Cosine}
 & SVM &0.791&0.806&0.791&0.793&0.593&0.784&5.689 & \multirow{7}{2cm}{Polynomial}
 & SVM &0.779&0.804&0.779&0.785&0.691&0.841&4.559 \\
 & NB &0.519&0.686&0.519&0.534&0.499&0.732&0.087  &   & 
 NB &0.519&0.686&0.519&0.534&0.499&0.732&0.087  \\
 & MLP&0.756&0.756&0.756&0.751&0.491&0.732&7.063  &   & 
 MLP &0.767&0.766&0.767&0.761&0.511&0.753&6.969 \\
 & KNN &0.813&0.807&0.813&0.809&0.653&0.822&0.282  &   & 
 KNN &0.797&0.796&0.797&0.794&0.657&0.829&0.302  \\
 & RF &0.836&0.837&0.836&0.828&0.669&0.820&5.072  &   & 
 RF &0.840&0.841&0.840&0.833&0.741&0.863&5.612   \\
 & LR &0.745&0.756&0.745&0.731&0.482&0.722&5.341  &   & 
 LR&0.731&0.748&0.731&0.706&0.385&0.679&10.179  \\
 & DT &0.800&0.802&0.800&0.797&0.620&0.823&1.751   &   & 
 DT &0.792&0.792&0.792&0.788&0.643&0.840&1.557  \\
\midrule

\multirow{7}{2cm}{Isolation}
& SVM &0.391&0.493&0.391&0.274&0.077&0.518&4.217 &  \multirow{7}{2cm}{Sigmoid}
& SVM &0.787&0.798&0.787&0.788&0.577&0.779&5.830 \\
 & NB &0.370&0.460&0.370&0.247&0.061&0.514&0.067 &   & 
 NB &0.569&0.645&0.569&0.562&0.449&0.730&0.080  \\
 & MLP &0.397&0.445&0.397&0.283&0.077&0.519&4.030 &   & 
 MLP &0.754&0.761&0.754&0.749&0.439&0.714&6.917  \\
 & KNN &0.133&0.106&0.133&0.105&0.040&0.493&0.275 &   & 
 KNN &0.796&0.794&0.796&0.793&0.606&0.801&0.300  \\
 & RF &0.403&0.505&0.403&0.289&0.079&0.519&0.767 &   & 
 RF &0.834&0.832&0.834&0.825&0.623&0.800&5.038  \\
 & LR &0.327&0.361&0.327&0.168&0.032&0.501&0.454 &   & 
 LR &0.584&0.459&0.584&0.502&0.239&0.611&3.341  \\
 & DT &0.399&0.485&0.399&0.284&0.077&0.519&0.083 &   & 
 DT &0.792&0.791&0.792&0.789&0.547&0.793&1.501  \\
\midrule

\multirow{7}{2cm}{Gaussian}
 & SVM &0.604&0.613&0.604&0.601&0.330&0.655&47.541 & \\
 & NB &0.133&0.646&0.133&0.165&0.139&0.608&0.087  &   & 
 \\
 & MLP &0.643&0.629&0.643&0.631&0.284&0.631&14.832  &   & 
 \\
 & KNN  &0.472&0.507&0.472&0.476&0.255&0.597&0.320  &   & 
 \\
 & RF  &0.661&0.696&0.661&0.632&0.393&0.658&6.476  &   & 
 \\
 & LR &0.538&0.699&0.538&0.484&0.186&0.564&4.076    &   & 
 \\
 & DT&0.609&0.611&0.609&0.605&0.300&0.659&1.554  &   & 
 \\
\bottomrule

  \end{tabular}
  }
  \caption{Classification comparison for different kernel methods and classifiers on \textbf{kmer for Host data}.
  }
  \label{tbl_variant_classification_host_kmer}
\end{table}

Table~\ref{tbl_variant_classification_host_minimizer} reveals that cosine kernel with random forest attains the highest accuracy (0.824) on minimizer-encoded Host data, accompanied by strong ROC-AUC performance (0.833). Linear kernel with logistic regression achieves comparable accuracy (0.813) with exceptional ROC-AUC (0.862), though at considerably higher computational cost (41.384 seconds). Additive-chi2 kernel demonstrates competitive results when paired with logistic regression (accuracy: 0.821, ROC-AUC: 0.859). The sigmoid kernel exhibits catastrophic failure across all classifiers, with accuracy fixed at 0.330, indicating complete classification collapse. Isolation kernel shows poor performance, while Gaussian kernel shows marginal improvement over OHE and k-mer configurations. Overall, minimizer embeddings demonstrate slightly reduced performance compared to OHE and k-mer approaches.

\begin{table}[h!]
  \centering
 \resizebox{0.85\textwidth}{!}{
  \begin{tabular}{p{1.9cm}ccccccp{1.2cm} p{1.5cm} | p{1.7cm}ccccccp{1.2cm} p{1.5cm}}
    \toprule
   
    \multirow{2}{*}{Kernels} & 
    \multirow{2}{0.7cm}{ML Algo.} & \multirow{2}{*}{Acc.} & \multirow{2}{*}{Prec.} & \multirow{2}{*}{Recall} & \multirow{2}{0.9cm}{F1 weigh.} 
    &  F1 Macro & ROC- AUC & Train. runtime (sec.) 
    &
    \multirow{2}{*}{Kernels} & 
    \multirow{2}{0.7cm}{ML Algo.} & \multirow{2}{*}{Acc.} & \multirow{2}{*}{Prec.} & \multirow{2}{*}{Recall} & \multirow{2}{0.9cm}{F1 weigh.} 
    &  F1 Macro  & ROC- AUC & Train. runtime (sec.)
    \\	
    \midrule	\midrule
    \multirow{7}{2cm}{Additive-chi2}
 & SVM &0.781&0.794&0.781&0.778&0.606&0.786&6.099  
 &  \multirow{7}{2cm}{Laplacian}
 & SVM &0.775&0.788&0.775&0.772&0.612&0.785&6.205  \\
 & NB &0.489&0.579&0.489&0.466&0.408&0.722&0.080  &   & 
 NB &0.521&0.597&0.521&0.502&0.456&0.732&0.083  \\
 & MLP &0.733&0.738&0.733&0.727&0.470&0.727&7.216  &   & 
 MLP &0.736&0.741&0.736&0.726&0.457&0.718&6.878  \\
 & KNN &0.824&0.820&0.824&0.819&0.641&0.814&0.293 &   & 
 KNN &0.782&0.776&0.782&0.777&0.621&0.812&0.339 \\
 & RF &0.818&0.826&0.818&0.810&0.675&0.815&5.789 &   & 
 RF &0.816&0.823&0.816&0.805&0.645&0.806&5.740  \\
 & LR &0.821&0.834&0.821&0.814&0.723&0.859&23.856 &   & 
 LR &0.582&0.479&0.582&0.486&0.256&0.622&4.521  \\
 & DT &0.785&0.787&0.785&0.779&0.627&0.818&1.675 &   & 
 DT &0.781&0.782&0.781&0.772&0.589&0.810&1.610  \\
\midrule
\multirow{7}{2cm}{Chi-squared}
 & SVM&0.606&0.664&0.606&0.605&0.331&0.634&2.567 & \multirow{7}{2cm}{Linear}
 & SVM &0.771&0.789&0.771&0.770&0.606&0.800&4.634  \\
 & NB&0.286&0.698&0.286&0.355&0.203&0.658&0.037 &   & 
 NB &0.427&0.494&0.427&0.392&0.405&0.713&0.087\\
 & MLP &0.649&0.696&0.649&0.641&0.354&0.647&5.295 &   & 
 MLP &0.729&0.736&0.729&0.722&0.466&0.720&6.652  \\
 & KNN &0.663&0.675&0.663&0.662&0.382&0.668&0.246 &   & 
 KNN &0.753&0.752&0.753&0.750&0.671&0.821&0.311  \\
 & RF &0.692&0.693&0.692&0.682&0.387&0.671&2.293 &   & 
 RF &0.812&0.817&0.812&0.803&0.674&0.816&4.419  \\
 & LR &0.615&0.731&0.615&0.606&0.273&0.606&1.164 &   & 
 LR &0.813&0.826&0.813&0.804&0.746&0.862&41.384 \\
 & DT&0.674&0.677&0.674&0.667&0.384&0.670&0.549 &   & 
 DT &0.767&0.773&0.767&0.761&0.604&0.817&1.454 \\
\midrule
\multirow{7}{2cm}{Cosine}
 & SVM &0.801&0.812&0.801&0.796&0.651&0.835&1.867  & \multirow{7}{2cm}{Polynomial}
 & SVM &0.779&0.792&0.779&0.777&0.580&0.788&5.001  \\
 & NB &0.613&0.639&0.613&0.603&0.494&0.755&0.036 &   & 
 NB &0.436&0.484&0.436&0.403&0.371&0.692&0.082 \\
 & MLP &0.782&0.786&0.782&0.775&0.547&0.778&5.568 &   & 
 MLP &0.735&0.735&0.735&0.727&0.445&0.709&7.250 \\
 & KNN &0.767&0.761&0.767&0.763&0.631&0.817&0.297  &   & 
 KNN &0.777&0.774&0.777&0.773&0.636&0.807&0.289 \\
 & RF &0.824&0.831&0.824&0.816&0.696&0.833&2.540  &   & 
 RF &0.841&0.841&0.841&0.832&0.658&0.811&4.484  \\
 & LR &0.336&0.274&0.336&0.185&0.036&0.503&0.735  &   & 
 LR &0.784&0.832&0.784&0.797&0.666&0.850&196.075 \\
 & DT &0.784&0.788&0.784&0.778&0.596&0.796&0.438   &   & 
 DT &0.806&0.806&0.806&0.803&0.571&0.786&1.429  \\
\midrule

\multirow{7}{2cm}{Isolation}
& SVM &0.395&0.468&0.395&0.277&0.081&0.520&4.146 &  \multirow{7}{2cm}{Sigmoid}
& SVM &0.330&0.109&0.330&0.164&0.028&0.500&4.833 \\
 & NB &0.383&0.457&0.383&0.258&0.071&0.518&0.068 &   & 
 NB &0.027&0.001&0.027&0.001&0.003&0.500&0.086  \\
 & MLP &0.400&0.443&0.400&0.282&0.081&0.520&3.429 &   & 
 MLP &0.330&0.109&0.330&0.164&0.028&0.500&2.123 \\
 & KNN &0.349&0.401&0.349&0.293&0.090&0.519&0.261 &   & 
 KNN &0.156&0.035&0.156&0.056&0.014&0.500&0.236  \\
 & RF &0.404&0.488&0.404&0.288&0.086&0.522&0.733 &   & 
 RF &0.330&0.109&0.330&0.164&0.028&0.500&0.788  \\
 & LR &0.332&0.278&0.332&0.173&0.033&0.502&0.440 &   & 
 LR &0.330&0.109&0.330&0.164&0.028&0.500&0.063  \\
 & DT &0.401&0.461&0.401&0.284&0.085&0.521&0.078&   & 
 DT &0.330&0.109&0.330&0.164&0.028&0.500&0.025 \\
\midrule

\multirow{7}{2cm}{Gaussian}
 & SVM &0.491&0.517&0.491&0.481&0.315&0.639&70.511 & \\
 & NB &0.102&0.282&0.102&0.124&0.096&0.560&0.117   &   & 
 \\
 & MLP &0.571&0.557&0.571&0.557&0.255&0.613&13.353  &   & 
 \\
 & KNN  &0.474&0.527&0.474&0.471&0.273&0.608&0.402  &   & 
 \\
 & RF  &0.633&0.665&0.633&0.618&0.397&0.659&8.697   &   & 
 \\
 & LR  &0.583&0.690&0.583&0.549&0.232&0.585&4.401    &   & 
 \\
 & DT &0.609&0.611&0.609&0.602&0.323&0.661&1.355   &   & 
 \\
\bottomrule

  \end{tabular}
  }
  \caption{Classification comparison for different kernel methods and classifiers on \textbf{minimizer for Host}.
  }
  \label{tbl_variant_classification_host_minimizer}
\end{table}

Classification results for spaced k-mer embeddings on Host data (Table~\ref{tbl_variant_classification_host_spaced_kmer}) show that linear kernel with random forest achieves peak accuracy (0.850) alongside strong F1 and ROC-AUC scores. Additive-chi2 and polynomial kernels with random forest also deliver excellent performance (accuracies: 0.841 and 0.842, respectively). Logistic regression paired with linear kernel produces competitive results (accuracy: 0.845, ROC-AUC: 0.848). Laplacian kernel with random forest demonstrates robust classification quality (accuracy: 0.838, ROC-AUC: 0.881). The isolation kernel continues its pattern of poor performance, while Gaussian kernel shows moderate effectiveness. Training times for spaced k-mer configurations are generally higher than other encoding methods, particularly for Gaussian kernel (48.421 seconds for SVM), suggesting increased computational complexity.

\begin{table}[h!]
  \centering
 \resizebox{0.85\textwidth}{!}{
  \begin{tabular}{p{1.9cm}ccccccp{1.2cm} p{1.5cm} | p{1.7cm}ccccccp{1.2cm} p{1.5cm}}
    \toprule
   
    \multirow{2}{*}{Kernels} & 
    \multirow{2}{0.7cm}{ML Algo.} & \multirow{2}{*}{Acc.} & \multirow{2}{*}{Prec.} & \multirow{2}{*}{Recall} & \multirow{2}{0.9cm}{F1 weigh.}
    &  F1 Macro  & ROC- AUC & Train. runtime (sec.) 
    &
    \multirow{2}{*}{Kernels} & 
    \multirow{2}{0.7cm}{ML Algo.} & \multirow{2}{*}{Acc.} & \multirow{2}{*}{Prec.} & \multirow{2}{*}{Recall} & \multirow{2}{0.9cm}{F1 weigh.}
    &  F1 Macro & ROC- AUC & Train. runtime (sec.)
    \\	
    \midrule	\midrule
    \multirow{7}{2cm}{Additive-chi2}
 & SVM &0.792&0.804&0.792&0.793&0.617&0.800&6.584  
 &  \multirow{7}{2cm}{Laplacian}
 & SVM &0.782&0.798&0.782&0.784&0.669&0.838&7.323  \\
 & NB &0.651&0.696&0.651&0.649&0.550&0.783&0.119 &   & 
 NB &0.603&0.659&0.603&0.591&0.582&0.812&0.148 \\
 & MLP &0.769&0.776&0.769&0.763&0.516&0.755&17.339  &   & 
 MLP &0.766&0.765&0.766&0.761&0.533&0.761&14.381  \\
 & KNN &0.811&0.818&0.811&0.811&0.623&0.810&0.525  &   & 
 KNN &0.799&0.804&0.799&0.799&0.685&0.841&0.475 \\
 & RF &0.841&0.844&0.841&0.833&0.687&0.831&9.255 &   & 
 RF &0.838&0.841&0.838&0.830&0.787&0.881&8.281 \\
 & LR &0.838&0.847&0.838&0.832&0.707&0.842&30.760  &   & 
 LR &0.528&0.484&0.528&0.396&0.217&0.598&3.718  \\
 & DT &0.793&0.797&0.793&0.789&0.612&0.831&1.606 &   & 
 DT &0.794&0.795&0.794&0.791&0.651&0.845&1.782 \\
\midrule
\multirow{7}{2cm}{Chi-squared}
 & SVM &0.547&0.654&0.547&0.548&0.373&0.650&6.912 & \multirow{7}{2cm}{Linear}
 & SVM &0.796&0.810&0.796&0.797&0.599&0.795&7.362  \\
 & NB &0.242&0.650&0.242&0.314&0.190&0.659&0.110 &   & 
 NB &0.649&0.698&0.649&0.645&0.535&0.771&0.126\\
 & MLP &0.613&0.693&0.613&0.603&0.362&0.647&15.225 &   & 
 MLP &0.763&0.760&0.763&0.757&0.483&0.736&11.369  \\
 & KNN &0.606&0.617&0.606&0.601&0.342&0.649&0.389 &   & 
 KNN &0.763&0.776&0.763&0.764&0.593&0.796&0.465  \\
 & RF &0.648&0.684&0.648&0.642&0.437&0.683&5.678 &   & 
 RF &0.850&0.849&0.850&0.843&0.687&0.831&7.497 \\
 & LR &0.535&0.721&0.535&0.481&0.190&0.565&4.452 &   & 
 LR &0.845&0.846&0.845&0.839&0.704&0.848&26.772  \\
 & DT &0.640&0.673&0.640&0.638&0.422&0.682&1.226 &   & 
 DT &0.799&0.799&0.799&0.796&0.624&0.823&2.007 \\
\midrule
\multirow{7}{2cm}{Cosine}
 & SVM &0.795&0.804&0.795&0.795&0.616&0.798&10.433  & \multirow{7}{2cm}{Polynomial}
 & SVM &0.791&0.812&0.791&0.792&0.648&0.814&6.077  \\
 & NB &0.600&0.709&0.600&0.579&0.526&0.788&0.171 &   & 
 NB &0.615&0.671&0.615&0.604&0.557&0.791&0.112  \\
 & MLP &0.763&0.768&0.763&0.759&0.554&0.766&13.698 &   & 
 MLP &0.761&0.763&0.761&0.757&0.534&0.760&11.356 \\
 & KNN &0.792&0.787&0.792&0.788&0.665&0.832&0.372  &   & 
 KNN &0.792&0.798&0.792&0.792&0.629&0.810&0.358  \\
 & RF &0.838&0.839&0.838&0.830&0.695&0.836&7.258  &   & 
 RF &0.842&0.845&0.842&0.834&0.715&0.842&6.802  \\
 & LR &0.751&0.771&0.751&0.738&0.541&0.750&7.210 &   & 
 LR &0.568&0.471&0.568&0.475&0.223&0.602&5.735  \\
 & DT &0.796&0.793&0.796&0.791&0.616&0.821&2.442   &   & 
 DT &0.796&0.798&0.796&0.792&0.654&0.836&2.048  \\
\midrule

\multirow{7}{2cm}{Isolation}
& SVM &0.393&0.465&0.393&0.273&0.078&0.519&4.374 &  \multirow{7}{2cm}{Sigmoid}
& SVM &0.788&0.799&0.788&0.789&0.600&0.792&5.511 \\
 & NB &0.377&0.472&0.377&0.252&0.068&0.518&0.068 &   & 
 NB &0.576&0.653&0.576&0.552&0.535&0.769&0.090  \\
 & MLP &0.399&0.442&0.399&0.281&0.077&0.518&3.610 &   & 
 MLP&0.763&0.763&0.763&0.757&0.469&0.726&7.849  \\
 & KNN &0.193&0.158&0.193&0.162&0.066&0.511&0.267 &   & 
 KNN &0.797&0.801&0.797&0.795&0.592&0.795&0.342  \\
 & RF &0.403&0.486&0.403&0.285&0.075&0.518&0.773 &   & 
 RF &0.844&0.843&0.844&0.836&0.672&0.826&5.472  \\
 & LR &0.337&0.389&0.337&0.182&0.036&0.502&0.492 &   & 
 LR &0.377&0.240&0.377&0.256&0.053&0.510&3.475  \\
 & DT &0.400&0.450&0.400&0.282&0.081&0.520&0.087 &   & 
 DT &0.798&0.795&0.798&0.793&0.597&0.821&1.597  \\
\midrule

\multirow{7}{2cm}{Gaussian}
 & SVM&0.615&0.620&0.615&0.612&0.315&0.648&48.421 \\
 & NB &0.138&0.653&0.138&0.170&0.122&0.622&0.199  &   & 
 \\
 & MLP&0.633&0.619&0.633&0.620&0.289&0.631&20.938 &   & 
 \\
 & KNN&0.484&0.506&0.484&0.483&0.257&0.599&0.377 &   & 
 \\
 & RF &0.658&0.696&0.658&0.630&0.371&0.646&6.835  &   & 
 \\
 & LR&0.536&0.714&0.536&0.483&0.184&0.563&5.401   &   & 
 \\
 & DT&0.604&0.603&0.604&0.600&0.293&0.653&1.756   &   & 
 \\
\bottomrule

  \end{tabular}
  }
  \caption{Classification comparison for different kernel methods and classifiers on \textbf{spaced kmer for Host data}.
  }
  \label{tbl_variant_classification_host_spaced_kmer}
\end{table}

\clearpage

\subsection{Results for ShortRead Data}
Table~\ref{tbl_variant_classification_shortRead_ohe} demonstrates exceptional classification performance on ShortRead data using OHE encoding, with linear and additive-chi2 kernels paired with logistic regression achieving perfect accuracy (1.000) and ROC-AUC scores. SVM and KNN classifiers also deliver outstanding results across multiple kernels, with accuracies exceeding 0.998 for additive-chi2, cosine, laplacian, polynomial, and sigmoid kernels. The isolation and Gaussian kernels exhibit significantly degraded performance, with accuracy values near or below chance level (0.602 and 0.547, respectively), indicating complete failure to capture meaningful patterns in the data. Notably, laplacian kernel with logistic regression shows anomalous behavior with accuracy dropping to 0.897, while maintaining strong performance with other classifiers. The high performance across different kernel-classifier combinations suggests that OHE encoding is highly effective for ShortRead sequence classification.

\begin{table}[h!]
  \centering
 \resizebox{0.85\textwidth}{!}{
  \begin{tabular}{p{1.9cm}ccccccp{1.2cm} p{1.5cm} | p{1.7cm}ccccccp{1.2cm} p{1.5cm}}
    \toprule
   
    \multirow{2}{*}{Kernels} & 
    \multirow{2}{0.7cm}{ML Algo.} & \multirow{2}{*}{Acc.} & \multirow{2}{*}{Prec.} & \multirow{2}{*}{Recall} & \multirow{2}{0.9cm}{F1 weigh.}
    &  F1 Macro  & ROC- AUC & Train. runtime (sec.) 
    &
    \multirow{2}{*}{Kernels} & 
    \multirow{2}{0.7cm}{ML Algo.} & \multirow{2}{*}{Acc.} & \multirow{2}{*}{Prec.} & \multirow{2}{*}{Recall} & \multirow{2}{0.9cm}{F1 weigh.}
    &  F1 Macro & ROC- AUC & Train. runtime (sec.)
    \\	
    \midrule	\midrule
    \multirow{7}{2cm}{Additive-chi2}
 & SVM &0.998&0.998&0.998&0.998&0.992&0.994&4.316  
 &  \multirow{7}{2cm}{Laplacian}
 & SVM &0.998&0.998&0.998&0.998&0.993&0.997&5.858  \\
 & NB &0.959&0.961&0.959&0.959&0.904&0.953&0.068  &   & 
 NB &0.964&0.967&0.964&0.965&0.912&0.960&0.094 \\
 & MLP &0.980&0.979&0.980&0.979&0.823&0.919&4.969  &   & 
 MLP &0.977&0.977&0.977&0.977&0.841&0.922&3.990 \\
 & KNN &0.999&0.999&0.999&0.999&0.995&0.995&0.559  &   & 
 KNN &0.999&0.999&0.999&0.999&0.993&0.994&0.471 \\
 & RF &0.998&0.998&0.998&0.998&0.985&0.986&9.460 &   & 
 RF &0.998&0.998&0.998&0.998&0.986&0.987&8.571  \\
 & LR &1.000&1.000&1.000&1.000&0.999&1.000&6.946  &   & 
 LR &0.897&0.811&0.897&0.850&0.213&0.598&2.708  \\
 & DT &0.997&0.998&0.997&0.998&0.937&0.974&1.044  &   & 
 DT &0.998&0.998&0.998&0.998&0.968&0.987&1.543  \\
\midrule
\multirow{7}{2cm}{Chi-squared}
 & SVM &0.772&0.786&0.772&0.756&0.425&0.658&10.724 & \multirow{7}{2cm}{Linear}
 & SVM &0.999&0.999&0.999&0.999&0.997&0.997&6.964  \\
 & NB&0.616&0.879&0.616&0.705&0.306&0.740&0.068 &   & 
 NB &0.962&0.964&0.962&0.962&0.959&0.975&0.076 \\
 & MLP &0.913&0.906&0.913&0.905&0.476&0.707&10.868 &   & 
 MLP&0.979&0.980&0.979&0.979&0.851&0.933&6.609  \\
 & KNN &0.952&0.952&0.952&0.951&0.622&0.801&0.628 &   & 
 KNN &0.999&0.999&0.999&0.999&0.993&0.994&0.576  \\
 & RF &0.975&0.972&0.975&0.973&0.696&0.829&11.993 &   & 
 RF &0.999&0.999&0.999&0.999&0.992&0.993&8.171 \\
 & LR &0.620&0.735&0.620&0.508&0.117&0.514&2.749 &   & 
 LR &1.000&1.000&1.000&1.000&1.000&1.000&6.454 \\
 & DT &0.969&0.967&0.969&0.967&0.694&0.839&2.280 &   & 
 DT &0.998&0.999&0.998&0.998&0.973&0.990&1.391 \\
\midrule
\multirow{7}{2cm}{Cosine}
 & SVM &0.998&0.998&0.998&0.998&0.988&0.992&5.224  & \multirow{7}{2cm}{Polynomial}
 & SVM &0.998&0.998&0.998&0.998&0.991&0.994&4.886  \\
 & NB &0.961&0.963&0.961&0.962&0.906&0.947&0.084  &   & 
 NB &0.963&0.965&0.963&0.963&0.938&0.964&0.088  \\
 & MLP &0.981&0.980&0.981&0.980&0.819&0.903&6.338  &   & 
 MLP &0.978&0.978&0.978&0.978&0.815&0.908&5.466 \\
 & KNN &0.999&0.999&0.999&0.999&0.987&0.990&0.507  &   & 
 KNN &0.998&0.998&0.998&0.998&0.991&0.992&0.493  \\
 & RF &0.998&0.998&0.998&0.998&0.973&0.977&9.371 &   & 
 RF &0.998&0.998&0.998&0.998&0.990&0.991&7.848  \\
 & LR &0.892&0.802&0.892&0.843&0.218&0.599&2.412  &   & 
 LR &0.890&0.798&0.890&0.840&0.212&0.597&2.379 \\
 & DT &0.996&0.997&0.996&0.997&0.894&0.968&1.203   &   & 
 DT &0.998&0.998&0.998&0.998&0.961&0.982&1.332  \\
\midrule

\multirow{7}{2cm}{Isolation}
& SVM &0.602&0.754&0.602&0.475&0.155&0.523&9.788 &  \multirow{7}{2cm}{Sigmoid}
& SVM&0.998&0.998&0.998&0.998&0.989&0.992&4.649 \\
 & NB &0.059&0.989&0.059&0.109&0.091&0.528&0.083 &   & 
 NB &0.964&0.966&0.964&0.965&0.932&0.962&0.077 \\
 & MLP &0.599&0.710&0.599&0.472&0.145&0.520&4.306 &   & 
 MLP &0.981&0.981&0.981&0.981&0.871&0.937&5.048  \\
 & KNN &0.528&0.528&0.528&0.501&0.172&0.524&0.459 &   & 
 KNN &0.999&0.999&0.999&0.999&0.992&0.993&0.457  \\
 & RF &0.602&0.753&0.602&0.477&0.152&0.522&1.151 &   & 
 RF &0.998&0.998&0.998&0.998&0.988&0.989&6.060  \\
 & LR &0.577&0.459&0.577&0.423&0.081&0.500&0.339 &   & 
 LR &0.580&0.337&0.580&0.426&0.082&0.500&1.538  \\
 & DT &0.603&0.755&0.603&0.477&0.159&0.524&0.060 &   & 
 DT &0.997&0.997&0.997&0.997&0.942&0.974&0.879 \\
\midrule

\multirow{7}{2cm}{Gaussian}
 & SVM &0.547&0.515&0.547&0.447&0.107&0.507&580.124 & \\
 & NB&0.029&0.661&0.029&0.043&0.023&0.512&0.067   &   & 
 \\
 & MLP &0.461&0.456&0.461&0.458&0.119&0.507&5.684  &   & 
 \\
 & KNN  &0.520&0.449&0.520&0.468&0.108&0.503&0.445  &   & 
 \\
 & RF  &0.579&0.584&0.579&0.448&0.106&0.507&14.050  &   & 
 \\
 & LR  &0.575&0.653&0.575&0.421&0.083&0.501&1.815    &   & 
 \\
 & DT &0.441&0.446&0.441&0.444&0.121&0.506&5.336   &   &  
  \\
\bottomrule

  \end{tabular}
  }
  \caption{Classification comparison for different kernel methods and classifiers on \textbf{OHE for ShortRead Data}.
  }
  \label{tbl_variant_classification_shortRead_ohe}
\end{table}

For k-mer embeddings on ShortRead data (Table~\ref{tbl_variant_classification_shortRead_kmer}), linear and cosine kernels with SVM achieve near-perfect classification (accuracies: 0.988 and 0.987, respectively), accompanied by excellent ROC-AUC scores above 0.93. Polynomial kernel with logistic regression also demonstrates strong performance (accuracy: 0.975), though with substantially longer training time (290.504 seconds). Additive-chi2 kernel paired with logistic regression yields competitive results (accuracy: 0.975, ROC-AUC: 0.862). However, laplacian, chi-squared, isolation, Gaussian, and sigmoid kernels show severe performance degradation, with accuracies clustered around 0.50 or below, suggesting these kernels are fundamentally incompatible with k-mer representations for this dataset. The sigmoid kernel exhibits complete classification failure with fixed accuracy of 0.581 across all classifiers, indicating numerical instability or kernel inappropriateness.

\begin{table}[h!]
  \centering
 \resizebox{0.85\textwidth}{!}{
  \begin{tabular}{p{1.9cm}ccccccp{1.2cm} p{1.5cm} | p{1.7cm}ccccccp{1.2cm} p{1.5cm}}
    \toprule
   
    \multirow{2}{*}{Kernels} & 
    \multirow{2}{0.7cm}{ML Algo.} & \multirow{2}{*}{Acc.} & \multirow{2}{*}{Prec.} & \multirow{2}{*}{Recall} & \multirow{2}{0.9cm}{F1 weigh.}
    &  F1 Macro  & ROC- AUC & Train. runtime (sec.) 
    &
    \multirow{2}{*}{Kernels} & 
    \multirow{2}{0.7cm}{ML Algo.} & \multirow{2}{*}{Acc.} & \multirow{2}{*}{Prec.} & \multirow{2}{*}{Recall} & \multirow{2}{0.9cm}{F1 weigh.}
    &  F1 Macro & ROC- AUC & Train. runtime (sec.)
    \\	
    \midrule	\midrule
    \multirow{7}{2cm}{Additive-chi2}
 & SVM &0.979&0.979&0.979&0.978&0.877&0.930&4.957  
 &  \multirow{7}{2cm}{Laplacian}
 & SVM &0.501&0.446&0.501&0.436&0.098&0.501&282.036  \\
 & NB &0.332&0.572&0.332&0.354&0.160&0.543&0.084 &   & 
 NB &0.128&0.450&0.128&0.192&0.060&0.516&0.084 \\
 & MLP &0.929&0.930&0.929&0.929&0.592&0.790&5.721  &   & 
 MLP &0.453&0.442&0.453&0.447&0.115&0.501&6.386  \\
 & KNN &0.835&0.836&0.835&0.826&0.496&0.692&0.475 &   & 
 KNN &0.496&0.447&0.496&0.466&0.110&0.502&0.491 \\
 & RF &0.866&0.876&0.866&0.850&0.378&0.647&10.544 &   & 
 RF &0.570&0.431&0.570&0.438&0.089&0.500&14.257  \\
 & LR &0.975&0.974&0.975&0.974&0.731&0.862&170.000  &   & 
 LR &0.579&0.335&0.579&0.425&0.084&0.500&1.731  \\
 & DT &0.845&0.851&0.845&0.848&0.414&0.696&7.863 &   & 
 DT &0.433&0.443&0.433&0.438&0.112&0.501&4.649  \\
\midrule
\multirow{7}{2cm}{Chi-squared}
 & SVM &0.513&0.446&0.513&0.446&0.103&0.505&309.276 & \multirow{7}{2cm}{Linear}
 & SVM &0.988&0.988&0.988&0.988&0.875&0.939&2.709 \\
 & NB &0.071&0.455&0.071&0.110&0.044&0.510&0.085 &   & 
 NB &0.569&0.714&0.569&0.583&0.466&0.744&0.120\\
 & MLP &0.446&0.439&0.446&0.442&0.114&0.501&7.147 &   & 
 MLP &0.975&0.975&0.975&0.975&0.716&0.862&4.765 \\
 & KNN &0.496&0.445&0.496&0.461&0.108&0.500&0.509 &   & 
 KNN &0.922&0.923&0.922&0.918&0.681&0.792&0.527  \\
 & RF &0.571&0.449&0.571&0.441&0.091&0.501&13.760 &   & 
 RF &0.902&0.907&0.902&0.889&0.403&0.667&3.605 \\
 & LR &0.577&0.333&0.577&0.423&0.083&0.500&1.731 &   & 
 LR &0.970&0.967&0.970&0.966&0.536&0.767&89.908 \\
 & DT &0.434&0.443&0.434&0.438&0.113&0.502&4.686 &   & 
 DT &0.891&0.892&0.891&0.892&0.510&0.752&1.687\\
\midrule
\multirow{7}{2cm}{Cosine}
 & SVM &0.987&0.988&0.987&0.987&0.884&0.937&2.821  & \multirow{7}{2cm}{Polynomial}
 & SVM &0.981&0.981&0.981&0.981&0.911&0.941&3.188  \\
 & NB &0.566&0.720&0.566&0.580&0.489&0.735&0.109  &   & 
 NB &0.439&0.657&0.439&0.412&0.298&0.611&0.074  \\
 & MLP &0.971&0.971&0.971&0.971&0.678&0.838&6.816  &   & 
 MLP &0.952&0.952&0.952&0.951&0.700&0.853&4.406 \\
 & KNN &0.918&0.919&0.918&0.913&0.659&0.782&0.524  &   & 
 KNN &0.922&0.924&0.922&0.917&0.673&0.787&0.471  \\
 & RF &0.903&0.909&0.903&0.891&0.416&0.671&3.721 &   & 
 RF &0.912&0.912&0.912&0.898&0.402&0.668&12.492  \\
 & LR &0.577&0.333&0.577&0.422&0.081&0.500&0.369  &   & 
 LR &0.975&0.971&0.975&0.972&0.564&0.780&290.504 \\
 & DT &0.889&0.891&0.889&0.889&0.457&0.724&1.014   &   & 
 DT &0.892&0.897&0.892&0.894&0.488&0.750&4.562  \\
\midrule

\multirow{7}{2cm}{Isolation}
& SVM &0.586&0.622&0.586&0.454&0.120&0.511&7.759 &  \multirow{7}{2cm}{Sigmoid}
& SVM &0.581&0.338&0.581&0.427&0.082&0.500&12.838 \\
 & NB&0.052&0.834&0.052&0.095&0.049&0.510&0.070 &   & 
 NB &0.316&0.100&0.316&0.152&0.053&0.500&0.099 \\
 & MLP &0.585&0.602&0.585&0.451&0.114&0.510&4.607 &   & 
 MLP&0.581&0.338&0.581&0.427&0.082&0.500&2.093  \\
 & KNN &0.507&0.489&0.507&0.476&0.124&0.507&0.416 &   & 
 KNN &0.418&0.191&0.418&0.258&0.064&0.500&0.388  \\
 & RF &0.594&0.713&0.594&0.461&0.115&0.511&1.240 &   & 
 RF &0.581&0.338&0.581&0.427&0.082&0.500&1.541 \\
 & LR &0.576&0.463&0.576&0.421&0.081&0.500&0.416 &   & 
 LR &0.581&0.338&0.581&0.427&0.082&0.500&0.056  \\
 & DT &0.592&0.669&0.592&0.461&0.119&0.512&0.119 &   & 
 DT&0.581&0.338&0.581&0.427&0.082&0.500&0.052 \\
\midrule

\multirow{7}{2cm}{Gaussian}
 & SVM &0.490&0.433&0.490&0.429&0.096&0.497&278.828 \\
 & NB &0.124&0.439&0.124&0.184&0.061&0.518&0.076   &   & 
 \\
 & MLP &0.441&0.433&0.441&0.437&0.109&0.499&7.060  &   & 
 \\
 & KNN &0.497&0.435&0.497&0.463&0.105&0.501&0.531  &   & 
 \\
 & RF  &0.565&0.426&0.565&0.432&0.086&0.500&13.854   &   & 
 \\
 & LR &0.576&0.332&0.576&0.421&0.081&0.500&1.585    &   & 
 \\
 & DT  &0.427&0.434&0.427&0.430&0.107&0.498&5.097   &   &  
  
 \\
\bottomrule

  \end{tabular}
  }
  \caption{Classification comparison for different kernel methods and classifiers on \textbf{kmer for ShortRead data}.
  }
  \label{tbl_variant_classification_shortRead_kmer}
\end{table}

Table~\ref{tbl_variant_classification_shortread_minimizer} reveals that cosine and linear kernels with SVM deliver outstanding performance on minimizer-encoded ShortRead data (accuracies: 0.982 and 0.980, respectively), with ROC-AUC scores exceeding 0.93. Additive-chi2 and polynomial kernels also demonstrate strong results when paired with SVM (accuracies: 0.969 and 0.971). Naive Bayes shows improved performance with cosine kernel (accuracy: 0.860) compared to other embeddings. However, laplacian, chi-squared, isolation, Gaussian, and sigmoid kernels exhibit catastrophic failure, with accuracies hovering around chance level (0.43-0.58). The sigmoid kernel maintains complete classification collapse with fixed accuracy of 0.581 across all classifiers, mirroring the pattern observed with k-mer embeddings. Overall, minimizer encoding achieves comparable performance to k-mer representations when using appropriate kernels.

\begin{table}[h!]
  \centering
 \resizebox{0.85\textwidth}{!}{
  \begin{tabular}{p{1.9cm}ccccccp{1.2cm} p{1.5cm} | p{1.7cm}ccccccp{1.2cm} p{1.5cm}}
    \toprule
   
    \multirow{2}{*}{Kernels} & 
    \multirow{2}{0.7cm}{ML Algo.} & \multirow{2}{*}{Acc.} & \multirow{2}{*}{Prec.} & \multirow{2}{*}{Recall} & \multirow{2}{0.9cm}{F1 weigh.}
    &  F1 Macro  & ROC- AUC & Train. runtime (sec.) 
    &
    \multirow{2}{*}{Kernels} & 
    \multirow{2}{0.7cm}{ML Algo.} & \multirow{2}{*}{Acc.} & \multirow{2}{*}{Prec.} & \multirow{2}{*}{Recall} & \multirow{2}{0.9cm}{F1 weigh.}
    &  F1 Macro & ROC- AUC & Train. runtime (sec.)
    \\	
    \midrule	\midrule
    \multirow{7}{2cm}{Additive-chi2}
 & SVM &0.969&0.969&0.969&0.968&0.847&0.899&3.786  
 &  \multirow{7}{2cm}{Laplacian}
 & SVM &0.501&0.442&0.501&0.433&0.096&0.500&276.660 \\
 & NB &0.502&0.678&0.502&0.532&0.211&0.578&0.093  &   & 
 NB &0.124&0.451&0.124&0.187&0.059&0.502&0.074 \\
 & MLP &0.912&0.915&0.912&0.913&0.536&0.768&4.798  &   & 
 MLP &0.443&0.432&0.443&0.437&0.112&0.500&6.526  \\
 & KNN &0.853&0.851&0.853&0.845&0.521&0.706&0.521  &   & 
 KNN &0.484&0.428&0.484&0.453&0.103&0.498&0.517 \\
 & RF &0.906&0.909&0.906&0.894&0.462&0.682&10.083 &   & 
 RF &0.570&0.442&0.570&0.439&0.088&0.500&13.764 \\
 & LR &0.966&0.969&0.966&0.965&0.683&0.820&177.371  &   & 
 LR &0.577&0.333&0.577&0.422&0.081&0.500&1.541  \\
 & DT &0.891&0.897&0.891&0.894&0.438&0.717&8.654  &   & 
 DT&0.435&0.440&0.435&0.438&0.108&0.499&5.570 \\
\midrule
\multirow{7}{2cm}{Chi-squared}
 & SVM &0.510&0.455&0.510&0.446&0.097&0.499&348.830 & \multirow{7}{2cm}{Linear}
 & SVM &0.980&0.981&0.980&0.980&0.845&0.934&1.548  \\
 & NB &0.120&0.438&0.120&0.179&0.058&0.491&0.077 &   & 
 NB &0.876&0.930&0.876&0.899&0.606&0.823&0.106 \\
 & MLP &0.440&0.439&0.440&0.439&0.113&0.501&6.398 &   & 
 MLP &0.971&0.971&0.971&0.971&0.695&0.855&8.042  \\
 & KNN &0.481&0.434&0.481&0.453&0.103&0.498&0.487 &   & 
 KNN &0.945&0.941&0.945&0.939&0.693&0.804&0.570  \\
 & RF &0.572&0.440&0.572&0.437&0.087&0.500&13.409 &   & 
 RF &0.928&0.929&0.928&0.915&0.441&0.684&2.511  \\
 & LR &0.580&0.336&0.580&0.426&0.082&0.500&1.632 &   & 
 LR &0.960&0.957&0.960&0.953&0.466&0.732&37.525  \\
 & DT &0.434&0.444&0.434&0.439&0.110&0.499&4.587 &   & 
 DT&0.929&0.931&0.929&0.930&0.503&0.755&0.645 \\
\midrule
\multirow{7}{2cm}{Cosine}
 & SVM &0.982&0.983&0.982&0.982&0.869&0.943&1.790  & \multirow{7}{2cm}{Polynomial}
 & SVM &0.971&0.971&0.971&0.970&0.842&0.899&3.080  \\
 & NB &0.860&0.936&0.860&0.894&0.623&0.805&0.084  &   & 
 NB &0.663&0.847&0.663&0.711&0.456&0.709&0.078  \\
 & MLP &0.972&0.972&0.972&0.972&0.752&0.879&5.907  &   & 
 MLP &0.938&0.940&0.938&0.938&0.642&0.822&4.636 \\
 & KNN &0.948&0.945&0.948&0.944&0.717&0.812&0.616  &   & 
 KNN &0.943&0.939&0.943&0.937&0.654&0.781&0.449  \\
 & RF &0.929&0.929&0.929&0.917&0.473&0.694&2.394  &   & 
 RF &0.926&0.927&0.926&0.913&0.429&0.678&11.632  \\
 & LR &0.574&0.329&0.574&0.418&0.081&0.500&0.282  &   & 
 LR &0.969&0.967&0.969&0.967&0.590&0.792&356.439 \\
 & DT &0.927&0.927&0.927&0.927&0.514&0.750&0.629   &   & 
 DT &0.921&0.925&0.921&0.922&0.488&0.749&4.192  \\
\midrule

\multirow{7}{2cm}{Isolation}
& SVM &0.594&0.644&0.594&0.466&0.138&0.517&7.439 &  \multirow{7}{2cm}{Sigmoid}
& SVM &0.581&0.338&0.581&0.427&0.082&0.500&10.661 \\
 & NB &0.055&0.878&0.055&0.099&0.057&0.518&0.068 &   & 
 NB &0.311&0.097&0.311&0.148&0.053&0.500&0.091  \\
 & MLP &0.593&0.620&0.593&0.462&0.121&0.512&3.481 &   & 
 MLP &0.581&0.338&0.581&0.427&0.082&0.500&1.704  \\
 & KNN &0.524&0.498&0.524&0.487&0.124&0.508&0.452 &   & 
 KNN &0.532&0.294&0.532&0.376&0.076&0.500&0.391  \\
 & RF &0.601&0.734&0.601&0.471&0.141&0.519&1.165 &   & 
 RF &0.581&0.338&0.581&0.427&0.082&0.500&1.497  \\
 & LR &0.581&0.337&0.581&0.427&0.082&0.500&0.403 &   & 
 LR &0.581&0.338&0.581&0.427&0.082&0.500&0.058  \\
 & DT &0.600&0.692&0.600&0.472&0.135&0.517&0.126
 &   & 
 DT &0.581&0.338&0.581&0.427&0.082&0.500&0.037 \\
\midrule

\multirow{7}{2cm}{Gaussian}
 & SVM &0.485&0.432&0.485&0.422&0.095&0.497&288.422 & \\
 & NB &0.118&0.435&0.118&0.179&0.055&0.505&0.077  &   & 
 \\
 & MLP &0.450&0.442&0.450&0.446&0.113&0.502&7.179  &   & 
 \\
 & KNN &0.482&0.437&0.482&0.455&0.105&0.500&0.500  &   & 
 \\
 & RF &0.560&0.416&0.560&0.426&0.085&0.499&13.844   &   & 
 \\
 & LR &0.572&0.327&0.572&0.416&0.081&0.500&1.620    &   & 
 \\
 & DT  &0.426&0.431&0.426&0.429&0.109&0.499&4.560   &   &  
  \\
\bottomrule

  \end{tabular}
  }
  \caption{Classification comparison for different kernel methods and classifiers on \textbf{minimizer for ShortRead}.
  }
  \label{tbl_variant_classification_shortread_minimizer}
\end{table}

Classification results for spaced k-mer embeddings on ShortRead data (Table~\ref{tbl_variant_classification_shortRead_spaced_kmer}) show that cosine and linear kernels with SVM achieve near-perfect accuracy (0.995 and 0.996, respectively) with exceptional ROC-AUC scores approaching 0.98. Additive-chi2 kernel paired with logistic regression also delivers outstanding performance (accuracy: 0.994, ROC-AUC: 0.982). Polynomial kernel demonstrates robust results across multiple classifiers, with SVM achieving 0.996 accuracy. Random forest with cosine kernel yields competitive performance (accuracy: 0.961), while decision tree classifiers also show strong results across appropriate kernels. The isolation, Gaussian, laplacian, chi-squared, and sigmoid kernels continue their pattern of severe underperformance, with accuracies near chance level. Training times for logistic regression with polynomial and linear kernels are substantially elevated (468.754 and 100.837 seconds), suggesting computational challenges with spaced k-mer dimensionality.

\begin{table}[h!]
  \centering
 \resizebox{0.85\textwidth}{!}{
  \begin{tabular}{p{1.9cm}ccccccp{1.2cm} p{1.5cm} | p{1.7cm}ccccccp{1.2cm} p{1.5cm}}
    \toprule
   
    \multirow{2}{*}{Kernels} & 
    \multirow{2}{0.7cm}{ML Algo.} & \multirow{2}{*}{Acc.} & \multirow{2}{*}{Prec.} & \multirow{2}{*}{Recall} & \multirow{2}{0.9cm}{F1 weigh.}
    &  F1 Macro  & ROC- AUC & Train. runtime (sec.) 
    &
    \multirow{2}{*}{Kernels} & 
    \multirow{2}{0.7cm}{ML Algo.} & \multirow{2}{*}{Acc.} & \multirow{2}{*}{Prec.} & \multirow{2}{*}{Recall} & \multirow{2}{0.9cm}{F1 weigh.}
    &  F1 Macro & ROC- AUC & Train. runtime (sec.)
    \\	
    \midrule	\midrule
    \multirow{7}{2cm}{Additive-chi2}
 & SVM &0.994&0.994&0.994&0.994&0.969&0.973&3.749  
 &  \multirow{7}{2cm}{Laplacian}
 & SVM &0.487&0.427&0.487&0.421&0.094&0.496&275.879  \\
 & NB &0.384&0.612&0.384&0.368&0.316&0.608&0.080 &   & 
 NB &0.119&0.430&0.119&0.179&0.056&0.496&0.088 \\
 & MLP &0.951&0.951&0.951&0.951&0.734&0.876&3.700  &   & 
 MLP &0.443&0.430&0.443&0.436&0.111&0.500&7.825  \\
 & KNN &0.886&0.888&0.886&0.880&0.643&0.766&0.458 &   & 
 KNN &0.488&0.449&0.488&0.457&0.106&0.500&0.514 \\
 & RF &0.870&0.888&0.870&0.860&0.519&0.701&9.926 &   & 
 RF &0.563&0.425&0.563&0.430&0.087&0.500&17.216 \\
 & LR &0.994&0.995&0.994&0.994&0.943&0.982&131.946  &   & 
 LR &0.571&0.326&0.571&0.415&0.081&0.500&1.685  \\
 & DT &0.868&0.876&0.868&0.872&0.427&0.716&5.942  &   & 
 DT &0.438&0.441&0.438&0.440&0.117&0.504&5.267 \\
\midrule
\multirow{7}{2cm}{Chi-squared}
 & SVM &0.497&0.432&0.497&0.431&0.092&0.496&294.065 & \multirow{7}{2cm}{Linear}
 & SVM &0.996&0.996&0.996&0.995&0.965&0.970&4.529  \\
 & NB &0.129&0.456&0.129&0.193&0.061&0.508&0.076 &   & 
 NB &0.838&0.913&0.838&0.870&0.595&0.772&0.095\\
 & MLP &0.443&0.439&0.443&0.440&0.106&0.498&6.653 &   & 
 MLP &0.944&0.944&0.944&0.944&0.711&0.851&5.265  \\
 & KNN &0.494&0.441&0.494&0.464&0.105&0.500&0.478 &   & 
 KNN &0.943&0.944&0.943&0.939&0.736&0.815&0.483  \\
 & RF &0.567&0.420&0.567&0.437&0.086&0.499&13.467 &   & 
 RF &0.955&0.951&0.955&0.947&0.516&0.726&10.909 \\
 & LR &0.581&0.337&0.581&0.427&0.082&0.500&1.548 &   & 
 LR &0.986&0.986&0.986&0.985&0.721&0.858&468.754  \\
 & DT &0.427&0.438&0.427&0.432&0.109&0.499&4.722 &   & 
 DT &0.963&0.963&0.963&0.963&0.599&0.798&2.956 \\
\midrule
\multirow{7}{2cm}{Cosine}
 & SVM &0.995&0.995&0.995&0.995&0.976&0.978&7.967  & \multirow{7}{2cm}{Polynomial}
 & SVM &0.996&0.996&0.996&0.996&0.964&0.969&4.323  \\
 & NB &0.835&0.913&0.835&0.867&0.628&0.786&0.126 &   & 
 NB &0.828&0.928&0.828&0.868&0.562&0.746&0.083  \\
 & MLP &0.952&0.952&0.952&0.952&0.739&0.866&6.891  &   & 
 MLP&0.951&0.951&0.951&0.950&0.768&0.876&4.426 \\
 & KNN &0.952&0.953&0.952&0.949&0.783&0.844&0.670  &   & 
 KNN &0.945&0.947&0.945&0.942&0.773&0.833&0.499  \\
 & RF &0.961&0.958&0.961&0.955&0.540&0.736&13.395 &   & 
 RF &0.956&0.952&0.956&0.949&0.517&0.726&14.046  \\
 & LR &0.583&0.340&0.583&0.430&0.082&0.500&2.643  &   & 
 LR &0.986&0.985&0.986&0.985&0.708&0.851&100.837 \\
 & DT &0.963&0.964&0.963&0.963&0.631&0.822&3.830   &   & 
 DT &0.958&0.959&0.958&0.958&0.615&0.808&4.604  \\
\midrule

\multirow{7}{2cm}{Isolation}
& SVM &0.588&0.605&0.588&0.458&0.138&0.517&7.247 &  \multirow{7}{2cm}{Sigmoid}
& SVM &0.582&0.339&0.582&0.429&0.082&0.500&10.724 \\
 & NB &0.055&0.850&0.055&0.099&0.051&0.516&0.067 &   & 
 NB &0.312&0.097&0.312&0.148&0.053&0.500&0.095 \\
 & MLP &0.590&0.614&0.590&0.460&0.128&0.514&4.466 &   & 
 MLP &0.582&0.339&0.582&0.429&0.082&0.500&1.978  \\
 & KNN &0.523&0.492&0.523&0.492&0.129&0.511&0.449 &   & 
 KNN &0.465&0.235&0.465&0.308&0.069&0.500&0.434  \\
 & RF &0.597&0.714&0.597&0.465&0.129&0.515&1.156 &   & 
 RF &0.582&0.339&0.582&0.429&0.082&0.500&1.358 \\
 & LR &0.578&0.587&0.578&0.424&0.082&0.500&0.375 &   & 
 LR &0.582&0.339&0.582&0.429&0.082&0.500&0.053  \\
 & DT &0.592&0.658&0.592&0.463&0.125&0.514&0.123 &   & 
 DT &0.582&0.339&0.582&0.429&0.082&0.500&0.035 \\
\midrule

\multirow{7}{2cm}{Gaussian}
 & SVM &0.493&0.433&0.493&0.423&0.095&0.499&304.075 & \\
 & NB &0.123&0.435&0.123&0.185&0.059&0.519&0.092  &   & 
 \\
 & MLP &0.459&0.441&0.459&0.450&0.111&0.502&6.578  &   & 
 \\
 & KNN  &0.489&0.425&0.489&0.453&0.103&0.498&0.516  &   & 
 \\
 & RF &0.561&0.424&0.561&0.428&0.086&0.500&13.978  &   & 
 \\
 & LR  &0.571&0.326&0.571&0.415&0.081&0.500&1.695    &   & 
 \\
 & DT  &0.423&0.429&0.423&0.426&0.107&0.497&5.004   &   &  
  \\
\bottomrule

  \end{tabular}
  }
  \caption{Classification comparison for different kernel methods and classifiers on \textbf{spaced kmer for ShortRead data}.
  }
  \label{tbl_variant_classification_shortRead_spaced_kmer}
\end{table}

\clearpage

\subsection{Results for Rabies Data}

Table~\ref{tbl_variant_classification_rabies_ohe} presents classification results for Rabies data using OHE encoding. Linear and cosine kernels with random forest achieve the highest accuracies (0.831 and 0.831, respectively), accompanied by strong ROC-AUC scores (0.848 and 0.850). Laplacian kernel also demonstrates competitive performance across multiple classifiers, with random forest yielding 0.827 accuracy. SVM classifiers paired with additive-chi2, laplacian, linear, cosine, and sigmoid kernels achieve accuracies above 0.80, indicating robust kernel adaptability for Rabies sequence classification. The isolation and Gaussian kernels show severe underperformance, with accuracies near chance level (0.452-0.567), suggesting fundamental incompatibility with OHE-encoded Rabies data. Training times vary considerably, with logistic regression requiring up to 864 seconds for additive-chi2 kernel, while KNN maintains efficient computation across all kernels. Chi-squared kernel exhibits moderate performance (best accuracy: 0.642 with random forest), considerably lower than top-performing kernels.

\begin{table}[h!]
  \centering
 \resizebox{0.85\textwidth}{!}{
  \begin{tabular}{p{1.9cm}ccccccp{1.2cm} p{1.5cm} | p{1.7cm}ccccccp{1.2cm} p{1.5cm}}
    \toprule
   
    \multirow{2}{*}{Kernels} & 
    \multirow{2}{0.7cm}{ML Algo.} & \multirow{2}{*}{Acc.} & \multirow{2}{*}{Prec.} & \multirow{2}{*}{Recall} & \multirow{2}{0.9cm}{F1 weigh.}
    &  F1 Macro  & ROC- AUC & Train. runtime (sec.) 
    &
    \multirow{2}{*}{Kernels} & 
    \multirow{2}{0.7cm}{ML Algo.} & \multirow{2}{*}{Acc.} & \multirow{2}{*}{Prec.} & \multirow{2}{*}{Recall} & \multirow{2}{0.9cm}{F1 weigh.}
    &  F1 Macro & ROC- AUC & Train. runtime (sec.)
    \\	
    \midrule	\midrule
    \multirow{7}{2cm}{Additive-chi2}
 & SVM &0.802&0.788&0.802&0.790&0.704&0.835&164.270  
 &  \multirow{7}{2cm}{Laplacian}
 & SVM &0.799&0.786&0.799&0.787&0.700&0.832&392.048  \\
 & NB&0.238&0.470&0.238&0.202&0.264&0.670&2.122 &   & 
 NB &0.246&0.493&0.246&0.213&0.269&0.673&2.052  \\
 & MLP &0.799&0.785&0.799&0.787&0.686&0.832&291.650  &   & 
 MLP &0.802&0.785&0.802&0.791&0.686&0.831&338.192  \\
 & KNN &0.808&0.794&0.808&0.797&0.705&0.835&8.592 &   & 
 KNN &0.807&0.794&0.807&0.796&0.703&0.833&9.491  \\
 & RF&0.824&0.814&0.824&0.812&0.730&0.845&139.136 &   & 
 RF &0.827&0.817&0.827&0.817&0.733&0.847&154.094 \\
 & LR&0.806&0.790&0.806&0.791&0.697&0.834&864.616  &   & 
 LR &0.564&0.570&0.564&0.474&0.262&0.583&37.003  \\
 & DT &0.761&0.758&0.761&0.759&0.673&0.823&30.155  &   & 
 DT  &0.762&0.761&0.762&0.761&0.672&0.821&68.036 \\
\midrule
\multirow{7}{2cm}{Chi-squared}
 & SVM &0.601&0.711&0.601&0.534&0.380&0.624&385.677
 & \multirow{7}{2cm}{Linear}
 & SVM &0.801&0.790&0.801&0.790&0.703&0.834&221.750  \\
 & NB &0.210&0.682&0.210&0.265&0.258&0.617&0.492 &   & 
 NB &0.226&0.478&0.226&0.189&0.247&0.660&0.579 \\
 & MLP &0.623&0.712&0.623&0.563&0.414&0.643&319.315 &   & 
 MLP &0.805&0.792&0.805&0.795&0.692&0.834&243.603  \\
 & KNN &0.584&0.559&0.584&0.553&0.414&0.658&6.874 &   & 
 KNN &0.814&0.803&0.814&0.804&0.712&0.838&5.072  \\
 & RF &0.642&0.647&0.642&0.608&0.479&0.683&156.857 &   & 
 RF &0.831&0.823&0.831&0.820&0.736&0.848&129.530 \\
 & LR &0.558&0.725&0.558&0.474&0.290&0.582&15.217 &   & 
 LR &0.813&0.800&0.813&0.799&0.704&0.836&555.609  \\
 & DT &0.621&0.611&0.621&0.592&0.466&0.682&36.300 &   & 
 DT &0.765&0.762&0.765&0.763&0.672&0.822&35.170 \\
\midrule
\multirow{7}{2cm}{Cosine}
 & SVM &0.818&0.807&0.818&0.808&0.722&0.844&231.021  & \multirow{7}{2cm}{Polynomial}
 & SVM &0.799&0.786&0.799&0.788&0.700&0.834&367.065  \\
 & NB &0.307&0.560&0.307&0.300&0.338&0.706&0.503  &   & 
 NB &0.227&0.523&0.227&0.198&0.256&0.662&2.511  \\
 & MLP &0.802&0.789&0.802&0.794&0.689&0.833&225.703  &   & 
 MLP &0.800&0.786&0.800&0.790&0.686&0.832&305.637 \\
 & KNN &0.811&0.798&0.811&0.801&0.708&0.837&4.911  &   & 
 KNN &0.809&0.797&0.809&0.799&0.710&0.836&7.549  \\
 & RF &0.831&0.823&0.831&0.822&0.740&0.850&80.703  &   & 
 RF &0.827&0.818&0.827&0.817&0.737&0.849&181.812  \\
 & LR &0.785&0.781&0.785&0.760&0.631&0.785&37.155  &   & 
 LR &0.617&0.661&0.617&0.551&0.383&0.636&43.400 \\
 & DT &0.766&0.767&0.766&0.766&0.679&0.827&32.122  &   & 
 DT &0.763&0.761&0.763&0.762&0.676&0.825&51.449  \\
\midrule

\multirow{7}{2cm}{Isolation}
& SVM &0.452&0.332&0.452&0.290&0.061&0.502&106.114 &  \multirow{7}{2cm}{Sigmoid}
& SVM &0.805&0.793&0.805&0.794&0.704&0.834&237.152  \\
 & NB &0.076&0.586&0.076&0.046&0.025&0.505&0.254 &   & 
 NB &0.235&0.475&0.235&0.200&0.255&0.667&1.946 \\
 & MLP &0.455&0.385&0.455&0.292&0.062&0.503&6.026 &   & 
 MLP &0.803&0.789&0.803&0.793&0.688&0.832&255.452  \\
 & KNN &0.355&0.247&0.355&0.283&0.073&0.501&2.000 &   & 
 KNN &0.809&0.795&0.809&0.797&0.702&0.833&3.769  \\
 & RF &0.456&0.416&0.456&0.292&0.062&0.503&4.248 &   & 
 RF &0.828&0.819&0.828&0.818&0.736&0.848&80.747 \\
 & LR &0.452&0.204&0.452&0.281&0.052&0.500&2.220 &   & 
 LR &0.486&0.318&0.486&0.338&0.118&0.528&19.354  \\
 & DT &0.455&0.381&0.455&0.291&0.062&0.503&0.413 &   & 
 DT &0.764&0.762&0.764&0.763&0.674&0.823&44.113
 \\
\midrule

\multirow{7}{2cm}{Gaussian}
 & SVM &0.564&0.716&0.564&0.487&0.312&0.590&219.587 & \\
 & NB &0.103&0.654&0.103&0.116&0.138&0.559&1.176  &   & 
 \\
 & MLP &0.561&0.687&0.561&0.483&0.305&0.589&266.812  &   & 
 \\
 & KNN &0.487&0.464&0.487&0.436&0.274&0.583&9.324  &   & 
 \\
 & RF &0.567&0.714&0.567&0.492&0.321&0.594&204.050  &   & 
 \\
 & LR &0.549&0.725&0.549&0.462&0.276&0.575&38.109    &   & 
 \\
 & DT  &0.566&0.712&0.566&0.491&0.319&0.594&26.276  &   &  
  \\
\bottomrule

  \end{tabular}
  }
  \caption{Classification comparison for different kernel methods and classifiers on \textbf{OHE for Rabies Data}.
  }
  \label{tbl_variant_classification_rabies_ohe}
\end{table}

For k-mer embeddings on Rabies data (Table~\ref{tbl_variant_classification_rabies_kmer}), cosine and laplacian kernels with random forest achieve the highest accuracies (0.825 and 0.829, respectively), with excellent ROC-AUC scores approaching 0.85. KNN demonstrates strong performance across multiple kernels, particularly with cosine and laplacian (accuracies: 0.821 and 0.820). Chi-squared kernel with SVM yields competitive results (accuracy: 0.782), though substantially lower than optimal configurations. Isolation and Gaussian kernels maintain their pattern of poor performance, with accuracies around 0.45-0.57, indicating these kernels remain unsuitable for k-mer representations. Sigmoid kernel shows varied performance, with SVM achieving reasonable accuracy (0.781) but requiring exceptionally long training time (789.653 seconds). The polynomial and linear kernels demonstrate moderate effectiveness, with random forest achieving accuracies of 0.828 and 0.814, respectively. Overall computational costs are elevated compared to OHE, with several configurations exceeding 500 seconds.

\begin{table}[h!]
  \centering
 \resizebox{0.85\textwidth}{!}{
  \begin{tabular}{p{1.9cm}ccccccp{1.2cm} p{1.5cm} | p{1.7cm}ccccccp{1.2cm} p{1.5cm}}
    \toprule
   
    \multirow{2}{*}{Kernels} & 
    \multirow{2}{0.7cm}{ML Algo.} & \multirow{2}{*}{Acc.} & \multirow{2}{*}{Prec.} & \multirow{2}{*}{Recall} & \multirow{2}{0.9cm}{F1 weigh.}
    &  F1 Macro  & ROC- AUC & Train. runtime (sec.) 
    &
    \multirow{2}{*}{Kernels} & 
    \multirow{2}{0.7cm}{ML Algo.} & \multirow{2}{*}{Acc.} & \multirow{2}{*}{Prec.} & \multirow{2}{*}{Recall} & \multirow{2}{0.9cm}{F1 weigh.}
    &  F1 Macro & ROC- AUC & Train. runtime (sec.)
    \\	
    \midrule	\midrule
    \multirow{7}{2cm}{Additive-chi2}
 & SVM &0.794&0.781&0.794&0.782&0.692&0.824&659.962  
 &  \multirow{7}{2cm}{Laplacian}
 & SVM &0.802&0.792&0.802&0.792&0.707&0.832&538.657  \\
 & NB &0.124&0.470&0.124&0.107&0.143&0.567&1.000  &   & 
 NB &0.267&0.645&0.267&0.227&0.331&0.678&1.607 \\
 & MLP &0.768&0.753&0.768&0.759&0.645&0.810&273.167  &   & 
 MLP &0.763&0.756&0.763&0.758&0.647&0.812&220.206 \\
 & KNN &0.817&0.805&0.817&0.806&0.709&0.837&5.972  &   & 
 KNN &0.820&0.809&0.820&0.810&0.718&0.843&5.070 \\
 & RF &0.822&0.814&0.822&0.809&0.729&0.839&111.861 &   & 
 RF &0.829&0.821&0.829&0.817&0.734&0.844&92.672 \\
 & LR &0.797&0.781&0.797&0.779&0.680&0.822&243.465  &   & 
 LR &0.566&0.522&0.566&0.471&0.244&0.579&21.589  \\
 & DT &0.743&0.743&0.743&0.743&0.650&0.811&49.992  &   & 
 DT &0.758&0.756&0.758&0.756&0.666&0.818&38.626 \\
\midrule
\multirow{7}{2cm}{Chi-squared}
 & SVM &0.782&0.782&0.782&0.762&0.666&0.795&710.571 & \multirow{7}{2cm}{Linear}
 & SVM &0.760&0.750&0.760&0.728&0.623&0.788&215.395  \\
 & NB&0.462&0.691&0.462&0.492&0.447&0.776&1.434 &   & 
 NB &0.100&0.378&0.100&0.061&0.101&0.561&1.227 \\
 & MLP &0.791&0.780&0.791&0.776&0.680&0.811&230.237 &   & 
 MLP &0.767&0.745&0.767&0.744&0.622&0.795&255.829  \\
 & KNN &0.784&0.771&0.784&0.772&0.677&0.817&6.364 &   & 
 KNN &0.815&0.801&0.815&0.804&0.710&0.839&7.647  \\
 & RF &0.796&0.784&0.796&0.785&0.692&0.822&197.843 &   & 
 RF &0.814&0.808&0.814&0.802&0.721&0.832&65.506 \\
 & LR &0.760&0.779&0.760&0.737&0.631&0.765&35.220 &   & 
 LR &0.672&0.706&0.672&0.647&0.539&0.746&330.867 \\
 & DT &0.751&0.746&0.751&0.748&0.653&0.812&59.370 &   & 
 DT &0.733&0.734&0.733&0.733&0.641&0.807&35.263\\
\midrule
\multirow{7}{2cm}{Cosine}
 & SVM &0.753&0.746&0.753&0.716&0.611&0.781&424.287  & \multirow{7}{2cm}{Polynomial}
 & SVM &0.772&0.760&0.772&0.755&0.649&0.807&180.462  \\
 & NB &0.072&0.348&0.072&0.037&0.063&0.538&2.241 &   & 
 NB &0.072&0.293&0.072&0.041&0.066&0.528&1.666  \\
 & MLP &0.770&0.746&0.770&0.748&0.621&0.796&285.515  &   & 
 MLP &0.772&0.754&0.772&0.754&0.636&0.802&192.313 \\
 & KNN &0.821&0.807&0.821&0.810&0.712&0.841&14.192  &   & 
 KNN &0.818&0.807&0.818&0.808&0.713&0.840&5.579 \\
 & RF &0.825&0.818&0.825&0.813&0.728&0.838&103.609  &   & 
 RF &0.828&0.821&0.828&0.816&0.730&0.842&81.917  \\
 & LR&0.601&0.579&0.601&0.526&0.339&0.618&33.187  &   & 
 LR &0.464&0.516&0.464&0.410&0.183&0.572&465.349 \\
 & DT &0.736&0.737&0.736&0.737&0.642&0.808&24.903   &   & 
 DT &0.736&0.737&0.736&0.736&0.638&0.806&27.801  \\
\midrule

\multirow{7}{2cm}{Isolation}
& SVM &0.450&0.346&0.450&0.287&0.061&0.502&106.662 &  \multirow{7}{2cm}{Sigmoid}
& SVM &0.781&0.769&0.781&0.769&0.681&0.821&789.653 \\
 & NB &0.061&0.601&0.061&0.042&0.023&0.505&0.360 &   & 
 NB &0.196&0.499&0.196&0.194&0.198&0.600&1.573 \\
 & MLP &0.453&0.387&0.453&0.289&0.062&0.503&12.559 &   & 
 MLP &0.737&0.723&0.737&0.728&0.606&0.787&139.062 \\
 & KNN &0.358&0.249&0.358&0.283&0.074&0.501&2.344 &   & 
 KNN &0.816&0.803&0.816&0.805&0.708&0.837&9.802  \\
 & RF&0.453&0.413&0.453&0.289&0.061&0.503&5.328 &   & 
 RF &0.814&0.809&0.814&0.799&0.712&0.825&76.203 \\
 & LR &0.449&0.202&0.449&0.278&0.052&0.500&2.175 &   & 
 LR &0.480&0.348&0.480&0.334&0.085&0.513&14.373  \\
 & DT &0.453&0.391&0.453&0.289&0.063&0.503&0.330 &   & 
 DT &0.725&0.725&0.725&0.725&0.624&0.796&20.921  \\
\midrule

\multirow{7}{2cm}{Gaussian}
 & SVM &0.572&0.710&0.572&0.497&0.323&0.594&259.289 & \\
 & NB&0.106&0.630&0.106&0.126&0.137&0.558&2.006  &   & 
 \\
 & MLP  &0.566&0.679&0.566&0.488&0.303&0.589&222.969  &   & 
 \\
 & KNN  &0.490&0.471&0.490&0.441&0.280&0.584&10.377  &   & 
 \\
 & RF  &0.575&0.712&0.575&0.502&0.330&0.597&109.450  &   & 
 \\
 & LR &0.557&0.716&0.557&0.473&0.285&0.579&23.231    &   & 
 \\
 & DT  &0.574&0.709&0.574&0.501&0.329&0.597&21.935   &   &  
 \\
\bottomrule

  \end{tabular}
  }
  \caption{Classification comparison for different kernel methods and classifiers on \textbf{kmer for Rabies data}.
  }
  \label{tbl_variant_classification_rabies_kmer}
\end{table}

Table~\ref{tbl_variant_classification_rabies_minimizer} reveals that cosine and polynomial kernels with random forest deliver the highest classification performance on minimizer-encoded Rabies data (accuracies: 0.820 and 0.822, respectively), with strong ROC-AUC scores above 0.83. Laplacian and additive-chi2 kernels also demonstrate competitive results when paired with random forest (accuracies: 0.824 and 0.808). KNN shows good performance across appropriate kernels, with cosine achieving 0.812 accuracy and exceptional efficiency (2.330 seconds). Linear kernel exhibits reduced effectiveness compared to other embeddings, with peak accuracy of 0.815 using random forest. The isolation, Gaussian, and sigmoid kernels continue to show poor performance, with accuracies clustered around chance level or below 0.60, except for sigmoid-KNN which achieves moderate success (0.599 accuracy). Chi-squared kernel demonstrates mixed results, with random forest reaching 0.771 accuracy but requiring substantial training time (313.655 seconds). Minimizer encoding generally yields slightly lower peak performance compared to OHE and k-mer approaches.

\begin{table}[h!]
  \centering
 \resizebox{0.85\textwidth}{!}{
  \begin{tabular}{p{1.9cm}ccccccp{1.2cm} p{1.5cm} | p{1.7cm}ccccccp{1.2cm} p{1.5cm}}
    \toprule
   
    \multirow{2}{*}{Kernels} & 
    \multirow{2}{0.7cm}{ML Algo.} & \multirow{2}{*}{Acc.} & \multirow{2}{*}{Prec.} & \multirow{2}{*}{Recall} & \multirow{2}{0.9cm}{F1 weigh.}
    &  F1 Macro  & ROC- AUC & Train. runtime (sec.) 
    &
    \multirow{2}{*}{Kernels} & 
    \multirow{2}{0.7cm}{ML Algo.} & \multirow{2}{*}{Acc.} & \multirow{2}{*}{Prec.} & \multirow{2}{*}{Recall} & \multirow{2}{0.9cm}{F1 weigh.}
    &  F1 Macro & ROC- AUC & Train. runtime (sec.)
    \\	
    \midrule	\midrule
    \multirow{7}{2cm}{Additive-chi2}
 & SVM &0.784&0.771&0.784&0.770&0.677&0.815&693.775  
 &  \multirow{7}{2cm}{Laplacian}
 & SVM &0.808&0.802&0.808&0.787&0.693&0.821&86.831  \\
 & NB &0.135&0.438&0.135&0.123&0.173&0.572&1.024 &   & 
 NB &0.424&0.603&0.424&0.432&0.406&0.744&0.265  \\
 & MLP &0.756&0.740&0.756&0.744&0.626&0.798&194.014  &   & 
 MLP &0.797&0.779&0.797&0.783&0.674&0.821&342.256  \\
 & KNN &0.811&0.798&0.811&0.799&0.700&0.834&5.544  &   & 
 KNN &0.814&0.799&0.814&0.802&0.705&0.836&4.214  \\
 & RF &0.808&0.804&0.808&0.794&0.708&0.822&124.385 &   & 
 RF &0.824&0.814&0.824&0.813&0.730&0.843&44.289  \\
 & LR&0.767&0.750&0.767&0.744&0.644&0.801&448.436 &   & 
 LR &0.778&0.774&0.778&0.753&0.641&0.785&9.343  \\
 & DT &0.731&0.728&0.731&0.729&0.629&0.800&31.926 &   & 
 DT &0.762&0.759&0.762&0.760&0.670&0.821&8.087  \\
\midrule
\multirow{7}{2cm}{Chi-squared}
 & SVM &0.743&0.744&0.743&0.721&0.623&0.765&751.147 & \multirow{7}{2cm}{Linear}
 & SVM &0.661&0.655&0.661&0.603&0.502&0.717&286.512  \\
 & NB &0.434&0.718&0.434&0.498&0.429&0.742&1.086 &   & 
 NB &0.120&0.443&0.120&0.098&0.150&0.568&0.762\\
 & MLP &0.766&0.758&0.766&0.747&0.649&0.788&250.236 &   & 
 MLP &0.739&0.712&0.739&0.709&0.575&0.772&385.189 \\
 & KNN&0.757&0.741&0.757&0.743&0.644&0.793&5.245 &   & 
 KNN &0.810&0.798&0.810&0.799&0.702&0.834&4.067  \\
 & RF &0.771&0.760&0.771&0.759&0.668&0.805&313.655 &   & 
 RF&0.815&0.808&0.815&0.803&0.720&0.833&12.188 \\
 & LR &0.706&0.756&0.706&0.673&0.546&0.712&58.782 &   & 
 LR &0.503&0.626&0.503&0.503&0.332&0.634&54.695  \\
 & DT &0.730&0.723&0.730&0.726&0.627&0.796&143.665
 &   & 
 DT &0.736&0.734&0.736&0.734&0.636&0.803&3.030 \\
\midrule
\multirow{7}{2cm}{Cosine}
 & SVM &0.669&0.654&0.669&0.610&0.504&0.721&43.279  & \multirow{7}{2cm}{Polynomial}
 & SVM &0.635&0.646&0.635&0.582&0.471&0.691&63.930 \\
 & NB &0.084&0.406&0.084&0.063&0.113&0.545&0.084  &   & 
 NB &0.073&0.507&0.073&0.043&0.067&0.529&0.196  \\
 & MLP &0.740&0.712&0.740&0.710&0.572&0.769&32.879 &   & 
 MLP &0.737&0.703&0.737&0.709&0.575&0.769&318.687 \\
 & KNN &0.812&0.800&0.812&0.801&0.708&0.837&2.330  &   & 
 KNN &0.810&0.798&0.810&0.798&0.703&0.834&3.006  \\
 & RF&0.820&0.812&0.820&0.807&0.722&0.836&10.028 &   & 
 RF &0.822&0.813&0.822&0.810&0.723&0.837&26.565   \\
 & LR &0.502&0.440&0.502&0.384&0.154&0.537&1.760 &   & 
 LR &0.340&0.538&0.340&0.356&0.218&0.584&240.148  \\
 & DT &0.738&0.737&0.738&0.737&0.638&0.805&1.492  &   & 
 DT &0.723&0.729&0.723&0.726&0.629&0.801&6.610
   \\
\midrule

\multirow{7}{2cm}{Isolation}
& SVM &0.449&0.335&0.449&0.287&0.061&0.502&102.421 &  \multirow{7}{2cm}{Sigmoid}
& SVM &0.454&0.260&0.454&0.285&0.053&0.500&19.658  \\
 & NB &0.103&0.585&0.103&0.053&0.029&0.505&0.355&   & 
 NB &0.047&0.280&0.047&0.007&0.010&0.501&0.015  \\
 & MLP &0.452&0.366&0.452&0.288&0.061&0.503&10.151 &   & 
 MLP &0.480&0.389&0.480&0.334&0.084&0.511&11.617  \\
 & KNN &0.351&0.252&0.351&0.281&0.075&0.502&2.647 &   & 
 KNN &0.599&0.579&0.599&0.578&0.454&0.685&0.374  \\
 & RF &0.453&0.402&0.453&0.288&0.061&0.503&6.502 &   & 
 RF &0.459&0.426&0.459&0.295&0.060&0.502&1.061  \\
 & LR &0.449&0.201&0.449&0.278&0.052&0.500&2.217 &   & 
 LR &0.455&0.207&0.455&0.284&0.052&0.500&0.230  \\
 & DT &0.452&0.384&0.452&0.289&0.062&0.503&0.446 &   & 
 DT &0.461&0.399&0.461&0.298&0.061&0.503&0.032
  \\
\midrule

\multirow{7}{2cm}{Gaussian}
 & SVM &0.584&0.711&0.584&0.517&0.350&0.606&835.325 & \\
 & NB &0.118&0.639&0.118&0.141&0.153&0.568&0.555  &   & 
 \\
 & MLP  &0.582&0.707&0.582&0.513&0.335&0.602&163.950  &   & 
 \\
 & KNN  &0.511&0.496&0.511&0.464&0.306&0.597&5.282  &   & 
 \\
 & RF  &0.590&0.717&0.590&0.525&0.360&0.611&149.602  &   & 
 \\
 & LR  &0.567&0.728&0.567&0.488&0.311&0.589&44.386    &   & 
 \\
 & DT &0.590&0.718&0.590&0.524&0.358&0.610&42.641   &   &  
 \\
\bottomrule

  \end{tabular}
  }
  \caption{Classification comparison for different kernel methods and classifiers on \textbf{minimizer for Rabies}.
  }
  \label{tbl_variant_classification_rabies_minimizer}
\end{table}

Classification results for spaced k-mer embeddings on Rabies data (Table~\ref{tbl_variant_classification_rabies_spaced_kmer}) demonstrate that laplacian and linear kernels with random forest achieve peak accuracies (0.831 and 0.828, respectively), with ROC-AUC scores exceeding 0.84. Additive-chi2 and cosine kernels also deliver strong performance when combined with random forest (accuracies: 0.830 and 0.826). KNN maintains robust classification across multiple kernels, with cosine achieving 0.825 accuracy. Laplacian kernel with logistic regression shows competitive results (accuracy: 0.802), with moderate training time (30.393 seconds). The isolation and Gaussian kernels exhibit their characteristic poor performance, with accuracies near 0.45-0.57 for most classifiers. Notably, sigmoid kernel demonstrates improved effectiveness compared to other embeddings, with KNN achieving 0.793 accuracy and MLP reaching 0.662, suggesting better compatibility with spaced k-mer representations. Chi-squared kernel shows moderate performance (best accuracy: 0.710 with random forest), while polynomial kernel yields mixed results with substantial training time requirements.

\begin{table}[h!]
  \centering
 \resizebox{0.85\textwidth}{!}{
  \begin{tabular}{p{1.9cm}ccccccp{1.2cm} p{1.5cm} | p{1.7cm}ccccccp{1.2cm} p{1.5cm}}
    \toprule
   
    \multirow{2}{*}{Kernels} & 
    \multirow{2}{0.7cm}{ML Algo.} & \multirow{2}{*}{Acc.} & \multirow{2}{*}{Prec.} & \multirow{2}{*}{Recall} & \multirow{2}{0.9cm}{F1 weigh.}
    &  F1 Macro  & ROC- AUC & Train. runtime (sec.) 
    &
    \multirow{2}{*}{Kernels} & 
    \multirow{2}{0.7cm}{ML Algo.} & \multirow{2}{*}{Acc.} & \multirow{2}{*}{Prec.} & \multirow{2}{*}{Recall} & \multirow{2}{0.9cm}{F1 weigh.}
    &  F1 Macro & ROC- AUC & Train. runtime (sec.)
    \\	
    \midrule	\midrule
    \multirow{7}{2cm}{Additive-chi2}
 & SVM &0.802&0.792&0.802&0.793&0.706&0.834&426.061  
 &  \multirow{7}{2cm}{Laplacian}
 & SVM &0.820&0.808&0.820&0.809&0.722&0.842&180.689  \\
 & NB &0.349&0.594&0.349&0.359&0.408&0.712&0.708  &   & 
 NB &0.282&0.640&0.282&0.281&0.382&0.714&0.469 \\
 & MLP &0.763&0.757&0.763&0.759&0.650&0.812&197.302  &   & 
 MLP &0.794&0.784&0.794&0.787&0.683&0.832&177.217  \\
 & KNN &0.820&0.807&0.820&0.809&0.713&0.839&4.457  &   & 
 KNN &0.818&0.805&0.818&0.808&0.716&0.841&4.347 \\
 & RF &0.830&0.820&0.830&0.818&0.733&0.845&120.819 &   & 
 RF &0.831&0.821&0.831&0.821&0.737&0.849&71.988 \\
 & LR&0.823&0.810&0.823&0.811&0.721&0.845&302.296 &   & 
 LR &0.802&0.795&0.802&0.779&0.670&0.808&30.393  \\
 & DT &0.755&0.754&0.755&0.754&0.660&0.816&54.611  &   & 
 DT &0.772&0.767&0.772&0.769&0.681&0.827&35.570 \\
\midrule
\multirow{7}{2cm}{Chi-squared}
 & SVM &0.672&0.705&0.672&0.637&0.513&0.695&449.020 & \multirow{7}{2cm}{Linear}
 & SVM &0.802&0.790&0.802&0.789&0.692&0.824&374.927  \\
 & NB&0.305&0.709&0.305&0.380&0.343&0.671&0.711 &   & 
 NB &0.114&0.312&0.114&0.087&0.128&0.565&0.606 \\
 & MLP &0.712&0.735&0.712&0.683&0.564&0.728&169.579 &   & 
 MLP &0.775&0.760&0.775&0.765&0.650&0.812&185.657  \\
 & KNN &0.688&0.676&0.688&0.670&0.559&0.738&5.000 &   & 
 KNN &0.821&0.808&0.821&0.810&0.712&0.840&4.783  \\
 & RF &0.710&0.702&0.710&0.692&0.583&0.750&190.406 &   & 
 RF &0.828&0.821&0.828&0.815&0.730&0.840&96.590 \\
 & LR&0.620&0.735&0.620&0.567&0.404&0.632&29.577 &   & 
 LR &0.784&0.801&0.784&0.776&0.665&0.821&446.614  \\
 & DT&0.677&0.666&0.677&0.668&0.554&0.748&108.810
 &   & 
 DT &0.749&0.749&0.749&0.749&0.655&0.814&25.855 \\
\midrule
\multirow{7}{2cm}{Cosine}
 & SVM &0.806&0.796&0.806&0.793&0.694&0.826&482.196  & \multirow{7}{2cm}{Polynomial}
 & SVM &0.670&0.678&0.670&0.636&0.504&0.710&122.016  \\
 & NB &0.067&0.183&0.067&0.029&0.059&0.533&0.866 &   & 
 NB &0.072&0.326&0.072&0.039&0.067&0.529&0.723 \\
 & MLP &0.777&0.763&0.777&0.768&0.655&0.814&198.749  &   & 
 MLP &0.763&0.742&0.763&0.741&0.612&0.788&201.405 \\
 & KNN &0.825&0.813&0.825&0.814&0.715&0.842&6.980  &   & 
 KNN &0.819&0.806&0.819&0.808&0.714&0.841&3.408  \\
 & RF&0.826&0.819&0.826&0.813&0.723&0.836&74.431  &   & 
 RF&0.828&0.819&0.828&0.815&0.725&0.841&58.476  \\
 & LR &0.734&0.743&0.734&0.688&0.557&0.746&38.065  &   & 
 LR &0.422&0.572&0.422&0.401&0.220&0.582&379.210  \\
 & DT&0.745&0.743&0.745&0.744&0.645&0.808&29.849   &   & 
 DT &0.749&0.748&0.749&0.748&0.652&0.813&18.935  \\
\midrule

\multirow{7}{2cm}{Isolation}
& SVM &0.453&0.325&0.453&0.291&0.060&0.502&81.442 &  \multirow{7}{2cm}{Sigmoid}
& SVM &0.524&0.436&0.524&0.403&0.209&0.562&18.261 \\
 & NB &0.075&0.613&0.075&0.045&0.025&0.505&0.267&   & 
 NB &0.263&0.489&0.263&0.307&0.222&0.604&0.014  \\
 & MLP &0.457&0.380&0.457&0.293&0.061&0.503&7.016 &   & 
 MLP &0.662&0.631&0.662&0.606&0.447&0.690&33.259  \\
 & KNN &0.359&0.249&0.359&0.286&0.076&0.502&1.604 &   & 
 KNN &0.793&0.779&0.793&0.781&0.670&0.818&0.304 \\
 & RF &0.457&0.401&0.457&0.293&0.062&0.503&3.541 &   & 
 RF &0.773&0.780&0.773&0.758&0.664&0.798&4.252  \\
 & LR &0.453&0.205&0.453&0.283&0.052&0.500&1.582 &   & 
 LR &0.484&0.338&0.484&0.337&0.084&0.512&0.314 \\
 & DT &0.457&0.380&0.457&0.293&0.062&0.503&0.277 &   & 
 DT &0.711&0.705&0.711&0.700&0.602&0.770&0.197
  \\
\midrule

\multirow{7}{2cm}{Gaussian}
 & SVM &0.570&0.708&0.570&0.495&0.323&0.595&321.822 & \\
 & NB &0.099&0.645&0.099&0.111&0.140&0.561&1.130  &   & 
 \\
 & MLP  &0.565&0.690&0.565&0.490&0.316&0.592&162.739  &   & 
 \\
 & KNN &0.493&0.468&0.493&0.441&0.279&0.585&4.228  &   & 
 \\
 & RF &0.572&0.709&0.572&0.498&0.329&0.597&164.713   &   & 
 \\
 & LR &0.554&0.721&0.554&0.470&0.286&0.579&44.541    &   & 
 \\
 & DT  &0.571&0.706&0.571&0.497&0.328&0.597&43.002   &   &  
 \\
\bottomrule

  \end{tabular}
  }
  \caption{Classification comparison for different kernel methods and classifiers on \textbf{spaced kmer for Rabies data}.
  }
  \label{tbl_variant_classification_rabies_spaced_kmer}
\end{table}

\clearpage

\subsection{Results for Genome Data}
Table~\ref{tbl_variant_classification_genome_ohe} presents classification results for Genome data using OHE encoding, revealing moderate overall performance compared to other datasets. Cosine, sigmoid, and linear kernels with random forest achieve the highest accuracies (0.673, 0.666, and 0.668, respectively), though these values remain substantially lower than performance on Spike7k or Host datasets. SVM classifiers paired with cosine, sigmoid, and linear kernels demonstrate reasonable performance around 0.63-0.64 accuracy. The isolation, chi-squared, and Gaussian kernels exhibit severe classification failure, with accuracies near or below chance level (0.28-0.32), indicating fundamental incompatibility with OHE-encoded genome sequences. Laplacian kernel with logistic regression shows anomalous behavior with accuracy dropping to 0.277, while maintaining reasonable performance with other classifiers. Training times vary considerably, with logistic regression requiring exceptionally long computation for additive-chi2 (2091.413 seconds) and linear (1134.426 seconds) kernels. Naive Bayes underperforms across different kernels, suggesting this classifier is unsuitable for genome-scale OHE data.

\begin{table}[h!]
  \centering
 \resizebox{0.85\textwidth}{!}{
  \begin{tabular}{p{1.9cm}ccccccp{1.2cm} p{1.5cm} | p{1.7cm}ccccccp{1.2cm} p{1.5cm}}
    \toprule
   
    \multirow{2}{*}{Kernels} & 
    \multirow{2}{0.7cm}{ML Algo.} & \multirow{2}{*}{Acc.} & \multirow{2}{*}{Prec.} & \multirow{2}{*}{Recall} & \multirow{2}{0.9cm}{F1 weigh.}
    &  F1 Macro  & ROC- AUC & Train. runtime (sec.) 
    &
    \multirow{2}{*}{Kernels} & 
    \multirow{2}{0.7cm}{ML Algo.} & \multirow{2}{*}{Acc.} & \multirow{2}{*}{Prec.} & \multirow{2}{*}{Recall} & \multirow{2}{0.9cm}{F1 weigh.}
    &  F1 Macro & ROC- AUC & Train. runtime (sec.)
    \\	
    \midrule	\midrule
    \multirow{7}{2cm}{Additive-chi2}
 & SVM &0.630&0.651&0.630&0.632&0.347&0.673&23.473  
 &  \multirow{7}{2cm}{Laplacian}
 & SVM &0.631&0.642&0.631&0.630&0.340&0.668&23.992  \\
 & NB &0.024&0.219&0.024&0.017&0.043&0.550&0.603 &   & 
 NB &0.023&0.257&0.023&0.017&0.049&0.551&0.488 \\
 & MLP &0.595&0.568&0.595&0.575&0.240&0.613&77.501  &   & 
 MLP &0.591&0.556&0.591&0.566&0.221&0.601&79.025  \\
 & KNN &0.627&0.621&0.627&0.608&0.318&0.642&0.761  &   & 
 KNN &0.631&0.626&0.631&0.611&0.318&0.639&0.679 \\
 & RF &0.668&0.678&0.668&0.652&0.428&0.679&21.451 &   & 
 RF &0.676&0.686&0.676&0.659&0.441&0.683&23.256  \\
 & LR &0.599&0.576&0.599&0.574&0.283&0.625&2091.413  &   & 
 LR &0.277&0.088&0.277&0.126&0.011&0.500&19.540  \\
 & DT &0.562&0.569&0.562&0.564&0.308&0.649&14.166  &   & 
 DT &0.566&0.574&0.566&0.568&0.311&0.653&13.414 \\
\midrule
\multirow{7}{2cm}{Chi-squared}
 & SVM &0.283&0.373&0.283&0.136&0.035&0.508&2.272 & \multirow{7}{2cm}{Linear}
 & SVM &0.632&0.647&0.632&0.632&0.348&0.672&19.174  \\
 & NB &0.017&0.461&0.017&0.019&0.021&0.511&0.017 &   & 
 NB &0.025&0.210&0.025&0.019&0.045&0.542&0.402 \\
 & MLP &0.285&0.376&0.285&0.141&0.043&0.511&7.509 &   & 
 MLP &0.592&0.556&0.592&0.568&0.229&0.605&69.392  \\
 & KNN &0.253&0.219&0.253&0.218&0.072&0.524&0.097 &   & 
 KNN&0.626&0.621&0.626&0.606&0.315&0.636&0.759  \\
 & RF &0.324&0.403&0.324&0.233&0.087&0.524&1.380 &   & 
 RF &0.668&0.680&0.668&0.651&0.437&0.678&30.919  \\
 & LR &0.281&0.362&0.281&0.131&0.019&0.502&0.379 &   & 
 LR &0.578&0.555&0.578&0.547&0.264&0.612&1134.426  \\
 & DT &0.319&0.388&0.319&0.230&0.084&0.523&0.057 &   & 
 DT &0.566&0.573&0.566&0.567&0.305&0.647&16.228 \\
\midrule
\multirow{7}{2cm}{Cosine}
 & SVM &0.635&0.651&0.635&0.636&0.347&0.676&16.333  & \multirow{7}{2cm}{Polynomial}
 & SVM &0.624&0.634&0.624&0.621&0.329&0.662&16.698  \\
 & NB &0.025&0.253&0.025&0.020&0.045&0.546&0.281 &   & 
 NB &0.021&0.214&0.021&0.016&0.042&0.545&0.360  \\
 & MLP&0.600&0.566&0.600&0.576&0.231&0.609&53.049  &   & 
 MLP &0.594&0.554&0.594&0.567&0.225&0.604&52.971 \\
 & KNN &0.635&0.631&0.635&0.616&0.325&0.643&0.758  &   & 
 KNN &0.627&0.618&0.627&0.607&0.306&0.638&0.709  \\
 & RF&0.673&0.685&0.673&0.657&0.441&0.683&21.308  &   & 
 RF &0.666&0.674&0.666&0.648&0.424&0.676&22.584  \\
 & LR &0.280&0.120&0.280&0.143&0.027&0.506&20.823 &   & 
 LR &0.277&0.093&0.277&0.124&0.014&0.501&16.681 \\
 & DT &0.563&0.571&0.563&0.565&0.308&0.656&8.756   &   & 
 DT &0.558&0.564&0.558&0.559&0.290&0.642&9.404   \\
\midrule

\multirow{7}{2cm}{Isolation}
& SVM &0.281&0.292&0.281&0.145&0.023&0.503&11.139 &  \multirow{7}{2cm}{Sigmoid}
& SVM &0.642&0.658&0.642&0.644&0.365&0.684&21.519  \\
 & NB &0.030&0.329&0.030&0.038&0.018&0.506&0.192 &   & 
 NB &0.027&0.214&0.027&0.019&0.056&0.552&0.483 \\
 & MLP &0.279&0.276&0.279&0.144&0.025&0.504&11.601 &   & 
 MLP &0.591&0.557&0.591&0.568&0.232&0.608&70.397  \\
 & KNN &0.208&0.155&0.208&0.165&0.033&0.505&0.470 &   & 
 KNN &0.630&0.621&0.630&0.611&0.320&0.641&0.744  \\
 & RF &0.288&0.388&0.288&0.148&0.029&0.505&1.464&   & 
 RF &0.666&0.674&0.666&0.648&0.422&0.674&19.287 \\
 & LR &0.278&0.250&0.278&0.123&0.011&0.500&1.821 &   & 
 LR &0.282&0.079&0.282&0.124&0.011&0.500&16.022  \\
 & DT &0.278&0.239&0.278&0.140&0.020&0.502&0.251 &   & 
 DT &0.555&0.561&0.555&0.556&0.299&0.643&8.041
 \\
\midrule

\multirow{7}{2cm}{Gaussian}
 & SVM &0.324&0.573&0.324&0.231&0.132&0.543&929.006 & \\
 & NB &0.021&0.344&0.021&0.023&0.062&0.528&0.303  &   & 
 \\
 & MLP &0.272&0.247&0.272&0.214&0.081&0.524&77.877  &   & 
 \\
 & KNN  &0.204&0.184&0.204&0.172&0.069&0.519&0.721  &   & 
 \\
 & RF &0.325&0.585&0.325&0.236&0.149&0.543&36.319   &   & 
 \\
 & LR  &0.302&0.561&0.302&0.168&0.025&0.504&17.754    &   & 
 \\
 & DT  &0.199&0.203&0.199&0.200&0.102&0.543&21.241   &   &  
 \\
\bottomrule

  \end{tabular}
  }

  \caption{Classification comparison for different kernel methods and classifiers on \textbf{OHE for Genome Data}.
  }
  \label{tbl_variant_classification_genome_ohe}
\end{table}

For k-mer embeddings on Genome data (Table~\ref{tbl_variant_classification_genome_kmer}), linear and cosine kernels with SVM achieve the highest accuracies (0.858 and 0.852, respectively), representing substantial improvement over OHE encoding. Additive-chi2 and laplacian kernels also demonstrate strong performance with SVM (accuracies: 0.846 and 0.848). Polynomial kernel paired with logistic regression yields competitive results (accuracy: 0.774), though with extremely long training time (5079.515 seconds), indicating severe computational challenges. Random forest performance varies across kernels, with modest accuracies ranging from 0.68-0.73. The isolation, chi-squared, Gaussian, and sigmoid kernels show catastrophic failure, with accuracies near or below 0.30, maintaining the pattern observed with OHE encoding. Notably, sigmoid kernel exhibits complete classification collapse with fixed accuracy of 0.276 across all classifiers, indicating numerical instability. The k-mer representation substantially enhances classification capability for appropriate kernel-classifier combinations.

\begin{table}[h!]
  \centering
 \resizebox{0.85\textwidth}{!}{
  \begin{tabular}{p{1.9cm}ccccccp{1.2cm} p{1.5cm} | p{1.7cm}ccccccp{1.2cm} p{1.5cm}}
    \toprule
   
    \multirow{2}{*}{Kernels} & 
    \multirow{2}{0.7cm}{ML Algo.} & \multirow{2}{*}{Acc.} & \multirow{2}{*}{Prec.} & \multirow{2}{*}{Recall} & \multirow{2}{0.9cm}{F1 weigh.}
    &  F1 Macro  & ROC- AUC & Train. runtime (sec.) 
    &
    \multirow{2}{*}{Kernels} & 
    \multirow{2}{0.7cm}{ML Algo.} & \multirow{2}{*}{Acc.} & \multirow{2}{*}{Prec.} & \multirow{2}{*}{Recall} & \multirow{2}{0.9cm}{F1 weigh.}
    &  F1 Macro & ROC- AUC & Train. runtime (sec.)
    \\	
    \midrule	\midrule
    \multirow{7}{2cm}{Additive-chi2}
 & SVM &0.846&0.847&0.846&0.843&0.740&0.861&23.669  
 &  \multirow{7}{2cm}{Laplacian}
 & SVM &0.848&0.851&0.848&0.845&0.718&0.832&18.319  \\
 & NB &0.076&0.423&0.076&0.092&0.132&0.585&0.429 &   & 
 NB &0.649&0.731&0.649&0.671&0.543&0.741&0.408 \\
 & MLP &0.708&0.704&0.708&0.702&0.361&0.683&75.605  &   & 
 MLP &0.670&0.671&0.670&0.668&0.315&0.657&41.940  \\
 & KNN&0.670&0.678&0.670&0.652&0.444&0.692&1.018  &   & 
 KNN &0.770&0.774&0.770&0.757&0.554&0.742&0.748 \\
 & RF &0.684&0.759&0.684&0.648&0.395&0.649&32.336&   & 
 RF &0.729&0.787&0.729&0.700&0.460&0.679&32.163  \\
 & LR&0.773&0.731&0.773&0.729&0.292&0.633&107.704  &   & 
 LR &0.726&0.729&0.726&0.676&0.260&0.610&23.627  \\
 & DT &0.552&0.561&0.552&0.554&0.289&0.649&10.967 &   & 
 DT &0.606&0.615&0.606&0.608&0.336&0.666&12.955 \\
\midrule
\multirow{7}{2cm}{Chi-squared}
 & SVM &0.286&0.115&0.286&0.137&0.015&0.502&2.883 & \multirow{7}{2cm}{Linear}
 & SVM &0.858&0.860&0.858&0.856&0.744&0.873&12.923  \\
 & NB &0.091&0.117&0.091&0.083&0.017&0.524&0.014 &   & 
 NB &0.257&0.663&0.257&0.302&0.328&0.692&0.498\\
 & MLP &0.294&0.184&0.294&0.181&0.028&0.507&25.091 &   & 
 MLP &0.753&0.741&0.753&0.744&0.387&0.693&221.908  \\
 & KNN &0.269&0.234&0.269&0.235&0.098&0.534&0.090 &   & 
 KNN &0.728&0.733&0.728&0.712&0.496&0.715&1.271  \\
 & RF &0.312&0.293&0.312&0.292&0.176&0.569&2.281 &   & 
 RF &0.689&0.774&0.689&0.653&0.419&0.660&8.764 \\
 & LR &0.286&0.115&0.286&0.137&0.015&0.501&0.336 &   & 
 LR &0.856&0.851&0.856&0.847&0.664&0.810&41.262 \\
 & DT &0.263&0.261&0.263&0.260&0.149&0.563&0.079 &   & 
 DT &0.583&0.587&0.583&0.583&0.315&0.656&3.059
 \\
\midrule
\multirow{7}{2cm}{Cosine}
 & SVM &0.852&0.854&0.852&0.851&0.711&0.860&19.627  & \multirow{7}{2cm}{Polynomial}
 & SVM &0.851&0.855&0.851&0.850&0.763&0.864&19.152  \\
 & NB &0.245&0.656&0.245&0.291&0.318&0.680&0.999  &   & 
 NB &0.175&0.482&0.175&0.221&0.282&0.640&0.509 \\
 & MLP &0.756&0.742&0.756&0.745&0.393&0.696&219.173  &   & 
 MLP &0.695&0.701&0.695&0.695&0.351&0.675&34.380 \\
 & KNN &0.733&0.744&0.733&0.719&0.503&0.718&1.131  &   & 
 KNN&0.710&0.718&0.710&0.693&0.475&0.703&0.815  \\
 & RF &0.690&0.772&0.690&0.655&0.406&0.656&14.585  &   & 
 RF &0.676&0.768&0.676&0.640&0.392&0.646&29.625  \\
 & LR &0.276&0.076&0.276&0.119&0.011&0.500&5.893 &   & 
 LR &0.774&0.747&0.774&0.736&0.337&0.659&5079.515 \\
 & DT &0.560&0.567&0.560&0.562&0.297&0.646&3.463  &   & 
 DT&0.569&0.576&0.569&0.570&0.320&0.655&12.245  \\
\midrule

\multirow{7}{2cm}{Isolation}
& SVM &0.291&0.390&0.291&0.158&0.034&0.506&11.197&  \multirow{7}{2cm}{Sigmoid}
& SVM &0.276&0.076&0.276&0.119&0.011&0.500&26.593 \\
 & NB &0.035&0.449&0.035&0.047&0.017&0.505&0.187 &   & 
 NB &0.086&0.007&0.086&0.014&0.004&0.500&1.070 \\
 & MLP &0.287&0.323&0.287&0.153&0.028&0.505&9.804 &   & 
 MLP &0.276&0.076&0.276&0.119&0.011&0.500&14.011  \\
 & KNN &0.209&0.159&0.209&0.165&0.028&0.503&0.440 &   & 
 KNN &0.199&0.048&0.199&0.076&0.008&0.500&1.104  \\
 & RF &0.295&0.445&0.295&0.161&0.031&0.506&1.487 &   & 
 RF &0.276&0.076&0.276&0.119&0.011&0.500&5.003 \\
 & LR &0.281&0.243&0.281&0.127&0.012&0.500&1.902 &   & 
 LR &0.276&0.076&0.276&0.119&0.011&0.500&0.924  \\
 & DT &0.288&0.352&0.288&0.155&0.028&0.505&0.212&   & 
 DT &0.276&0.076&0.276&0.119&0.011&0.500&0.110  \\
\midrule

\multirow{7}{2cm}{Gaussian}
 & SVM &0.326&0.568&0.326&0.233&0.140&0.547&1410.142 & \\
 & NB &0.023&0.379&0.023&0.028&0.066&0.532&1.223  &   & 
 \\
 & MLP  &0.263&0.257&0.263&0.209&0.076&0.521&304.053  &   & 
 \\
 & KNN  &0.224&0.203&0.224&0.188&0.081&0.524&2.195  &   & 
 \\
 & RF  &0.326&0.598&0.326&0.238&0.158&0.546&120.125  &   & 
 \\
 & LR  &0.298&0.527&0.298&0.162&0.022&0.503&59.320    &   & 
 \\
 & DT  &0.204&0.208&0.204&0.204&0.111&0.546&66.209   &   &  
 \\
\bottomrule

  \end{tabular}
  }
  \caption{Classification comparison for different kernel methods and classifiers on \textbf{kmer for Genome data}.
  }
  \label{tbl_variant_classification_genome_kmer}
\end{table}

Table~\ref{tbl_variant_classification_genome_minimizer} reveals that laplacian kernel with SVM achieves the highest accuracy (0.704) on minimizer-encoded Genome data, though performance remains substantially below k-mer encoding results. Chi-squared kernel with KNN demonstrates unexpected strong performance (accuracy: 0.793), representing an anomaly given poor chi-squared performance across other configurations. Polynomial and additive-chi2 kernels with SVM yield moderate accuracies (0.619 and 0.609, respectively). The isolation, Gaussian, and sigmoid kernels maintain their pattern of severe underperformance, with accuracies clustered around 0.28 or below, indicating fundamental unsuitability for minimizer representations. Cosine and linear kernels show mixed results across classifiers, with peak accuracies around 0.60 for appropriate configurations. Training times are generally moderate, though polynomial kernel with logistic regression requires exceptionally long computation (4727.354 seconds). Overall, minimizer encoding demonstrates reduced classification effectiveness compared to k-mer representations for genome sequences, with most configurations achieving modest performance.

\begin{table}[h!]
  \centering
 \resizebox{0.85\textwidth}{!}{
  \begin{tabular}{p{1.9cm}ccccccp{1.2cm} p{1.5cm} | p{1.7cm}ccccccp{1.2cm} p{1.5cm}}
    \toprule
   
    \multirow{2}{*}{Kernels} & 
    \multirow{2}{0.7cm}{ML Algo.} & \multirow{2}{*}{Acc.} & \multirow{2}{*}{Prec.} & \multirow{2}{*}{Recall} & \multirow{2}{0.9cm}{F1 weigh.}
    &  F1 Macro  & ROC- AUC & Train. runtime (sec.) 
    &
    \multirow{2}{*}{Kernels} & 
    \multirow{2}{0.7cm}{ML Algo.} & \multirow{2}{*}{Acc.} & \multirow{2}{*}{Prec.} & \multirow{2}{*}{Recall} & \multirow{2}{0.9cm}{F1 weigh.}
    &  F1 Macro & ROC- AUC & Train. runtime (sec.)
    \\	
    \midrule	\midrule
    \multirow{7}{2cm}{Additive-chi2}
 & SVM &0.609&0.603&0.609&0.596&0.510&0.773&15.843  
 &  \multirow{7}{2cm}{Laplacian}
 & SVM &0.704&0.706&0.704&0.700&0.626&0.791&55.339  \\
 & NB &0.094&0.323&0.094&0.108&0.155&0.603&0.289  &   & 
 NB &0.482&0.584&0.482&0.496&0.499&0.717&0.943 \\
 & MLP &0.538&0.500&0.538&0.508&0.257&0.627&362.872  &   & 
 MLP&0.554&0.554&0.554&0.552&0.268&0.632&67.028  \\
 & KNN &0.521&0.538&0.521&0.497&0.369&0.651&0.626  &   & 
 KNN &0.635&0.638&0.635&0.618&0.479&0.704&1.255 \\
 & RF &0.599&0.698&0.599&0.570&0.427&0.659&12.784 &   & 
 RF &0.602&0.717&0.602&0.574&0.408&0.651&60.583  \\
 & LR &0.566&0.521&0.566&0.513&0.227&0.596&24.497  &   & 
 LR &0.609&0.608&0.609&0.549&0.203&0.586&59.755  \\
 & DT &0.436&0.443&0.436&0.437&0.279&0.637&2.562
  &   & 
 DT &0.429&0.435&0.429&0.430&0.263&0.628&20.096  \\
\midrule
\multirow{7}{2cm}{Chi-squared}
 & SVM &0.524&0.436&0.524&0.403&0.209&0.562&18.261 & \multirow{7}{2cm}{Linear}
 & SVM &0.606&0.589&0.606&0.571&0.491&0.745&35.667  \\
 & NB &0.263&0.489&0.263&0.307&0.222&0.604&0.014 &   & 
 NB &0.199&0.466&0.199&0.224&0.292&0.677&1.236\\
 & MLP &0.662&0.631&0.662&0.606&0.447&0.690&33.259 &   & 
 MLP&0.567&0.518&0.567&0.524&0.283&0.632&243.401  \\
 & KNN &0.793&0.779&0.793&0.781&0.670&0.818&0.304 &   & 
 KNN &0.552&0.557&0.552&0.530&0.386&0.664&0.948  \\
 & RF &0.773&0.780&0.773&0.758&0.664&0.798&4.252 &   & 
 RF &0.616&0.726&0.616&0.590&0.451&0.672&7.789 \\
 & LR &0.484&0.338&0.484&0.337&0.084&0.512&0.314 &   & 
 LR &0.585&0.552&0.585&0.542&0.371&0.662&36.660 \\
 & DT &0.711&0.705&0.711&0.700&0.602&0.770&0.197
 &   & 
 DT &0.451&0.456&0.451&0.451&0.296&0.647&1.882
\\
\midrule
\multirow{7}{2cm}{Cosine}
 & SVM &0.602&0.592&0.602&0.567&0.493&0.739&36.318  & \multirow{7}{2cm}{Polynomial}
 & SVM &0.619&0.645&0.619&0.625&0.544&0.791&27.262  \\
 & NB &0.194&0.466&0.194&0.217&0.285&0.674&0.669 &   & 
 NB &0.131&0.365&0.131&0.156&0.233&0.618&0.495  \\
 & MLP &0.551&0.492&0.551&0.506&0.272&0.631&191.742  &   & 
 MLP &0.502&0.487&0.502&0.492&0.233&0.613&49.752 \\
 & KNN &0.568&0.568&0.568&0.546&0.397&0.669&0.972  &   & 
 KNN &0.539&0.538&0.539&0.515&0.336&0.643&0.585  \\
 & RF &0.608&0.732&0.608&0.581&0.433&0.663&5.336 &   & 
 RF &0.581&0.714&0.581&0.548&0.375&0.638&25.444  \\
 & LR &0.278&0.078&0.278&0.121&0.011&0.500&2.852 &   & 
 LR &0.560&0.538&0.560&0.521&0.250&0.616&4727.354 \\
 & DT &0.453&0.461&0.453&0.455&0.300&0.649&1.270  &   & 
 DT &0.423&0.430&0.423&0.425&0.256&0.627&7.355   \\
\midrule

\multirow{7}{2cm}{Isolation}
& SVM &0.282&0.311&0.282&0.147&0.028&0.505&11.243 &  \multirow{7}{2cm}{Sigmoid}
& SVM &0.277&0.077&0.277&0.120&0.011&0.500&28.613 \\
 & NB &0.029&0.353&0.029&0.038&0.011&0.503&0.195 &   & 
 NB &0.087&0.008&0.087&0.014&0.004&0.500&0.880 \\
 & MLP &0.280&0.282&0.280&0.144&0.023&0.503&9.275 &   & 
 MLP &0.277&0.077&0.277&0.120&0.011&0.500&11.214  \\
 & KNN &0.200&0.144&0.200&0.157&0.025&0.502&0.466 &   & 
 KNN &0.220&0.054&0.220&0.086&0.009&0.500&0.829  \\
 & RF &0.286&0.369&0.286&0.146&0.021&0.503&1.566 &   & 
 RF &0.277&0.077&0.277&0.120&0.011&0.500&3.414  \\
 & LR &0.277&0.252&0.277&0.122&0.011&0.500&1.913 &   & 
 LR &0.277&0.077&0.277&0.120&0.011&0.500&0.737  \\
 & DT &0.279&0.271&0.279&0.141&0.021&0.503&0.261
 &   & 
 DT &0.277&0.077&0.277&0.120&0.011&0.500&0.080  \\
\midrule

\multirow{7}{2cm}{Gaussian}
 & SVM &0.337&0.658&0.337&0.241&0.125&0.536&328.055 & \\
 & NB &0.017&0.392&0.017&0.018&0.044&0.516&0.776  &   & 
 \\
 & MLP  &0.286&0.284&0.286&0.226&0.085&0.524&144.268 &   & 
 \\
 & KNN  &0.223&0.170&0.223&0.170&0.044&0.509&1.198  &   & 
 \\
 & RF  &0.331&0.591&0.331&0.247&0.136&0.538&58.119  &   & 
 \\
 & LR  &0.304&0.553&0.304&0.172&0.025&0.504&28.159    &   & 
 \\
 & DT  &0.209&0.211&0.209&0.209&0.098&0.538&19.748   &   &  
 \\
\bottomrule

  \end{tabular}
  }
  \caption{Classification comparison for different kernel methods and classifiers on \textbf{minimizer for Genome}.
  }
  \label{tbl_variant_classification_genome_minimizer}
\end{table}

Classification results for spaced k-mer embeddings on Genome data (Table~\ref{tbl_variant_classification_genome_spaced_kmer}) demonstrate exceptional performance improvement, with linear kernel paired with logistic regression achieving peak accuracy (0.961) and outstanding ROC-AUC (0.948). Polynomial kernel with logistic regression also delivers excellent results (accuracy: 0.963, ROC-AUC: 0.943), though requiring substantial training time (123.578 seconds). Additive-chi2, cosine, and laplacian kernels with SVM achieve accuracies exceeding 0.94, indicating robust kernel adaptability for spaced k-mer representations. Laplacian kernel with naive Bayes shows remarkably strong performance (accuracy: 0.850), contrasting sharply with naive Bayes underperformance on other encodings. The isolation, chi-squared, Gaussian, and sigmoid kernels continue their pattern of severe failure, with accuracies near or below 0.47, maintaining incompatibility across all genome encoding methods. Sigmoid kernel exhibits complete classification collapse with fixed accuracy of 0.276 across all classifiers. Spaced k-mer encoding emerges as the most effective representation for genome classification, substantially outperforming OHE, k-mer, and minimizer approaches when paired with appropriate kernels.

\begin{table}[h!]
  \centering
 \resizebox{0.85\textwidth}{!}{
  \begin{tabular}{p{1.9cm}ccccccp{1.2cm} p{1.5cm} | p{1.7cm}ccccccp{1.2cm} p{1.5cm}}
    \toprule
   
    \multirow{2}{*}{Kernels} & 
    \multirow{2}{0.7cm}{ML Algo.} & \multirow{2}{*}{Acc.} & \multirow{2}{*}{Prec.} & \multirow{2}{*}{Recall} & \multirow{2}{0.9cm}{F1 weigh.}
    &  F1 Macro  & ROC- AUC & Train. runtime (sec.) 
    &
    \multirow{2}{*}{Kernels} & 
    \multirow{2}{0.7cm}{ML Algo.} & \multirow{2}{*}{Acc.} & \multirow{2}{*}{Prec.} & \multirow{2}{*}{Recall} & \multirow{2}{0.9cm}{F1 weigh.}
    &  F1 Macro & ROC- AUC & Train. runtime (sec.)
    \\	
    \midrule	\midrule
    \multirow{7}{2cm}{Additive-chi2}
 & SVM &0.944&0.942&0.944&0.941&0.892&0.932&29.904  
 &  \multirow{7}{2cm}{Laplacian}
 & SVM &0.951&0.951&0.951&0.950&0.895&0.931&18.776  \\
 & NB &0.245&0.591&0.245&0.290&0.381&0.680&0.780 &   & 
 NB &0.850&0.887&0.850&0.860&0.763&0.854&0.386 \\
 & MLP &0.788&0.794&0.788&0.789&0.486&0.744&73.071  &   & 
 MLP &0.813&0.821&0.813&0.814&0.524&0.774&49.666  \\
 & KNN &0.718&0.738&0.718&0.701&0.496&0.715&0.983  &   & 
 KNN &0.862&0.866&0.862&0.856&0.704&0.827&0.810 \\
 & RF&0.732&0.798&0.732&0.709&0.477&0.689&37.285 &   & 
 RF &0.825&0.851&0.825&0.804&0.588&0.744&36.846 \\
 & LR &0.920&0.912&0.920&0.909&0.676&0.819&120.697  &   & 
 LR &0.806&0.768&0.806&0.756&0.294&0.634&27.203  \\
 & DT &0.617&0.628&0.617&0.620&0.333&0.672&13.004  &   & 
 DT &0.736&0.743&0.736&0.737&0.425&0.717&12.642  \\
\midrule
\multirow{7}{2cm}{Chi-squared}
 & SVM &0.441&0.397&0.441&0.361&0.065&0.523&6.297 & \multirow{7}{2cm}{Linear}
 & SVM &0.945&0.945&0.945&0.943&0.902&0.938&16.846 \\
 & NB &0.111&0.410&0.111&0.129&0.034&0.543&0.020 &   & 
 NB &0.270&0.651&0.270&0.311&0.403&0.692&0.630\\
 & MLP &0.488&0.400&0.488&0.414&0.100&0.541&12.814 &   & 
 MLP &0.784&0.792&0.784&0.785&0.482&0.745&52.468  \\
 & KNN &0.468&0.444&0.468&0.436&0.195&0.583&0.108 &   & 
 KNN &0.796&0.802&0.796&0.785&0.600&0.766&0.931  \\
 & RF &0.539&0.523&0.539&0.514&0.301&0.627&3.028&   & 
 RF &0.779&0.817&0.779&0.744&0.477&0.694&29.177 \\
 & LR &0.396&0.291&0.396&0.281&0.035&0.511&0.577 &   & 
 LR &0.961&0.961&0.961&0.959&0.912&0.948&117.378  \\
 & DT &0.472&0.468&0.472&0.465&0.250&0.618&0.142
 &   & 
 DT &0.698&0.704&0.698&0.699&0.402&0.705&10.364 \\
\midrule
\multirow{7}{2cm}{Cosine}
 & SVM &0.948&0.948&0.948&0.947&0.909&0.941&28.529  & \multirow{7}{2cm}{Polynomial}
 & SVM &0.945&0.945&0.945&0.943&0.891&0.927&15.734  \\
 & NB &0.300&0.669&0.300&0.330&0.418&0.697&0.744  &   & 
 NB &0.267&0.655&0.267&0.299&0.395&0.684&0.668 \\
 & MLP &0.793&0.799&0.793&0.793&0.494&0.749&55.129  &   & 
 MLP&0.790&0.796&0.790&0.791&0.473&0.739&55.134 \\
 & KNN &0.807&0.815&0.807&0.798&0.609&0.774&0.983  &   & 
 KNN &0.805&0.813&0.805&0.794&0.611&0.770&0.905  \\
 & RF &0.786&0.827&0.786&0.753&0.480&0.695&42.687  &   & 
 RF &0.791&0.825&0.791&0.758&0.484&0.696&34.621  \\
 & LR &0.276&0.077&0.276&0.120&0.011&0.500&25.902  &   & 
 LR &0.963&0.964&0.963&0.962&0.909&0.943&123.578  \\
 & DT &0.705&0.712&0.705&0.706&0.389&0.699&12.238  &   & 
 DT &0.700&0.704&0.700&0.700&0.387&0.697&15.952  \\
\midrule

\multirow{7}{2cm}{Isolation}
& SVM &0.296&0.423&0.296&0.161&0.033&0.506&10.898 &  \multirow{7}{2cm}{Sigmoid}
& SVM &0.276&0.076&0.276&0.120&0.011&0.500&40.752 \\
 & NB &0.041&0.569&0.041&0.062&0.022&0.506&0.191 &   & 
 NB &0.088&0.008&0.088&0.014&0.004&0.500&1.434 \\
 & MLP &0.292&0.374&0.292&0.158&0.029&0.505&9.863 &   & 
 MLP &0.276&0.076&0.276&0.120&0.011&0.500&18.267  \\
 & KNN &0.215&0.170&0.215&0.173&0.034&0.505&0.456 &   & 
 KNN &0.218&0.054&0.218&0.085&0.009&0.500&1.014 \\
 & RF &0.302&0.477&0.302&0.168&0.031&0.506&1.391 &   & 
 RF&0.276&0.076&0.276&0.120&0.011&0.500&3.556 \\
 & LR &0.285&0.250&0.285&0.130&0.012&0.500&1.723 &   & 
 LR &0.276&0.076&0.276&0.120&0.011&0.500&0.816  \\
 & DT &0.295&0.388&0.295&0.161&0.027&0.504&0.235 &   & 
 DT &0.276&0.076&0.276&0.120&0.011&0.500&0.098 \\
\midrule

\multirow{7}{2cm}{Gaussian}
 & SVM &0.325&0.575&0.325&0.229&0.126&0.539&1620.742 \\
 & NB &0.018&0.383&0.018&0.022&0.045&0.520&0.942  &   & 
 \\
 & MLP  &0.265&0.251&0.265&0.208&0.065&0.516&165.962  &   & 
 \\
 & KNN  &0.205&0.185&0.205&0.174&0.062&0.515&0.839  &   & 
 \\
 & RF  &0.322&0.587&0.322&0.233&0.139&0.538&56.842  &   & 
 \\
 & LR  &0.303&0.547&0.303&0.168&0.023&0.504&28.340    &   & 
 \\
 & DT  &0.200&0.205&0.200&0.202&0.096&0.538&41.742
   &   &  
 \\
\bottomrule

  \end{tabular}
  }
  \caption{Classification comparison for different kernel methods and classifiers on  \textbf{spaced kmer for Genome data}.
  }
  \label{tbl_variant_classification_genome_spaced_kmer}
\end{table}

\clearpage

\subsection{Results For Breast Cancer Data}
The OHE embedding results in Table~\ref{tbl_variant_classification_breastcancer_ohe} demonstrate strong performance across multiple kernel-classifier combinations, with the Chi-squared kernel achieving the highest accuracy of 89.8\% when paired with Random Forest. Notably, the Isolation kernel shows robust results across different classifiers, with Naive Bayes and Random Forest both exceeding 88\% accuracy and ROC-AUC scores above 0.75. The Gaussian kernel exhibits relatively poor performance, particularly with KNN, achieving only 9.9\% accuracy, suggesting that this kernel struggles with the high-dimensional sparse representations inherent in OHE. Training times remain relatively low across all combinations, with most models completing in under 2 seconds, though MLP requires more time for convergence.

\begin{table}[h!]
  \centering
 \resizebox{0.85\textwidth}{!}{
  \begin{tabular}{p{1.9cm}ccccccp{1.2cm} p{1.5cm} | p{1.7cm}ccccccp{1.2cm} p{1.5cm}}
    \toprule
   
    \multirow{2}{*}{Kernels} & 
    \multirow{2}{0.7cm}{ML Algo.} & \multirow{2}{*}{Acc.} & \multirow{2}{*}{Prec.} & \multirow{2}{*}{Recall} & \multirow{2}{0.9cm}{F1 weigh.}
    &  F1 Macro  & ROC- AUC & Train. runtime (sec.) 
    &
    \multirow{2}{*}{Kernels} & 
    \multirow{2}{0.7cm}{ML Algo.} & \multirow{2}{*}{Acc.} & \multirow{2}{*}{Prec.} & \multirow{2}{*}{Recall} & \multirow{2}{0.9cm}{F1 weigh.}
    &  F1 Macro & ROC- AUC & Train. runtime (sec.)
    \\	
    \midrule	\midrule
    \multirow{7}{2cm}{Additive-chi2}
 & SVM &0.848&0.832&0.848&0.839&0.495&0.698&0.152  
 &  \multirow{7}{2cm}{Laplacian}
 & SVM &0.883&0.878&0.883&0.878&0.566&0.747&0.154  \\
 & NB &0.780&0.804&0.780&0.786&0.443&0.676&0.007 &   & 
 NB &0.793&0.833&0.793&0.806&0.436&0.691&0.008 \\
 & MLP &0.659&0.757&0.659&0.695&0.365&0.625&1.183  &   & 
 MLP &0.632&0.783&0.632&0.684&0.351&0.627&1.322 \\
 & KNN &0.531&0.844&0.531&0.614&0.334&0.640&0.014 &   & 
 KNN&0.635&0.863&0.635&0.702&0.384&0.672&0.015 \\
 & RF&0.836&0.788&0.836&0.795&0.405&0.608&0.755 &   & 
 RF &0.852&0.812&0.852&0.814&0.414&0.609&0.730 \\
 & LR &0.878&0.876&0.878&0.872&0.553&0.742&0.166 &   & 
 LR &0.807&0.652&0.807&0.721&0.223&0.500&0.052  \\
 & DT &0.827&0.829&0.827&0.827&0.449&0.686&0.408  &   & 
 DT &0.844&0.853&0.844&0.848&0.490&0.710&0.410 \\
\midrule
\multirow{7}{2cm}{Chi-squared}
 & SVM &0.820&0.823&0.820&0.817&0.523&0.700&0.212 & \multirow{7}{2cm}{Linear}
 & SVM &0.863&0.848&0.863&0.854&0.527&0.715&0.154  \\
 & NB &0.879&0.882&0.879&0.872&0.547&0.763&0.007 &   & 
 NB &0.765&0.798&0.765&0.778&0.422&0.673&0.007 \\
 & MLP &0.676&0.806&0.676&0.720&0.454&0.708&1.903 &   & 
 MLP &0.655&0.743&0.655&0.688&0.352&0.612&1.589  \\
 & KNN &0.213&0.821&0.213&0.231&0.163&0.538&0.014 &   & 
 KNN &0.620&0.850&0.620&0.681&0.400&0.671&0.014  \\
 & RF &0.898&0.895&0.898&0.890&0.642&0.786&0.893&   & 
 RF &0.828&0.780&0.828&0.784&0.398&0.604&0.799 \\
 & LR &0.787&0.620&0.787&0.694&0.220&0.500&0.063&   & 
 LR &0.880&0.872&0.880&0.872&0.577&0.747&0.145 \\
 & DT &0.863&0.868&0.863&0.864&0.550&0.757&0.283
&   & 
 DT &0.839&0.838&0.839&0.837&0.503&0.719&0.416\\
\midrule
\multirow{7}{2cm}{Cosine}
 & SVM &0.861&0.853&0.861&0.856&0.527&0.722&0.165  & \multirow{7}{2cm}{Polynomial}
 & SVM&0.846&0.833&0.846&0.838&0.511&0.711&0.145  \\
 & NB &0.838&0.830&0.838&0.831&0.497&0.703&0.007 &   & 
 NB &0.768&0.793&0.768&0.777&0.418&0.674&0.008 \\
 & MLP&0.628&0.773&0.628&0.682&0.358&0.625&1.601 &   & 
 MLP &0.726&0.766&0.726&0.743&0.406&0.642&1.365 \\
 & KNN &0.813&0.852&0.813&0.827&0.451&0.717&0.015  &   & 
 KNN &0.616&0.849&0.616&0.677&0.377&0.669&0.013 \\
 & RF &0.835&0.780&0.835&0.788&0.369&0.584&0.736 &   & 
 RF &0.825&0.776&0.825&0.779&0.394&0.599&0.741  \\
 & LR &0.860&0.820&0.860&0.832&0.435&0.651&0.078  &   & 
 LR &0.778&0.606&0.778&0.681&0.219&0.500&0.053 \\
 & DT &0.834&0.837&0.834&0.835&0.447&0.686&0.421  &   & 
 DT&0.822&0.828&0.822&0.823&0.460&0.696&0.439  \\
\midrule

\multirow{7}{2cm}{Isolation}
& SVM &0.854&0.849&0.854&0.850&0.547&0.732&0.057 &  \multirow{7}{2cm}{Sigmoid}
& SVM &0.853&0.870&0.853&0.858&0.536&0.750&0.045  \\
 & NB &0.857&0.879&0.857&0.865&0.576&0.782&0.008 &   & 
 NB &0.857&0.850&0.857&0.851&0.523&0.735&0.004  \\
 & MLP &0.851&0.852&0.851&0.850&0.569&0.747&1.867 &   & 
 MLP &0.784&0.815&0.784&0.796&0.448&0.685&1.225 \\
 & KNN &0.825&0.785&0.825&0.777&0.372&0.587&0.012 &   & 
 KNN &0.663&0.861&0.663&0.727&0.406&0.680&0.015  \\
 & RF &0.889&0.885&0.889&0.882&0.583&0.755&0.224 &   & 
 RF &0.865&0.828&0.865&0.840&0.462&0.662&0.560  \\
 & LR &0.793&0.629&0.793&0.701&0.221&0.500&0.021 &   & 
 LR &0.798&0.637&0.798&0.708&0.222&0.500&0.027  \\
 & DT &0.871&0.878&0.871&0.874&0.566&0.770&0.055 &   & 
 DT &0.839&0.852&0.839&0.844&0.499&0.716&0.160
  \\
\midrule

\multirow{7}{2cm}{Gaussian}
 & SVM &0.578&0.685&0.578&0.622&0.265&0.536&0.188 & \\
 & NB &0.847&0.868&0.847&0.856&0.519&0.750&0.007  &   & 
 \\
 & MLP &0.579&0.686&0.579&0.621&0.273&0.538&1.393  &   & 
 \\
 & KNN  &0.099&0.648&0.099&0.050&0.083&0.497&0.012  &   & 
 \\
 & RF  &0.796&0.707&0.796&0.719&0.280&0.531&0.868  &   & 
 \\
 & LR  &0.784&0.615&0.784&0.689&0.220&0.500&0.055    &   & 
 \\
 & DT  &0.762&0.767&0.762&0.764&0.361&0.623&0.308   &   & 
 \\
\bottomrule

  \end{tabular}
  }

  \caption{Classification comparison for different kernel methods and classifiers on \textbf{OHE for Breast Cancer}.
  }
  \label{tbl_variant_classification_breastcancer_ohe}
\end{table}

The Spike2Vec embedding results in Table~\ref{tbl_variant_classification_breastcancer_kmer} yield competitive results, with several kernel-classifier pairs achieving accuracies above 88\%. The Chi-squared and Isolation kernels demonstrate particular effectiveness, both reaching 89.1\% accuracy with Random Forest. Interestingly, the Cosine kernel with Naive Bayes achieves 89.1\% accuracy—the highest for this encoding—while maintaining exceptional training efficiency at 0.008 seconds. The Linear kernel shows strong performance with Logistic Regression (88.3\% accuracy), suggesting that kmer features exhibit good linear separability. Similar to OHE, the Gaussian kernel underperforms significantly, indicating that kmer's characteristic feature distributions are not well-suited to Gaussian kernel transformations.

\begin{table}[h!]
  \centering
 \resizebox{0.85\textwidth}{!}{
  \begin{tabular}{p{1.9cm}ccccccp{1.2cm} p{1.5cm} | p{1.7cm}ccccccp{1.2cm} p{1.5cm}}
    \toprule
   
    \multirow{2}{*}{Kernels} & 
    \multirow{2}{0.7cm}{ML Algo.} & \multirow{2}{*}{Acc.} & \multirow{2}{*}{Prec.} & \multirow{2}{*}{Recall} & \multirow{2}{0.9cm}{F1 weigh.}
    &  F1 Macro  & ROC- AUC & Train. runtime (sec.) 
    &
    \multirow{2}{*}{Kernels} & 
    \multirow{2}{0.7cm}{ML Algo.} & \multirow{2}{*}{Acc.} & \multirow{2}{*}{Prec.} & \multirow{2}{*}{Recall} & \multirow{2}{0.9cm}{F1 weigh.}
    &  F1 Macro & ROC- AUC & Train. runtime (sec.)
    \\	
    \midrule	\midrule
    \multirow{7}{2cm}{Additive-chi2}
 & SVM&0.801&0.836&0.801&0.813&0.499&0.718&0.150  
 &  \multirow{7}{2cm}{Laplacian}
 & SVM &0.840&0.858&0.840&0.848&0.555&0.746&0.180  \\
 & NB&0.853&0.854&0.853&0.849&0.494&0.728&0.008  &   & 
 NB &0.850&0.865&0.850&0.853&0.508&0.736&0.008 \\
 & MLP &0.619&0.775&0.619&0.672&0.381&0.655&1.446  &   & 
 MLP&0.617&0.774&0.617&0.670&0.371&0.643&1.486  \\
 & KNN &0.868&0.840&0.868&0.851&0.483&0.701&0.013  &   & 
 KNN &0.881&0.854&0.881&0.866&0.494&0.705&0.013 \\
 & RF &0.874&0.852&0.874&0.856&0.560&0.715&0.667 &   & 
 RF &0.872&0.845&0.872&0.851&0.516&0.685&0.731 \\
 & LR &0.886&0.879&0.886&0.880&0.560&0.757&0.152  &   & 
 LR &0.796&0.635&0.796&0.706&0.222&0.500&0.044  \\
 & DT&0.833&0.846&0.833&0.838&0.499&0.727&0.365 &   & 
 DT &0.834&0.851&0.834&0.841&0.496&0.723&0.355 \\
\midrule
\multirow{7}{2cm}{Chi-squared}
 & SVM &0.811&0.820&0.811&0.813&0.475&0.683&0.160 & \multirow{7}{2cm}{Linear}
 & SVM&0.788&0.825&0.788&0.801&0.473&0.697&0.176  \\
 & NB &0.862&0.875&0.862&0.860&0.508&0.738&0.008&   & 
 NB &0.869&0.861&0.869&0.863&0.531&0.742&0.008\\
 & MLP &0.699&0.812&0.699&0.737&0.450&0.699&1.761 &   & 
 MLP &0.618&0.786&0.618&0.671&0.373&0.655&1.538  \\
 & KNN &0.618&0.811&0.618&0.637&0.341&0.614&0.015 &   & 
 KNN &0.867&0.836&0.867&0.848&0.482&0.693&0.014  \\
 & RF&0.891&0.880&0.891&0.883&0.570&0.753&0.899&   & 
 RF &0.862&0.838&0.862&0.842&0.476&0.685&0.698 \\
 & LR &0.795&0.651&0.795&0.705&0.225&0.501&0.061 &   & 
 LR&0.883&0.870&0.883&0.871&0.548&0.732&0.152\\
 & DT&0.847&0.854&0.847&0.849&0.518&0.740&0.314&   & 
 DT&0.831&0.839&0.831&0.834&0.504&0.716&0.337
 \\
\midrule
\multirow{7}{2cm}{Cosine}
 & SVM &0.855&0.874&0.855&0.861&0.526&0.741&0.199  & \multirow{7}{2cm}{Polynomial}
 & SVM &0.800&0.835&0.800&0.813&0.490&0.709&0.192  \\
 & NB &0.891&0.882&0.891&0.882&0.507&0.732&0.008 &   & 
 NB&0.864&0.853&0.864&0.855&0.498&0.725&0.007  \\
 & MLP &0.565&0.802&0.565&0.636&0.338&0.643&1.642  &   & 
 MLP &0.613&0.768&0.613&0.669&0.353&0.628&1.607 \\
 & KNN&0.877&0.875&0.877&0.874&0.504&0.737&0.017  &   & 
 KNN &0.865&0.835&0.865&0.848&0.481&0.694&0.014 \\
 & RF &0.879&0.855&0.879&0.861&0.494&0.692&0.661  &   & 
 RF &0.854&0.822&0.854&0.831&0.455&0.674&0.701  \\
 & LR &0.858&0.825&0.858&0.827&0.431&0.642&0.079  &   & 
 LR &0.777&0.604&0.777&0.679&0.219&0.500&0.047  \\
 & DT&0.845&0.859&0.845&0.851&0.491&0.719&0.352  &   & 
 DT &0.840&0.842&0.840&0.838&0.475&0.727&0.298  \\
\midrule

\multirow{7}{2cm}{Isolation}
& SVM &0.857&0.845&0.857&0.847&0.516&0.712&0.054 &  \multirow{7}{2cm}{Sigmoid}
& SVM &0.799&0.844&0.799&0.817&0.470&0.709&0.187 \\
 & NB &0.882&0.897&0.882&0.882&0.572&0.784&0.008 &   & 
 NB &0.865&0.866&0.865&0.861&0.516&0.736&0.007 \\
 & MLP &0.856&0.851&0.856&0.852&0.547&0.734&1.881 &   & 
 MLP&0.662&0.794&0.662&0.710&0.364&0.645&1.384 \\
 & KNN &0.834&0.791&0.834&0.794&0.420&0.621&0.013 &   & 
 KNN &0.878&0.849&0.878&0.862&0.476&0.696&0.013 \\
 & RF &0.891&0.887&0.891&0.885&0.573&0.765&0.206 &   & 
 RF &0.873&0.847&0.873&0.855&0.462&0.689&0.687 \\
 & LR &0.778&0.606&0.778&0.681&0.219&0.500&0.018 &   & 
 LR &0.800&0.640&0.800&0.711&0.222&0.500&0.039  \\
 & DT &0.874&0.878&0.874&0.874&0.526&0.747&0.025
 &   & 
 DT &0.856&0.862&0.856&0.858&0.486&0.720&0.339 \\
\midrule

\multirow{7}{2cm}{Gaussian}
 & SVM&0.570&0.711&0.570&0.623&0.278&0.564&0.183 & \\
 & NB &0.845&0.864&0.845&0.853&0.503&0.740&0.007  &   & 
 \\
 & MLP &0.558&0.679&0.558&0.604&0.255&0.524&1.374  &   & 
 \\
 & KNN  &0.126&0.491&0.126&0.094&0.104&0.518&0.013  &   & 
 \\
 & RF &0.805&0.723&0.805&0.730&0.276&0.528&0.886  &   & 
 \\
 & LR  &0.794&0.631&0.794&0.703&0.221&0.500&0.055    &   & 
 \\
 & DT  &0.775&0.778&0.775&0.776&0.373&0.628&0.347   &   & 
 \\
\bottomrule

  \end{tabular}
  }

  \caption{Classification comparison for different kernel methods and classifiers on \textbf{kmer for Breast Cancer}.
  }
  \label{tbl_variant_classification_breastcancer_kmer}
\end{table}

The minimizer embedding results in Table~\ref{tbl_variant_classification_breastcancer_minimizer} reveal more variable performance patterns compared to OHE and kmer representations. The Isolation kernel emerges as the strongest performer, achieving 87.3\% accuracy with Random Forest and demonstrating reasonable results across multiple classifiers. Chi-squared and Cosine kernels also show robust performance, with Random Forest achieving approximately 85\% accuracy in both cases. Notably, the performance gap between best and worst kernels is more pronounced in this encoding, with the Gaussian kernel's SVM achieving only 47.6\% accuracy. The overall stable training times across encodings suggest that minimizer representations maintain computational efficiency while potentially capturing different sequence characteristics than standard kmer approaches.

\begin{table}[h!]
  \centering
 \resizebox{0.85\textwidth}{!}{
  \begin{tabular}{p{1.9cm}ccccccp{1.2cm} p{1.5cm} | p{1.7cm}ccccccp{1.2cm} p{1.5cm}}
    \toprule
   
    \multirow{2}{*}{Kernels} & 
    \multirow{2}{0.7cm}{ML Algo.} & \multirow{2}{*}{Acc.} & \multirow{2}{*}{Prec.} & \multirow{2}{*}{Recall} & \multirow{2}{0.9cm}{F1 weigh.}
    &  F1 Macro  & ROC- AUC & Train. runtime (sec.) 
    &
    \multirow{2}{*}{Kernels} & 
    \multirow{2}{0.7cm}{ML Algo.} & \multirow{2}{*}{Acc.} & \multirow{2}{*}{Prec.} & \multirow{2}{*}{Recall} & \multirow{2}{0.9cm}{F1 weigh.}
    &  F1 Macro & ROC- AUC & Train. runtime (sec.)
    \\	
    \midrule	\midrule
    \multirow{7}{2cm}{Additive-chi2}
 & SVM &0.714&0.794&0.714&0.745&0.427&0.663&0.140  
 &  \multirow{7}{2cm}{Laplacian}
 & SVM &0.681&0.786&0.681&0.722&0.372&0.642&0.170  \\
 & NB &0.672&0.805&0.672&0.718&0.444&0.695&0.007 &   & 
 NB &0.659&0.810&0.659&0.707&0.439&0.697&0.007 \\
 & MLP&0.650&0.758&0.650&0.691&0.350&0.622&1.703  &   & 
 MLP &0.687&0.784&0.687&0.724&0.364&0.637&1.623  \\
 & KNN &0.596&0.779&0.596&0.653&0.336&0.611&0.013  &   & 
 KNN&0.456&0.785&0.456&0.542&0.285&0.582&0.014 \\
 & RF &0.844&0.811&0.844&0.817&0.452&0.649&0.686&   & 
 RF &0.851&0.813&0.851&0.822&0.439&0.649&0.739  \\
 & LR&0.849&0.825&0.849&0.832&0.449&0.676&0.147 &   & 
 LR &0.800&0.640&0.800&0.711&0.222&0.500&0.040  \\
 & DT &0.790&0.815&0.790&0.801&0.450&0.685&0.369 &   & 
 DT &0.804&0.812&0.804&0.806&0.416&0.672&0.328  \\
\midrule
\multirow{7}{2cm}{Chi-squared}
 & SVM &0.772&0.787&0.772&0.775&0.431&0.654&0.078 & \multirow{7}{2cm}{Linear}
 & SVM &0.745&0.808&0.745&0.767&0.429&0.678&0.133  \\
 & NB &0.728&0.811&0.728&0.756&0.446&0.700&0.008 &   & 
 NB &0.679&0.794&0.679&0.720&0.409&0.684&0.008 \\
 & MLP &0.752&0.805&0.752&0.774&0.447&0.676&1.927 &   & 
 MLP&0.718&0.782&0.718&0.743&0.402&0.651&1.932  \\
 & KNN &0.836&0.790&0.836&0.799&0.412&0.617&0.015 &   & 
 KNN &0.629&0.786&0.629&0.676&0.347&0.619&0.015 \\
 & RF&0.853&0.827&0.853&0.834&0.467&0.666&1.038 &   & 
 RF &0.853&0.822&0.853&0.830&0.462&0.664&0.741 \\
 & LR&0.820&0.758&0.820&0.763&0.352&0.569&0.069&   & 
 LR &0.851&0.835&0.851&0.841&0.508&0.703&0.200 \\
 & DT &0.827&0.841&0.827&0.832&0.498&0.716&0.615 &   & 
 DT &0.797&0.814&0.797&0.804&0.431&0.677&0.391\\
\midrule
\multirow{7}{2cm}{Cosine}
 & SVM &0.709&0.816&0.709&0.742&0.458&0.689&0.123  & \multirow{7}{2cm}{Polynomial}
 & SVM &0.717&0.778&0.717&0.739&0.404&0.653&0.119 \\
 & NB &0.688&0.787&0.688&0.726&0.381&0.659&0.007  &   & 
 NB &0.693&0.786&0.693&0.725&0.413&0.673&0.008  \\
 & MLP &0.681&0.791&0.681&0.723&0.406&0.664&2.118  &   & 
 MLP &0.673&0.771&0.673&0.710&0.376&0.641&1.761 \\
 & KNN &0.590&0.778&0.590&0.647&0.329&0.609&0.013  &   & 
 KNN &0.625&0.790&0.625&0.668&0.355&0.629&0.013 \\
 & RF &0.848&0.813&0.848&0.826&0.441&0.658&0.742  &   & 
 RF &0.842&0.806&0.842&0.817&0.447&0.652&0.738 \\
 & LR &0.839&0.800&0.839&0.803&0.414&0.624&0.081  &   & 
 LR &0.778&0.606&0.778&0.681&0.219&0.500&0.049 \\
 & DT &0.805&0.813&0.805&0.807&0.439&0.678&0.358  &   & 
 DT &0.805&0.814&0.805&0.808&0.469&0.697&0.358  \\
\midrule

\multirow{7}{2cm}{Isolation}
& SVM &0.818&0.822&0.818&0.818&0.465&0.696&0.052 &  \multirow{7}{2cm}{Sigmoid}
& SVM &0.700&0.808&0.700&0.738&0.404&0.655&0.160 \\
 & NB &0.832&0.846&0.832&0.837&0.459&0.705&0.008 &   & 
 NB &0.671&0.800&0.671&0.720&0.374&0.656&0.008 \\
 & MLP &0.822&0.834&0.822&0.826&0.461&0.694&1.918 &   & 
 MLP &0.681&0.789&0.681&0.721&0.383&0.655&1.795  \\
 & KNN &0.833&0.783&0.833&0.787&0.380&0.591&0.013 &   & 
 KNN &0.623&0.797&0.623&0.674&0.341&0.620&0.014 \\
 & RF &0.873&0.857&0.873&0.862&0.493&0.703&0.223 &   & 
 RF &0.865&0.836&0.865&0.842&0.464&0.669&0.714 \\
 & LR &0.799&0.638&0.799&0.709&0.222&0.500&0.021 &   & 
 LR&0.799&0.638&0.799&0.709&0.222&0.500&0.044  \\
 & DT &0.847&0.849&0.847&0.847&0.470&0.699&0.046 &   & 
 DT &0.815&0.835&0.815&0.823&0.478&0.725&0.446 \\
\midrule

\multirow{7}{2cm}{Gaussian}
 & SVM &0.476&0.639&0.476&0.535&0.272&0.521&0.185 & \\
 & NB &0.754&0.863&0.754&0.790&0.427&0.725&0.008  &   & 
 \\
 & MLP  &0.593&0.721&0.593&0.640&0.336&0.598&1.923  &   & 
 \\
 & KNN  &0.798&0.734&0.798&0.728&0.308&0.544&0.016  &   & 
 \\
 & RF &0.845&0.815&0.845&0.823&0.441&0.661&0.728  &   & 
 \\
 & LR &0.784&0.625&0.784&0.690&0.223&0.502&0.071    &   & 
 \\
 & DT  &0.845&0.841&0.845&0.842&0.469&0.703&0.172
   &   & 
 \\
\bottomrule

  \end{tabular}
  }

  \caption{Classification comparison for different kernel methods and classifiers on \textbf{minimizer for Breast Cancer}.
  }
  \label{tbl_variant_classification_breastcancer_minimizer}
\end{table}

Spaced kmer embedding results in Table~\ref{tbl_variant_classification_breastcancer_spaced_kmer} demonstrates excellent overall performance, with multiple configurations exceeding 88\% accuracy. The Cosine kernel achieves the highest accuracy of 88.8\% when paired with Random Forest, while the Chi-squared and Isolation kernels both reach 88.3\% with the same classifier. Logistic Regression shows notably strong performance with several kernels, particularly achieving 86.9\% accuracy with Additive-chi2 and 86.7\% with Linear kernels. The Gaussian kernel again exhibits poor performance, reinforcing the pattern observed across all encodings that Gaussian transformations are ill-suited to these genomic feature representations. The spaced kmer approach appears to offer a balanced trade-off between discriminative power and computational efficiency, with training times comparable to other encodings.

\begin{table}[h!]
  \centering
 \resizebox{0.85\textwidth}{!}{
  \begin{tabular}{p{1.9cm}ccccccp{1.2cm} p{1.5cm} | p{1.7cm}ccccccp{1.2cm} p{1.5cm}}
    \toprule
   
    \multirow{2}{*}{Kernels} & 
    \multirow{2}{0.7cm}{ML Algo.} & \multirow{2}{*}{Acc.} & \multirow{2}{*}{Prec.} & \multirow{2}{*}{Recall} & \multirow{2}{0.9cm}{F1 weigh.}
    &  F1 Macro  & ROC- AUC & Train. runtime (sec.) 
    &
    \multirow{2}{*}{Kernels} & 
    \multirow{2}{0.7cm}{ML Algo.} & \multirow{2}{*}{Acc.} & \multirow{2}{*}{Prec.} & \multirow{2}{*}{Recall} & \multirow{2}{0.9cm}{F1 weigh.}
    &  F1 Macro & ROC- AUC & Train. runtime (sec.)
    \\	
    \midrule	\midrule
    \multirow{7}{2cm}{Additive-chi2}
 & SVM&0.822&0.826&0.822&0.820&0.434&0.671&0.158  
 &  \multirow{7}{2cm}{Laplacian}
 & SVM &0.845&0.843&0.845&0.842&0.492&0.699&0.187  \\
 & NB &0.561&0.842&0.561&0.646&0.353&0.663&0.007  &   & 
 NB &0.569&0.834&0.569&0.652&0.365&0.659&0.008 \\
 & MLP &0.650&0.779&0.650&0.701&0.341&0.617&1.776  &   & 
 MLP &0.612&0.783&0.612&0.667&0.376&0.646&2.460  \\
 & KNN&0.688&0.832&0.688&0.731&0.391&0.662&0.013 &   & 
 KNN &0.765&0.828&0.765&0.776&0.415&0.654&0.016 \\
 & RF&0.871&0.860&0.871&0.862&0.503&0.719&0.609 &   & 
 RF &0.881&0.871&0.881&0.875&0.563&0.741&0.635 \\
 & LR &0.869&0.848&0.869&0.854&0.497&0.696&0.132  &   & 
 LR &0.792&0.629&0.792&0.701&0.221&0.500&0.039  \\
 & DT&0.835&0.844&0.835&0.837&0.452&0.695&0.216 &   & 
 DT &0.839&0.848&0.839&0.842&0.498&0.715&0.213 \\
\midrule
\multirow{7}{2cm}{Chi-squared}
 & SVM &0.837&0.833&0.837&0.833&0.504&0.712&0.144 & \multirow{7}{2cm}{Linear}
 & SVM &0.838&0.837&0.838&0.833&0.488&0.692&0.140  \\
 & NB&0.603&0.831&0.603&0.678&0.376&0.684&0.007&   & 
 NB &0.547&0.823&0.547&0.634&0.331&0.645&0.008\\
 & MLP &0.718&0.813&0.718&0.755&0.443&0.699&1.716 &   & 
 MLP &0.603&0.785&0.603&0.663&0.357&0.637&2.109  \\
 & KNN &0.825&0.771&0.825&0.778&0.380&0.592&0.013 &   & 
 KNN &0.669&0.826&0.669&0.714&0.367&0.642&0.013  \\
 & RF &0.883&0.877&0.883&0.877&0.564&0.747&0.662 &   & 
 RF &0.877&0.872&0.877&0.869&0.527&0.726&0.651 \\
 & LR&0.799&0.692&0.799&0.720&0.257&0.520&0.070 &   & 
 LR &0.867&0.839&0.867&0.845&0.488&0.679&0.113 \\
 & DT &0.841&0.856&0.841&0.847&0.498&0.723&0.225
&   & 
 DT &0.833&0.848&0.833&0.839&0.483&0.722&0.271\\
\midrule
\multirow{7}{2cm}{Cosine}
 & SVM &0.758&0.820&0.758&0.782&0.443&0.685&0.175  & \multirow{7}{2cm}{Polynomial}
 & SVM &0.810&0.808&0.810&0.806&0.447&0.672&0.154 \\
 & NB &0.624&0.842&0.624&0.693&0.277&0.639&0.008 &   & 
 NB &0.534&0.822&0.534&0.626&0.366&0.658&0.007  \\
 & MLP &0.610&0.804&0.610&0.674&0.387&0.665&1.818  &   & 
 MLP &0.619&0.780&0.619&0.675&0.394&0.649&1.748 \\
 & KNN &0.581&0.813&0.581&0.634&0.353&0.623&0.013 &   & 
 KNN &0.608&0.816&0.608&0.650&0.375&0.642&0.013  \\
 & RF &0.888&0.880&0.888&0.881&0.574&0.753&0.626  &   & 
 RF &0.866&0.857&0.866&0.856&0.524&0.725&0.609  \\
 & LR &0.827&0.777&0.827&0.781&0.374&0.597&0.072 &   & 
 LR &0.774&0.599&0.774&0.676&0.218&0.500&0.037 \\
 & DT &0.853&0.864&0.853&0.856&0.537&0.746&0.224  &   & 
 DT&0.829&0.833&0.829&0.830&0.488&0.708&0.219 \\
\midrule

\multirow{7}{2cm}{Isolation}
& SVM &0.849&0.849&0.849&0.845&0.507&0.716&0.061 &  \multirow{7}{2cm}{Sigmoid}
& SVM &0.815&0.816&0.815&0.809&0.430&0.671&0.154 \\
 & NB &0.869&0.879&0.869&0.871&0.520&0.751&0.008 &   & 
 NB &0.545&0.830&0.545&0.636&0.367&0.666&0.008 \\
 & MLP &0.836&0.842&0.836&0.834&0.494&0.713&2.171 &   & 
 MLP &0.604&0.785&0.604&0.661&0.365&0.640&1.843 \\
 & KNN &0.844&0.803&0.844&0.813&0.429&0.639&0.013 &   & 
 KNN &0.524&0.796&0.524&0.600&0.327&0.606&0.012 \\
 & RF &0.872&0.868&0.872&0.865&0.478&0.725&0.203 &   & 
 RF &0.874&0.865&0.874&0.868&0.517&0.730&0.633 \\
 & LR &0.790&0.625&0.790&0.698&0.221&0.500&0.017 &   & 
 LR &0.787&0.620&0.787&0.694&0.220&0.500&0.037  \\
 & DT &0.867&0.865&0.867&0.864&0.481&0.724&0.021 &   & 
 DT &0.843&0.853&0.843&0.846&0.498&0.721&0.218 \\
\midrule

\multirow{7}{2cm}{Gaussian}
 & SVM &0.581&0.700&0.581&0.627&0.289&0.562&0.185 & \\
 & NB &0.819&0.847&0.819&0.830&0.455&0.702&0.008  &   & 
 \\
 & MLP  &0.580&0.728&0.580&0.632&0.299&0.579&1.748  &   & 
 \\
 & KNN  &0.796&0.721&0.796&0.724&0.297&0.537&0.015  &   & 
 \\
 & RF  &0.862&0.833&0.862&0.844&0.458&0.689&0.888  &   & 
 \\
 & LR &0.785&0.664&0.785&0.692&0.232&0.504&0.062    &   & 
 \\
 & DT  &0.820&0.818&0.820&0.818&0.426&0.673&0.289   &   &
 \\
\bottomrule

  \end{tabular}
  }

  \caption{Classification comparison for different kernel methods and classifiers on \textbf{Spaced kmer for Breast Cancer}.
  }
  \label{tbl_variant_classification_breastcancer_spaced_kmer}
\end{table}

\clearpage

\section{Runtime Analysis}

\begin{figure}[h!]
\centering
    \begin{subfigure}{0.33\textwidth}
        \centering
        \includegraphics[scale=0.55]{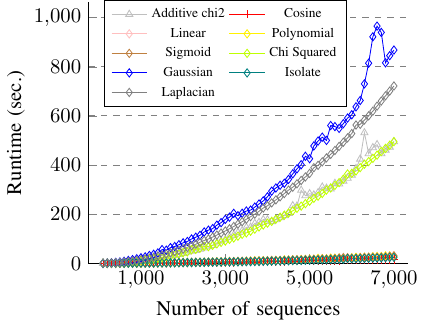}
        \caption{Kernel runtime}
        \label{fig_kernel_runtime}
    \end{subfigure}%
    \begin{subfigure}{0.33\textwidth}
        \centering
        \includegraphics[scale=0.55]{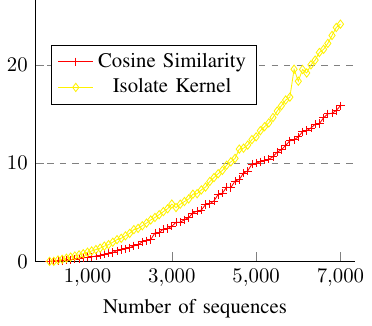}
        \caption{Zoomed Version}
        \label{fig_zoom_kernel}
    \end{subfigure}%
    \begin{subfigure}{0.33\textwidth}
        \centering
        \includegraphics[scale=0.55]{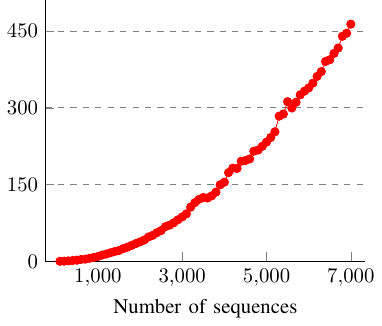}
    \caption{t-SNE runtime}
        \label{fig_data_tsne_runtime}
    \end{subfigure}%
    \caption{(a) and (b) shows the kernel computation Runtime for an increasing number of sequences (using Spike2Vec-based embedding) for the \textbf{Spike7k dataset}. (c) shows the t-SNE computation runtime with an increasing number of sequences (using Spike2Vec-based embedding). This figure is best seen in color.
 }
 \label{fig_data_kernel_runtime}
\end{figure}

Figure~\ref{fig_kernel_runtime} illustrates the computational time required for calculating various kernel matrices as the number of sequences increases, with a magnified view presented in Figure~\ref{fig_zoom_kernel}. The results demonstrate that the Cosine similarity kernel achieves the fastest computation time compared to all other kernel approaches. In contrast, the Gaussian kernel exhibits the highest computational cost. Additionally, the runtime scaling behavior for most kernels follows a linear pattern with respect to the number of input sequences.

Given that the Cosine similarity kernel demonstrates superior performance in general in both objective evaluation metrics and computational efficiency relative to alternative kernel methods for t-SNE, we present the t-SNE computation time using the Cosine similarity kernel across varying sequence counts in Figure~\ref{fig_data_tsne_runtime}. The results reveal a linear scaling relationship between runtime and the number of input sequences.



\section{Conclusion}\label{sec_conclusion}

This comprehensive investigation across six diverse biological datasets demonstrates that, in general, cosine similarity outperforms traditional Gaussian and recently proposed isolation kernels for t-SNE-based dimensionality reduction of biological sequences. Through rigorous evaluation combining subjective visualization assessment, objective neighborhood preservation metrics (AU C$_{RNX}$), and extensive downstream analysis via classification and clustering experiments, we establish that kernel selection fundamentally impacts both visualization quality and analytical task performance. The cosine similarity kernel achieves superior computational efficiency with linear scaling behavior while maintaining robust performance across multiple embedding strategies (OHE, Spike2Vec, minimizers, and spaced k-mers), making it particularly suitable for large-scale genomic analysis. Our classification experiments across seven machine learning algorithms reveal that appropriate kernel-embedding combinations can achieve higher accuracies on certain datasets, while clustering analyses expose substantial performance variations based on kernel choice. Notably, the Gaussian kernel's underperformance across different datasets reinforces that Euclidean distance-based approaches inadequately capture categorical sequence dissimilarity. These findings provide practitioners with evidence-based guidance for method selection in biological sequence analysis pipelines, emphasizing that the angular similarity captured by cosine kernel better represents relationships in high-dimensional genomic feature spaces than distance-based alternatives.









\bibliography{references}

\end{document}